\documentclass[12pt]{article}

\makeatletter
\def\newleaf{\newpage
\newcount\tmp
\tmp=\c@page
\divide\tmp by 2
\multiply\tmp by 2
\ifnum\c@page=\tmp
~\newpage
\fi
}
\makeatother

\expandafter\ifx\csname proofnewpage\endcsname\relax

\fi

\def\color[#1]#2{}

\long\def\nop#1{}

\def\comment{\edef\cps{\the\parskip} \parskip=0.5cm \begingroup \tt}

\def\separator{\vskip 1cm\hrule\vskip 1cm}

\def\noproof#1{
\let\oproof=\proof
\let\oqed=\qed
\def\proof{{#1}\iffalse}
\let\qed=\fi
}
\def\showproof{
\let\proof=\oproof
\let\qed=\oqed
}

\begingroup
\makeatletter
\global\def\fakelabel#1#2{
\expandafter\ifx\csname fakelabelsome\endcsname\relax
\let\fakelabelsome\par
\AtEndDocument{\typeout{}}
\fi
\AtEndDocument{\typeout{NOTE: #1 is a fake label, marked #2}\typeout{}}
\@ifundefined {r@#1}
{\global\@namedef{r@#1}{#2}}
{}
}
\endgroup

\long\def\figurearrow#1#2{
\newbox\before
\setbox\before=\hbox{#1}
\newbox\after
\setbox\after=\hbox{#2}
\newdimen\vdim
\vdim=\ht\before
\ifdim\vdim<\the\ht\after
  \vdim=\ht\after
\fi
\hbox{
\vbox to \the\vdim{\vfill\box\before\vfill}
\vbox to \the\vdim{\vfill\hbox to 3cm{\hfill\Huge$\Rightarrow$\hfill}\vfill}
\vbox to \the\vdim{\vfill\box\after\vfill}
}
}

\expandafter\ifx\csname shortcite\endcsname\relax
\let\shortcite=\cite
\fi

\newbox\current

\long\def\plframebox#1{
\setbox\current\vbox{#1}		

\vbox to \ht\current {\hrule\vss
\hbox to \wd\current {%
\vrule \hss\box\current\hss \vrule}
\vss\hrule }
}

\long\def\eatpar#1{%
\ifx#1\par                      
\let\nextmove=\eatpar           
\else
\let\nextmove=#1
\fi
\noexpand\nextmove
}

\def\modifymargins#1#2{
\newdimen\addtoh
\newdimen\addtow
\addtoh=#1
\addtow=#2

\advance\topmargin by -\addtoh
\multiply\addtoh by 2
\advance\textheight by \addtoh

\advance\oddsidemargin by -\addtow
\advance\evensidemargin by -\addtow
\multiply\addtow by 2
\advance\textwidth by \addtow
}

\begingroup
\catcode`\~=11
\gdef\centertilde#1{\lower #1pt\hbox{~}}
\endgroup

\newcount\currenttime
\newcount\hour
\newcount\minute

\def\printtime{%
\currenttime=\time
\hour=\currenttime
\divide\hour by 60
\minute=-\hour
\multiply\minute by 60
\advance\minute by \currenttime
\the\hour:\ifnum\minute<10 0\fi\the\minute
}

\begingroup
\makeatletter
\global\let\@@date=\@date
\gdef\@date{\@@date\ --- \printtime}
\endgroup

\def\oggi{\number\day\space 
\ifcase\month\or
Gennaio\or Febbraio\or Marzo\or Aprile\or Maggio\or Giugno\or
Luglio\or Agosto\or Settembre\or Ottobre\or Novembre\or Dicembre\fi
\space \number\year}

\newcounter{rmexample}

\def\proof{\noindent {\sl Proof.\ \ }}

\def\qed{\hfill{\boxit{}}
  \ifdim\lastskip<\medskipamount \removelastskip\penalty55\medskip\fi}
\def\qedn#1{\hfill{\boxit{}$_#1$}
  \ifdim\lastskip<\medskipamount \removelastskip\penalty55\medskip\fi}
\long\def\boxit#1{\vbox{\hrule\hbox{\vrule\kern3pt
                  \vbox{\kern3pt#1\kern3pt}\kern3pt\vrule}\hrule}}

\def\true{{\sf true}}

\def\bh#1{\if#1{}{\rm BH}\else\mbox{BH$_{#1}$}\fi}

\def\profont{\sf}

\def\x3c{{\profont x3c}}

\def\possnewtheorem#1#2{
\expandafter\ifx\csname #1\endcsname\relax
\newtheorem{#1}{#2}
\fi
}

\def\possnewtheoremthree#1[#2]#3{
\expandafter\ifx\csname #1\endcsname\relax
\newtheorem{#1}[#2]{#3}
\fi
}

\possnewtheorem{theorem}{Theorem}
\possnewtheorem{corollary}{Corollary}
\possnewtheorem{lemma}{Lemma}
\possnewtheoremthree{proposition}[theorem]{Proposition}
\possnewtheorem{definition}{Definition}
\possnewtheorem{question}{Question}
\possnewtheorem{example}{Example}
\possnewtheorem{nontheorem}{Counterexample}
\possnewtheorem{property}{Property}
\possnewtheorem{assumption}{Assumption}
\possnewtheorem{conjecture}{Conjecture}
\possnewtheorem{notation}{Notation}
\newenvironment{theorem*}[1]{{\noindent \bf Theorem~#1}\begin{it}}{\end{it}\

}

\def\nat{*_{NR}}

\def\after#1#2{#1~{\sf after}~#2}

\def\txt{txt}\ifx\outformat\txt\fi

\iffalse

\expandafter\ifx\csname ttytty\endcsname\relax
{\catcode`\ \active\gdef\charspace{\catcode`\ \active\let ~}}
\else
{\catcode`\ \active\gdef\charspace{\catcode`\ \active\let \asciispace}}
\fi
\def\v{\begingroup\obeylines\hsize 20cm\parskip -0.2cm\charspace\tt}
\def\vv{\endgroup}

\else

\catcode`@11
\def\vfont#1{\font\fonttoshow{#1}\fam\z@\fonttoshow}
\catcode`@12
\def\ttspace{\begingroup\vfont{cmtex10}\char32\endgroup}
{\catcode`\ \active\gdef\charspace{\catcode`\ \active\let \ttspace}}
\def\v{\begingroup\obeylines\parskip -0.2cm\vfont{cmtt10}\charspace}
\def\vv{\endgroup}

\fi

\newenvironment{hfigure}{\refstepcounter{figure}\vbox\bgroup\begin{center}}{\end{center}\egroup}
\def\hcaption#1{\par Figure \thefigure: #1}

\def\draft{\begingroup\em [draft]}
\def\enddraft{[/draft]\endgroup}
\newif\draft
\let\enddraft=\fi

\ifx\outformat\txt\long\def\displaytext#1{\hskip 5cm {\bf #1}}\else
\let\displaytext=\nop\fi

\def\rev{\mathrm{rev}}		
\def\lex{\mathrm{lex}}		
\def\sev{\mathrm{sev}}		
\def\nat{\mathrm{nat}}		
\def\res{\mathrm{res}}		
\def\rad{\mathrm{rad}}		
\def\vrad{\mathrm{vrad}}	
\def\psev{\mathrm{psev}}	
\def\msev{\mathrm{msev}}	
\def\dsev{\mathrm{dsev}}	
\def\full{\mathrm{full}}	

\def\flatorder{C_\epsilon}
\def\formulaorder#1{C_{#1}}
\def\last{\omega}

\makeatletter
\DeclareRobustCommand\imin{\mathop{\operator@font imin}}
\DeclareRobustCommand\imax{\mathop{\operator@font imax}}
\makeatother

\title{Iterated belief revision: from postulates to abilities}
\author{Paolo Liberatore%
\footnote{DIAG, Sapienza University of Rome, Italy.
\tt liberato@diag.uniroma1.it
}
}

\begin{document}

\maketitle

\begin{abstract}

The belief revision field is opulent in new proposals and indigent in analyses
of existing approaches. Much work hinge on postulates, employed as syntactic
characterizations: some revision mechanism is equivalent to some properties.
Postulates constraint specific revision instances: certain revisions update
certain beliefs in a certain way. As an example, if the revision is consistent
with the current beliefs, it is incorporated with no other change. A postulate
like this tells what revisions must do and neglect what they can do. Can they
reach a certain state of beliefs? Can they reach all possible states of
beliefs? Can they reach all possible states of beliefs from no previous belief?
Can they reach a dogmatic state of beliefs, where everything not believed is
impossible? Can they make two conditions equally believed? An application where
every possible state of beliefs is sensible requires each state of beliefs to
be reachable. An application where conditions may be equally believed requires
such a belief state to be reachable. An application where beliefs may become
dogmatic requires a way to make them dogmatic. Such doxastic states need to be
reached in a way or another. Not in specific way, as dictated by a typical
belief revision postulate. This is an ability, not a constraint: the ability of
being plastic, equating, dogmatic. Amnesic, correcting, believer, damascan,
learnable are other abilities. Each revision mechanism owns some of these
abilities and lacks the others: lexicographic, natural, restrained, very
radical, full meet, radical, severe, moderate severe, deep severe, plain severe
and deep severe revisions, each of these revisions is proved to possess certain
abilities.

\end{abstract}


\section{Introduction}
\label{section-introduction}

Just because belief revision in formal logic started with
postulates~\cite{alch-gard-maki-85,gard-88}, postulates do not tell everything
about a belief revision formal framework. They neglect its abilities. They
overlook what it can do, rather than what it must do.

Postulates constraint specific revisions: if the current beliefs are related to
their revisions in a certain way, they change in a certain way. They entail the
new belief~\cite{alch-gard-maki-85,gard-88}. They no longer do it if a
following revision contradicts it~\cite{darw-pear-97}. They change
independently on separate languages~\cite{pari-99}.

An alternative name for belief revision postulates is ``characterizing
properties'' or ``syntactic axiomatization''. A revising mechanism is
characterized or axiomatized by certain properties. A revising mechanism is to
retain the implications of the beliefs as much as possible; it is characterized
by the eight postulates by Alchurron, Gardenfors and
Makinson~\shortcite{alch-gard-maki-85,gard-88}. A revising mechanism is to
retain the strongest beliefs; it is characterized by the four postulates by
Darwich and Pearl~\shortcite{darw-pear-97}.

The strength of the beliefs tells more than the beliefs only. It tells how to
revise: firm beliefs tend to stay, doubted beliefs are prone to leave. It is
the entire doxastic state: not only what is believed, also how it changes. It
tells everything about beliefs from this point on.

Mario and Giulia are both believed trustable. One may not. Maybe none is, but
this is even more disbelieved. The most believable situation is that both are
trustable. Less believable is that one is not. Even less is that none is. This
is an order of the possible situations: ``both are'' is more believed than
``Mario is, Giulia is not'' and ``Mario is not, Giulia is''; which are still
more believed than ``none is''. This order of situations is the doxastic state.

The sun will shine and the beachfront restaurant is open today. The restaurant
is open year-round. The weather forecast is sometimes wrong. The situation
where the sun shines and the restaurant is closed is less believed than the
opposite: ``sun, not open'' is less believed than ``cloudy, open''. This time,
no two situations are equally believed. It is a different doxastic state.

Seeing clouds on the horizon changes the beliefs. Failing to open the website
of the restaurant changes them in another. Beliefs change in every possible
way. The least believed situation with clouds and a closed-down restaurant may
become the most.

\def\textnot{\mbox{not-}}

Abstracting individual conditions by symbols: sun is $s$, open is $o$. The
situation $s,o$ is more believed than $\textnot s,o$, still more believed than
$s,\textnot o$, more believed than $\textnot s,\textnot o$. This is the initial
doxastic state. Revising by $\textnot s$ changes this order. Revising by
$\textnot o$ changes it in another. Revisions change the doxastic state.

Doxastic states change. They change in every possible way.

A revision mechanism unable to change beliefs in every possible way is
insufficient.


The wind brings in the clouds, the weather forecast change, the website of the
restaurant is unreachable, the restaurant road advertisement missing, its sign
is broken. No sun, no lunch. This must be the case, this is now clear. The
other possibilities are just unrealistic. Not only $\textnot s$ and $\textnot
o$ are believed, all other combinations of $s$ and $o$ are equally disbelieved.

Radical changes like this may be accepted gradually and refused all-at-once. Or
the contrary: a single realization may invert all beliefs when small changes
are rejected. It is not a matter of specific sequences of revisions. It is a
matter of reaching a doxastic state in a way or another.

Postulates do not tell this. Postulates tell how seeing the cloud change the
beliefs, how seeing the restaurant sign broken change the beliefs, how other
specific revisions change the beliefs. They do not tell about reaching a state
of beliefs with clouds, hunger and nothing else is possible. They tell about
specific revisions, not about the reachable states of beliefs. They do not tell
about reaching them in a way or another, gradually or suddenly.

Some belief revision mechanisms pass all postulate tests with flying colors,
but exclude such doxastic states. Every single revision or sequence of
revisions meets the rules. Yet, none is able to change the doxastic state this
way. They never reach the dogmatic conclusion that there will be no sun and
no lunch. Lexicographic revision is an example~\cite{spoh-88,naya-94}.


A different revision mechanism may do it. Full meet revision is an
example~\cite{libe-97-c,libe-23}. At the same time, full meet revision fails to
reach other doxastic states. Lexicographic and full meet revision together do
it.




It is not a matter of satisfying rules. It is not a matter of how one or two
specific revisions change specific doxastic states: one or two specific
revisions may not reach such a firm conclusion. It is a matter of how revisions
can change beliefs. Not how they change, but how they can change. Which
doxastic states can be turned in which others by some sequence of revisions.
Not the rules they meet, but the abilities they possess.

Beliefs may change in every possible way; the revision mechanism must be
plastic. Beliefs may be dogmatic: something is believed so firmly that
everything else is totally disbelieved; the revision mechanism must be
dogmatic. Beliefs may be learned from scratch; the revision mechanism must be
learnable. Beliefs may become equally entrenched; the revision mechanism must
be equating. The following Section~\ref{section-list} lists some abilities of
interest.

Planning a summer vacation leads to a website reporting the five best beaches
of Elefonisos; ten are common for the other Greek islands. A state of complete
ignorance about the island turns into a state of believing that it has few
beaches, but they are beautiful as evidenced by the pictures. From all
situations equally doubted to $f, b$ more believed than $\textnot f, b$, more
believed than $f, \textnot b$, still more believed than $\textnot f, \textnot
b$. A revision turning the doxastic state devoid of beliefs into an arbitrary
doxastic state possesses the learnable ability.

Seeing a Moon eclipse in a lifetime is believed more likely than seeing a Sun
eclipse. A new job leads to foreign cities where both will take place.
Differently believed situations turn into equally so. A revision equalizing the
strength of belief possesses the equating ability.

A revision equalizing the strength of all beliefs is amnesic. A revision
inverting them is damascan. A revision inverting two is correcting. Amnesic,
damascan, correcting, equating, learnable, dogmatic, plastic. Some revisions
possess some of these abilities and not the others. Some are learnable and not
damascan. They are fitted for the Greek island, they are unfitted for the
eclipses.

The belief revision mechanisms are introduced in
Section~\ref{section-operators}: natural, lexicographic, restrained, radical,
very radical, full meet, severe, moderate severe, deep severe and plain severe.
The abilities are in Section~\ref{section-list}: fully plastic, plastic,
amnesic, equating, dogmatic, believer, learnable, damascan, correcting.
Section~\ref{section-results} tells which of these abilities each revision
possesses. After some concluding comments in Section~\ref{section-conclusions},
all mathematical results are given in the Appendices.

\section{Belief revision}
\label{section-operators}

The doxastic state is a connected preorder $C$ between propositional models
over a finite alphabet. Other forms of the doxastic state%
{}%
{}~\cite{arav-etal-18,andr-etal-02,gura-kodi-18,souz-etal-19,brew-89,nebe-91,
benf-etal-93,kuts-19,andr-etal-02,saue-etal-22,doms-etal-11}
{}%
are not considered. A connected preorder $C$ is written as the sequence of its
equivalence classes
{} $[C(0), C(1), \ldots, C(\last - 1), C(\last)]$,
where $\last$ is the index of the last, the class comprising the least believed
models. The $\last$ notation abstracts from the differing numbers: $C(\last)$
is the last class of $C$ and $G(\last)$ the last of $G$ even if these doxastic
states differ in their number of classes: $\last$ is their last, $\last - 1$
their second-to-last.

Figure~\ref{figure-order} is a graphical depiction of a connected preorder. The
most believed situations $C(0)$ are at the top, the least $C(\last)$ at the
bottom.

A propositional interpretation $I$ is less than another $J$ if $i < j$ where $I
\in C(i)$, $J \in C(j)$; such indexes $i$ and $j$ uniquely exist because $C$ is
a partition. This condition is denoted $I <_C J$. Less than or equal to is
denoted $I \leq_C J$. Equality is $I \equiv_C J$.

\begin{hfigure}
%
%
\setlength{\unitlength}{3750sp}%
\begin{picture}(1224,2124)(5089,-5173)
\thinlines
{\color[rgb]{0,0,0}\put(5101,-3361){\line( 1, 0){1200}}
}%
{\color[rgb]{0,0,0}\put(5101,-3661){\line( 1, 0){1200}}
}%
{\color[rgb]{0,0,0}\put(5101,-3961){\line( 1, 0){1200}}
}%
{\color[rgb]{0,0,0}\put(5101,-4561){\line( 1, 0){1200}}
}%
{\color[rgb]{0,0,0}\put(5101,-5161){\framebox(1200,2100){}}
}%
{\color[rgb]{0,0,0}\put(5101,-4861){\line( 1, 0){1200}}
}%
\put(5701,-3286){\makebox(0,0)[b]{\smash{\fontsize{9}{10.8}
\usefont{T1}{cmr}{m}{n}{\color[rgb]{0,0,0}$C(0)$}%
}}}
\put(5701,-3586){\makebox(0,0)[b]{\smash{\fontsize{9}{10.8}
\usefont{T1}{cmr}{m}{n}{\color[rgb]{0,0,0}$C(1)$}%
}}}
\put(5701,-3886){\makebox(0,0)[b]{\smash{\fontsize{9}{10.8}
\usefont{T1}{cmr}{m}{n}{\color[rgb]{0,0,0}$C(2)$}%
}}}
\put(5701,-4336){\makebox(0,0)[b]{\smash{\fontsize{9}{10.8}
\usefont{T1}{cmr}{m}{n}{\color[rgb]{0,0,0}$\vdots$}%
}}}
\put(5701,-5086){\makebox(0,0)[b]{\smash{\fontsize{9}{10.8}
\usefont{T1}{cmr}{m}{n}{\color[rgb]{0,0,0}$C(\last)$}%
}}}
\put(5701,-4786){\makebox(0,0)[b]{\smash{\fontsize{9}{10.8}
\usefont{T1}{cmr}{m}{n}{\color[rgb]{0,0,0}$C(\last-1)$}%
}}}
\end{picture}%
\nop{
 +----------+
 |   C(0)   |
 +----------+
 |   C(1)   |
 +----------+
 |   C(2)   |
 +----------+
 |          |
 |          |
 |   ...    |
 |          |
 |          |
 +----------+
 |  C(o-1)  |
 +----------+
 |   C(o)   |
 +----------+
}
\label{figure-order}
\hcaption{A doxastic state}
\end{hfigure}

An equivalence class is a list of models or a formula satisfied by them and no
other. More generally, a formula may take the place of a set of models. For
example,
{} $C(i) \cap (x \vee \neg y)$
stands for
{} $C(i) \cap \{\{x,y\}, \{x, \neg y\}, \{\neg x, \neg y\}\}$,
where each model is written as the set of literals it satisfies.

Similar to the omission of multiplication in algebraic expressions like
$a(b+c)$ stands for $a \times (b+c)$, conjunctions are omitted in Boolean
formulae and $\neg a(b \vee c)$ stands for $\neg a \wedge (b \vee c)$.

Two orders are especially relevant: the flat order $\flatorder = [\true]$ and
the order of a formula $\formulaorder A = [A, \true \backslash A]$.

Revisions change the doxastic state. Each revision changes the doxastic state
its way. It changes $C$ into $C \rev(A)$, where $A$ is a Boolean formula and
$\rev()$ a revision mechanism like natural revision $\nat()$, lexicographic
revision $\lex()$ or restrained revision $\rev()$.

\

Revisions differ on two orthogonal dimensions: contingent revisions apply to
the current situation only, not all; epiphanic revisions cancel some unrelated
belief.

\begin{description}

\item[contingent]

revisions may apply to the current situation or to all of them;

Bruno may believe that his cat is under the rusted pickup just because he also
believes that his cat is around there; he does not if he learns that it has
been found in the train station; at the same time, he may believe hearing
meowing, no matter where his cat is~\cite{libe-25};

a hunter may believe a bird is red because a postman told having seen a red
bird in a thicket; when no bird is found there, believing it red vanishes as
well; redness is believed only as long as the presence of the bird in the
thicket; at the same time, a zoologist may read that exotic, red birds started
to populate the area, regardless of what the postman said~\cite{libe-23};

\item[epiphanic]

revisions may erase beliefs in unrelated situations;

believing a situation $I$ more than another $J$ may vanish when a still less
believed situation $Z$ becomes the most; such a radical inversion from $I < J <
Z$ to $Z < I$ and $Z< J$ may suggest ignorance of $I$ and $J$, eroding their
differing strength of belief;

believing that Mario comes from Brighton more than it comes from London may be
doubted when he is found to be a pathological liar; a change in a belief in
something may undermine the belief in something else;

learning that Napoleon was not a short man may suggest a need to study his
personal life rather than trusting what believed for granted, such that he
never married.

\end{description}

\

A revision is either contingent or not contingent: it applies to the currently
most believed situations or to all of them; it is yes or no. A revision may be
more or less epiphanic; it may erase few beliefs or many. At one end of the
scale, it only erases what previously most believed, like severe revision. At
the other, it erases all doubtful beliefs, like very radical revision. Other
revisions are in the middle.

Revisions class on whether they are contingent and how much they are epiphanic.

\begin{description}

\item[Natural revision]

is contingent: the most believed situations supporting the new belief $A$
become the most believed of all. Graphically, the top class of $A$ moves to the
top, as shown in Figure~\ref{figure-natural-changes}.

The change does not affect any other situation. Revising does not erase
unrelated beliefs. Natural revision is not epiphanic.

\begin{hfigure}
\long\def\ttytex#1#2{#1}
\ttytex{
\begin{tabular}{ccc}
\setlength{\unitlength}{3750sp}%
\begin{picture}(1224,1824)(5089,-4873)
\thinlines
{\color[rgb]{0,0,0}\put(5101,-3361){\line( 1, 0){1200}}
}%
{\color[rgb]{0,0,0}\put(5101,-4861){\framebox(1200,1800){}}
}%
{\color[rgb]{0,0,0}\put(5101,-3961){\line( 1, 0){1200}}
}%
{\color[rgb]{0,0,0}\put(5101,-4261){\line( 1, 0){1200}}
}%
{\color[rgb]{0,0,0}\put(5101,-4561){\line( 1, 0){1200}}
}%
{\color[rgb]{0,0,0}\put(5401,-4486){\framebox(750,1050){}}
}%
{\color[rgb]{0,0,0}\put(5101,-3661){\line( 1, 0){1200}}
}%
\put(5851,-3886){\makebox(0,0)[b]{\smash{\fontsize{9}{10.8}
\usefont{T1}{cmr}{m}{n}{\color[rgb]{0,0,0}$A$}%
}}}
\end{picture}%
&
\setlength{\unitlength}{3750sp}%
\begin{picture}(1104,1104)(5659,-4303)
\thinlines
{\color[rgb]{0,0,0}\multiput(6391,-3391)(-9.47368,4.73684){20}{\makebox(2.1167,14.8167){\tiny.}}
\put(6211,-3301){\line( 0,-1){ 45}}
\put(6211,-3346){\line(-1, 0){180}}
\put(6031,-3346){\line( 0,-1){ 90}}
\put(6031,-3436){\line( 1, 0){180}}
\put(6211,-3436){\line( 0,-1){ 45}}
\multiput(6211,-3481)(9.47368,4.73684){20}{\makebox(2.1167,14.8167){\tiny.}}
}%
\end{picture}%
&
\setlength{\unitlength}{3750sp}%
\begin{picture}(1224,2049)(5089,-4873)
\thinlines
{\color[rgb]{0,0,0}\put(5101,-3361){\line( 1, 0){1200}}
}%
{\color[rgb]{0,0,0}\put(5101,-3661){\line( 1, 0){1200}}
}%
{\color[rgb]{0,0,0}\put(5101,-4861){\framebox(1200,1800){}}
}%
{\color[rgb]{0,0,0}\put(5101,-3961){\line( 1, 0){1200}}
}%
{\color[rgb]{0,0,0}\put(5101,-4261){\line( 1, 0){1200}}
}%
{\color[rgb]{0,0,0}\put(5101,-4561){\line( 1, 0){1200}}
}%
{\color[rgb]{0,0,0}\put(5401,-3061){\framebox(750,225){}}
}%
{\color[rgb]{0,0,0}\put(5401,-4486){\framebox(750,1050){}}
}%
{\color[rgb]{0,0,0}\multiput(5401,-3586)(7.89474,7.89474){20}{\makebox(2.2222,15.5556){\tiny.}}
}%
{\color[rgb]{0,0,0}\multiput(5401,-3661)(8.03571,8.03571){29}{\makebox(2.2222,15.5556){\tiny.}}
}%
{\color[rgb]{0,0,0}\multiput(5401,-3511)(8.33333,8.33333){10}{\makebox(2.2222,15.5556){\tiny.}}
}%
{\color[rgb]{0,0,0}\multiput(5476,-3661)(8.03571,8.03571){29}{\makebox(2.2222,15.5556){\tiny.}}
}%
{\color[rgb]{0,0,0}\multiput(5551,-3661)(8.03571,8.03571){29}{\makebox(2.2222,15.5556){\tiny.}}
}%
{\color[rgb]{0,0,0}\multiput(5626,-3661)(8.03571,8.03571){29}{\makebox(2.2222,15.5556){\tiny.}}
}%
{\color[rgb]{0,0,0}\multiput(5701,-3661)(8.03571,8.03571){29}{\makebox(2.2222,15.5556){\tiny.}}
}%
{\color[rgb]{0,0,0}\multiput(5776,-3661)(8.03571,8.03571){29}{\makebox(2.2222,15.5556){\tiny.}}
}%
{\color[rgb]{0,0,0}\multiput(5851,-3661)(8.03571,8.03571){29}{\makebox(2.2222,15.5556){\tiny.}}
}%
{\color[rgb]{0,0,0}\multiput(5926,-3661)(8.03571,8.03571){29}{\makebox(2.2222,15.5556){\tiny.}}
}%
{\color[rgb]{0,0,0}\multiput(6001,-3661)(7.89474,7.89474){20}{\makebox(2.2222,15.5556){\tiny.}}
}%
{\color[rgb]{0,0,0}\multiput(6076,-3661)(8.33333,8.33333){10}{\makebox(2.2222,15.5556){\tiny.}}
}%
\end{picture}%
\end{tabular}
}{
                             +------+ min(A)
                             +------+
+-------------+          +-------------+
|             |          |             |
+-------------+          +-------------+
|   +------+  |          |   +------+  |
+---|------|--+          +---+      +--+
|   |   A  |  |    =>
+---|------|--+          +---+------+--+
|   |      |  |          |   |      |  |
+---|------|--+          +---|------|--+
|   +------+  |          |   |      |  |
+-------------+          +---|------|--+
|             |          |   +------+  |
+-------------+          +-------------+
|             |          |             |
+-------------+          +-------------+
|             |          |             |
+-------------+          +-------------+
                         |             |
                         +-------------+
  current.fig              natural.fig
}
\label{figure-natural-changes}
\hcaption{Natural revision}
\end{hfigure}

\item[Lexicographic revision]

makes all situations supporting the new belief more believed than all others.
Not only the currently most believed situations like natural revision, but all
of them. It is not contingent.

Like natural revision, lexicographic revision does not erase unrelated beliefs.
It is not epiphanic.

\begin{hfigure}
\long\def\ttytex#1#2{#1}
\ttytex{
\begin{tabular}{ccc}
\setlength{\unitlength}{3750sp}%
\begin{picture}(1224,1824)(5089,-4873)
\thinlines
{\color[rgb]{0,0,0}\put(5101,-3361){\line( 1, 0){1200}}
}%
{\color[rgb]{0,0,0}\put(5101,-4861){\framebox(1200,1800){}}
}%
{\color[rgb]{0,0,0}\put(5101,-3961){\line( 1, 0){1200}}
}%
{\color[rgb]{0,0,0}\put(5101,-4261){\line( 1, 0){1200}}
}%
{\color[rgb]{0,0,0}\put(5101,-4561){\line( 1, 0){1200}}
}%
{\color[rgb]{0,0,0}\put(5401,-4486){\framebox(750,1050){}}
}%
{\color[rgb]{0,0,0}\put(5101,-3661){\line( 1, 0){1200}}
}%
\put(5851,-3886){\makebox(0,0)[b]{\smash{\fontsize{9}{10.8}
\usefont{T1}{cmr}{m}{n}{\color[rgb]{0,0,0}$A$}%
}}}
\end{picture}%
&
\setlength{\unitlength}{3750sp}%
\begin{picture}(1104,1104)(5659,-4303)
\thinlines
{\color[rgb]{0,0,0}\multiput(6391,-3391)(-9.47368,4.73684){20}{\makebox(2.1167,14.8167){\tiny.}}
\put(6211,-3301){\line( 0,-1){ 45}}
\put(6211,-3346){\line(-1, 0){180}}
\put(6031,-3346){\line( 0,-1){ 90}}
\put(6031,-3436){\line( 1, 0){180}}
\put(6211,-3436){\line( 0,-1){ 45}}
\multiput(6211,-3481)(9.47368,4.73684){20}{\makebox(2.1167,14.8167){\tiny.}}
}%
\end{picture}%
&
\setlength{\unitlength}{3750sp}%
\begin{picture}(1224,2949)(5089,-4873)
\thinlines
{\color[rgb]{0,0,0}\put(5101,-3361){\line( 1, 0){1200}}
}%
{\color[rgb]{0,0,0}\put(5101,-4861){\framebox(1200,1800){}}
}%
{\color[rgb]{0,0,0}\put(5101,-4561){\line( 1, 0){1200}}
}%
{\color[rgb]{0,0,0}\put(5401,-4486){\framebox(750,1050){}}
}%
{\color[rgb]{0,0,0}\put(5401,-2986){\framebox(750,1050){}}
}%
{\color[rgb]{0,0,0}\put(5401,-2161){\line( 1, 0){750}}
}%
{\color[rgb]{0,0,0}\put(5401,-2461){\line( 1, 0){750}}
}%
{\color[rgb]{0,0,0}\put(5401,-2761){\line( 1, 0){750}}
}%
{\color[rgb]{0,0,0}\put(5101,-3661){\line( 1, 0){300}}
}%
{\color[rgb]{0,0,0}\put(6151,-3661){\line( 1, 0){150}}
}%
{\color[rgb]{0,0,0}\put(5101,-3961){\line( 1, 0){300}}
}%
{\color[rgb]{0,0,0}\put(5101,-4261){\line( 1, 0){300}}
}%
{\color[rgb]{0,0,0}\put(6151,-3961){\line( 1, 0){150}}
}%
{\color[rgb]{0,0,0}\put(6151,-4261){\line( 1, 0){150}}
}%
\put(5851,-2386){\makebox(0,0)[b]{\smash{\fontsize{9}{10.8}
\usefont{T1}{cmr}{m}{n}{\color[rgb]{0,0,0}$A$}%
}}}
\end{picture}%
\end{tabular}
}{
                             +------+
                             |------|
                             |   A  |
                             |------|
                             |      |
                             |------|
                             +------+
+-------------+          +-------------+
|             |          |             |
+-------------+          +-------------+
|   +------+  |          |   +------+  |
+---|------|--+          +---|      |--+
|   |   A  |  |    =>    |   |      |  |
+---|------|--+          +---|      |--+
|   |      |  |          |   |      |  |
+---|------|--+          +---|      |--+
|   +------+  |          |   +------+  |
+-------------+          +-------------+
|             |          |             |
+-------------+          +-------------+
|             |          |             |
+-------------+          +-------------+
|             |          |             |
+-------------+          +-------------+
  current.fig           lexicographic.fig
}
\label{figure-lexicographic-revision}
\hcaption{Lexicographic revision}
\end{hfigure}

\item[Restrained revision]

is somehow in the middle between natural and lexicographic revision. Like
natural revision, it fully trusts the new belief in the currently most believed
scenarios. Similarly to lexicographic revision, it also trusts the new belief
in the other scenarios but very weakly, unlike lexicographic revisions: less
than the previous revisions.

Restrained revision is not epiphanic. It does not erase a belief unrelated with
the new one.

\begin{hfigure}
\long\def\ttytex#1#2{#1}
\ttytex{
\begin{tabular}{ccc}
\setlength{\unitlength}{3750sp}%
\begin{picture}(1224,1824)(5089,-4873)
\thinlines
{\color[rgb]{0,0,0}\put(5101,-3361){\line( 1, 0){1200}}
}%
{\color[rgb]{0,0,0}\put(5101,-4861){\framebox(1200,1800){}}
}%
{\color[rgb]{0,0,0}\put(5101,-3961){\line( 1, 0){1200}}
}%
{\color[rgb]{0,0,0}\put(5101,-4261){\line( 1, 0){1200}}
}%
{\color[rgb]{0,0,0}\put(5101,-4561){\line( 1, 0){1200}}
}%
{\color[rgb]{0,0,0}\put(5401,-4486){\framebox(750,1050){}}
}%
{\color[rgb]{0,0,0}\put(5101,-3661){\line( 1, 0){1200}}
}%
\put(5851,-3886){\makebox(0,0)[b]{\smash{\fontsize{9}{10.8}
\usefont{T1}{cmr}{m}{n}{\color[rgb]{0,0,0}$A$}%
}}}
\end{picture}%
&
\setlength{\unitlength}{3750sp}%
\begin{picture}(1104,1104)(5659,-4303)
\thinlines
{\color[rgb]{0,0,0}\multiput(6391,-3391)(-9.47368,4.73684){20}{\makebox(2.1167,14.8167){\tiny.}}
\put(6211,-3301){\line( 0,-1){ 45}}
\put(6211,-3346){\line(-1, 0){180}}
\put(6031,-3346){\line( 0,-1){ 90}}
\put(6031,-3436){\line( 1, 0){180}}
\put(6211,-3436){\line( 0,-1){ 45}}
\multiput(6211,-3481)(9.47368,4.73684){20}{\makebox(2.1167,14.8167){\tiny.}}
}%
\end{picture}%
&
\setlength{\unitlength}{3750sp}%
\begin{picture}(1224,3399)(5089,-6223)
\thinlines
{\color[rgb]{0,0,0}\put(5401,-4111){\framebox(750,300){}}
}%
{\color[rgb]{0,0,0}\put(5101,-6211){\framebox(1200,300){}}
}%
{\color[rgb]{0,0,0}\put(5101,-5611){\line( 1, 0){300}}
\put(5401,-5611){\line( 0,-1){225}}
\put(5401,-5836){\line( 1, 0){750}}
\put(6151,-5836){\line( 0, 1){225}}
\put(6151,-5611){\line( 1, 0){150}}
\put(6301,-5611){\line( 0,-1){300}}
\put(6301,-5911){\line(-1, 0){1200}}
\put(5101,-5911){\line( 0, 1){300}}
}%
{\color[rgb]{0,0,0}\put(5401,-5536){\framebox(750,225){}}
}%
{\color[rgb]{0,0,0}\put(5101,-5236){\framebox(300,300){}}
}%
{\color[rgb]{0,0,0}\put(6151,-5236){\framebox(150,300){}}
}%
{\color[rgb]{0,0,0}\put(5401,-4861){\framebox(750,300){}}
}%
{\color[rgb]{0,0,0}\put(5101,-4486){\framebox(300,300){}}
}%
{\color[rgb]{0,0,0}\put(6151,-4486){\framebox(150,300){}}
}%
{\color[rgb]{0,0,0}\put(5101,-3736){\line( 0, 1){300}}
\put(5101,-3436){\line( 1, 0){1200}}
\put(6301,-3436){\line( 0,-1){300}}
\put(6301,-3736){\line(-1, 0){150}}
\put(6151,-3736){\line( 0, 1){225}}
\put(6151,-3511){\line(-1, 0){750}}
\put(5401,-3511){\line( 0,-1){225}}
\put(5401,-3736){\line(-1, 0){300}}
}%
{\color[rgb]{0,0,0}\put(5101,-3436){\framebox(1200,300){}}
}%
{\color[rgb]{0,0,0}\put(5401,-3061){\framebox(750,225){}}
}%
\end{picture}%
\end{tabular}
}{
                             +------+ min(A)
                             +------+
                         +-------------+
                         |             |
                         +-------------+
+-------------+          |   +------+  |
|             |          +---+      +--+
+-------------+              +------+
|   +------+  |              |      |
+---|------|--+          +---+------+--+
|   |   A  |  |    =>    |   |      |  |
+---|------|--+          +---+      +--+
|   |      |  |              +------+
+---|------|--+              |      |
|   +------+  |          +---|------|--+
+-------------+          |   |      |  |
|             |          +---+      +--+
+-------------+              +------+
|             |              +------+
+-------------+          +---+      +--+
|             |          |   +------+  |
+-------------+          +-------------+
  current.fig            |             |
                         +-------------+
                         |             |
                         +-------------+
                         |             |
                         +-------------+
                          restrained.fig
}
\label{figure-restrained-revision}
\hcaption{Restrained revision}
\end{hfigure}

\item[Very radical revision]

is not contingent: it held the new belief in all situations, not only the most
believed ones.

At the same time, it erases all beliefs contrary to the new one. It is maximally
epiphanic.

\begin{hfigure}
\long\def\ttytex#1#2{#1}
\ttytex{
\begin{tabular}{ccc}
\setlength{\unitlength}{3750sp}%
\begin{picture}(1224,1824)(5089,-4873)
\thinlines
{\color[rgb]{0,0,0}\put(5101,-3361){\line( 1, 0){1200}}
}%
{\color[rgb]{0,0,0}\put(5101,-4861){\framebox(1200,1800){}}
}%
{\color[rgb]{0,0,0}\put(5101,-3961){\line( 1, 0){1200}}
}%
{\color[rgb]{0,0,0}\put(5101,-4261){\line( 1, 0){1200}}
}%
{\color[rgb]{0,0,0}\put(5101,-4561){\line( 1, 0){1200}}
}%
{\color[rgb]{0,0,0}\put(5401,-4486){\framebox(750,1050){}}
}%
{\color[rgb]{0,0,0}\put(5101,-3661){\line( 1, 0){1200}}
}%
\put(5851,-3886){\makebox(0,0)[b]{\smash{\fontsize{9}{10.8}
\usefont{T1}{cmr}{m}{n}{\color[rgb]{0,0,0}$A$}%
}}}
\end{picture}%
&
\setlength{\unitlength}{3750sp}%
\begin{picture}(1104,1104)(5659,-4303)
\thinlines
{\color[rgb]{0,0,0}\multiput(6391,-3391)(-9.47368,4.73684){20}{\makebox(2.1167,14.8167){\tiny.}}
\put(6211,-3301){\line( 0,-1){ 45}}
\put(6211,-3346){\line(-1, 0){180}}
\put(6031,-3346){\line( 0,-1){ 90}}
\put(6031,-3436){\line( 1, 0){180}}
\put(6211,-3436){\line( 0,-1){ 45}}
\multiput(6211,-3481)(9.47368,4.73684){20}{\makebox(2.1167,14.8167){\tiny.}}
}%
\end{picture}%
&
\setlength{\unitlength}{3750sp}%
\begin{picture}(1224,2949)(5089,-4873)
\thinlines
{\color[rgb]{0,0,0}\put(5101,-4861){\framebox(1200,1800){}}
}%
{\color[rgb]{0,0,0}\put(5401,-4486){\framebox(750,1050){}}
}%
{\color[rgb]{0,0,0}\put(5401,-2986){\framebox(750,1050){}}
}%
{\color[rgb]{0,0,0}\put(5401,-2161){\line( 1, 0){750}}
}%
{\color[rgb]{0,0,0}\put(5401,-2461){\line( 1, 0){750}}
}%
{\color[rgb]{0,0,0}\put(5401,-2761){\line( 1, 0){750}}
}%
\put(5851,-2386){\makebox(0,0)[b]{\smash{\fontsize{9}{10.8}
\usefont{T1}{cmr}{m}{n}{\color[rgb]{0,0,0}$A$}%
}}}
\end{picture}%
\end{tabular}
}{
                             +------+
                             |------|
                             |   A  |
                             |------|
                             |      |
                             |------|
                             +------+
+-------------+          +-------------+
|             |          |             |
+-------------+          |             |
|   +------+  |          |   +------+  |
+---|------|--+          |   |      |  |
|   |   A  |  |    =>    |   |      |  |
+---|------|--+          |   |      |  |
|   |      |  |          |   |      |  |
+---|------|--+          |   |      |  |
|   +------+  |          |   +------+  |
+-------------+          |             |
|             |          |             |
+-------------+          |             |
|             |          |             |
+-------------+          |             |
|             |          |             |
+-------------+          +-------------+
  current.fig            veryradical.fig
}
\label{figure-veryradical-revision}
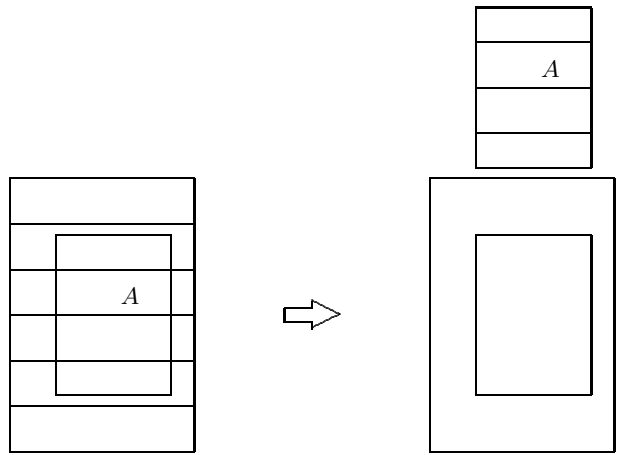
\hcaption{Very radical revision}
\end{hfigure}

\item[Full meet revision]

is contingent. It trusts the new belief only in the currently most believed
situations, like natural and restrained revision.

It is the contingent version of very radical revision: everything else is
disbelieved the same. It is maximally epiphanic.

\begin{hfigure}
\long\def\ttytex#1#2{#1}
\ttytex{
\begin{tabular}{ccc}
\setlength{\unitlength}{3750sp}%
\begin{picture}(1224,1824)(5089,-4873)
\thinlines
{\color[rgb]{0,0,0}\put(5101,-3361){\line( 1, 0){1200}}
}%
{\color[rgb]{0,0,0}\put(5101,-4861){\framebox(1200,1800){}}
}%
{\color[rgb]{0,0,0}\put(5101,-3961){\line( 1, 0){1200}}
}%
{\color[rgb]{0,0,0}\put(5101,-4261){\line( 1, 0){1200}}
}%
{\color[rgb]{0,0,0}\put(5101,-4561){\line( 1, 0){1200}}
}%
{\color[rgb]{0,0,0}\put(5401,-4486){\framebox(750,1050){}}
}%
{\color[rgb]{0,0,0}\put(5101,-3661){\line( 1, 0){1200}}
}%
\put(5851,-3886){\makebox(0,0)[b]{\smash{\fontsize{9}{10.8}
\usefont{T1}{cmr}{m}{n}{\color[rgb]{0,0,0}$A$}%
}}}
\end{picture}%
&
\setlength{\unitlength}{3750sp}%
\begin{picture}(1104,1104)(5659,-4303)
\thinlines
{\color[rgb]{0,0,0}\multiput(6391,-3391)(-9.47368,4.73684){20}{\makebox(2.1167,14.8167){\tiny.}}
\put(6211,-3301){\line( 0,-1){ 45}}
\put(6211,-3346){\line(-1, 0){180}}
\put(6031,-3346){\line( 0,-1){ 90}}
\put(6031,-3436){\line( 1, 0){180}}
\put(6211,-3436){\line( 0,-1){ 45}}
\multiput(6211,-3481)(9.47368,4.73684){20}{\makebox(2.1167,14.8167){\tiny.}}
}%
\end{picture}%
&
\setlength{\unitlength}{3750sp}%
\begin{picture}(1224,2049)(5089,-4873)
\thinlines
{\color[rgb]{0,0,0}\put(5101,-4861){\framebox(1200,1800){}}
}%
{\color[rgb]{0,0,0}\put(5401,-3061){\framebox(750,225){}}
}%
{\color[rgb]{0,0,0}\put(5401,-3661){\framebox(750,225){}}
}%
{\color[rgb]{0,0,0}\multiput(5401,-3511)(8.33333,8.33333){10}{\makebox(2.2222,15.5556){\tiny.}}
}%
{\color[rgb]{0,0,0}\multiput(6076,-3661)(8.33333,8.33333){10}{\makebox(2.2222,15.5556){\tiny.}}
}%
{\color[rgb]{0,0,0}\multiput(5401,-3586)(7.89474,7.89474){20}{\makebox(2.2222,15.5556){\tiny.}}
}%
{\color[rgb]{0,0,0}\multiput(5401,-3661)(8.03571,8.03571){29}{\makebox(2.2222,15.5556){\tiny.}}
}%
{\color[rgb]{0,0,0}\multiput(5476,-3661)(8.03571,8.03571){29}{\makebox(2.2222,15.5556){\tiny.}}
}%
{\color[rgb]{0,0,0}\multiput(5551,-3661)(8.03571,8.03571){29}{\makebox(2.2222,15.5556){\tiny.}}
}%
{\color[rgb]{0,0,0}\multiput(5626,-3661)(8.03571,8.03571){29}{\makebox(2.2222,15.5556){\tiny.}}
}%
{\color[rgb]{0,0,0}\multiput(5701,-3661)(8.03571,8.03571){29}{\makebox(2.2222,15.5556){\tiny.}}
}%
{\color[rgb]{0,0,0}\multiput(5776,-3661)(8.03571,8.03571){29}{\makebox(2.2222,15.5556){\tiny.}}
}%
{\color[rgb]{0,0,0}\multiput(5851,-3661)(8.03571,8.03571){29}{\makebox(2.2222,15.5556){\tiny.}}
}%
{\color[rgb]{0,0,0}\multiput(5926,-3661)(8.03571,8.03571){29}{\makebox(2.2222,15.5556){\tiny.}}
}%
{\color[rgb]{0,0,0}\multiput(6001,-3661)(7.89474,7.89474){20}{\makebox(2.2222,15.5556){\tiny.}}
}%
\end{picture}%
\end{tabular}
}{
                             +------+ min(A)
                             +------+
+-------------+          +-------------+
|             |          |             |
+-------------+          |             |
|   +------+  |          |   ////////  |
+---|------|--+          |   ////////  |
|   |   A  |  |    =>    |             |
+---|------|--+          |             |
|   |      |  |          |             |
+---|------|--+          |             |
|   +------+  |          |             |
+-------------+          |             |
|             |          |             |
+-------------+          |             |
|             |          |             |
+-------------+          |             |
|             |          |             |
+-------------+          +-------------+
  current.fig             fullmeet.fig
}
\label{figure-fullmeet-revision}
\hcaption{Full meet revision}
\end{hfigure}

\item[Radical revision]

differs from very radical only if the new belief accords with the currently
least believed situations. These are deemed impossible rather than just
currently unbelieved. They forever remain impossible, no matter of what new
beliefs say.

\begin{hfigure}
\long\def\ttytex#1#2{#1}
\ttytex{
\begin{tabular}{ccc}
\setlength{\unitlength}{3750sp}%
\begin{picture}(1224,1824)(5089,-4873)
\thinlines
{\color[rgb]{0,0,0}\put(5101,-3361){\line( 1, 0){1200}}
}%
{\color[rgb]{0,0,0}\put(5101,-3661){\line( 1, 0){1200}}
}%
{\color[rgb]{0,0,0}\put(5101,-4861){\framebox(1200,1800){}}
}%
{\color[rgb]{0,0,0}\put(5101,-3961){\line( 1, 0){1200}}
}%
{\color[rgb]{0,0,0}\put(5101,-4261){\line( 1, 0){1200}}
}%
{\color[rgb]{0,0,0}\put(5101,-4561){\line( 1, 0){1200}}
}%
{\color[rgb]{0,0,0}\put(5401,-4786){\framebox(750,1350){}}
}%
\put(5851,-3886){\makebox(0,0)[b]{\smash{\fontsize{9}{10.8}
\usefont{T1}{cmr}{m}{n}{\color[rgb]{0,0,0}$A$}%
}}}
\end{picture}%
&
\setlength{\unitlength}{3750sp}%
\begin{picture}(1104,1104)(5659,-4303)
\thinlines
{\color[rgb]{0,0,0}\multiput(6391,-3391)(-9.47368,4.73684){20}{\makebox(2.1167,14.8167){\tiny.}}
\put(6211,-3301){\line( 0,-1){ 45}}
\put(6211,-3346){\line(-1, 0){180}}
\put(6031,-3346){\line( 0,-1){ 90}}
\put(6031,-3436){\line( 1, 0){180}}
\put(6211,-3436){\line( 0,-1){ 45}}
\multiput(6211,-3481)(9.47368,4.73684){20}{\makebox(2.1167,14.8167){\tiny.}}
}%
\end{picture}%
&
\setlength{\unitlength}{3750sp}%
\begin{picture}(1974,2949)(5089,-4873)
\thinlines
{\color[rgb]{0,0,0}\put(6301,-2461){\line( 1, 0){750}}
}%
{\color[rgb]{0,0,0}\put(6301,-2761){\line( 1, 0){750}}
}%
{\color[rgb]{0,0,0}\put(6301,-3061){\line( 1, 0){750}}
}%
{\color[rgb]{0,0,0}\put(6301,-1936){\line( 1, 0){750}}
\put(7051,-1936){\line( 0,-1){1350}}
\put(7051,-3286){\line(-1, 0){750}}
\put(6301,-3286){\line( 0,-1){1575}}
\put(6301,-4861){\line(-1, 0){1200}}
\put(5101,-4861){\line( 0, 1){1800}}
\put(5101,-3061){\line( 1, 0){1200}}
\put(6301,-3061){\line( 0, 1){1125}}
}%
{\color[rgb]{0,0,0}\put(5401,-4786){\framebox(750,1350){}}
}%
{\color[rgb]{0,0,0}\put(6301,-2161){\line( 1, 0){750}}
}%
\put(6751,-2386){\makebox(0,0)[b]{\smash{\fontsize{9}{10.8}
\usefont{T1}{cmr}{m}{n}{\color[rgb]{0,0,0}$A$}%
}}}
\end{picture}%
\end{tabular}
}{
                                       +------+
                                       |------|
                                       |   A  |
                                       |------|
                                       |      |
+-------------+          +-------------+------|
|             |          |             +------+
+-------------+          |             |
|   +------+  |          |   +------+  |
+---|------|--+          |   |      |  |
|   |   A  |  |    =>    |   |      |  |
+---|------|--+          |   |      |  |
|   |      |  |          |   |      |  |
+---|------|--+          |   |      |  |
|   |      |  |          |   |      |  |
+---|------|--+          |   |      |  |
|   +------+  |          |   +------+  |
+-------------+          +-------------+
  current.fig              radical.fig
}
\label{figure-radical-revision}
\hcaption{Radical revision}
\end{hfigure}

\item[Severe revision]

is contingent: it trusts the new belief only in the currently most believed
situations.

It is epiphanic: it erases the beliefs made weaker than the new one.

\begin{hfigure}
\long\def\ttytex#1#2{#1}
\ttytex{
\begin{tabular}{ccc}
\setlength{\unitlength}{3750sp}%
\begin{picture}(1224,1824)(5089,-4873)
\thinlines
{\color[rgb]{0,0,0}\put(5101,-3361){\line( 1, 0){1200}}
}%
{\color[rgb]{0,0,0}\put(5101,-4861){\framebox(1200,1800){}}
}%
{\color[rgb]{0,0,0}\put(5101,-3961){\line( 1, 0){1200}}
}%
{\color[rgb]{0,0,0}\put(5101,-4261){\line( 1, 0){1200}}
}%
{\color[rgb]{0,0,0}\put(5101,-4561){\line( 1, 0){1200}}
}%
{\color[rgb]{0,0,0}\put(5401,-4486){\framebox(750,1050){}}
}%
{\color[rgb]{0,0,0}\put(5101,-3661){\line( 1, 0){1200}}
}%
\put(5851,-3886){\makebox(0,0)[b]{\smash{\fontsize{9}{10.8}
\usefont{T1}{cmr}{m}{n}{\color[rgb]{0,0,0}$A$}%
}}}
\end{picture}%
&
\setlength{\unitlength}{3750sp}%
\begin{picture}(1104,1104)(5659,-4303)
\thinlines
{\color[rgb]{0,0,0}\multiput(6391,-3391)(-9.47368,4.73684){20}{\makebox(2.1167,14.8167){\tiny.}}
\put(6211,-3301){\line( 0,-1){ 45}}
\put(6211,-3346){\line(-1, 0){180}}
\put(6031,-3346){\line( 0,-1){ 90}}
\put(6031,-3436){\line( 1, 0){180}}
\put(6211,-3436){\line( 0,-1){ 45}}
\multiput(6211,-3481)(9.47368,4.73684){20}{\makebox(2.1167,14.8167){\tiny.}}
}%
\end{picture}%
&
\setlength{\unitlength}{3750sp}%
\begin{picture}(1224,2049)(5089,-4873)
\thinlines
{\color[rgb]{0,0,0}\put(5101,-3961){\line( 1, 0){1200}}
}%
{\color[rgb]{0,0,0}\put(5101,-4261){\line( 1, 0){1200}}
}%
{\color[rgb]{0,0,0}\put(5101,-4561){\line( 1, 0){1200}}
}%
{\color[rgb]{0,0,0}\put(5101,-4861){\framebox(1200,1200){}}
}%
{\color[rgb]{0,0,0}\put(5101,-3661){\line( 0, 1){600}}
\put(5101,-3061){\line( 1, 0){1200}}
\put(6301,-3061){\line( 0,-1){600}}
\put(6301,-3661){\line(-1, 0){150}}
\put(6151,-3661){\line( 0, 1){225}}
\put(6151,-3436){\line(-1, 0){750}}
\put(5401,-3436){\line( 0,-1){225}}
\put(5401,-3661){\line(-1, 0){300}}
}%
{\color[rgb]{0,0,0}\put(5401,-3061){\framebox(750,225){}}
}%
{\color[rgb]{0,0,0}\put(5401,-4486){\framebox(750,825){}}
}%
{\color[rgb]{0,0,0}\multiput(5401,-3511)(8.33333,8.33333){10}{\makebox(2.2222,15.5556){\tiny.}}
}%
{\color[rgb]{0,0,0}\multiput(5401,-3586)(7.89474,7.89474){20}{\makebox(2.2222,15.5556){\tiny.}}
}%
{\color[rgb]{0,0,0}\multiput(5401,-3661)(8.03571,8.03571){29}{\makebox(2.2222,15.5556){\tiny.}}
}%
{\color[rgb]{0,0,0}\multiput(6076,-3661)(8.33333,8.33333){10}{\makebox(2.2222,15.5556){\tiny.}}
}%
{\color[rgb]{0,0,0}\multiput(6001,-3661)(7.89474,7.89474){20}{\makebox(2.2222,15.5556){\tiny.}}
}%
{\color[rgb]{0,0,0}\multiput(5476,-3661)(8.03571,8.03571){29}{\makebox(2.2222,15.5556){\tiny.}}
}%
{\color[rgb]{0,0,0}\multiput(5551,-3661)(8.03571,8.03571){29}{\makebox(2.2222,15.5556){\tiny.}}
}%
{\color[rgb]{0,0,0}\multiput(5626,-3661)(8.03571,8.03571){29}{\makebox(2.2222,15.5556){\tiny.}}
}%
{\color[rgb]{0,0,0}\multiput(5701,-3661)(8.03571,8.03571){29}{\makebox(2.2222,15.5556){\tiny.}}
}%
{\color[rgb]{0,0,0}\multiput(5776,-3661)(8.03571,8.03571){29}{\makebox(2.2222,15.5556){\tiny.}}
}%
{\color[rgb]{0,0,0}\multiput(5851,-3661)(8.03571,8.03571){29}{\makebox(2.2222,15.5556){\tiny.}}
}%
{\color[rgb]{0,0,0}\multiput(5926,-3661)(8.03571,8.03571){29}{\makebox(2.2222,15.5556){\tiny.}}
}%
\end{picture}%
\end{tabular}
}{
                             +------+ min(A)
                             +------+
+-------------+          +---+------+--+
|             |          |             |
+-------------+          |             |
|   +------+  |          |   +------+  |
+---|------|--+          +---+      +--+
|   |   A  |  |    =>
+---|------|--+          +---+------+--+
|   |      |  |          |   |      |  |
+---|------|--+          +---|------|--+
|   +------+  |          |   |      |  |
+-------------+          +---|------|--+
|             |          |   +------+  |
+-------------+          +-------------+
|             |          |             |
+-------------+          +-------------+
|             |          |             |
+-------------+          +-------------+
                         |             |
                         +-------------+
  current.fig              severe.fig
}
\label{figure-severe-revision}
\hcaption{Severe revision}
\end{hfigure}

\item[Plain severe revision]

is slightly more epiphanic than severe revision. It does not only erase
beliefs made weaker than the new, but also the slightly stronger.

\begin{hfigure}
\long\def\ttytex#1#2{#1}
\ttytex{
\begin{tabular}{ccc}
\setlength{\unitlength}{3750sp}%
\begin{picture}(1224,1824)(5089,-4873)
\thinlines
{\color[rgb]{0,0,0}\put(5101,-3361){\line( 1, 0){1200}}
}%
{\color[rgb]{0,0,0}\put(5101,-4861){\framebox(1200,1800){}}
}%
{\color[rgb]{0,0,0}\put(5101,-3961){\line( 1, 0){1200}}
}%
{\color[rgb]{0,0,0}\put(5101,-4261){\line( 1, 0){1200}}
}%
{\color[rgb]{0,0,0}\put(5101,-4561){\line( 1, 0){1200}}
}%
{\color[rgb]{0,0,0}\put(5401,-4486){\framebox(750,1050){}}
}%
{\color[rgb]{0,0,0}\put(5101,-3661){\line( 1, 0){1200}}
}%
\put(5851,-3886){\makebox(0,0)[b]{\smash{\fontsize{9}{10.8}
\usefont{T1}{cmr}{m}{n}{\color[rgb]{0,0,0}$A$}%
}}}
\end{picture}%
&
\setlength{\unitlength}{3750sp}%
\begin{picture}(1104,1104)(5659,-4303)
\thinlines
{\color[rgb]{0,0,0}\multiput(6391,-3391)(-9.47368,4.73684){20}{\makebox(2.1167,14.8167){\tiny.}}
\put(6211,-3301){\line( 0,-1){ 45}}
\put(6211,-3346){\line(-1, 0){180}}
\put(6031,-3346){\line( 0,-1){ 90}}
\put(6031,-3436){\line( 1, 0){180}}
\put(6211,-3436){\line( 0,-1){ 45}}
\multiput(6211,-3481)(9.47368,4.73684){20}{\makebox(2.1167,14.8167){\tiny.}}
}%
\end{picture}%
&
\setlength{\unitlength}{3750sp}%
\begin{picture}(1224,2049)(5089,-5473)
\thinlines
{\color[rgb]{0,0,0}\put(5101,-4561){\line( 1, 0){1200}}
}%
{\color[rgb]{0,0,0}\put(5101,-4861){\line( 1, 0){1200}}
}%
{\color[rgb]{0,0,0}\put(5101,-5161){\line( 1, 0){1200}}
}%
{\color[rgb]{0,0,0}\put(5101,-5461){\framebox(1200,1800){}}
}%
{\color[rgb]{0,0,0}\put(5401,-3661){\framebox(750,225){}}
}%
{\color[rgb]{0,0,0}\put(5401,-4261){\framebox(750,225){}}
}%
{\color[rgb]{0,0,0}\multiput(5401,-4186)(7.89474,7.89474){20}{\makebox(2.2222,15.5556){\tiny.}}
}%
{\color[rgb]{0,0,0}\multiput(5401,-4261)(8.03571,8.03571){29}{\makebox(2.2222,15.5556){\tiny.}}
}%
{\color[rgb]{0,0,0}\multiput(5476,-4261)(8.03571,8.03571){29}{\makebox(2.2222,15.5556){\tiny.}}
}%
{\color[rgb]{0,0,0}\multiput(5551,-4261)(8.03571,8.03571){29}{\makebox(2.2222,15.5556){\tiny.}}
}%
{\color[rgb]{0,0,0}\multiput(5626,-4261)(8.03571,8.03571){29}{\makebox(2.2222,15.5556){\tiny.}}
}%
{\color[rgb]{0,0,0}\multiput(5701,-4261)(8.03571,8.03571){29}{\makebox(2.2222,15.5556){\tiny.}}
}%
{\color[rgb]{0,0,0}\multiput(5776,-4261)(8.03571,8.03571){29}{\makebox(2.2222,15.5556){\tiny.}}
}%
{\color[rgb]{0,0,0}\multiput(5851,-4261)(8.03571,8.03571){29}{\makebox(2.2222,15.5556){\tiny.}}
}%
{\color[rgb]{0,0,0}\multiput(5926,-4261)(8.03571,8.03571){29}{\makebox(2.2222,15.5556){\tiny.}}
}%
{\color[rgb]{0,0,0}\multiput(5401,-4111)(8.33333,8.33333){10}{\makebox(2.2222,15.5556){\tiny.}}
}%
{\color[rgb]{0,0,0}\multiput(6001,-4261)(7.89474,7.89474){20}{\makebox(2.2222,15.5556){\tiny.}}
}%
{\color[rgb]{0,0,0}\multiput(6076,-4261)(8.33333,8.33333){10}{\makebox(2.2222,15.5556){\tiny.}}
}%
{\color[rgb]{0,0,0}\put(5401,-5086){\framebox(750,525){}}
}%
\put(5851,-4786){\makebox(0,0)[b]{\smash{\fontsize{9}{10.8}
\usefont{T1}{cmr}{m}{n}{\color[rgb]{0,0,0}$A$}%
}}}
\put(5776,-3586){\makebox(0,0)[b]{\smash{\fontsize{9}{10.8}
\usefont{T1}{cmr}{m}{n}{\color[rgb]{0,0,0}$A$}%
}}}
\end{picture}%
\end{tabular}
}{
                             +------+ min(A)
                             +------+
+-------------+          +-------------+
|             |          |             |
+-------------+          |             |
|   +------+  |          |   ////////  |
+---|------|--+          |   ////////  |
|   |   A  |  |          |             |
+---|------|--+          +---+------+--+
|   |      |  |    =>    |   |      |  |
+---|------|--+          +---|------|--+
|   +------+  |          |   +------+  |
+-------------+          +-------------+
|             |          |             |
+-------------+          +-------------+
|             |          |             |
+-------------+          +-------------+
|             |          |             |
+-------------+          +-------------+
  current.fig            plainsevere.fig
}
\label{figure-plainsevere-revision}
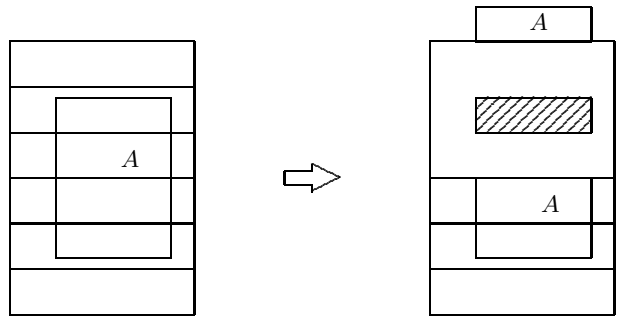
\hcaption{Plain severe revision}
\end{hfigure}

\item[Deep severe revision]

is the non-contingent version of severe revision: it trusts the new belief in
all situations, not just the most believed ones.

It is epiphanic: it erases the weakened beliefs.

\begin{hfigure}
\long\def\ttytex#1#2{#1}
\ttytex{
\begin{tabular}{ccc}
\setlength{\unitlength}{3750sp}%
\begin{picture}(1224,1824)(5089,-4873)
\thinlines
{\color[rgb]{0,0,0}\put(5101,-3361){\line( 1, 0){1200}}
}%
{\color[rgb]{0,0,0}\put(5101,-4861){\framebox(1200,1800){}}
}%
{\color[rgb]{0,0,0}\put(5101,-3961){\line( 1, 0){1200}}
}%
{\color[rgb]{0,0,0}\put(5101,-4261){\line( 1, 0){1200}}
}%
{\color[rgb]{0,0,0}\put(5101,-4561){\line( 1, 0){1200}}
}%
{\color[rgb]{0,0,0}\put(5401,-4486){\framebox(750,1050){}}
}%
{\color[rgb]{0,0,0}\put(5101,-3661){\line( 1, 0){1200}}
}%
\put(5851,-3886){\makebox(0,0)[b]{\smash{\fontsize{9}{10.8}
\usefont{T1}{cmr}{m}{n}{\color[rgb]{0,0,0}$A$}%
}}}
\end{picture}%
&
\setlength{\unitlength}{3750sp}%
\begin{picture}(1104,1104)(5659,-4303)
\thinlines
{\color[rgb]{0,0,0}\multiput(6391,-3391)(-9.47368,4.73684){20}{\makebox(2.1167,14.8167){\tiny.}}
\put(6211,-3301){\line( 0,-1){ 45}}
\put(6211,-3346){\line(-1, 0){180}}
\put(6031,-3346){\line( 0,-1){ 90}}
\put(6031,-3436){\line( 1, 0){180}}
\put(6211,-3436){\line( 0,-1){ 45}}
\multiput(6211,-3481)(9.47368,4.73684){20}{\makebox(2.1167,14.8167){\tiny.}}
}%
\end{picture}%
&
\setlength{\unitlength}{3750sp}%
\begin{picture}(1224,2949)(5089,-4873)
\thinlines
{\color[rgb]{0,0,0}\put(5101,-4561){\line( 1, 0){1200}}
}%
{\color[rgb]{0,0,0}\put(5401,-2161){\line( 1, 0){750}}
}%
{\color[rgb]{0,0,0}\put(5401,-2461){\line( 1, 0){750}}
}%
{\color[rgb]{0,0,0}\put(5401,-2761){\line( 1, 0){750}}
}%
{\color[rgb]{0,0,0}\put(5401,-2986){\framebox(750,1050){}}
}%
{\color[rgb]{0,0,0}\put(5401,-4486){\framebox(750,1050){}}
}%
{\color[rgb]{0,0,0}\put(5101,-4861){\framebox(1200,1800){}}
}%
\put(5851,-2386){\makebox(0,0)[b]{\smash{\fontsize{9}{10.8}
\usefont{T1}{cmr}{m}{n}{\color[rgb]{0,0,0}$A$}%
}}}
\end{picture}%
\end{tabular}
}{
                             +------+
                             |------|
                             |   A  |
                             |------|
                             |      |
                             |------|
                             +------+
+-------------+          +-------------+
|             |          |             |
+-------------+          |             |
|   +------+  |          |   +------+  |
+---|------|--+          |   |      |  |
|   |   A  |  |          |   |      |  |
+---|------|--+          |   |      |  |
|   |      |  |          |   |      |  |
+---|------|--+          |   |      |  |
|   +------+  |    =>    |   +------+  |
+-------------+          +-------------+
|             |          |             |
+-------------+          +-------------+
|             |          |             |
+-------------+          +-------------+
|             |          |             |
+-------------+          +-------------+
  current.fig             deepsevere.fig
}
\label{figure-deepsevere-revision}
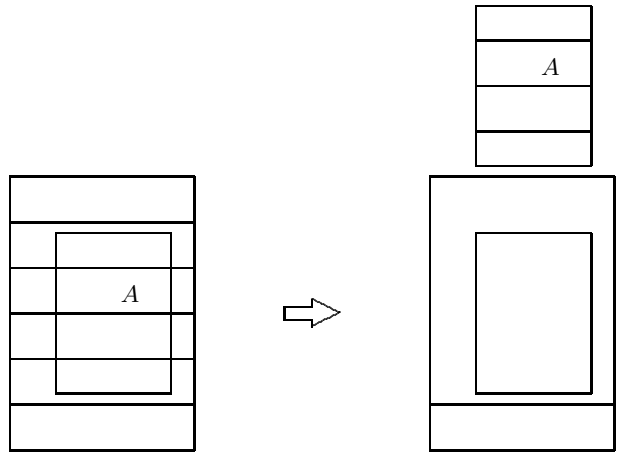
\hcaption{Deep severe revision}
\end{hfigure}

\item[Moderate severe revision]

is also not contingent. It trusts the new beliefs in all scenarios, not only
the currently most believed ones.

It is epiphanic like severe revision rather like deep severe revision. It
erases beliefs made weaker than the ones made the strongest.

\begin{hfigure}
\long\def\ttytex#1#2{#1}
\ttytex{
\begin{tabular}{ccc}
\setlength{\unitlength}{3750sp}%
\begin{picture}(1224,1824)(5089,-4873)
\thinlines
{\color[rgb]{0,0,0}\put(5101,-3361){\line( 1, 0){1200}}
}%
{\color[rgb]{0,0,0}\put(5101,-4861){\framebox(1200,1800){}}
}%
{\color[rgb]{0,0,0}\put(5101,-3961){\line( 1, 0){1200}}
}%
{\color[rgb]{0,0,0}\put(5101,-4261){\line( 1, 0){1200}}
}%
{\color[rgb]{0,0,0}\put(5101,-4561){\line( 1, 0){1200}}
}%
{\color[rgb]{0,0,0}\put(5401,-4486){\framebox(750,1050){}}
}%
{\color[rgb]{0,0,0}\put(5101,-3661){\line( 1, 0){1200}}
}%
\put(5851,-3886){\makebox(0,0)[b]{\smash{\fontsize{9}{10.8}
\usefont{T1}{cmr}{m}{n}{\color[rgb]{0,0,0}$A$}%
}}}
\end{picture}%
&
\setlength{\unitlength}{3750sp}%
\begin{picture}(1104,1104)(5659,-4303)
\thinlines
{\color[rgb]{0,0,0}\multiput(6391,-3391)(-9.47368,4.73684){20}{\makebox(2.1167,14.8167){\tiny.}}
\put(6211,-3301){\line( 0,-1){ 45}}
\put(6211,-3346){\line(-1, 0){180}}
\put(6031,-3346){\line( 0,-1){ 90}}
\put(6031,-3436){\line( 1, 0){180}}
\put(6211,-3436){\line( 0,-1){ 45}}
\multiput(6211,-3481)(9.47368,4.73684){20}{\makebox(2.1167,14.8167){\tiny.}}
}%
\end{picture}%
&
\setlength{\unitlength}{3750sp}%
\begin{picture}(1224,2949)(5089,-4873)
\thinlines
{\color[rgb]{0,0,0}\put(5101,-4861){\framebox(1200,1800){}}
}%
{\color[rgb]{0,0,0}\put(5101,-4561){\line( 1, 0){1200}}
}%
{\color[rgb]{0,0,0}\put(5401,-2161){\line( 1, 0){750}}
}%
{\color[rgb]{0,0,0}\put(5401,-2461){\line( 1, 0){750}}
}%
{\color[rgb]{0,0,0}\put(5401,-2761){\line( 1, 0){750}}
}%
{\color[rgb]{0,0,0}\put(5401,-2986){\framebox(750,1050){}}
}%
{\color[rgb]{0,0,0}\put(5401,-4486){\framebox(750,1050){}}
}%
{\color[rgb]{0,0,0}\put(5101,-3661){\line( 1, 0){300}}
}%
{\color[rgb]{0,0,0}\put(6151,-3661){\line( 1, 0){150}}
}%
{\color[rgb]{0,0,0}\put(5101,-3961){\line( 1, 0){300}}
}%
{\color[rgb]{0,0,0}\put(5101,-4261){\line( 1, 0){300}}
}%
{\color[rgb]{0,0,0}\put(6151,-3961){\line( 1, 0){150}}
}%
{\color[rgb]{0,0,0}\put(6151,-4261){\line( 1, 0){150}}
}%
\put(5851,-2386){\makebox(0,0)[b]{\smash{\fontsize{9}{10.8}
\usefont{T1}{cmr}{m}{n}{\color[rgb]{0,0,0}$A$}%
}}}
\end{picture}%
\end{tabular}
}{
                             +------+
                             |------|
                             |   A  |
                             |------|
                             |      |
                             |------|
                             +------+
+-------------+          +-------------+
|             |          |             |
+-------------+          |             |
|   +------+  |          |   +------+  |
+---|------|--+          +---+      +--+
|   |   A  |  |          |   |      |  |
+---|------|--+          +---|      |--+
|   |      |  |    =>    |   |      |  |
+---|------|--+          +---|      |--+
|   +------+  |          |   +------+  |
+-------------+          +-------------+
|             |          |             |
+-------------+          +-------------+
|             |          |             |
+-------------+          +-------------+
|             |          |             |
+-------------+          +-------------+
  current.fig           moderatesevere.fig
}
\label{figure-moderatesevere-revision}
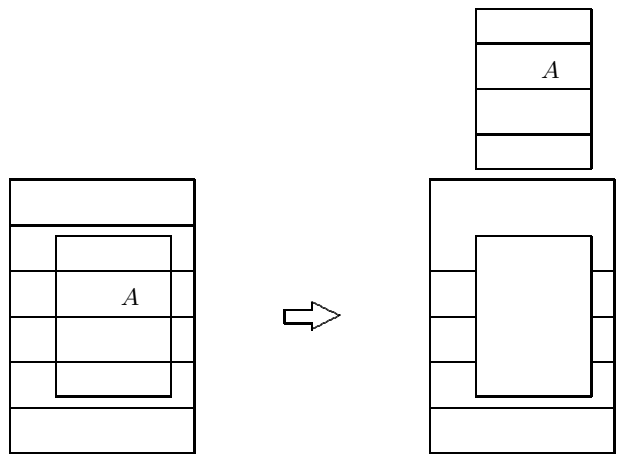
\hcaption{Moderate severe revision}
\end{hfigure}

\end{description}

\section{Abilities}
\label{section-list}

The root of mathematics is that numbers abstract from specific objects: two
sheeps plus three sheeps are five sheeps, two stars plus three stars are five
stars, two days plus three days are five days, and so on.

Mathematical calculations work regardless of what numbers stand for. Formal
logic deductions work regardless of what Boolean variables stand for. No matter
whether $a$ is ``the sky is cloudy'' and $b$ is ``the stars are visible'', or
$a$ is ``today is holiday'' and $b$ is ``Lorenzo goes to work'', or $a$ is
``the car starts'' and $b$ is ``battery is dead'', regardless of the meaning of
the variables $a$ and $b$, in all three cases $a$ and $a \rightarrow \neg b$
entail $\neg b$. What $a$ and $b$ stand for is irrelevant. Formal logic is only
for deductions that hold for all possible meaning of the variables.

Applied to belief revision: $I < J$ may mean that a round earth is more
believed that a flat one, but may also mean that cold fusion is more believed
feasible than not. In some contexts, every order of beliefs is possible.
Someone searching for minerals in an area may research for information whether
gold, quartz, cobalt, copper or iron are common in the country. Each mineral
can or cannot. Every combination may turn out to be possible. Starting from
complete ignorance, every book read, every article studied, every relation
considered adds information. Abstracting, as in mathematics: the flat doxastic
state turns into an arbitrary doxastic state by a sequence of belief revisions.
A form of belief revision that does not do it is insufficient; another revision
is needed, in addition or in its place, to reach an arbitrary doxastic state
from complete ignorance, the flat doxastic state. The learnable ability is
required.

The flat doxastic state may be unrealistic instead. A magazine article may
provide the position of a new politician on economy, foreign affairs and
welfare. While every combination of these is in principle possible, pacifism
is rarely associated with conservatism on welfare. Even before reading, their
combination is less believed than the combination of pacifism and support for
welfare. The initial doxastic state is not flat. The learning ability is
useless.

Useless in a context does not mean useless in all. The need for an ability in a
certain context prove it useful even if it is unneeded or even harmful in
others. Contrary to the postulates belief revision historically started
with~\cite{alch-gard-maki-85,kats-mend-91,darw-pear-97}, abilities are not
rules. The same ability may be needed, unneeded, useless or harmful.
Theorem~\ref{theorem-lexicographic-learnable} does not declare lexicographic
revision universally useful, but only as long as learning from scratch is
required. Similarly, Theorem~\ref{theorem-lexicographic-equating} does not
always exclude the lexicographic revision, only if all beliefs may be lost at
some point.

Each ability establishes that certain doxastic states are changed in certain
others by sequences of revisions: from the flat state to an arbitrary one is
the learnable ability, from an arbitrary state to the flat one is the amnesic
ability.

Lexicographic revision can make a situation less believed than a previously
less believed one; it possesses the correcting ability; it cannot make two
situations equally believed if they are not; it does not possess the equating
ability. The lack of an ability does not prevent a revision from being used. If
equating two models is required, it can be done by another revision, or a
combination of the two. Lexicographic revision cannot equalize models but full
meet revision can; lexicographic revision can invert the order between all
models while full meet revision cannot. They fill the lack of each other.

Reaching certain doxastic states from other doxastic states by a sequence of
revisions defines an ability. The plastic ability is reaching every state from
every state. The damascan ability is reaching the exact opposite of every
state. The amnesic ability is reaching the flat doxastic state from every other
state. Each of these abilities is or is not necessary depending on the specific
context of application.

\begin{description}

\item[fully plastic:] turning every doxastic state in every other

depending on the context, all doxastic states may make sense or not; Alessio
may be older than Ascanio or not, Belarus larger than Belgium or not; yet,
these conditions $a$ and $b$ do not compare arbitrarily: if $\{a,b\}$ is less
than $\{\neg a,b\}$, then Alessio is believed older than Ascanio; if $\{\neg a,
\neg b\}$ is less than $\{a, \neg b\}$ then Alessio is believed younger
instead; believed older or believed younger depends on the size of two
countries: older if $b$, younger if $\neg b$; older if larger, younger if
smaller; a nonsensical doxastic state; reaching it is not an ability; it is a
drawback;

other contexts require every possible doxastic state; nausea is believed a more
common symptom of heart failure in female than in man; the doxastic state
forbidden in the older--man/larger--country context is not just allowed, it is
an established medical fact; researchers may start with common beliefs about a
new disease that turn out to be false; an arbitrary doxastic state turns into
another arbitrary one;

\begin{definition}
\label{definition-fully-plastic}

A revision is fully plastic if a sequence of revisions turns an arbitrary
doxastic state into an arbitrary doxastic state.

\[
\forall C,G \exists R_1,\ldots,R_m . C \rev(R_1),\ldots,\rev(R_m) = G
\]

\end{definition}

\item[plastic:] turning every doxastic state in every non-flat one

revisions add new beliefs; a new belief is believed, a truism; revisions may
contradict old beliefs, but also carry new ones; the state of complete
ignorance, the flat doxastic state, is never reached by adding beliefs; other
forms of belief change such as contractions or withdrawals remove beliefs,
possibly all of them; revisions cannot because they always add beliefs;

yet, some revisions reach every non-flat doxastic state; this is the best a
revision can achieve, as proved by Lemma~\ref{theorem-amnesic};
they are plastic: very radical revision, severe revision, moderate severe
revision and deep severe revisions;

\begin{definition}
\label{definition-plastic}

A revision is plastic if a sequence of revisions turns an arbitrary doxastic
state into an arbitrary non-flat doxastic state.

\[
\forall C,G \not= \flatorder \exists R_1,\ldots,R_m .
C \rev(R_1),\ldots,\rev(R_m) = G
\]

\end{definition}

\item[learnable:] turning the flat doxastic state in every other one

learning from scratch, in short; acquiring information about a completely
unknown context; completely unknown is the flat doxastic state; completely
unknown also means that no doxastic state can be excluded, since no information
excludes it; every doxastic state is possible;

\begin{definition}
\label{definition-learnable}

A revision is learnable if a sequence of revisions turns the flat doxastic
state into an arbitrary doxastic state.

\[
\forall G \exists R_1,\ldots,R_m . \flatorder \rev(R_1),\ldots,\rev(R_m) = G
\]

\end{definition}

\item[correcting:] inverting the order between two models

the basic property of every revision is that the new information is believed;
it is made true in the most believed situations; if only one of them $I$
exists, it is less than all others: $I < J$, regardless of how $I$ and $J$
compared;

\begin{definition}
\label{definition-correcting}

A revision is correcting if a sequence of revisions inverts the order between
two arbitrary models.

\[
\forall C \forall I,J  \exists R_1,\ldots,R_m .
	I <_{C\rev(R_1),\ldots,\rev(R_m)} J
\]

\end{definition}

\item[damascan:] inverting the order between all models

changing mind completely may or may not be sensible depending on the context;
the more believed becomes the less, the less becomes the more; while reversing
the relative strength of belief in two situations always makes sense, a total
reverse may be not; however, in some specific context it expresses open
mindness: facts are accepted even if they contradict all current beliefs;

\begin{definition}
\label{definition-damascan}

A revision is damascan if a sequence of revisions inverts every doxastic
state.

\[
\forall C \exists R_1,\ldots,R_m .
	C \rev(R_1),\ldots,\rev(R_m) =
		[C(\last),\ldots,C(0)]
\]

\end{definition}

\item[equating:] believing two situations the same

this is similar to the correcting ability: new information may lead to believe
a situation as likely as another; it is however a separate ability because it
is not obvious for a revision; equating two models is losing the belief in one;
revision is acquiring new beliefs, not losing;

\begin{definition}
\label{definition-equating}

A revision is equating if a sequence of revisions makes two arbitrary models
equivalent.

\[
\forall C,I,J \exists R_1,\ldots,R_m .
	I \equiv_{C \rev(R_1),\ldots,\rev(R_m)} J
\]

\end{definition}

\item[amnesic:] reaching the flat doxastic state

the flat doxastic state is complete ignorance: all conditions are believed
equally possible; this is what happens when learning about new topics; yet, it
is the initial condition; arriving to a state of complete ignorance from a
state of existing beliefs seems unlikely; it is however what happens when
realizing of knowing nothing; new information disprove everything believed;

\begin{definition}
\label{definition-amnesic}

A revision is amnesic if a sequence of revisions turns every doxastic state
into the flat doxastic state.

\[
\forall C \exists R_1,\ldots,R_m . C \rev(R_1),\ldots,\rev(R_m) = \flatorder
\]

\end{definition}

\item[believer:] believing certain situations the most

revising is believing new information; yet, it is not believing equally in all
situations supported by the new information; even if Alessio turns out younger
than Ascanio; Belarus remains believed larger than Belgium; the situation where
Alessio is the youngest and Belarus is the smallest instead is still unbelieved
in spite of the correct order of age;

the believer ability is not ``I believe'', it is ``I only believe'': only
believe Saturn further to the Sun than Jupiter, not that it is further from
Earth; only believe Alessio younger than Ascanio, not smarter;

\begin{definition}
\label{definition-believer}

A revision is believer if a sequence of revisions produces a first class that
is an arbitrary set of models.

\[
\forall F \exists R_1,\ldots,R_m .
	C \rev(R_1),\ldots,\rev(R_m)(0) = F
\]

\end{definition}

\item[dogmatic:] believing certain situations and disbelieving all others

an astronomer believes the Earth round, the other shapes are impossible,
equally impossible; a Taliban believes Allah the only God, all others are
equally unreal; a fan believes Bob Dylan the greatest singer in the world, all
others are all plainly tuneless; dogmatism is not just believing something, is
totally excluding everything else.

\begin{definition}
\label{definition-dogmatic}

A revision is dogmatic if a sequence of revisions produces an arbitrary
two-class doxastic state.

\[
\forall F \exists R_1,\ldots,R_m .
	C \rev(R_1),\ldots,\rev(R_m) = [F, \true \backslash F]
\]

\end{definition}

\end{description}

The correcting and equating abilities have a weak version, where only some
models are corrected or equated: $I<J$ can be inverted or $I \equiv J$
established for some models $I$ and $J$; ``some models'' instead of ``for all'.
They are preferred when disproving an ability instead of proving.

Believer and dogmatic coincide with amnesic if the target first class comprises
all models. Yet, believing equally in all possible situations is believing
nothing: everything is equally possible; nothing is more believed than anything
else. Believing in nothing and being dogmatic on nothing are contradictions.
They are technically relevant in the corner case of equating the only two
models of an alphabet of a single variable, the same as believing them both
equally;

Abilities are not unrelated. Fully plastic implies amnesic and plastic. Which
implies learnable. A depiction is in Figure~\ref{figure-relations}. Arrows
stand for implications.

\begin{hfigure}
\setlength{\unitlength}{3750sp}%
\begin{picture}(5942,3835)(3174,-6136)
\thinlines
{\color[rgb]{0,0,0}\put(6274,-2572){\vector( 4,-3){552}}
}%
{\color[rgb]{0,0,0}\put(5341,-2491){\vector(-2,-1){1140}}
}%
{\color[rgb]{0,0,0}\put(3901,-3286){\vector( 0,-1){2400}}
}%
{\color[rgb]{0,0,0}\put(6526,-3211){\vector(-3,-2){1125}}
}%
{\color[rgb]{0,0,0}\put(6901,-3286){\vector( 0,-1){675}}
}%
{\color[rgb]{0,0,0}\put(7090,-3259){\vector( 4,-3){936}}
}%
{\color[rgb]{0,0,0}\put(8101,-4186){\vector( 0,-1){675}}
}%
{\color[rgb]{0,0,0}\put(5401,-4186){\vector( 0,-1){675}}
}%
{\color[rgb]{0,0,0}\put(4351,-3136){\line( 5, 1){1125}}
}%
{\color[rgb]{0,0,0}\put(6526,-3961){\line(-1, 1){1050}}
\multiput(5476,-2911)(5.00000,10.00000){31}{\makebox(2.2222,15.5556){\tiny.}}
\put(5626,-2611){\vector( 1, 2){0}}
}%
{\color[rgb]{0,0,0}\put(5776,-4936){\vector( 1, 0){1875}}
}%
{\color[rgb]{0,0,0}\put(5026,-5086){\vector(-3,-2){900}}
}%
\put(5851,-2461){\makebox(0,0)[b]{\smash{\fontsize{9}{10.8}
\usefont{T1}{cmr}{m}{n}{\color[rgb]{0,0,0}$fully\ plastic$}%
}}}
\put(3901,-3211){\makebox(0,0)[b]{\smash{\fontsize{9}{10.8}
\usefont{T1}{cmr}{m}{n}{\color[rgb]{0,0,0}$amnesic$}%
}}}
\put(6901,-4111){\makebox(0,0)[b]{\smash{\fontsize{9}{10.8}
\usefont{T1}{cmr}{m}{n}{\color[rgb]{0,0,0}$learnable$}%
}}}
\put(6826,-3211){\makebox(0,0)[b]{\smash{\fontsize{9}{10.8}
\usefont{T1}{cmr}{m}{n}{\color[rgb]{0,0,0}$plastic$}%
}}}
\put(5401,-4111){\makebox(0,0)[b]{\smash{\fontsize{9}{10.8}
\usefont{T1}{cmr}{m}{n}{\color[rgb]{0,0,0}$dogmatic$}%
}}}
\put(8101,-4111){\makebox(0,0)[b]{\smash{\fontsize{9}{10.8}
\usefont{T1}{cmr}{m}{n}{\color[rgb]{0,0,0}$damascan$}%
}}}
\put(8101,-5011){\makebox(0,0)[b]{\smash{\fontsize{9}{10.8}
\usefont{T1}{cmr}{m}{n}{\color[rgb]{0,0,0}$correcting$}%
}}}
\put(5401,-5011){\makebox(0,0)[b]{\smash{\fontsize{9}{10.8}
\usefont{T1}{cmr}{m}{n}{\color[rgb]{0,0,0}$believer$}%
}}}
\put(3901,-5836){\makebox(0,0)[b]{\smash{\fontsize{9}{10.8}
\usefont{T1}{cmr}{m}{n}{\color[rgb]{0,0,0}$equating$}%
}}}
\end{picture}%
\nop{
             fully plastic
           /     ^        \                              \space
         /       |          \                            \space
       /         |            \                          \space
      |          |             |
      V          |             V
   amnesic ------+          plastic
      |            \     /     |     \                   \space
      |              \ /       |       \                 \space
      |              / \       |         \               \space
      |             |    \     |          |
      |             V      \   V          V
      |         dogmatic   learnable   damascan
      |             |                     |
      |             |                     |
      |             |                     |
      |             V                     V
      |         believer ----------> correcting
      |        /
      |      /
      |    / 
      V   V
    equating
}
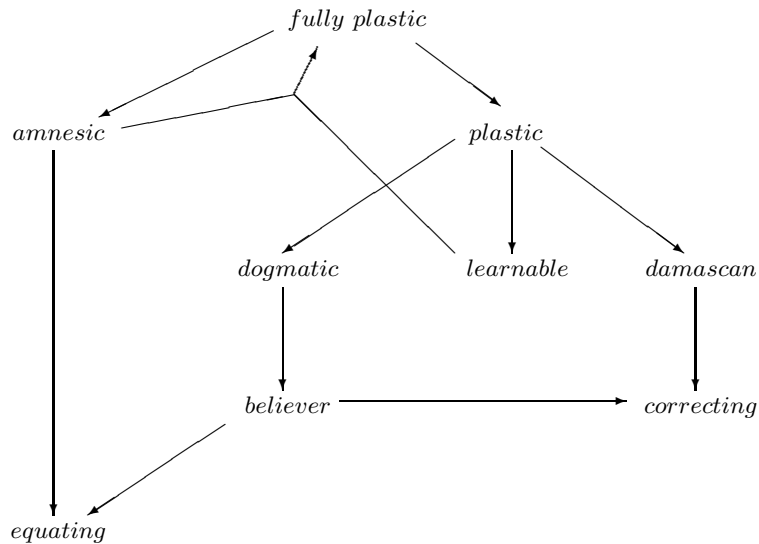
\hcaption{The relations between abilities}
\label{figure-relations}
\end{hfigure}

The only implication going up in the figure is that amnesic and learnable imply
fully plastic. The amnesic ability allows turning an arbitrary doxastic state
into the flat doxastic state, which can be turned into every other doxastic
state because of the learnable ability.

\section{Results}
\label{section-results}

\draft

\subsection{plastic.tex}

general results, not just on plasticity

\begin{description}

\item[theorem-amnesic]
No revision $\rev()$ satisfying
{} $(C \rev(A))(0) \models A$
and
{} $C \rev(\true) = C$
is amnesic.

\item[theorem-split]
If every class of $C \rev(A)$ is contained in a class of $C$,
the revision $\rev()$ is not equating.

\item[theorem-plastic-equating]
Every plastic change operator is also equating.

\item[theorem-believer-correcting]
Every believer revision is correcting.

\item[theorem-believer-equating]

Every believer revision is equating if the alphabet comprises at least two
variables.

\end{description}

\

\subsection{natural.tex}

\begin{description}

\item[lemma-natural-singleclass]
For every orders $C$ and $G$ such that every class of $G$ is contained in a
class of $C$, a sequence of single-class natural revisions turns $C$ into $G$.

\item[theorem-natural-learnable]
Natural revisions is learnable even when restricting to single-class revisions.

\item[theorem-natural-damascan]
Natural revisions is damascan even when restricting to single-class revisions.

\item[theorem-natural-equating]
Natural revision is not equating.

\item[theorem-natural-plastic]
Natural revision is not plastic.

\end{description}

\

\subsection{lexicographic.tex}

\begin{description}

\item[lemma-lexicographic-natural]
Lexicographic and natural revision coincide when the revision is contained in a
class of the order: $C \lex(A) = C \nat(A)$ if $A \subseteq C(i)$ for some $i$.

\item[theorem-lexicographic-learnable]
Lexicographic revisions is learnable.

\item[theorem-lexicographic-damascan]
Lexicographic revisions is damascan.

\item[theorem-lexicographic-equating]
Lexicographic revision is not equating.
Lexicographic revision is not plastic.

\end{description}

\subsection{restrained.tex}

\begin{description}

\item[lemma-radical-last]
Every non-flat order is flattened by radically revising it by a model $I$ of
its last class.

\item[theorem-radical-amnesic]
Radical revision is amnesic.

\item[lemma-radical-true]
Radically revising an order by $\true$ does not change it.

\item[theorem-radical-one-two]
$\flatorder \rad(A)$ comprises at most two classes.

\item[theorem-radical-two-two]
$[C(0),C(1)] \rad(A)$ comprises at most two classes.

\item[theorem-radical-learnable]
Radical revision is not learnable.

\item[theorem-radical-dogmatic]
Radical revision is dogmatic.

\item[theorem-radical-damascan]
Radical revision is not damascan.

\end{description}

\subsection{radical.tex}

\begin{description}

\item[lemma-radical-last]
Every non-flat order is flattened by radically revising it by a model $I$ of
its last class.

\item[theorem-radical-amnesic]
Radical revision is amnesic.

\item[lemma-radical-true]
Radically revising an order by $\true$ does not change it.

\item[theorem-radical-one-two]
$\flatorder \rad(A)$ comprises at most two classes.

\item[theorem-radical-two-two]
$[C(0),C(1)] \rad(A)$ comprises at most two classes.

\item[theorem-radical-learnable]
Radical revision is not learnable.

\item[theorem-radical-dogmatic]
Radical revision is dogmatic.

\item[theorem-radical-damascan]
Radical revision is not damascan.

\end{description}

\subsection{veryradical.tex}

\begin{description}

\item[lemma-plastic-veryradical]
Every non-flat arbitrary order $G$ has a sequence of very radical revisions
that turns every order $S$ into $G$.

\item[theorem-plastic-veryradical]
Very radical revision is plastic.

\item[theorem-veryradical-amnesic]
Very radical revision is not amnesic.

\item[corollary-veryradical-full]
Very radical revision is not fully plastic.

\end{description}

\subsection{fullmeet.tex}

\begin{description}

\item[theorem-fullmeet-amnesic]
Full meet revisions is amnesic if and only if the alphabet comprises at least
two symbols.

\item[theorem-fullmeet-learnable]
Full meet revision is not learnable.

\item[theorem-fullmeet-correcting]
Full meet revision is correcting.

\item[theorem-fullmeet-damascan]
Full meet revision is not damascan.

\item[theorem-fullmeet-dogmatic]
Full meet revision is dogmatic.

\end{description}

\subsection{severe.tex}

\begin{description}

\item[lemma-plastic-severe]
For every order $S$, a sequence of severe revisions each revising an order
containing it in a single class turns $S$ into an arbitrary non-flat order $G$.

\item[theorem-severe-plastic]
Severe revision is plastic.

\item[corollary-severe-amnesic]
Severe revision is not amnesic.

\item[corollary-severe-fully]
Severe revision is not fully plastic.

\end{description}

\subsection{moderatesevere.tex}

\begin{description}

\item[lemma-moderatesevere-severe]
Moderate severe revision and severe revision coincide when the revision is
contained in a class of the order: $C \sev(A) = C \msev(A)$ if $A \subseteq
C(i)$ for some $i$.

\item[theorem-moderatesevere-plastic]
Moderate severe revision is plastic.

\item[theorem-moderatesevere-amnesic]
Moderate severe revision is not amnesic.

\item[corollary-moderatesevere-fully]
Moderate severe revision is not fully plastic.

\end{description}

\subsection{deepsevere}

\begin{description}

\item[lemma-deepsevere-severe]
Deep severe revision and severe revision coincide when the revision is
contained in a class of the order: $C \sev(A) = C \dsev(A)$ if $A \subseteq
C(i)$ for some $i$.

\item[theorem-deepsevere-plastic]

\item[theorem-deepsevere-amnesic]
Deep severe revision is not amnesic.

\item[corollary-deepsevere-fully]
Deep severe revision is not fully plastic.

\end{description}

\subsection{plainsevere}

\begin{description}

\item[lemma-plainsevere-last]
Plainly severely revising an order $C$ by a subset $A \subseteq C(\last)$ of
its last class produces $\formulaorder A = [A, \true \backslash A]$.

\item[lemma-plainsevere-flatorder]
Plainly severely revising $\flatorder$ by $A$ produces
{} $\flatorder \psev(A) = [A, \true \backslash A]$.

\item[lemma-plainsevere-one-two]
The order $\flatorder \psev(A)$ comprises at most two classes.

\item[lemma-plainsevere-two-two]
If $C$ comprises two classes, so does $C \psev(A)$.

\item[theorem-plainsevere-learnable]
Plain severe revision is not learnable.

\item[theorem-plainsevere-amnesic]
Plain severe revision is not amnesic.

\item[theorem-plainsevere-dogmatic]
Plain severe revision is dogmatic.

\item[theorem-plainsevere-damascan]
Plain severe revision is not damascan.

\end{description}

\enddraft

Which revisions have which abilitities?

Natural revision is learnable but not equating and therefore not plastic.
Moderate severe revision is plastic but not amnesic and therefore not fully
plastic. Plain severe revision is not learnable. The abilities of all revisions
are established.

Revisions are assumed consistent: a revision $C \rev(A)$ is only applied when
$A$ is consistent. A single-class revision is completely contained in a class
of the doxastic state.

\begin{definition}
\label{definition-singleclass}

A revision $C \rev(A)$ is single-class if $A \subseteq C(i)$ for some $i$.

\end{definition}

The finite alphabet is assumed to comprise two or more variables. This is
relevant to the equating ability. It is reported explicitly in the lemmas and
theorems but not in the following summary.

\subsection{General results}

Theorem~\ref{theorem-amnesic} proves that no operator is amnesic if it has two
common properties of revisions: success and invariance to tautologies. Success
is fully believing the revision: $A$ obtains in all most believed scenarios $C
\rev(A)(0)$. Invariance to tautologies is not changing beliefs in response to
obvious statements: $C \rev(\true) = C$.

No revision is equating if it never merges classes, only splits them: every
class of $C \rev(A)$ is contained in some class of $C$. This is proved by
Theorem~\ref{theorem-split}.

The implications between abilities in Figure~\ref{figure-relations} are
trivial, except three: plastic revisions are also equating, shown in
Theorem~\ref{theorem-plastic-equating}, believer revisions are correcting,
proved by Theorem~\ref{theorem-believer-correcting}, believer revisions are
equating, as stated by Theorem~\ref{theorem-believer-equating}.

\subsection{The abilities of the revisions}

The considered revisions are of three kinds:

\begin{itemize}

\item the non-epiphanic natural, lexicographic and restrained revisions are
learnable, damascan and not equating;

\item the epiphanic very radical, severe, moderate severe and deep severe
revisions are plastic, equating and not amnesic;

\item the epiphanic plain severe, full meet and radical revisions are equating,
dogmatic, not learnable and not damascan; they differ on amnesic.

\end{itemize}

\begin{hfigure}
\setlength{\unitlength}{3750sp}%
\begin{picture}(5942,3835)(3174,-6136)
\thinlines
{\color[rgb]{0,0,0}\put(6274,-2572){\vector( 4,-3){552}}
}%
{\color[rgb]{0,0,0}\put(5341,-2491){\vector(-2,-1){1140}}
}%
{\color[rgb]{0,0,0}\put(3901,-3286){\vector( 0,-1){2400}}
}%
{\color[rgb]{0,0,0}\put(6526,-3211){\vector(-3,-2){1125}}
}%
{\color[rgb]{0,0,0}\put(6901,-3286){\vector( 0,-1){675}}
}%
{\color[rgb]{0,0,0}\put(7090,-3259){\vector( 4,-3){936}}
}%
{\color[rgb]{0,0,0}\put(8101,-4186){\vector( 0,-1){675}}
}%
{\color[rgb]{0,0,0}\put(5401,-4186){\vector( 0,-1){675}}
}%
{\color[rgb]{0,0,0}\put(4351,-3136){\line( 5, 1){1125}}
}%
{\color[rgb]{0,0,0}\put(6526,-3961){\line(-1, 1){1050}}
\multiput(5476,-2911)(5.00000,10.00000){31}{\makebox(2.2222,15.5556){\tiny.}}
\put(5626,-2611){\vector( 1, 2){0}}
}%
{\color[rgb]{0,0,0}\put(5776,-4936){\vector( 1, 0){1875}}
}%
{\color[rgb]{0,0,0}\put(5026,-5086){\vector(-3,-2){900}}
}%
\put(5851,-2461){\makebox(0,0)[b]{\smash{\fontsize{9}{10.8}
\usefont{T1}{cmr}{m}{n}{\color[rgb]{0,0,0}$fully\ plastic$}%
}}}
\put(3901,-3211){\makebox(0,0)[b]{\smash{\fontsize{9}{10.8}
\usefont{T1}{cmr}{m}{n}{\color[rgb]{0,0,0}$amnesic$}%
}}}
\put(6901,-4111){\makebox(0,0)[b]{\smash{\fontsize{9}{10.8}
\usefont{T1}{cmr}{m}{n}{\color[rgb]{0,0,0}$learnable$}%
}}}
\put(6826,-3211){\makebox(0,0)[b]{\smash{\fontsize{9}{10.8}
\usefont{T1}{cmr}{m}{n}{\color[rgb]{0,0,0}$plastic$}%
}}}
\put(5401,-4111){\makebox(0,0)[b]{\smash{\fontsize{9}{10.8}
\usefont{T1}{cmr}{m}{n}{\color[rgb]{0,0,0}$dogmatic$}%
}}}
\put(8101,-4111){\makebox(0,0)[b]{\smash{\fontsize{9}{10.8}
\usefont{T1}{cmr}{m}{n}{\color[rgb]{0,0,0}$damascan$}%
}}}
\put(8101,-5011){\makebox(0,0)[b]{\smash{\fontsize{9}{10.8}
\usefont{T1}{cmr}{m}{n}{\color[rgb]{0,0,0}$correcting$}%
}}}
\put(5401,-5011){\makebox(0,0)[b]{\smash{\fontsize{9}{10.8}
\usefont{T1}{cmr}{m}{n}{\color[rgb]{0,0,0}$believer$}%
}}}
\put(3901,-5836){\makebox(0,0)[b]{\smash{\fontsize{9}{10.8}
\usefont{T1}{cmr}{m}{n}{\color[rgb]{0,0,0}$equating$}%
}}}
{\color[rgb]{0,0,0}\multiput(8851,-3811)(10.49020,-6.29412){4}{\makebox(2.2222,15.5556){\tiny.}}
\multiput(8883,-3829)(9.61540,-6.41027){4}{\makebox(2.2222,15.5556){\tiny.}}
\multiput(8912,-3848)(8.72130,-7.26775){4}{\makebox(2.2222,15.5556){\tiny.}}
\multiput(8938,-3870)(11.75000,-11.75000){3}{\makebox(2.2222,15.5556){\tiny.}}
\multiput(8962,-3893)(10.45080,-12.54096){3}{\makebox(2.2222,15.5556){\tiny.}}
\multiput(8983,-3918)(9.23075,-13.84613){3}{\makebox(2.2222,15.5556){\tiny.}}
\multiput(9001,-3946)(5.76470,-9.60783){4}{\makebox(2.2222,15.5556){\tiny.}}
\multiput(9018,-3975)(5.06667,-10.13333){4}{\makebox(2.2222,15.5556){\tiny.}}
\multiput(9032,-4006)(4.39080,-10.97700){4}{\makebox(2.2222,15.5556){\tiny.}}
\multiput(9045,-4039)(3.80000,-11.40000){4}{\makebox(2.2222,15.5556){\tiny.}}
\multiput(9057,-4073)(3.07843,-12.31373){4}{\makebox(2.2222,15.5556){\tiny.}}
\multiput(9066,-4110)(3.07843,-12.31373){4}{\makebox(2.2222,15.5556){\tiny.}}
\multiput(9075,-4147)(2.22523,-13.35140){4}{\makebox(2.2222,15.5556){\tiny.}}
\multiput(9082,-4187)(2.21620,-13.29720){4}{\makebox(2.2222,15.5556){\tiny.}}
\multiput(9088,-4227)(2.26127,-13.56760){4}{\makebox(2.2222,15.5556){\tiny.}}
\multiput(9093,-4268)(2.36037,-14.16220){4}{\makebox(2.2222,15.5556){\tiny.}}
\put(9097,-4311){\line( 0,-1){ 43}}
\put(9100,-4354){\line( 0,-1){ 44}}
\put(9102,-4398){\line( 0,-1){ 44}}
\put(9104,-4442){\line( 0,-1){ 44}}
\put(9104,-4486){\line( 0,-1){ 44}}
\put(9104,-4530){\line( 0,-1){ 44}}
\put(9102,-4574){\line( 0,-1){ 44}}
\put(9100,-4618){\line( 0,-1){ 43}}
\multiput(9097,-4661)(-2.36037,-14.16220){4}{\makebox(2.2222,15.5556){\tiny.}}
\multiput(9093,-4704)(-2.26127,-13.56760){4}{\makebox(2.2222,15.5556){\tiny.}}
\multiput(9088,-4745)(-2.21620,-13.29720){4}{\makebox(2.2222,15.5556){\tiny.}}
\multiput(9082,-4785)(-2.22523,-13.35140){4}{\makebox(2.2222,15.5556){\tiny.}}
\multiput(9075,-4825)(-3.07843,-12.31373){4}{\makebox(2.2222,15.5556){\tiny.}}
\multiput(9066,-4862)(-3.07843,-12.31373){4}{\makebox(2.2222,15.5556){\tiny.}}
\multiput(9057,-4899)(-3.80000,-11.40000){4}{\makebox(2.2222,15.5556){\tiny.}}
\multiput(9045,-4933)(-4.39080,-10.97700){4}{\makebox(2.2222,15.5556){\tiny.}}
\multiput(9032,-4966)(-5.06667,-10.13333){4}{\makebox(2.2222,15.5556){\tiny.}}
\multiput(9018,-4997)(-5.76470,-9.60783){4}{\makebox(2.2222,15.5556){\tiny.}}
\multiput(9001,-5026)(-9.23075,-13.84613){3}{\makebox(2.2222,15.5556){\tiny.}}
\multiput(8983,-5054)(-10.45080,-12.54096){3}{\makebox(2.2222,15.5556){\tiny.}}
\multiput(8962,-5079)(-11.75000,-11.75000){3}{\makebox(2.2222,15.5556){\tiny.}}
\multiput(8938,-5102)(-8.72130,-7.26775){4}{\makebox(2.2222,15.5556){\tiny.}}
\multiput(8912,-5124)(-9.61540,-6.41027){4}{\makebox(2.2222,15.5556){\tiny.}}
\multiput(8883,-5143)(-10.49020,-6.29412){4}{\makebox(2.2222,15.5556){\tiny.}}
\multiput(8851,-5161)(-13.80000,-6.90000){3}{\makebox(2.2222,15.5556){\tiny.}}
\multiput(8823,-5174)(-15.00000,-6.00000){3}{\makebox(2.2222,15.5556){\tiny.}}
\multiput(8793,-5186)(-11.00000,-3.66667){4}{\makebox(2.2222,15.5556){\tiny.}}
\multiput(8760,-5197)(-11.76470,-2.94117){4}{\makebox(2.2222,15.5556){\tiny.}}
\multiput(8725,-5207)(-12.62747,-3.15687){4}{\makebox(2.2222,15.5556){\tiny.}}
\multiput(8687,-5216)(-13.25490,-3.31372){4}{\makebox(2.2222,15.5556){\tiny.}}
\multiput(8647,-5225)(-14.32433,-2.38739){4}{\makebox(2.2222,15.5556){\tiny.}}
\multiput(8604,-5232)(-11.27027,-1.87838){5}{\makebox(2.2222,15.5556){\tiny.}}
\multiput(8559,-5240)(-11.91892,-1.98649){5}{\makebox(2.2222,15.5556){\tiny.}}
\multiput(8511,-5246)(-12.40540,-2.06757){5}{\makebox(2.2222,15.5556){\tiny.}}
\multiput(8461,-5252)(-12.85135,-2.14189){5}{\makebox(2.2222,15.5556){\tiny.}}
\multiput(8409,-5257)(-13.58108,-2.26351){5}{\makebox(2.2222,15.5556){\tiny.}}
\put(8354,-5262){\line(-1, 0){ 56}}
\put(8298,-5266){\line(-1, 0){ 59}}
\put(8239,-5270){\line(-1, 0){ 61}}
\put(8178,-5273){\line(-1, 0){ 62}}
\put(8116,-5276){\line(-1, 0){ 64}}
\put(8052,-5278){\line(-1, 0){ 66}}
\put(7986,-5281){\line(-1, 0){ 67}}
\put(7919,-5283){\line(-1, 0){ 68}}
\put(7851,-5284){\line(-1, 0){ 69}}
\put(7782,-5285){\line(-1, 0){ 69}}
\put(7713,-5286){\line(-1, 0){ 71}}
\put(7642,-5287){\line(-1, 0){ 70}}
\put(7572,-5287){\line(-1, 0){ 71}}
\put(7501,-5288){\line(-1, 0){ 71}}
\put(7430,-5287){\line(-1, 0){ 70}}
\put(7360,-5287){\line(-1, 0){ 71}}
\put(7289,-5286){\line(-1, 0){ 69}}
\put(7220,-5285){\line(-1, 0){ 69}}
\put(7151,-5284){\line(-1, 0){ 68}}
\put(7083,-5283){\line(-1, 0){ 67}}
\put(7016,-5281){\line(-1, 0){ 66}}
\put(6950,-5278){\line(-1, 0){ 64}}
\put(6886,-5276){\line(-1, 0){ 62}}
\put(6824,-5273){\line(-1, 0){ 61}}
\put(6763,-5270){\line(-1, 0){ 59}}
\put(6704,-5266){\line(-1, 0){ 56}}
\multiput(6648,-5262)(-13.58108,2.26351){5}{\makebox(2.2222,15.5556){\tiny.}}
\multiput(6593,-5257)(-12.85135,2.14189){5}{\makebox(2.2222,15.5556){\tiny.}}
\multiput(6541,-5252)(-12.40540,2.06757){5}{\makebox(2.2222,15.5556){\tiny.}}
\multiput(6491,-5246)(-11.91892,1.98649){5}{\makebox(2.2222,15.5556){\tiny.}}
\multiput(6443,-5240)(-11.27027,1.87838){5}{\makebox(2.2222,15.5556){\tiny.}}
\multiput(6398,-5232)(-14.32433,2.38739){4}{\makebox(2.2222,15.5556){\tiny.}}
\multiput(6355,-5225)(-13.25490,3.31372){4}{\makebox(2.2222,15.5556){\tiny.}}
\multiput(6315,-5216)(-12.62747,3.15687){4}{\makebox(2.2222,15.5556){\tiny.}}
\multiput(6277,-5207)(-11.76470,2.94117){4}{\makebox(2.2222,15.5556){\tiny.}}
\multiput(6242,-5197)(-11.00000,3.66667){4}{\makebox(2.2222,15.5556){\tiny.}}
\multiput(6209,-5186)(-15.00000,6.00000){3}{\makebox(2.2222,15.5556){\tiny.}}
\multiput(6179,-5174)(-13.80000,6.90000){3}{\makebox(2.2222,15.5556){\tiny.}}
\multiput(6151,-5161)(-10.49020,6.29412){4}{\makebox(2.2222,15.5556){\tiny.}}
\multiput(6119,-5143)(-9.61540,6.41027){4}{\makebox(2.2222,15.5556){\tiny.}}
\multiput(6090,-5124)(-8.72130,7.26775){4}{\makebox(2.2222,15.5556){\tiny.}}
\multiput(6064,-5102)(-11.75000,11.75000){3}{\makebox(2.2222,15.5556){\tiny.}}
\multiput(6040,-5079)(-10.45080,12.54096){3}{\makebox(2.2222,15.5556){\tiny.}}
\multiput(6019,-5054)(-9.23075,13.84613){3}{\makebox(2.2222,15.5556){\tiny.}}
\multiput(6001,-5026)(-5.76470,9.60783){4}{\makebox(2.2222,15.5556){\tiny.}}
\multiput(5984,-4997)(-5.06667,10.13333){4}{\makebox(2.2222,15.5556){\tiny.}}
\multiput(5970,-4966)(-4.39080,10.97700){4}{\makebox(2.2222,15.5556){\tiny.}}
\multiput(5957,-4933)(-3.80000,11.40000){4}{\makebox(2.2222,15.5556){\tiny.}}
\multiput(5945,-4899)(-3.07843,12.31373){4}{\makebox(2.2222,15.5556){\tiny.}}
\multiput(5936,-4862)(-3.07843,12.31373){4}{\makebox(2.2222,15.5556){\tiny.}}
\multiput(5927,-4825)(-2.22523,13.35140){4}{\makebox(2.2222,15.5556){\tiny.}}
\multiput(5920,-4785)(-2.21620,13.29720){4}{\makebox(2.2222,15.5556){\tiny.}}
\multiput(5914,-4745)(-2.26127,13.56760){4}{\makebox(2.2222,15.5556){\tiny.}}
\multiput(5909,-4704)(-2.36037,14.16220){4}{\makebox(2.2222,15.5556){\tiny.}}
\put(5905,-4661){\line( 0, 1){ 43}}
\put(5902,-4618){\line( 0, 1){ 44}}
\put(5900,-4574){\line( 0, 1){ 44}}
\put(5898,-4530){\line( 0, 1){ 44}}
\put(5898,-4486){\line( 0, 1){ 44}}
\put(5898,-4442){\line( 0, 1){ 44}}
\put(5900,-4398){\line( 0, 1){ 44}}
\put(5902,-4354){\line( 0, 1){ 43}}
\multiput(5905,-4311)(2.36037,14.16220){4}{\makebox(2.2222,15.5556){\tiny.}}
\multiput(5909,-4268)(2.26127,13.56760){4}{\makebox(2.2222,15.5556){\tiny.}}
\multiput(5914,-4227)(2.21620,13.29720){4}{\makebox(2.2222,15.5556){\tiny.}}
\multiput(5920,-4187)(2.22523,13.35140){4}{\makebox(2.2222,15.5556){\tiny.}}
\multiput(5927,-4147)(3.07843,12.31373){4}{\makebox(2.2222,15.5556){\tiny.}}
\multiput(5936,-4110)(3.07843,12.31373){4}{\makebox(2.2222,15.5556){\tiny.}}
\multiput(5945,-4073)(3.80000,11.40000){4}{\makebox(2.2222,15.5556){\tiny.}}
\multiput(5957,-4039)(4.39080,10.97700){4}{\makebox(2.2222,15.5556){\tiny.}}
\multiput(5970,-4006)(5.06667,10.13333){4}{\makebox(2.2222,15.5556){\tiny.}}
\multiput(5984,-3975)(5.76470,9.60783){4}{\makebox(2.2222,15.5556){\tiny.}}
\multiput(6001,-3946)(9.23075,13.84613){3}{\makebox(2.2222,15.5556){\tiny.}}
\multiput(6019,-3918)(10.45080,12.54096){3}{\makebox(2.2222,15.5556){\tiny.}}
\multiput(6040,-3893)(11.75000,11.75000){3}{\makebox(2.2222,15.5556){\tiny.}}
\multiput(6064,-3870)(8.72130,7.26775){4}{\makebox(2.2222,15.5556){\tiny.}}
\multiput(6090,-3848)(9.61540,6.41027){4}{\makebox(2.2222,15.5556){\tiny.}}
\multiput(6119,-3829)(10.49020,6.29412){4}{\makebox(2.2222,15.5556){\tiny.}}
\multiput(6151,-3811)(13.80000,6.90000){3}{\makebox(2.2222,15.5556){\tiny.}}
\multiput(6179,-3798)(15.00000,6.00000){3}{\makebox(2.2222,15.5556){\tiny.}}
\multiput(6209,-3786)(11.00000,3.66667){4}{\makebox(2.2222,15.5556){\tiny.}}
\multiput(6242,-3775)(11.76470,2.94117){4}{\makebox(2.2222,15.5556){\tiny.}}
\multiput(6277,-3765)(12.62747,3.15687){4}{\makebox(2.2222,15.5556){\tiny.}}
\multiput(6315,-3756)(13.25490,3.31372){4}{\makebox(2.2222,15.5556){\tiny.}}
\multiput(6355,-3747)(14.32433,2.38739){4}{\makebox(2.2222,15.5556){\tiny.}}
\multiput(6398,-3740)(11.27027,1.87838){5}{\makebox(2.2222,15.5556){\tiny.}}
\multiput(6443,-3732)(11.91892,1.98649){5}{\makebox(2.2222,15.5556){\tiny.}}
\multiput(6491,-3726)(12.40540,2.06757){5}{\makebox(2.2222,15.5556){\tiny.}}
\multiput(6541,-3720)(12.85135,2.14189){5}{\makebox(2.2222,15.5556){\tiny.}}
\multiput(6593,-3715)(13.58108,2.26351){5}{\makebox(2.2222,15.5556){\tiny.}}
\put(6648,-3710){\line( 1, 0){ 56}}
\put(6704,-3706){\line( 1, 0){ 59}}
\put(6763,-3702){\line( 1, 0){ 61}}
\put(6824,-3699){\line( 1, 0){ 62}}
\put(6886,-3696){\line( 1, 0){ 64}}
\put(6950,-3694){\line( 1, 0){ 66}}
\put(7016,-3691){\line( 1, 0){ 67}}
\put(7083,-3689){\line( 1, 0){ 68}}
\put(7151,-3688){\line( 1, 0){ 69}}
\put(7220,-3687){\line( 1, 0){ 69}}
\put(7289,-3686){\line( 1, 0){ 71}}
\put(7360,-3685){\line( 1, 0){ 70}}
\put(7430,-3685){\line( 1, 0){ 71}}
\put(7501,-3684){\line( 1, 0){ 71}}
\put(7572,-3685){\line( 1, 0){ 70}}
\put(7642,-3685){\line( 1, 0){ 71}}
\put(7713,-3686){\line( 1, 0){ 69}}
\put(7782,-3687){\line( 1, 0){ 69}}
\put(7851,-3688){\line( 1, 0){ 68}}
\put(7919,-3689){\line( 1, 0){ 67}}
\put(7986,-3691){\line( 1, 0){ 66}}
\put(8052,-3694){\line( 1, 0){ 64}}
\put(8116,-3696){\line( 1, 0){ 62}}
\put(8178,-3699){\line( 1, 0){ 61}}
\put(8239,-3702){\line( 1, 0){ 59}}
\put(8298,-3706){\line( 1, 0){ 56}}
\multiput(8354,-3710)(13.58108,-2.26351){5}{\makebox(2.2222,15.5556){\tiny.}}
\multiput(8409,-3715)(12.85135,-2.14189){5}{\makebox(2.2222,15.5556){\tiny.}}
\multiput(8461,-3720)(12.40540,-2.06757){5}{\makebox(2.2222,15.5556){\tiny.}}
\multiput(8511,-3726)(11.91892,-1.98649){5}{\makebox(2.2222,15.5556){\tiny.}}
\multiput(8559,-3732)(11.27027,-1.87838){5}{\makebox(2.2222,15.5556){\tiny.}}
\multiput(8604,-3740)(14.32433,-2.38739){4}{\makebox(2.2222,15.5556){\tiny.}}
\multiput(8647,-3747)(13.25490,-3.31372){4}{\makebox(2.2222,15.5556){\tiny.}}
\multiput(8687,-3756)(12.62747,-3.15687){4}{\makebox(2.2222,15.5556){\tiny.}}
\multiput(8725,-3765)(11.76470,-2.94117){4}{\makebox(2.2222,15.5556){\tiny.}}
\multiput(8760,-3775)(11.00000,-3.66667){4}{\makebox(2.2222,15.5556){\tiny.}}
\multiput(8793,-3786)(15.00000,-6.00000){3}{\makebox(2.2222,15.5556){\tiny.}}
\multiput(8823,-3798)(13.80000,-6.90000){3}{\makebox(2.2222,15.5556){\tiny.}}
}%
\end{picture}%
\nop{
             fully plastic
           /     ^        \                              \space
         /       |          \                            \space
       /         |            \                          \space
      |          |             |
      V          |             V
   amnesic ------+          plastic
      |            \     /     |     \                   \space
      |              \ /       |       \                 \space
      |              / \       |         \               \space
      |             |    +----------------------+
      |             V    | \   V          V     |
      |         dogmatic | learnable   damascan |
      |             |    |                |     |
      |             |    |                |     |
      |             |    |                |     |
      |             V    |                V     |
      |         believer |           correcting |
      |        /         |                      |
      |      /           +----------------------+
      |    / 
      V   V
    equating
}
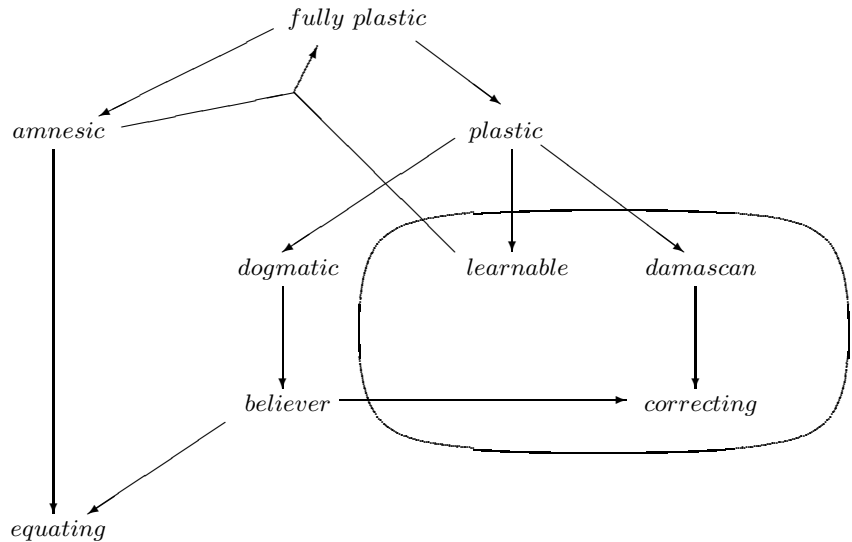
\hcaption{Natural, lexicographic and restrained revisions}
\label{figure-relations-nonepiphanic}
\end{hfigure}

Lemma~\ref{lemma-natural-singleclass}, Theorem~\ref{theorem-natural-learnable}
and Theorem~\ref{theorem-natural-damascan} prove that natural revision is
learnable and damascan even when restricting to sequences of single-class
revisions. Lexicographic and restrained revisions are proved to coincide with
it on single-class revisions and are therefore learnable and damascan as well:
Lemma~\ref{lemma-lexicographic-natural}, Lemma~\ref{lemma-restrained-natural},
Theorem~\ref{theorem-lexicographic-learnable},
Theorem~\ref{theorem-restrained-learnable},
Theorem~\ref{theorem-lexicographic-damascan} and
Theorem~\ref{theorem-restrained-damascan}.

Theorem~\ref{theorem-natural-equating},
Theorem~\ref{theorem-lexicographic-equating} and
Theorem~\ref{theorem-restrained-equating} prove that natural, lexicographic and
restrained revisions are not equating. Therefore, they do not have the implying
abilities: believer, dogmatic, amnesic, plastic and fully plastic.

\begin{hfigure}
\setlength{\unitlength}{3750sp}%
\begin{picture}(5942,3835)(3174,-6136)
\thinlines
{\color[rgb]{0,0,0}\put(6274,-2572){\vector( 4,-3){552}}
}%
{\color[rgb]{0,0,0}\put(5341,-2491){\vector(-2,-1){1140}}
}%
{\color[rgb]{0,0,0}\put(3901,-3286){\vector( 0,-1){2400}}
}%
{\color[rgb]{0,0,0}\put(6526,-3211){\vector(-3,-2){1125}}
}%
{\color[rgb]{0,0,0}\put(6901,-3286){\vector( 0,-1){675}}
}%
{\color[rgb]{0,0,0}\put(7090,-3259){\vector( 4,-3){936}}
}%
{\color[rgb]{0,0,0}\put(8101,-4186){\vector( 0,-1){675}}
}%
{\color[rgb]{0,0,0}\put(5401,-4186){\vector( 0,-1){675}}
}%
{\color[rgb]{0,0,0}\put(4351,-3136){\line( 5, 1){1125}}
}%
{\color[rgb]{0,0,0}\put(6526,-3961){\line(-1, 1){1050}}
\multiput(5476,-2911)(5.00000,10.00000){31}{\makebox(2.2222,15.5556){\tiny.}}
\put(5626,-2611){\vector( 1, 2){0}}
}%
{\color[rgb]{0,0,0}\put(5776,-4936){\vector( 1, 0){1875}}
}%
{\color[rgb]{0,0,0}\put(5026,-5086){\vector(-3,-2){900}}
}%
\put(5851,-2461){\makebox(0,0)[b]{\smash{\fontsize{9}{10.8}
\usefont{T1}{cmr}{m}{n}{\color[rgb]{0,0,0}$fully\ plastic$}%
}}}
\put(3901,-3211){\makebox(0,0)[b]{\smash{\fontsize{9}{10.8}
\usefont{T1}{cmr}{m}{n}{\color[rgb]{0,0,0}$amnesic$}%
}}}
\put(6901,-4111){\makebox(0,0)[b]{\smash{\fontsize{9}{10.8}
\usefont{T1}{cmr}{m}{n}{\color[rgb]{0,0,0}$learnable$}%
}}}
\put(6826,-3211){\makebox(0,0)[b]{\smash{\fontsize{9}{10.8}
\usefont{T1}{cmr}{m}{n}{\color[rgb]{0,0,0}$plastic$}%
}}}
\put(5401,-4111){\makebox(0,0)[b]{\smash{\fontsize{9}{10.8}
\usefont{T1}{cmr}{m}{n}{\color[rgb]{0,0,0}$dogmatic$}%
}}}
\put(8101,-4111){\makebox(0,0)[b]{\smash{\fontsize{9}{10.8}
\usefont{T1}{cmr}{m}{n}{\color[rgb]{0,0,0}$damascan$}%
}}}
\put(8101,-5011){\makebox(0,0)[b]{\smash{\fontsize{9}{10.8}
\usefont{T1}{cmr}{m}{n}{\color[rgb]{0,0,0}$correcting$}%
}}}
\put(5401,-5011){\makebox(0,0)[b]{\smash{\fontsize{9}{10.8}
\usefont{T1}{cmr}{m}{n}{\color[rgb]{0,0,0}$believer$}%
}}}
\put(3901,-5836){\makebox(0,0)[b]{\smash{\fontsize{9}{10.8}
\usefont{T1}{cmr}{m}{n}{\color[rgb]{0,0,0}$equating$}%
}}}
{\color[rgb]{0,0,0}\multiput(7576,-2836)(12.00000,-4.00000){5}{\makebox(2.2222,15.5556){\tiny.}}
\multiput(7624,-2852)(11.55172,-4.62069){5}{\makebox(2.2222,15.5556){\tiny.}}
\multiput(7670,-2871)(11.20000,-5.60000){5}{\makebox(2.2222,15.5556){\tiny.}}
\multiput(7715,-2893)(10.55147,-6.33089){5}{\makebox(2.2222,15.5556){\tiny.}}
\multiput(7758,-2917)(10.77205,-6.46323){5}{\makebox(2.2222,15.5556){\tiny.}}
\multiput(7801,-2943)(10.44230,-6.96153){5}{\makebox(2.2222,15.5556){\tiny.}}
\multiput(7842,-2972)(10.28000,-7.71000){5}{\makebox(2.2222,15.5556){\tiny.}}
\multiput(7883,-3003)(10.08197,-8.40165){5}{\makebox(2.2222,15.5556){\tiny.}}
\multiput(7923,-3037)(9.62500,-9.62500){5}{\makebox(2.2222,15.5556){\tiny.}}
\multiput(7963,-3074)(9.75000,-9.75000){5}{\makebox(2.2222,15.5556){\tiny.}}
\multiput(8003,-3112)(8.00000,-8.00000){6}{\makebox(2.2222,15.5556){\tiny.}}
\multiput(8042,-3153)(7.42622,-8.91146){6}{\makebox(2.2222,15.5556){\tiny.}}
\multiput(8081,-3196)(7.72132,-9.26558){6}{\makebox(2.2222,15.5556){\tiny.}}
\multiput(8120,-3242)(7.55122,-9.43903){6}{\makebox(2.2222,15.5556){\tiny.}}
\multiput(8158,-3289)(7.92196,-9.90245){6}{\makebox(2.2222,15.5556){\tiny.}}
\multiput(8197,-3339)(7.63200,-10.17600){6}{\makebox(2.2222,15.5556){\tiny.}}
\multiput(8235,-3390)(7.82400,-10.43200){6}{\makebox(2.2222,15.5556){\tiny.}}
\multiput(8273,-3443)(7.32308,-10.98462){6}{\makebox(2.2222,15.5556){\tiny.}}
\multiput(8311,-3497)(6.30770,-9.46155){7}{\makebox(2.2222,15.5556){\tiny.}}
\multiput(8350,-3553)(6.35897,-9.53845){7}{\makebox(2.2222,15.5556){\tiny.}}
\multiput(8387,-3611)(6.41025,-9.61538){7}{\makebox(2.2222,15.5556){\tiny.}}
\multiput(8425,-3669)(6.04412,-10.07353){7}{\makebox(2.2222,15.5556){\tiny.}}
\multiput(8462,-3729)(6.04412,-10.07353){7}{\makebox(2.2222,15.5556){\tiny.}}
\multiput(8499,-3789)(6.19118,-10.31864){7}{\makebox(2.2222,15.5556){\tiny.}}
\multiput(8536,-3851)(6.07353,-10.12256){7}{\makebox(2.2222,15.5556){\tiny.}}
\multiput(8572,-3912)(6.22058,-10.36764){7}{\makebox(2.2222,15.5556){\tiny.}}
\multiput(8608,-3975)(5.26667,-10.53333){7}{\makebox(2.2222,15.5556){\tiny.}}
\multiput(8642,-4037)(5.33333,-10.66667){7}{\makebox(2.2222,15.5556){\tiny.}}
\multiput(8676,-4100)(5.33333,-10.66667){7}{\makebox(2.2222,15.5556){\tiny.}}
\multiput(8710,-4163)(5.20000,-10.40000){7}{\makebox(2.2222,15.5556){\tiny.}}
\multiput(8742,-4225)(5.16667,-10.33333){7}{\makebox(2.2222,15.5556){\tiny.}}
\multiput(8773,-4287)(5.13333,-10.26667){7}{\makebox(2.2222,15.5556){\tiny.}}
\multiput(8803,-4349)(5.00000,-10.00000){7}{\makebox(2.2222,15.5556){\tiny.}}
\multiput(8831,-4410)(5.88000,-11.76000){6}{\makebox(2.2222,15.5556){\tiny.}}
\multiput(8858,-4470)(4.82758,-12.06895){6}{\makebox(2.2222,15.5556){\tiny.}}
\multiput(8883,-4530)(4.63448,-11.58620){6}{\makebox(2.2222,15.5556){\tiny.}}
\multiput(8906,-4588)(4.60690,-11.51725){6}{\makebox(2.2222,15.5556){\tiny.}}
\multiput(8928,-4646)(3.74000,-11.22000){6}{\makebox(2.2222,15.5556){\tiny.}}
\multiput(8947,-4702)(3.64000,-10.92000){6}{\makebox(2.2222,15.5556){\tiny.}}
\multiput(8964,-4757)(2.71764,-10.87056){6}{\makebox(2.2222,15.5556){\tiny.}}
\multiput(8979,-4811)(3.23530,-12.94120){5}{\makebox(2.2222,15.5556){\tiny.}}
\multiput(8991,-4863)(2.08785,-12.52710){5}{\makebox(2.2222,15.5556){\tiny.}}
\multiput(9000,-4913)(2.03377,-12.20265){5}{\makebox(2.2222,15.5556){\tiny.}}
\put(9007,-4962){\line( 0,-1){ 47}}
\put(9010,-5009){\line( 0,-1){ 45}}
\put(9011,-5054){\line( 0,-1){ 44}}
\multiput(9008,-5098)(-2.32433,-13.94600){4}{\makebox(2.2222,15.5556){\tiny.}}
\multiput(9002,-5140)(-3.33333,-13.33333){4}{\makebox(2.2222,15.5556){\tiny.}}
\multiput(8992,-5180)(-4.23333,-12.70000){4}{\makebox(2.2222,15.5556){\tiny.}}
\multiput(8979,-5218)(-6.13333,-12.26667){4}{\makebox(2.2222,15.5556){\tiny.}}
\multiput(8961,-5255)(-7.00000,-11.66667){4}{\makebox(2.2222,15.5556){\tiny.}}
\multiput(8940,-5290)(-8.44000,-11.25333){4}{\makebox(2.2222,15.5556){\tiny.}}
\multiput(8915,-5324)(-10.16667,-10.16667){4}{\makebox(2.2222,15.5556){\tiny.}}
\multiput(8885,-5355)(-8.82788,-7.35656){5}{\makebox(2.2222,15.5556){\tiny.}}
\multiput(8851,-5386)(-9.26830,-7.41464){4}{\makebox(2.2222,15.5556){\tiny.}}
\multiput(8823,-5408)(-10.38460,-6.92307){4}{\makebox(2.2222,15.5556){\tiny.}}
\multiput(8792,-5429)(-11.42157,-6.85294){4}{\makebox(2.2222,15.5556){\tiny.}}
\multiput(8758,-5450)(-11.76470,-7.05882){4}{\makebox(2.2222,15.5556){\tiny.}}
\multiput(8722,-5470)(-13.06667,-6.53333){4}{\makebox(2.2222,15.5556){\tiny.}}
\multiput(8683,-5490)(-10.30000,-5.15000){5}{\makebox(2.2222,15.5556){\tiny.}}
\multiput(8641,-5509)(-11.12070,-4.44828){5}{\makebox(2.2222,15.5556){\tiny.}}
\multiput(8597,-5528)(-11.68102,-4.67241){5}{\makebox(2.2222,15.5556){\tiny.}}
\multiput(8550,-5546)(-12.60000,-4.20000){5}{\makebox(2.2222,15.5556){\tiny.}}
\multiput(8500,-5564)(-10.56000,-3.52000){6}{\makebox(2.2222,15.5556){\tiny.}}
\multiput(8447,-5581)(-10.92000,-3.64000){6}{\makebox(2.2222,15.5556){\tiny.}}
\multiput(8392,-5598)(-11.90588,-2.97647){6}{\makebox(2.2222,15.5556){\tiny.}}
\multiput(8333,-5615)(-12.04706,-3.01177){6}{\makebox(2.2222,15.5556){\tiny.}}
\multiput(8273,-5631)(-12.80000,-3.20000){6}{\makebox(2.2222,15.5556){\tiny.}}
\multiput(8209,-5647)(-10.98040,-2.74510){7}{\makebox(2.2222,15.5556){\tiny.}}
\multiput(8143,-5663)(-11.45098,-2.86275){7}{\makebox(2.2222,15.5556){\tiny.}}
\multiput(8074,-5679)(-11.85897,-2.37179){7}{\makebox(2.2222,15.5556){\tiny.}}
\multiput(8003,-5694)(-12.17948,-2.43590){7}{\makebox(2.2222,15.5556){\tiny.}}
\multiput(7930,-5709)(-12.50000,-2.50000){7}{\makebox(2.2222,15.5556){\tiny.}}
\multiput(7855,-5724)(-11.12637,-2.22527){8}{\makebox(2.2222,15.5556){\tiny.}}
\multiput(7777,-5739)(-11.44401,-1.90734){8}{\makebox(2.2222,15.5556){\tiny.}}
\multiput(7697,-5753)(-11.74517,-1.95753){8}{\makebox(2.2222,15.5556){\tiny.}}
\multiput(7615,-5768)(-12.00000,-2.00000){8}{\makebox(2.2222,15.5556){\tiny.}}
\multiput(7531,-5782)(-12.13900,-2.02317){8}{\makebox(2.2222,15.5556){\tiny.}}
\multiput(7446,-5796)(-12.41699,-2.06950){8}{\makebox(2.2222,15.5556){\tiny.}}
\multiput(7359,-5810)(-11.10811,-1.85135){9}{\makebox(2.2222,15.5556){\tiny.}}
\multiput(7270,-5824)(-11.20946,-1.86824){9}{\makebox(2.2222,15.5556){\tiny.}}
\multiput(7180,-5837)(-11.47297,-1.91216){9}{\makebox(2.2222,15.5556){\tiny.}}
\multiput(7088,-5851)(-11.45270,-1.90878){9}{\makebox(2.2222,15.5556){\tiny.}}
\multiput(6996,-5864)(-11.69595,-1.94932){9}{\makebox(2.2222,15.5556){\tiny.}}
\multiput(6902,-5877)(-11.69595,-1.94932){9}{\makebox(2.2222,15.5556){\tiny.}}
\multiput(6808,-5890)(-11.81756,-1.96959){9}{\makebox(2.2222,15.5556){\tiny.}}
\multiput(6713,-5903)(-11.93919,-1.98986){9}{\makebox(2.2222,15.5556){\tiny.}}
\multiput(6617,-5916)(-12.04054,-2.00676){9}{\makebox(2.2222,15.5556){\tiny.}}
\multiput(6520,-5928)(-11.91892,-1.98649){9}{\makebox(2.2222,15.5556){\tiny.}}
\multiput(6424,-5940)(-12.04054,-2.00676){9}{\makebox(2.2222,15.5556){\tiny.}}
\multiput(6327,-5952)(-12.04054,-2.00676){9}{\makebox(2.2222,15.5556){\tiny.}}
\multiput(6230,-5964)(-12.02027,-2.00338){9}{\makebox(2.2222,15.5556){\tiny.}}
\multiput(6133,-5975)(-12.02027,-2.00338){9}{\makebox(2.2222,15.5556){\tiny.}}
\multiput(6036,-5986)(-11.89865,-1.98311){9}{\makebox(2.2222,15.5556){\tiny.}}
\multiput(5940,-5997)(-11.89865,-1.98311){9}{\makebox(2.2222,15.5556){\tiny.}}
\multiput(5844,-6008)(-11.75676,-1.95946){9}{\makebox(2.2222,15.5556){\tiny.}}
\multiput(5749,-6018)(-11.75676,-1.95946){9}{\makebox(2.2222,15.5556){\tiny.}}
\multiput(5654,-6028)(-11.51351,-1.91892){9}{\makebox(2.2222,15.5556){\tiny.}}
\multiput(5561,-6038)(-11.49324,-1.91554){9}{\makebox(2.2222,15.5556){\tiny.}}
\multiput(5468,-6047)(-11.25000,-1.87500){9}{\makebox(2.2222,15.5556){\tiny.}}
\multiput(5377,-6056)(-11.10811,-1.85135){9}{\makebox(2.2222,15.5556){\tiny.}}
\multiput(5287,-6064)(-10.98649,-1.83108){9}{\makebox(2.2222,15.5556){\tiny.}}
\multiput(5198,-6072)(-12.27799,-2.04633){8}{\makebox(2.2222,15.5556){\tiny.}}
\put(5111,-6080){\line(-1, 0){ 86}}
\put(5025,-6087){\line(-1, 0){ 84}}
\put(4941,-6093){\line(-1, 0){ 82}}
\put(4859,-6099){\line(-1, 0){ 81}}
\put(4778,-6104){\line(-1, 0){ 78}}
\put(4700,-6109){\line(-1, 0){ 77}}
\put(4623,-6113){\line(-1, 0){ 74}}
\put(4549,-6117){\line(-1, 0){ 72}}
\put(4477,-6120){\line(-1, 0){ 70}}
\put(4407,-6122){\line(-1, 0){ 67}}
\put(4340,-6123){\line(-1, 0){ 66}}
\put(4274,-6124){\line(-1, 0){ 62}}
\put(4212,-6124){\line(-1, 0){ 61}}
\put(4151,-6123){\line(-1, 0){ 58}}
\put(4093,-6121){\line(-1, 0){ 56}}
\put(4037,-6118){\line(-1, 0){ 53}}
\multiput(3984,-6115)(-12.60810,2.10135){5}{\makebox(2.2222,15.5556){\tiny.}}
\multiput(3933,-6110)(-11.91892,1.98649){5}{\makebox(2.2222,15.5556){\tiny.}}
\multiput(3885,-6104)(-11.43243,1.90540){5}{\makebox(2.2222,15.5556){\tiny.}}
\multiput(3839,-6098)(-11.02703,1.83784){5}{\makebox(2.2222,15.5556){\tiny.}}
\multiput(3795,-6090)(-13.97437,2.79487){4}{\makebox(2.2222,15.5556){\tiny.}}
\multiput(3753,-6082)(-13.33333,3.33333){4}{\makebox(2.2222,15.5556){\tiny.}}
\multiput(3713,-6072)(-12.20000,4.06667){4}{\makebox(2.2222,15.5556){\tiny.}}
\multiput(3676,-6061)(-12.30000,4.10000){5}{\makebox(2.2222,15.5556){\tiny.}}
\multiput(3627,-6044)(-11.33620,4.53448){5}{\makebox(2.2222,15.5556){\tiny.}}
\multiput(3582,-6025)(-10.40000,5.20000){5}{\makebox(2.2222,15.5556){\tiny.}}
\multiput(3541,-6003)(-9.81617,5.88970){5}{\makebox(2.2222,15.5556){\tiny.}}
\multiput(3502,-5979)(-11.62667,8.72000){4}{\makebox(2.2222,15.5556){\tiny.}}
\multiput(3467,-5953)(-10.88523,9.07103){4}{\makebox(2.2222,15.5556){\tiny.}}
\multiput(3435,-5925)(-10.00000,10.00000){4}{\makebox(2.2222,15.5556){\tiny.}}
\multiput(3406,-5894)(-9.09837,10.91804){4}{\makebox(2.2222,15.5556){\tiny.}}
\multiput(3379,-5861)(-8.10257,12.15385){4}{\makebox(2.2222,15.5556){\tiny.}}
\multiput(3354,-5825)(-7.38237,12.30394){4}{\makebox(2.2222,15.5556){\tiny.}}
\multiput(3332,-5788)(-6.60000,13.20000){4}{\makebox(2.2222,15.5556){\tiny.}}
\multiput(3313,-5748)(-4.24137,10.60344){5}{\makebox(2.2222,15.5556){\tiny.}}
\multiput(3295,-5706)(-3.60000,10.80000){5}{\makebox(2.2222,15.5556){\tiny.}}
\multiput(3280,-5663)(-3.80000,11.40000){5}{\makebox(2.2222,15.5556){\tiny.}}
\multiput(3266,-5617)(-2.94117,11.76470){5}{\makebox(2.2222,15.5556){\tiny.}}
\multiput(3254,-5570)(-2.98530,11.94120){5}{\makebox(2.2222,15.5556){\tiny.}}
\multiput(3243,-5522)(-2.12838,12.77025){5}{\makebox(2.2222,15.5556){\tiny.}}
\multiput(3234,-5471)(-2.11488,12.68925){5}{\makebox(2.2222,15.5556){\tiny.}}
\multiput(3227,-5420)(-2.18920,13.13520){5}{\makebox(2.2222,15.5556){\tiny.}}
\multiput(3221,-5367)(-2.22297,13.33785){5}{\makebox(2.2222,15.5556){\tiny.}}
\put(3216,-5313){\line( 0, 1){ 55}}
\put(3212,-5258){\line( 0, 1){ 56}}
\put(3210,-5202){\line( 0, 1){ 57}}
\put(3208,-5145){\line( 0, 1){ 57}}
\put(3208,-5088){\line( 0, 1){ 58}}
\put(3208,-5030){\line( 0, 1){ 58}}
\put(3210,-4972){\line( 0, 1){ 58}}
\put(3212,-4914){\line( 0, 1){ 58}}
\put(3215,-4856){\line( 0, 1){ 58}}
\multiput(3218,-4798)(1.87568,11.25408){6}{\makebox(2.2222,15.5556){\tiny.}}
\multiput(3223,-4741)(1.90810,11.44860){6}{\makebox(2.2222,15.5556){\tiny.}}
\multiput(3228,-4683)(1.84864,11.09184){6}{\makebox(2.2222,15.5556){\tiny.}}
\multiput(3234,-4627)(1.85406,11.12436){6}{\makebox(2.2222,15.5556){\tiny.}}
\multiput(3241,-4571)(1.85946,11.15676){6}{\makebox(2.2222,15.5556){\tiny.}}
\multiput(3249,-4515)(2.24325,13.45950){5}{\makebox(2.2222,15.5556){\tiny.}}
\multiput(3257,-4461)(2.69230,13.46150){5}{\makebox(2.2222,15.5556){\tiny.}}
\multiput(3267,-4407)(2.59615,12.98075){5}{\makebox(2.2222,15.5556){\tiny.}}
\multiput(3277,-4355)(2.55770,12.78850){5}{\makebox(2.2222,15.5556){\tiny.}}
\multiput(3288,-4304)(3.11765,12.47060){5}{\makebox(2.2222,15.5556){\tiny.}}
\multiput(3300,-4254)(3.95000,11.85000){5}{\makebox(2.2222,15.5556){\tiny.}}
\multiput(3314,-4206)(3.95000,11.85000){5}{\makebox(2.2222,15.5556){\tiny.}}
\multiput(3328,-4158)(3.77500,11.32500){5}{\makebox(2.2222,15.5556){\tiny.}}
\multiput(3344,-4113)(4.50000,11.25000){5}{\makebox(2.2222,15.5556){\tiny.}}
\multiput(3362,-4068)(4.24137,10.60344){5}{\makebox(2.2222,15.5556){\tiny.}}
\multiput(3380,-4026)(5.25000,10.50000){5}{\makebox(2.2222,15.5556){\tiny.}}
\multiput(3401,-3984)(5.75735,9.59558){5}{\makebox(2.2222,15.5556){\tiny.}}
\multiput(3423,-3945)(5.88970,9.81617){5}{\makebox(2.2222,15.5556){\tiny.}}
\multiput(3447,-3906)(8.88000,11.84000){4}{\makebox(2.2222,15.5556){\tiny.}}
\multiput(3473,-3870)(7.12195,8.90244){5}{\makebox(2.2222,15.5556){\tiny.}}
\multiput(3501,-3834)(7.35655,8.82786){5}{\makebox(2.2222,15.5556){\tiny.}}
\multiput(3532,-3800)(8.12500,8.12500){5}{\makebox(2.2222,15.5556){\tiny.}}
\multiput(3565,-3768)(9.24590,7.70492){5}{\makebox(2.2222,15.5556){\tiny.}}
\multiput(3601,-3736)(10.93497,8.74797){4}{\makebox(2.2222,15.5556){\tiny.}}
\multiput(3634,-3710)(11.25333,8.44000){4}{\makebox(2.2222,15.5556){\tiny.}}
\multiput(3668,-3685)(12.23077,8.15384){4}{\makebox(2.2222,15.5556){\tiny.}}
\multiput(3705,-3661)(9.81617,5.88970){5}{\makebox(2.2222,15.5556){\tiny.}}
\multiput(3744,-3637)(10.70000,5.35000){5}{\makebox(2.2222,15.5556){\tiny.}}
\multiput(3786,-3614)(10.55147,6.33089){5}{\makebox(2.2222,15.5556){\tiny.}}
\multiput(3829,-3590)(11.40000,5.70000){5}{\makebox(2.2222,15.5556){\tiny.}}
\multiput(3875,-3568)(12.10000,6.05000){5}{\makebox(2.2222,15.5556){\tiny.}}
\multiput(3924,-3545)(12.67243,5.06897){5}{\makebox(2.2222,15.5556){\tiny.}}
\multiput(3974,-3523)(10.65518,4.26207){6}{\makebox(2.2222,15.5556){\tiny.}}
\multiput(4027,-3501)(11.00000,4.40000){6}{\makebox(2.2222,15.5556){\tiny.}}
\multiput(4082,-3479)(11.34482,4.53793){6}{\makebox(2.2222,15.5556){\tiny.}}
\multiput(4139,-3457)(12.06000,4.02000){6}{\makebox(2.2222,15.5556){\tiny.}}
\multiput(4199,-3436)(12.30000,4.10000){6}{\makebox(2.2222,15.5556){\tiny.}}
\multiput(4260,-3414)(10.55000,3.51667){7}{\makebox(2.2222,15.5556){\tiny.}}
\multiput(4323,-3392)(10.80000,3.60000){7}{\makebox(2.2222,15.5556){\tiny.}}
\multiput(4388,-3371)(11.10000,3.70000){7}{\makebox(2.2222,15.5556){\tiny.}}
\multiput(4455,-3350)(11.45000,3.81667){7}{\makebox(2.2222,15.5556){\tiny.}}
\multiput(4524,-3328)(11.55000,3.85000){7}{\makebox(2.2222,15.5556){\tiny.}}
\multiput(4594,-3307)(11.90000,3.96667){7}{\makebox(2.2222,15.5556){\tiny.}}
\multiput(4666,-3285)(12.27452,3.06863){7}{\makebox(2.2222,15.5556){\tiny.}}
\multiput(4739,-3264)(10.78991,2.69748){8}{\makebox(2.2222,15.5556){\tiny.}}
\multiput(4814,-3243)(10.92437,2.73109){8}{\makebox(2.2222,15.5556){\tiny.}}
\multiput(4890,-3222)(11.05883,2.76471){8}{\makebox(2.2222,15.5556){\tiny.}}
\multiput(4967,-3201)(11.19327,2.79832){8}{\makebox(2.2222,15.5556){\tiny.}}
\multiput(5045,-3180)(11.19327,2.79832){8}{\makebox(2.2222,15.5556){\tiny.}}
\multiput(5123,-3159)(11.46219,2.86555){8}{\makebox(2.2222,15.5556){\tiny.}}
\multiput(5203,-3138)(11.42857,2.85714){8}{\makebox(2.2222,15.5556){\tiny.}}
\multiput(5283,-3118)(11.46219,2.86555){8}{\makebox(2.2222,15.5556){\tiny.}}
\multiput(5363,-3097)(11.56303,2.89076){8}{\makebox(2.2222,15.5556){\tiny.}}
\multiput(5444,-3077)(11.56303,2.89076){8}{\makebox(2.2222,15.5556){\tiny.}}
\multiput(5525,-3057)(11.52941,2.88235){8}{\makebox(2.2222,15.5556){\tiny.}}
\multiput(5606,-3038)(11.52941,2.88235){8}{\makebox(2.2222,15.5556){\tiny.}}
\multiput(5687,-3019)(11.52941,2.88235){8}{\makebox(2.2222,15.5556){\tiny.}}
\multiput(5768,-3000)(11.62089,2.32418){8}{\makebox(2.2222,15.5556){\tiny.}}
\multiput(5849,-2982)(11.36134,2.84034){8}{\makebox(2.2222,15.5556){\tiny.}}
\multiput(5929,-2964)(11.45604,2.29121){8}{\makebox(2.2222,15.5556){\tiny.}}
\multiput(6009,-2947)(11.31869,2.26374){8}{\makebox(2.2222,15.5556){\tiny.}}
\multiput(6088,-2930)(11.15384,2.23077){8}{\makebox(2.2222,15.5556){\tiny.}}
\multiput(6166,-2914)(11.12637,2.22527){8}{\makebox(2.2222,15.5556){\tiny.}}
\multiput(6244,-2899)(12.66025,2.53205){7}{\makebox(2.2222,15.5556){\tiny.}}
\multiput(6320,-2884)(12.62820,2.52564){7}{\makebox(2.2222,15.5556){\tiny.}}
\multiput(6396,-2870)(12.35135,2.05856){7}{\makebox(2.2222,15.5556){\tiny.}}
\multiput(6470,-2857)(12.16217,2.02703){7}{\makebox(2.2222,15.5556){\tiny.}}
\multiput(6543,-2845)(11.81082,1.96847){7}{\makebox(2.2222,15.5556){\tiny.}}
\multiput(6614,-2834)(11.62162,1.93694){7}{\makebox(2.2222,15.5556){\tiny.}}
\multiput(6684,-2824)(11.43243,1.90541){7}{\makebox(2.2222,15.5556){\tiny.}}
\multiput(6753,-2815)(11.10810,1.85135){7}{\makebox(2.2222,15.5556){\tiny.}}
\multiput(6820,-2806)(12.87568,2.14595){6}{\makebox(2.2222,15.5556){\tiny.}}
\put(6885,-2799){\line( 1, 0){ 64}}
\put(6949,-2794){\line( 1, 0){ 62}}
\put(7011,-2789){\line( 1, 0){ 60}}
\put(7071,-2786){\line( 1, 0){ 58}}
\put(7129,-2784){\line( 1, 0){ 57}}
\put(7186,-2784){\line( 1, 0){ 55}}
\put(7241,-2785){\line( 1, 0){ 53}}
\put(7294,-2787){\line( 1, 0){ 51}}
\multiput(7345,-2791)(12.16215,-2.02703){5}{\makebox(2.2222,15.5556){\tiny.}}
\multiput(7394,-2797)(11.95945,-1.99324){5}{\makebox(2.2222,15.5556){\tiny.}}
\multiput(7442,-2804)(11.49038,-2.29807){5}{\makebox(2.2222,15.5556){\tiny.}}
\multiput(7488,-2813)(11.23530,-2.80883){5}{\makebox(2.2222,15.5556){\tiny.}}
\multiput(7533,-2824)(10.82353,-2.70588){5}{\makebox(2.2222,15.5556){\tiny.}}
}%
\end{picture}%
\nop{
             fully plastic
           /     ^        \                              \space
         /       |          \                            \space
       /      +---------------------------------+
      |       |  |             |                |
      V       |  |             V                |
   amnesic ---|--+          plastic             |
      |       |    \     /     |     \          |
      |       |      \ /       |       \        |
      |       |      / \       |         \      |
      |       |     |    \     |          |     |
      |       |     V      \   V          V     |
      |       | dogmatic   learnable   damascan |
      |       |     |                     |     |
      |       |     |                     |     |
  +-----------+     |                     |     |
  |   |             V                     V     |
  |   |         believer ----------> correcting |
  |   |        /                                |
  |   |      /                                  |
  |   |    /                                    |
  |   V   V                                     |
  | equating                                    |
  |                                             |
  +---------------------------------------------+
}
\hcaption{Very radical, severe, moderate severe and deep severe revisions}
\label{figure-relations-epiphanic}
\end{hfigure}

Theorem~\ref{theorem-plastic-veryradical} proves that very radical revision is
plastic. The same is proved for single-class revisions for severe revisions in
Lemma~\ref{lemma-moderatesevere-severe}. Since moderate and deep severe
revisions coincide with it by Lemma~\ref{lemma-moderatesevere-severe} and
Lemma~\ref{lemma-deepsevere-severe}, they are all plastic:
Theorem~\ref{theorem-severe-plastic},
Theorem~\ref{theorem-moderatesevere-plastic} and
Theorem~\ref{theorem-deepsevere-plastic}.

These revisions are not amnesic and therefore not fully plastic either. Each
revision requires its own proof: Theorem~\ref{theorem-veryradical-amnesic},
Corollary~\ref{corollary-severe-amnesic},
Theorem~\ref{theorem-moderatesevere-amnesic} and
Theorem~\ref{theorem-deepsevere-amnesic}.

\begin{hfigure}
\setlength{\unitlength}{3750sp}%
\begin{picture}(5942,3835)(3174,-6136)
\thinlines
{\color[rgb]{0,0,0}\put(6274,-2572){\vector( 4,-3){552}}
}%
{\color[rgb]{0,0,0}\put(5341,-2491){\vector(-2,-1){1140}}
}%
{\color[rgb]{0,0,0}\put(3901,-3286){\vector( 0,-1){2400}}
}%
{\color[rgb]{0,0,0}\put(6526,-3211){\vector(-3,-2){1125}}
}%
{\color[rgb]{0,0,0}\put(6901,-3286){\vector( 0,-1){675}}
}%
{\color[rgb]{0,0,0}\put(7090,-3259){\vector( 4,-3){936}}
}%
{\color[rgb]{0,0,0}\put(8101,-4186){\vector( 0,-1){675}}
}%
{\color[rgb]{0,0,0}\put(5401,-4186){\vector( 0,-1){675}}
}%
{\color[rgb]{0,0,0}\put(4351,-3136){\line( 5, 1){1125}}
}%
{\color[rgb]{0,0,0}\put(6526,-3961){\line(-1, 1){1050}}
\multiput(5476,-2911)(5.00000,10.00000){31}{\makebox(2.2222,15.5556){\tiny.}}
\put(5626,-2611){\vector( 1, 2){0}}
}%
{\color[rgb]{0,0,0}\put(5776,-4936){\vector( 1, 0){1875}}
}%
{\color[rgb]{0,0,0}\put(5026,-5086){\vector(-3,-2){900}}
}%
\put(5851,-2461){\makebox(0,0)[b]{\smash{\fontsize{9}{10.8}
\usefont{T1}{cmr}{m}{n}{\color[rgb]{0,0,0}$fully\ plastic$}%
}}}
\put(3901,-3211){\makebox(0,0)[b]{\smash{\fontsize{9}{10.8}
\usefont{T1}{cmr}{m}{n}{\color[rgb]{0,0,0}$amnesic$}%
}}}
\put(6901,-4111){\makebox(0,0)[b]{\smash{\fontsize{9}{10.8}
\usefont{T1}{cmr}{m}{n}{\color[rgb]{0,0,0}$learnable$}%
}}}
\put(6826,-3211){\makebox(0,0)[b]{\smash{\fontsize{9}{10.8}
\usefont{T1}{cmr}{m}{n}{\color[rgb]{0,0,0}$plastic$}%
}}}
\put(5401,-4111){\makebox(0,0)[b]{\smash{\fontsize{9}{10.8}
\usefont{T1}{cmr}{m}{n}{\color[rgb]{0,0,0}$dogmatic$}%
}}}
\put(8101,-4111){\makebox(0,0)[b]{\smash{\fontsize{9}{10.8}
\usefont{T1}{cmr}{m}{n}{\color[rgb]{0,0,0}$damascan$}%
}}}
\put(8101,-5011){\makebox(0,0)[b]{\smash{\fontsize{9}{10.8}
\usefont{T1}{cmr}{m}{n}{\color[rgb]{0,0,0}$correcting$}%
}}}
\put(5401,-5011){\makebox(0,0)[b]{\smash{\fontsize{9}{10.8}
\usefont{T1}{cmr}{m}{n}{\color[rgb]{0,0,0}$believer$}%
}}}
\put(3901,-5836){\makebox(0,0)[b]{\smash{\fontsize{9}{10.8}
\usefont{T1}{cmr}{m}{n}{\color[rgb]{0,0,0}$equating$}%
}}}
{\color[rgb]{0,0,0}\multiput(6751,-4411)(12.93103,-5.17241){4}{\makebox(2.2222,15.5556){\tiny.}}
\multiput(6790,-4426)(13.62070,-5.44828){4}{\makebox(2.2222,15.5556){\tiny.}}
\multiput(6831,-4442)(13.80000,-4.60000){4}{\makebox(2.2222,15.5556){\tiny.}}
\multiput(6872,-4457)(10.80000,-3.60000){5}{\makebox(2.2222,15.5556){\tiny.}}
\multiput(6915,-4472)(11.25000,-3.75000){5}{\makebox(2.2222,15.5556){\tiny.}}
\multiput(6960,-4487)(11.70000,-3.90000){5}{\makebox(2.2222,15.5556){\tiny.}}
\multiput(7007,-4502)(11.92500,-3.97500){5}{\makebox(2.2222,15.5556){\tiny.}}
\multiput(7055,-4517)(12.60000,-4.20000){5}{\makebox(2.2222,15.5556){\tiny.}}
\multiput(7106,-4532)(10.82352,-2.70588){6}{\makebox(2.2222,15.5556){\tiny.}}
\multiput(7160,-4546)(11.01176,-2.75294){6}{\makebox(2.2222,15.5556){\tiny.}}
\multiput(7215,-4560)(11.62352,-2.90588){6}{\makebox(2.2222,15.5556){\tiny.}}
\multiput(7273,-4575)(11.95294,-2.98823){6}{\makebox(2.2222,15.5556){\tiny.}}
\multiput(7333,-4589)(12.32942,-3.08236){6}{\makebox(2.2222,15.5556){\tiny.}}
\multiput(7395,-4603)(12.80770,-2.56154){6}{\makebox(2.2222,15.5556){\tiny.}}
\multiput(7459,-4616)(11.02563,-2.20513){7}{\makebox(2.2222,15.5556){\tiny.}}
\multiput(7525,-4630)(11.34615,-2.26923){7}{\makebox(2.2222,15.5556){\tiny.}}
\multiput(7593,-4644)(11.47437,-2.29487){7}{\makebox(2.2222,15.5556){\tiny.}}
\multiput(7662,-4657)(11.50642,-2.30128){7}{\makebox(2.2222,15.5556){\tiny.}}
\multiput(7731,-4671)(11.86487,-1.97748){7}{\makebox(2.2222,15.5556){\tiny.}}
\multiput(7802,-4684)(11.82692,-2.36538){7}{\makebox(2.2222,15.5556){\tiny.}}
\multiput(7873,-4698)(12.02703,-2.00451){7}{\makebox(2.2222,15.5556){\tiny.}}
\multiput(7945,-4711)(11.82692,-2.36538){7}{\makebox(2.2222,15.5556){\tiny.}}
\multiput(8016,-4725)(11.86487,-1.97748){7}{\makebox(2.2222,15.5556){\tiny.}}
\multiput(8087,-4738)(11.63462,-2.32692){7}{\makebox(2.2222,15.5556){\tiny.}}
\multiput(8157,-4751)(11.50642,-2.30128){7}{\makebox(2.2222,15.5556){\tiny.}}
\multiput(8226,-4765)(11.15385,-2.23077){7}{\makebox(2.2222,15.5556){\tiny.}}
\multiput(8293,-4778)(11.02563,-2.20513){7}{\makebox(2.2222,15.5556){\tiny.}}
\multiput(8359,-4792)(12.65384,-2.53077){6}{\makebox(2.2222,15.5556){\tiny.}}
\multiput(8422,-4806)(12.03846,-2.40769){6}{\makebox(2.2222,15.5556){\tiny.}}
\multiput(8482,-4819)(11.38824,-2.84706){6}{\makebox(2.2222,15.5556){\tiny.}}
\multiput(8539,-4833)(10.87058,-2.71765){6}{\makebox(2.2222,15.5556){\tiny.}}
\multiput(8593,-4848)(12.58822,-3.14706){5}{\makebox(2.2222,15.5556){\tiny.}}
\multiput(8643,-4862)(11.40000,-3.80000){5}{\makebox(2.2222,15.5556){\tiny.}}
\multiput(8689,-4876)(10.57500,-3.52500){5}{\makebox(2.2222,15.5556){\tiny.}}
\multiput(8731,-4891)(12.06897,-4.82759){4}{\makebox(2.2222,15.5556){\tiny.}}
\multiput(8767,-4906)(10.53333,-5.26667){4}{\makebox(2.2222,15.5556){\tiny.}}
\multiput(8799,-4921)(12.86765,-7.72059){3}{\makebox(2.2222,15.5556){\tiny.}}
\multiput(8825,-4936)(10.00000,-8.00000){3}{\makebox(2.2222,15.5556){\tiny.}}
\multiput(8845,-4952)(14.50000,-14.50000){2}{\makebox(2.2222,15.5556){\tiny.}}
\multiput(8859,-4967)(8.40000,-16.80000){2}{\makebox(2.2222,15.5556){\tiny.}}
\multiput(8867,-4984)(2.64860,-15.89160){2}{\makebox(2.2222,15.5556){\tiny.}}
\multiput(8869,-5000)(-5.60000,-16.80000){2}{\makebox(2.2222,15.5556){\tiny.}}
\multiput(8864,-5017)(-12.00000,-16.00000){2}{\makebox(2.2222,15.5556){\tiny.}}
\multiput(8852,-5033)(-9.00000,-9.00000){3}{\makebox(2.2222,15.5556){\tiny.}}
\multiput(8834,-5051)(-12.92310,-8.61540){3}{\makebox(2.2222,15.5556){\tiny.}}
\multiput(8808,-5068)(-10.49020,-6.29412){4}{\makebox(2.2222,15.5556){\tiny.}}
\multiput(8776,-5086)(-12.00000,-6.00000){3}{\makebox(2.2222,15.5556){\tiny.}}
\multiput(8752,-5098)(-13.96550,-5.58620){3}{\makebox(2.2222,15.5556){\tiny.}}
\multiput(8724,-5109)(-15.00000,-6.00000){3}{\makebox(2.2222,15.5556){\tiny.}}
\multiput(8694,-5121)(-11.10000,-3.70000){4}{\makebox(2.2222,15.5556){\tiny.}}
\multiput(8661,-5133)(-12.40000,-4.13333){4}{\makebox(2.2222,15.5556){\tiny.}}
\multiput(8624,-5146)(-13.30000,-4.43333){4}{\makebox(2.2222,15.5556){\tiny.}}
\multiput(8584,-5159)(-10.65000,-3.55000){5}{\makebox(2.2222,15.5556){\tiny.}}
\multiput(8541,-5172)(-11.58822,-2.89706){5}{\makebox(2.2222,15.5556){\tiny.}}
\multiput(8495,-5185)(-12.58822,-3.14706){5}{\makebox(2.2222,15.5556){\tiny.}}
\multiput(8445,-5199)(-13.35295,-3.33824){5}{\makebox(2.2222,15.5556){\tiny.}}
\multiput(8392,-5214)(-11.20000,-2.80000){6}{\makebox(2.2222,15.5556){\tiny.}}
\multiput(8336,-5228)(-11.85882,-2.96470){6}{\makebox(2.2222,15.5556){\tiny.}}
\multiput(8277,-5244)(-12.37648,-3.09412){6}{\makebox(2.2222,15.5556){\tiny.}}
\multiput(8215,-5259)(-10.98040,-2.74510){7}{\makebox(2.2222,15.5556){\tiny.}}
\multiput(8149,-5275)(-11.49020,-2.87255){7}{\makebox(2.2222,15.5556){\tiny.}}
\multiput(8080,-5292)(-11.80392,-2.95098){7}{\makebox(2.2222,15.5556){\tiny.}}
\multiput(8009,-5309)(-12.47058,-3.11765){7}{\makebox(2.2222,15.5556){\tiny.}}
\multiput(7934,-5327)(-10.95799,-2.73950){8}{\makebox(2.2222,15.5556){\tiny.}}
\multiput(7857,-5345)(-11.36134,-2.84034){8}{\makebox(2.2222,15.5556){\tiny.}}
\multiput(7777,-5363)(-11.79831,-2.94958){8}{\makebox(2.2222,15.5556){\tiny.}}
\multiput(7694,-5382)(-12.19780,-2.43956){8}{\makebox(2.2222,15.5556){\tiny.}}
\multiput(7609,-5401)(-10.82353,-2.70588){9}{\makebox(2.2222,15.5556){\tiny.}}
\multiput(7522,-5421)(-11.27404,-2.25481){9}{\makebox(2.2222,15.5556){\tiny.}}
\multiput(7432,-5440)(-11.44117,-2.86029){9}{\makebox(2.2222,15.5556){\tiny.}}
\multiput(7340,-5461)(-11.77885,-2.35577){9}{\makebox(2.2222,15.5556){\tiny.}}
\multiput(7246,-5481)(-12.04327,-2.40865){9}{\makebox(2.2222,15.5556){\tiny.}}
\multiput(7150,-5502)(-12.16346,-2.43269){9}{\makebox(2.2222,15.5556){\tiny.}}
\multiput(7053,-5523)(-11.02564,-2.20513){10}{\makebox(2.2222,15.5556){\tiny.}}
\multiput(6954,-5544)(-11.15384,-2.23077){10}{\makebox(2.2222,15.5556){\tiny.}}
\multiput(6854,-5566)(-11.34616,-2.26923){10}{\makebox(2.2222,15.5556){\tiny.}}
\multiput(6752,-5587)(-11.36752,-2.27350){10}{\makebox(2.2222,15.5556){\tiny.}}
\multiput(6650,-5609)(-11.58120,-2.31624){10}{\makebox(2.2222,15.5556){\tiny.}}
\multiput(6546,-5631)(-11.58120,-2.31624){10}{\makebox(2.2222,15.5556){\tiny.}}
\multiput(6442,-5653)(-11.55983,-2.31197){10}{\makebox(2.2222,15.5556){\tiny.}}
\multiput(6338,-5674)(-11.68803,-2.33761){10}{\makebox(2.2222,15.5556){\tiny.}}
\multiput(6233,-5696)(-11.68803,-2.33761){10}{\makebox(2.2222,15.5556){\tiny.}}
\multiput(6128,-5718)(-11.77350,-2.35470){10}{\makebox(2.2222,15.5556){\tiny.}}
\multiput(6022,-5739)(-11.66667,-2.33333){10}{\makebox(2.2222,15.5556){\tiny.}}
\multiput(5917,-5760)(-11.55983,-2.31197){10}{\makebox(2.2222,15.5556){\tiny.}}
\multiput(5813,-5781)(-11.66667,-2.33333){10}{\makebox(2.2222,15.5556){\tiny.}}
\multiput(5708,-5802)(-11.43162,-2.28632){10}{\makebox(2.2222,15.5556){\tiny.}}
\multiput(5605,-5822)(-11.43162,-2.28632){10}{\makebox(2.2222,15.5556){\tiny.}}
\multiput(5502,-5842)(-11.21794,-2.24359){10}{\makebox(2.2222,15.5556){\tiny.}}
\multiput(5401,-5862)(-11.19658,-2.23932){10}{\makebox(2.2222,15.5556){\tiny.}}
\multiput(5300,-5881)(-10.98291,-2.19658){10}{\makebox(2.2222,15.5556){\tiny.}}
\multiput(5201,-5900)(-12.21154,-2.44231){9}{\makebox(2.2222,15.5556){\tiny.}}
\multiput(5103,-5918)(-12.02027,-2.00338){9}{\makebox(2.2222,15.5556){\tiny.}}
\multiput(5007,-5935)(-11.77703,-1.96284){9}{\makebox(2.2222,15.5556){\tiny.}}
\multiput(4913,-5952)(-11.51351,-1.91892){9}{\makebox(2.2222,15.5556){\tiny.}}
\multiput(4821,-5968)(-11.25000,-1.87500){9}{\makebox(2.2222,15.5556){\tiny.}}
\multiput(4731,-5983)(-11.12838,-1.85473){9}{\makebox(2.2222,15.5556){\tiny.}}
\multiput(4642,-5998)(-12.13900,-2.02317){8}{\makebox(2.2222,15.5556){\tiny.}}
\multiput(4557,-6012)(-11.83784,-1.97297){8}{\makebox(2.2222,15.5556){\tiny.}}
\multiput(4474,-6025)(-11.53669,-1.92278){8}{\makebox(2.2222,15.5556){\tiny.}}
\multiput(4393,-6037)(-11.09653,-1.84942){8}{\makebox(2.2222,15.5556){\tiny.}}
\multiput(4315,-6048)(-12.43243,-2.07207){7}{\makebox(2.2222,15.5556){\tiny.}}
\multiput(4240,-6058)(-12.08108,-2.01351){7}{\makebox(2.2222,15.5556){\tiny.}}
\multiput(4167,-6067)(-11.40540,-1.90090){7}{\makebox(2.2222,15.5556){\tiny.}}
\multiput(4098,-6075)(-11.05405,-1.84234){7}{\makebox(2.2222,15.5556){\tiny.}}
\put(4031,-6082){\line(-1, 0){ 63}}
\multiput(3968,-6087)(-11.83784,-1.97297){6}{\makebox(2.2222,15.5556){\tiny.}}
\put(3908,-6092){\line(-1, 0){ 58}}
\put(3850,-6095){\line(-1, 0){ 54}}
\put(3796,-6098){\line(-1, 0){ 51}}
\put(3745,-6099){\line(-1, 0){ 47}}
\put(3698,-6098){\line(-1, 0){ 45}}
\put(3653,-6097){\line(-1, 0){ 41}}
\multiput(3612,-6094)(-12.54053,2.09009){4}{\makebox(2.2222,15.5556){\tiny.}}
\multiput(3574,-6090)(-11.62163,1.93694){4}{\makebox(2.2222,15.5556){\tiny.}}
\multiput(3539,-6085)(-16.05770,3.21154){3}{\makebox(2.2222,15.5556){\tiny.}}
\multiput(3507,-6078)(-15.05880,3.76470){3}{\makebox(2.2222,15.5556){\tiny.}}
\multiput(3477,-6070)(-13.05000,4.35000){3}{\makebox(2.2222,15.5556){\tiny.}}
\multiput(3451,-6061)(-10.13333,5.06667){4}{\makebox(2.2222,15.5556){\tiny.}}
\multiput(3420,-6047)(-12.92310,8.61540){3}{\makebox(2.2222,15.5556){\tiny.}}
\multiput(3394,-6030)(-10.86885,9.05738){3}{\makebox(2.2222,15.5556){\tiny.}}
\multiput(3373,-6011)(-8.34000,11.12000){3}{\makebox(2.2222,15.5556){\tiny.}}
\multiput(3356,-5989)(-6.20000,12.40000){3}{\makebox(2.2222,15.5556){\tiny.}}
\multiput(3344,-5964)(-3.38235,13.52940){3}{\makebox(2.2222,15.5556){\tiny.}}
\put(3337,-5937){\line( 0, 1){ 30}}
\put(3335,-5907){\line( 0, 1){ 32}}
\multiput(3336,-5875)(1.94593,11.67560){4}{\makebox(2.2222,15.5556){\tiny.}}
\multiput(3342,-5840)(3.19607,12.78427){4}{\makebox(2.2222,15.5556){\tiny.}}
\multiput(3353,-5802)(4.46667,13.40000){4}{\makebox(2.2222,15.5556){\tiny.}}
\multiput(3367,-5762)(4.32758,10.81894){5}{\makebox(2.2222,15.5556){\tiny.}}
\multiput(3385,-5719)(5.60000,11.20000){5}{\makebox(2.2222,15.5556){\tiny.}}
\multiput(3407,-5674)(5.95000,11.90000){5}{\makebox(2.2222,15.5556){\tiny.}}
\multiput(3432,-5627)(5.85882,9.76470){6}{\makebox(2.2222,15.5556){\tiny.}}
\multiput(3461,-5578)(6.19412,10.32353){6}{\makebox(2.2222,15.5556){\tiny.}}
\multiput(3493,-5527)(6.56470,10.94117){6}{\makebox(2.2222,15.5556){\tiny.}}
\multiput(3527,-5473)(6.17948,9.26923){7}{\makebox(2.2222,15.5556){\tiny.}}
\multiput(3565,-5418)(6.88000,9.17333){7}{\makebox(2.2222,15.5556){\tiny.}}
\multiput(3605,-5362)(7.16000,9.54667){7}{\makebox(2.2222,15.5556){\tiny.}}
\multiput(3647,-5304)(7.50000,10.00000){7}{\makebox(2.2222,15.5556){\tiny.}}
\multiput(3692,-5244)(7.56000,10.08000){7}{\makebox(2.2222,15.5556){\tiny.}}
\multiput(3738,-5184)(8.08130,10.10163){7}{\makebox(2.2222,15.5556){\tiny.}}
\multiput(3786,-5123)(7.05227,8.81534){8}{\makebox(2.2222,15.5556){\tiny.}}
\multiput(3835,-5061)(7.23344,9.04180){8}{\makebox(2.2222,15.5556){\tiny.}}
\multiput(3886,-4998)(7.47073,8.96487){8}{\makebox(2.2222,15.5556){\tiny.}}
\multiput(3938,-4935)(7.47073,8.96487){8}{\makebox(2.2222,15.5556){\tiny.}}
\multiput(3990,-4872)(7.59953,9.11943){8}{\makebox(2.2222,15.5556){\tiny.}}
\multiput(4043,-4808)(7.58783,9.10539){8}{\makebox(2.2222,15.5556){\tiny.}}
\multiput(4097,-4745)(7.52927,9.03513){8}{\makebox(2.2222,15.5556){\tiny.}}
\multiput(4150,-4682)(7.51757,9.02109){8}{\makebox(2.2222,15.5556){\tiny.}}
\multiput(4204,-4620)(7.38876,8.86651){8}{\makebox(2.2222,15.5556){\tiny.}}
\multiput(4257,-4559)(7.38876,8.86651){8}{\makebox(2.2222,15.5556){\tiny.}}
\multiput(4310,-4498)(7.31850,8.78220){8}{\makebox(2.2222,15.5556){\tiny.}}
\multiput(4363,-4438)(7.11944,8.54333){8}{\makebox(2.2222,15.5556){\tiny.}}
\multiput(4415,-4380)(8.15573,9.78688){7}{\makebox(2.2222,15.5556){\tiny.}}
\multiput(4466,-4323)(7.92350,9.50820){7}{\makebox(2.2222,15.5556){\tiny.}}
\multiput(4516,-4268)(7.77322,9.32786){7}{\makebox(2.2222,15.5556){\tiny.}}
\multiput(4565,-4214)(8.33333,8.33333){7}{\makebox(2.2222,15.5556){\tiny.}}
\multiput(4613,-4162)(8.08333,8.08333){7}{\makebox(2.2222,15.5556){\tiny.}}
\multiput(4660,-4112)(9.30000,9.30000){6}{\makebox(2.2222,15.5556){\tiny.}}
\multiput(4705,-4064)(8.90000,8.90000){6}{\makebox(2.2222,15.5556){\tiny.}}
\multiput(4749,-4019)(8.60000,8.60000){6}{\makebox(2.2222,15.5556){\tiny.}}
\multiput(4791,-3975)(8.20000,8.20000){6}{\makebox(2.2222,15.5556){\tiny.}}
\multiput(4832,-3934)(9.75000,9.75000){5}{\makebox(2.2222,15.5556){\tiny.}}
\multiput(4872,-3896)(9.25000,9.25000){5}{\makebox(2.2222,15.5556){\tiny.}}
\multiput(4910,-3860)(8.87500,8.87500){5}{\makebox(2.2222,15.5556){\tiny.}}
\multiput(4947,-3826)(9.12295,7.60246){5}{\makebox(2.2222,15.5556){\tiny.}}
\multiput(4983,-3795)(8.58197,7.15165){5}{\makebox(2.2222,15.5556){\tiny.}}
\multiput(5017,-3766)(11.04000,8.28000){4}{\makebox(2.2222,15.5556){\tiny.}}
\multiput(5050,-3741)(10.88000,8.16000){4}{\makebox(2.2222,15.5556){\tiny.}}
\multiput(5083,-3717)(10.38460,6.92307){4}{\makebox(2.2222,15.5556){\tiny.}}
\multiput(5114,-3696)(10.39217,6.23530){4}{\makebox(2.2222,15.5556){\tiny.}}
\multiput(5145,-3677)(10.40000,5.20000){4}{\makebox(2.2222,15.5556){\tiny.}}
\multiput(5176,-3661)(10.10000,5.05000){5}{\makebox(2.2222,15.5556){\tiny.}}
\multiput(5217,-3642)(13.70000,4.56667){4}{\makebox(2.2222,15.5556){\tiny.}}
\multiput(5258,-3628)(13.72550,3.43137){4}{\makebox(2.2222,15.5556){\tiny.}}
\multiput(5299,-3617)(13.35137,2.22523){4}{\makebox(2.2222,15.5556){\tiny.}}
\multiput(5339,-3610)(13.18920,2.19820){4}{\makebox(2.2222,15.5556){\tiny.}}
\put(5379,-3606){\line( 1, 0){ 40}}
\multiput(5419,-3607)(12.86487,-2.14414){4}{\makebox(2.2222,15.5556){\tiny.}}
\multiput(5458,-3611)(13.35137,-2.22523){4}{\makebox(2.2222,15.5556){\tiny.}}
\multiput(5498,-3618)(12.70587,-3.17647){4}{\makebox(2.2222,15.5556){\tiny.}}
\multiput(5536,-3628)(13.10000,-4.36667){4}{\makebox(2.2222,15.5556){\tiny.}}
\multiput(5575,-3642)(12.75863,-5.10345){4}{\makebox(2.2222,15.5556){\tiny.}}
\multiput(5613,-3658)(12.66667,-6.33333){4}{\makebox(2.2222,15.5556){\tiny.}}
\multiput(5651,-3677)(12.40197,-7.44118){4}{\makebox(2.2222,15.5556){\tiny.}}
\multiput(5689,-3698)(12.69607,-7.61764){4}{\makebox(2.2222,15.5556){\tiny.}}
\multiput(5727,-3721)(9.28845,-6.19230){5}{\makebox(2.2222,15.5556){\tiny.}}
\multiput(5764,-3746)(9.16000,-6.87000){5}{\makebox(2.2222,15.5556){\tiny.}}
\multiput(5801,-3773)(9.28000,-6.96000){5}{\makebox(2.2222,15.5556){\tiny.}}
\multiput(5838,-3801)(9.17682,-7.34146){5}{\makebox(2.2222,15.5556){\tiny.}}
\multiput(5875,-3830)(9.27050,-7.72542){5}{\makebox(2.2222,15.5556){\tiny.}}
\multiput(5912,-3861)(9.27050,-7.72542){5}{\makebox(2.2222,15.5556){\tiny.}}
\multiput(5949,-3892)(9.27050,-7.72542){5}{\makebox(2.2222,15.5556){\tiny.}}
\multiput(5986,-3923)(9.54097,-7.95081){5}{\makebox(2.2222,15.5556){\tiny.}}
\multiput(6024,-3955)(9.39345,-7.82787){5}{\makebox(2.2222,15.5556){\tiny.}}
\multiput(6061,-3987)(9.39345,-7.82787){5}{\makebox(2.2222,15.5556){\tiny.}}
\multiput(6098,-4019)(9.57318,-7.65854){5}{\makebox(2.2222,15.5556){\tiny.}}
\multiput(6136,-4050)(9.57318,-7.65854){5}{\makebox(2.2222,15.5556){\tiny.}}
\multiput(6174,-4081)(9.45123,-7.56098){5}{\makebox(2.2222,15.5556){\tiny.}}
\multiput(6212,-4111)(9.72000,-7.29000){5}{\makebox(2.2222,15.5556){\tiny.}}
\multiput(6251,-4140)(9.72000,-7.29000){5}{\makebox(2.2222,15.5556){\tiny.}}
\multiput(6290,-4169)(9.86538,-6.57692){5}{\makebox(2.2222,15.5556){\tiny.}}
\multiput(6329,-4196)(9.92308,-6.61538){5}{\makebox(2.2222,15.5556){\tiny.}}
\multiput(6369,-4222)(10.11030,-6.06618){5}{\makebox(2.2222,15.5556){\tiny.}}
\multiput(6409,-4247)(10.18383,-6.11030){5}{\makebox(2.2222,15.5556){\tiny.}}
\multiput(6450,-4271)(10.07353,-6.04411){5}{\makebox(2.2222,15.5556){\tiny.}}
\multiput(6491,-4294)(10.60000,-5.30000){5}{\makebox(2.2222,15.5556){\tiny.}}
\multiput(6533,-4316)(10.60000,-5.30000){5}{\makebox(2.2222,15.5556){\tiny.}}
\multiput(6576,-4336)(10.60000,-5.30000){5}{\makebox(2.2222,15.5556){\tiny.}}
\multiput(6619,-4356)(10.90517,-4.36207){5}{\makebox(2.2222,15.5556){\tiny.}}
\multiput(6662,-4375)(11.03448,-4.41379){5}{\makebox(2.2222,15.5556){\tiny.}}
\multiput(6706,-4393)(11.25000,-4.50000){5}{\makebox(2.2222,15.5556){\tiny.}}
}%
\end{picture}%
\nop{
             fully plastic
           /     ^        \                              \space
         /       |          \                            \space
       /         |            \                          \space
      |          |             |
      V          |             V
   amnesic ------+          plastic
      |            \     /     |     \                   \space
      |              \ /       |       \                 \space
      |              / \       |         \               \space
      |    +-------------+     |          |
      |    |        V    | \   V          V
      |    |    dogmatic | learnable   damascan
      |    |        |    |                |
      |    |        |    +----------------------+
      |    |        |                     |     |
      |    |        V                     V     |
      |    |    believer ----------> correcting |
      |    |   /                                |
  +--------+ /                                  |
  |   |    /                                    |
  |   V   V                                     |
  | equating                                    |
  |                                             |
  +---------------------------------------------+
}
\hcaption{Plain severe and full meet revisions}
\label{figure-relations-fullmeet}
\end{hfigure}

Plain severe and full meet revisions are not learnable by
Theorem~\ref{theorem-plainsevere-learnable} and
Theorem~\ref{theorem-fullmeet-learnable}.
They are not amnesic by
Theorem~\ref{theorem-plainsevere-amnesic} and
Theorem~\ref{theorem-fullmeet-amnesic}.
They are not damascan by
Theorem~\ref{theorem-plainsevere-damascan} and
Theorem~\ref{theorem-fullmeet-damascan}.
They are dogmatic by
Theorem~\ref{theorem-plainsevere-dogmatic} and
Theorem~\ref{theorem-fullmeet-dogmatic}.

\begin{hfigure}
\setlength{\unitlength}{3750sp}%
\begin{picture}(5942,3835)(3174,-6136)
\thinlines
{\color[rgb]{0,0,0}\put(6274,-2572){\vector( 4,-3){552}}
}%
{\color[rgb]{0,0,0}\put(5341,-2491){\vector(-2,-1){1140}}
}%
{\color[rgb]{0,0,0}\put(3901,-3286){\vector( 0,-1){2400}}
}%
{\color[rgb]{0,0,0}\put(6526,-3211){\vector(-3,-2){1125}}
}%
{\color[rgb]{0,0,0}\put(6901,-3286){\vector( 0,-1){675}}
}%
{\color[rgb]{0,0,0}\put(7090,-3259){\vector( 4,-3){936}}
}%
{\color[rgb]{0,0,0}\put(8101,-4186){\vector( 0,-1){675}}
}%
{\color[rgb]{0,0,0}\put(5401,-4186){\vector( 0,-1){675}}
}%
{\color[rgb]{0,0,0}\put(4351,-3136){\line( 5, 1){1125}}
}%
{\color[rgb]{0,0,0}\put(6526,-3961){\line(-1, 1){1050}}
\multiput(5476,-2911)(5.00000,10.00000){31}{\makebox(2.2222,15.5556){\tiny.}}
\put(5626,-2611){\vector( 1, 2){0}}
}%
{\color[rgb]{0,0,0}\put(5776,-4936){\vector( 1, 0){1875}}
}%
{\color[rgb]{0,0,0}\put(5026,-5086){\vector(-3,-2){900}}
}%
\put(5851,-2461){\makebox(0,0)[b]{\smash{\fontsize{9}{10.8}
\usefont{T1}{cmr}{m}{n}{\color[rgb]{0,0,0}$fully\ plastic$}%
}}}
\put(3901,-3211){\makebox(0,0)[b]{\smash{\fontsize{9}{10.8}
\usefont{T1}{cmr}{m}{n}{\color[rgb]{0,0,0}$amnesic$}%
}}}
\put(6901,-4111){\makebox(0,0)[b]{\smash{\fontsize{9}{10.8}
\usefont{T1}{cmr}{m}{n}{\color[rgb]{0,0,0}$learnable$}%
}}}
\put(6826,-3211){\makebox(0,0)[b]{\smash{\fontsize{9}{10.8}
\usefont{T1}{cmr}{m}{n}{\color[rgb]{0,0,0}$plastic$}%
}}}
\put(5401,-4111){\makebox(0,0)[b]{\smash{\fontsize{9}{10.8}
\usefont{T1}{cmr}{m}{n}{\color[rgb]{0,0,0}$dogmatic$}%
}}}
\put(8101,-4111){\makebox(0,0)[b]{\smash{\fontsize{9}{10.8}
\usefont{T1}{cmr}{m}{n}{\color[rgb]{0,0,0}$damascan$}%
}}}
\put(8101,-5011){\makebox(0,0)[b]{\smash{\fontsize{9}{10.8}
\usefont{T1}{cmr}{m}{n}{\color[rgb]{0,0,0}$correcting$}%
}}}
\put(5401,-5011){\makebox(0,0)[b]{\smash{\fontsize{9}{10.8}
\usefont{T1}{cmr}{m}{n}{\color[rgb]{0,0,0}$believer$}%
}}}
\put(3901,-5836){\makebox(0,0)[b]{\smash{\fontsize{9}{10.8}
\usefont{T1}{cmr}{m}{n}{\color[rgb]{0,0,0}$equating$}%
}}}
{\color[rgb]{0,0,0}\multiput(6076,-4036)(12.58620,-5.03448){5}{\makebox(2.2222,15.5556){\tiny.}}
\multiput(6126,-4057)(12.58620,-5.03448){5}{\makebox(2.2222,15.5556){\tiny.}}
\multiput(6176,-4078)(12.80172,-5.12069){5}{\makebox(2.2222,15.5556){\tiny.}}
\multiput(6227,-4099)(10.34482,-4.13793){6}{\makebox(2.2222,15.5556){\tiny.}}
\multiput(6279,-4119)(10.41380,-4.16552){6}{\makebox(2.2222,15.5556){\tiny.}}
\multiput(6331,-4140)(10.75862,-4.30345){6}{\makebox(2.2222,15.5556){\tiny.}}
\multiput(6385,-4161)(11.28000,-3.76000){6}{\makebox(2.2222,15.5556){\tiny.}}
\multiput(6441,-4181)(11.27586,-4.51034){6}{\makebox(2.2222,15.5556){\tiny.}}
\multiput(6498,-4202)(11.70000,-3.90000){6}{\makebox(2.2222,15.5556){\tiny.}}
\multiput(6556,-4223)(12.00000,-4.00000){6}{\makebox(2.2222,15.5556){\tiny.}}
\multiput(6616,-4243)(12.42000,-4.14000){6}{\makebox(2.2222,15.5556){\tiny.}}
\multiput(6678,-4264)(10.60000,-3.53333){7}{\makebox(2.2222,15.5556){\tiny.}}
\multiput(6742,-4284)(10.80000,-3.60000){7}{\makebox(2.2222,15.5556){\tiny.}}
\multiput(6807,-4305)(10.95000,-3.65000){7}{\makebox(2.2222,15.5556){\tiny.}}
\multiput(6873,-4326)(11.20000,-3.73333){7}{\makebox(2.2222,15.5556){\tiny.}}
\multiput(6941,-4346)(11.55000,-3.85000){7}{\makebox(2.2222,15.5556){\tiny.}}
\multiput(7011,-4367)(11.76470,-2.94117){7}{\makebox(2.2222,15.5556){\tiny.}}
\multiput(7081,-4387)(11.85000,-3.95000){7}{\makebox(2.2222,15.5556){\tiny.}}
\multiput(7153,-4408)(12.23530,-3.05883){7}{\makebox(2.2222,15.5556){\tiny.}}
\multiput(7226,-4428)(12.27452,-3.06863){7}{\makebox(2.2222,15.5556){\tiny.}}
\multiput(7299,-4449)(12.54902,-3.13725){7}{\makebox(2.2222,15.5556){\tiny.}}
\multiput(7374,-4469)(12.43137,-3.10784){7}{\makebox(2.2222,15.5556){\tiny.}}
\multiput(7448,-4490)(12.54902,-3.13725){7}{\makebox(2.2222,15.5556){\tiny.}}
\multiput(7523,-4510)(10.78991,-2.69748){8}{\makebox(2.2222,15.5556){\tiny.}}
\multiput(7598,-4531)(12.54902,-3.13725){7}{\makebox(2.2222,15.5556){\tiny.}}
\multiput(7673,-4551)(12.43137,-3.10784){7}{\makebox(2.2222,15.5556){\tiny.}}
\multiput(7747,-4572)(12.39215,-3.09804){7}{\makebox(2.2222,15.5556){\tiny.}}
\multiput(7821,-4592)(12.07843,-3.01961){7}{\makebox(2.2222,15.5556){\tiny.}}
\multiput(7893,-4612)(11.85000,-3.95000){7}{\makebox(2.2222,15.5556){\tiny.}}
\multiput(7965,-4633)(11.60785,-2.90196){7}{\makebox(2.2222,15.5556){\tiny.}}
\multiput(8034,-4653)(11.60785,-2.90196){7}{\makebox(2.2222,15.5556){\tiny.}}
\multiput(8103,-4673)(10.95000,-3.65000){7}{\makebox(2.2222,15.5556){\tiny.}}
\multiput(8169,-4694)(12.54000,-4.18000){6}{\makebox(2.2222,15.5556){\tiny.}}
\multiput(8232,-4714)(12.36000,-4.12000){6}{\makebox(2.2222,15.5556){\tiny.}}
\multiput(8294,-4734)(11.64000,-3.88000){6}{\makebox(2.2222,15.5556){\tiny.}}
\multiput(8352,-4754)(11.28000,-3.76000){6}{\makebox(2.2222,15.5556){\tiny.}}
\multiput(8408,-4774)(10.34482,-4.13793){6}{\makebox(2.2222,15.5556){\tiny.}}
\multiput(8460,-4794)(12.06898,-4.82759){5}{\makebox(2.2222,15.5556){\tiny.}}
\multiput(8508,-4814)(11.42243,-4.56897){5}{\makebox(2.2222,15.5556){\tiny.}}
\multiput(8553,-4834)(10.20000,-5.10000){5}{\makebox(2.2222,15.5556){\tiny.}}
\multiput(8594,-4854)(11.76470,-7.05882){4}{\makebox(2.2222,15.5556){\tiny.}}
\multiput(8630,-4874)(10.78430,-6.47058){4}{\makebox(2.2222,15.5556){\tiny.}}
\multiput(8662,-4894)(9.38460,-6.25640){4}{\makebox(2.2222,15.5556){\tiny.}}
\multiput(8690,-4913)(11.40985,-9.50821){3}{\makebox(2.2222,15.5556){\tiny.}}
\multiput(8712,-4933)(9.25000,-9.25000){3}{\makebox(2.2222,15.5556){\tiny.}}
\multiput(8730,-4952)(6.00000,-10.00000){3}{\makebox(2.2222,15.5556){\tiny.}}
\multiput(8742,-4972)(7.51720,-18.79300){2}{\makebox(2.2222,15.5556){\tiny.}}
\multiput(8749,-4991)(3.13510,-18.81060){2}{\makebox(2.2222,15.5556){\tiny.}}
\multiput(8751,-5010)(-3.80770,-19.03850){2}{\makebox(2.2222,15.5556){\tiny.}}
\multiput(8747,-5029)(-9.60000,-19.20000){2}{\makebox(2.2222,15.5556){\tiny.}}
\multiput(8737,-5048)(-7.56100,-9.45125){3}{\makebox(2.2222,15.5556){\tiny.}}
\multiput(8722,-5067)(-10.86885,-9.05738){3}{\makebox(2.2222,15.5556){\tiny.}}
\multiput(8701,-5086)(-9.44000,-7.08000){3}{\makebox(2.2222,15.5556){\tiny.}}
\multiput(8682,-5100)(-11.54410,-6.92646){3}{\makebox(2.2222,15.5556){\tiny.}}
\multiput(8659,-5114)(-13.20000,-6.60000){3}{\makebox(2.2222,15.5556){\tiny.}}
\multiput(8633,-5128)(-14.40000,-7.20000){3}{\makebox(2.2222,15.5556){\tiny.}}
\multiput(8604,-5142)(-10.80460,-4.32184){4}{\makebox(2.2222,15.5556){\tiny.}}
\multiput(8572,-5156)(-12.06897,-4.82759){4}{\makebox(2.2222,15.5556){\tiny.}}
\multiput(8536,-5171)(-13.40000,-4.46667){4}{\makebox(2.2222,15.5556){\tiny.}}
\multiput(8496,-5185)(-10.57500,-3.52500){5}{\makebox(2.2222,15.5556){\tiny.}}
\multiput(8454,-5200)(-11.77500,-3.92500){5}{\makebox(2.2222,15.5556){\tiny.}}
\multiput(8407,-5216)(-12.22500,-4.07500){5}{\makebox(2.2222,15.5556){\tiny.}}
\multiput(8358,-5232)(-13.12500,-4.37500){5}{\makebox(2.2222,15.5556){\tiny.}}
\multiput(8305,-5248)(-11.29412,-2.82353){6}{\makebox(2.2222,15.5556){\tiny.}}
\multiput(8249,-5264)(-12.09412,-3.02353){6}{\makebox(2.2222,15.5556){\tiny.}}
\multiput(8189,-5281)(-12.47058,-3.11764){6}{\makebox(2.2222,15.5556){\tiny.}}
\multiput(8127,-5298)(-11.05882,-2.76470){7}{\makebox(2.2222,15.5556){\tiny.}}
\multiput(8061,-5316)(-11.52942,-2.88235){7}{\makebox(2.2222,15.5556){\tiny.}}
\multiput(7992,-5334)(-11.84313,-2.96078){7}{\makebox(2.2222,15.5556){\tiny.}}
\multiput(7921,-5352)(-12.50980,-3.12745){7}{\makebox(2.2222,15.5556){\tiny.}}
\multiput(7846,-5371)(-10.99160,-2.74790){8}{\makebox(2.2222,15.5556){\tiny.}}
\multiput(7769,-5390)(-11.42857,-2.85714){8}{\makebox(2.2222,15.5556){\tiny.}}
\multiput(7689,-5410)(-11.69749,-2.92437){8}{\makebox(2.2222,15.5556){\tiny.}}
\multiput(7607,-5430)(-12.10084,-3.02521){8}{\makebox(2.2222,15.5556){\tiny.}}
\multiput(7522,-5450)(-12.26891,-3.06723){8}{\makebox(2.2222,15.5556){\tiny.}}
\multiput(7436,-5471)(-11.08824,-2.77206){9}{\makebox(2.2222,15.5556){\tiny.}}
\multiput(7347,-5492)(-11.32353,-2.83088){9}{\makebox(2.2222,15.5556){\tiny.}}
\multiput(7256,-5513)(-11.44117,-2.86029){9}{\makebox(2.2222,15.5556){\tiny.}}
\multiput(7164,-5534)(-11.82353,-2.95588){9}{\makebox(2.2222,15.5556){\tiny.}}
\multiput(7069,-5556)(-11.92308,-2.38462){9}{\makebox(2.2222,15.5556){\tiny.}}
\multiput(6974,-5577)(-12.05883,-3.01471){9}{\makebox(2.2222,15.5556){\tiny.}}
\multiput(6877,-5599)(-10.94017,-2.18803){10}{\makebox(2.2222,15.5556){\tiny.}}
\multiput(6779,-5621)(-11.04701,-2.20940){10}{\makebox(2.2222,15.5556){\tiny.}}
\multiput(6680,-5643)(-11.13248,-2.22650){10}{\makebox(2.2222,15.5556){\tiny.}}
\multiput(6580,-5664)(-11.15384,-2.23077){10}{\makebox(2.2222,15.5556){\tiny.}}
\multiput(6480,-5686)(-11.26069,-2.25214){10}{\makebox(2.2222,15.5556){\tiny.}}
\multiput(6379,-5708)(-11.23931,-2.24786){10}{\makebox(2.2222,15.5556){\tiny.}}
\multiput(6278,-5729)(-11.36752,-2.27350){10}{\makebox(2.2222,15.5556){\tiny.}}
\multiput(6176,-5751)(-11.23931,-2.24786){10}{\makebox(2.2222,15.5556){\tiny.}}
\multiput(6075,-5772)(-11.21794,-2.24359){10}{\makebox(2.2222,15.5556){\tiny.}}
\multiput(5974,-5792)(-11.23931,-2.24786){10}{\makebox(2.2222,15.5556){\tiny.}}
\multiput(5873,-5813)(-11.11111,-2.22222){10}{\makebox(2.2222,15.5556){\tiny.}}
\multiput(5773,-5833)(-11.08974,-2.21795){10}{\makebox(2.2222,15.5556){\tiny.}}
\multiput(5673,-5852)(-10.98291,-2.19658){10}{\makebox(2.2222,15.5556){\tiny.}}
\multiput(5574,-5871)(-12.09135,-2.41827){9}{\makebox(2.2222,15.5556){\tiny.}}
\multiput(5477,-5889)(-12.09135,-2.41827){9}{\makebox(2.2222,15.5556){\tiny.}}
\multiput(5380,-5907)(-11.89865,-1.98311){9}{\makebox(2.2222,15.5556){\tiny.}}
\multiput(5285,-5924)(-11.75676,-1.95946){9}{\makebox(2.2222,15.5556){\tiny.}}
\multiput(5191,-5940)(-11.63514,-1.93919){9}{\makebox(2.2222,15.5556){\tiny.}}
\multiput(5098,-5956)(-11.25000,-1.87500){9}{\makebox(2.2222,15.5556){\tiny.}}
\multiput(5008,-5971)(-11.08784,-1.84797){9}{\makebox(2.2222,15.5556){\tiny.}}
\multiput(4919,-5984)(-12.39383,-2.06564){8}{\makebox(2.2222,15.5556){\tiny.}}
\multiput(4832,-5997)(-12.09266,-2.01544){8}{\makebox(2.2222,15.5556){\tiny.}}
\multiput(4747,-6009)(-11.65251,-1.94209){8}{\makebox(2.2222,15.5556){\tiny.}}
\multiput(4665,-6020)(-11.46719,-1.91120){8}{\makebox(2.2222,15.5556){\tiny.}}
\multiput(4584,-6029)(-11.05020,-1.84170){8}{\makebox(2.2222,15.5556){\tiny.}}
\multiput(4506,-6038)(-12.51352,-2.08559){7}{\makebox(2.2222,15.5556){\tiny.}}
\put(4430,-6045){\line(-1, 0){ 73}}
\put(4357,-6051){\line(-1, 0){ 71}}
\put(4286,-6056){\line(-1, 0){ 68}}
\put(4218,-6059){\line(-1, 0){ 66}}
\put(4152,-6061){\line(-1, 0){ 63}}
\put(4089,-6062){\line(-1, 0){ 60}}
\put(4029,-6061){\line(-1, 0){ 58}}
\put(3971,-6059){\line(-1, 0){ 55}}
\multiput(3916,-6056)(-12.89190,2.14865){5}{\makebox(2.2222,15.5556){\tiny.}}
\multiput(3864,-6050)(-12.40540,2.06757){5}{\makebox(2.2222,15.5556){\tiny.}}
\multiput(3814,-6044)(-11.73077,2.34615){5}{\makebox(2.2222,15.5556){\tiny.}}
\multiput(3767,-6035)(-11.29808,2.25962){5}{\makebox(2.2222,15.5556){\tiny.}}
\multiput(3722,-6025)(-14.35293,3.58823){4}{\makebox(2.2222,15.5556){\tiny.}}
\multiput(3679,-6014)(-13.30000,4.43333){4}{\makebox(2.2222,15.5556){\tiny.}}
\multiput(3639,-6001)(-12.64367,5.05747){4}{\makebox(2.2222,15.5556){\tiny.}}
\multiput(3601,-5986)(-11.10000,5.55000){5}{\makebox(2.2222,15.5556){\tiny.}}
\multiput(3556,-5965)(-10.07353,6.04411){5}{\makebox(2.2222,15.5556){\tiny.}}
\multiput(3515,-5942)(-9.75000,6.50000){5}{\makebox(2.2222,15.5556){\tiny.}}
\multiput(3476,-5916)(-8.72950,7.27458){5}{\makebox(2.2222,15.5556){\tiny.}}
\multiput(3441,-5887)(-8.12500,8.12500){5}{\makebox(2.2222,15.5556){\tiny.}}
\multiput(3408,-5855)(-7.37705,8.85246){5}{\makebox(2.2222,15.5556){\tiny.}}
\multiput(3378,-5820)(-6.99000,9.32000){5}{\makebox(2.2222,15.5556){\tiny.}}
\multiput(3351,-5782)(-6.17647,10.29412){5}{\makebox(2.2222,15.5556){\tiny.}}
\multiput(3326,-5741)(-5.45000,10.90000){5}{\makebox(2.2222,15.5556){\tiny.}}
\multiput(3303,-5698)(-4.74137,11.85344){5}{\makebox(2.2222,15.5556){\tiny.}}
\multiput(3283,-5651)(-4.20000,12.60000){5}{\makebox(2.2222,15.5556){\tiny.}}
\multiput(3265,-5601)(-3.27940,13.11760){5}{\makebox(2.2222,15.5556){\tiny.}}
\multiput(3250,-5549)(-2.75294,11.01176){6}{\makebox(2.2222,15.5556){\tiny.}}
\multiput(3236,-5494)(-2.32308,11.61540){6}{\makebox(2.2222,15.5556){\tiny.}}
\multiput(3224,-5436)(-2.00000,12.00000){6}{\makebox(2.2222,15.5556){\tiny.}}
\multiput(3214,-5376)(-2.05406,12.32436){6}{\makebox(2.2222,15.5556){\tiny.}}
\multiput(3206,-5314)(-2.14594,12.87564){6}{\makebox(2.2222,15.5556){\tiny.}}
\put(3199,-5249){\line( 0, 1){ 67}}
\put(3194,-5182){\line( 0, 1){ 69}}
\put(3190,-5113){\line( 0, 1){ 70}}
\put(3188,-5043){\line( 0, 1){ 72}}
\put(3186,-4971){\line( 0, 1){ 74}}
\put(3186,-4897){\line( 0, 1){ 75}}
\put(3187,-4822){\line( 0, 1){ 75}}
\put(3189,-4747){\line( 0, 1){ 77}}
\put(3192,-4670){\line( 0, 1){ 77}}
\put(3195,-4593){\line( 0, 1){ 77}}
\put(3200,-4516){\line( 0, 1){ 78}}
\put(3205,-4438){\line( 0, 1){ 78}}
\put(3211,-4360){\line( 0, 1){ 77}}
\put(3217,-4283){\line( 0, 1){ 77}}
\multiput(3223,-4206)(2.11712,12.70270){7}{\makebox(2.2222,15.5556){\tiny.}}
\multiput(3231,-4129)(2.05857,12.35140){7}{\makebox(2.2222,15.5556){\tiny.}}
\multiput(3238,-4054)(2.06307,12.37840){7}{\makebox(2.2222,15.5556){\tiny.}}
\multiput(3246,-3979)(2.00902,12.05410){7}{\makebox(2.2222,15.5556){\tiny.}}
\multiput(3254,-3906)(1.98648,11.91890){7}{\makebox(2.2222,15.5556){\tiny.}}
\multiput(3263,-3834)(1.93243,11.59460){7}{\makebox(2.2222,15.5556){\tiny.}}
\multiput(3272,-3764)(1.87838,11.27030){7}{\makebox(2.2222,15.5556){\tiny.}}
\multiput(3281,-3696)(1.85135,11.10810){7}{\makebox(2.2222,15.5556){\tiny.}}
\multiput(3290,-3629)(2.12972,12.77832){6}{\makebox(2.2222,15.5556){\tiny.}}
\multiput(3300,-3565)(2.06486,12.38916){6}{\makebox(2.2222,15.5556){\tiny.}}
\multiput(3310,-3503)(1.96756,11.80536){6}{\makebox(2.2222,15.5556){\tiny.}}
\multiput(3320,-3444)(2.27692,11.38460){6}{\makebox(2.2222,15.5556){\tiny.}}
\multiput(3331,-3387)(2.20000,11.00000){6}{\makebox(2.2222,15.5556){\tiny.}}
\multiput(3342,-3332)(2.55770,12.78850){5}{\makebox(2.2222,15.5556){\tiny.}}
\multiput(3353,-3281)(3.05883,12.23530){5}{\makebox(2.2222,15.5556){\tiny.}}
\multiput(3365,-3232)(2.88235,11.52940){5}{\makebox(2.2222,15.5556){\tiny.}}
\multiput(3377,-3186)(4.63333,13.90000){4}{\makebox(2.2222,15.5556){\tiny.}}
\multiput(3390,-3144)(4.46667,13.40000){4}{\makebox(2.2222,15.5556){\tiny.}}
\multiput(3404,-3104)(4.89657,12.24142){4}{\makebox(2.2222,15.5556){\tiny.}}
\multiput(3418,-3067)(5.40000,10.80000){4}{\makebox(2.2222,15.5556){\tiny.}}
\multiput(3433,-3034)(5.20000,10.40000){4}{\makebox(2.2222,15.5556){\tiny.}}
\multiput(3449,-3003)(9.00000,13.50000){3}{\makebox(2.2222,15.5556){\tiny.}}
\multiput(3467,-2976)(9.24000,12.32000){3}{\makebox(2.2222,15.5556){\tiny.}}
\multiput(3485,-2951)(10.25000,10.25000){3}{\makebox(2.2222,15.5556){\tiny.}}
\multiput(3505,-2930)(10.86885,9.05738){3}{\makebox(2.2222,15.5556){\tiny.}}
\multiput(3526,-2911)(12.57690,8.38460){3}{\makebox(2.2222,15.5556){\tiny.}}
\multiput(3551,-2894)(13.20000,6.60000){3}{\makebox(2.2222,15.5556){\tiny.}}
\multiput(3577,-2880)(14.39655,5.75862){3}{\makebox(2.2222,15.5556){\tiny.}}
\multiput(3606,-2869)(15.52940,3.88235){3}{\makebox(2.2222,15.5556){\tiny.}}
\multiput(3637,-2861)(15.97295,2.66216){3}{\makebox(2.2222,15.5556){\tiny.}}
\put(3669,-2856){\line( 1, 0){ 34}}
\put(3703,-2854){\line( 1, 0){ 36}}
\put(3739,-2855){\line( 1, 0){ 38}}
\multiput(3777,-2858)(13.02703,-2.17117){4}{\makebox(2.2222,15.5556){\tiny.}}
\multiput(3816,-2865)(13.71793,-2.74359){4}{\makebox(2.2222,15.5556){\tiny.}}
\multiput(3857,-2874)(10.82353,-2.70588){5}{\makebox(2.2222,15.5556){\tiny.}}
\multiput(3900,-2886)(10.95000,-3.65000){5}{\makebox(2.2222,15.5556){\tiny.}}
\multiput(3944,-2900)(11.55000,-3.85000){5}{\makebox(2.2222,15.5556){\tiny.}}
\multiput(3990,-2916)(11.76725,-4.70690){5}{\makebox(2.2222,15.5556){\tiny.}}
\multiput(4037,-2935)(12.06898,-4.82759){5}{\makebox(2.2222,15.5556){\tiny.}}
\multiput(4085,-2955)(12.10000,-6.05000){5}{\makebox(2.2222,15.5556){\tiny.}}
\multiput(4134,-2978)(10.08000,-5.04000){6}{\makebox(2.2222,15.5556){\tiny.}}
\multiput(4185,-3002)(10.32000,-5.16000){6}{\makebox(2.2222,15.5556){\tiny.}}
\multiput(4236,-3029)(10.48000,-5.24000){6}{\makebox(2.2222,15.5556){\tiny.}}
\multiput(4288,-3056)(10.96000,-5.48000){6}{\makebox(2.2222,15.5556){\tiny.}}
\multiput(4342,-3085)(10.58824,-6.35294){6}{\makebox(2.2222,15.5556){\tiny.}}
\multiput(4396,-3115)(10.91176,-6.54706){6}{\makebox(2.2222,15.5556){\tiny.}}
\multiput(4451,-3147)(10.91176,-6.54706){6}{\makebox(2.2222,15.5556){\tiny.}}
\multiput(4506,-3179)(11.14706,-6.68824){6}{\makebox(2.2222,15.5556){\tiny.}}
\multiput(4562,-3212)(11.23530,-6.74118){6}{\makebox(2.2222,15.5556){\tiny.}}
\multiput(4618,-3246)(11.23530,-6.74118){6}{\makebox(2.2222,15.5556){\tiny.}}
\multiput(4674,-3280)(9.55882,-5.73529){7}{\makebox(2.2222,15.5556){\tiny.}}
\multiput(4731,-3315)(9.55882,-5.73529){7}{\makebox(2.2222,15.5556){\tiny.}}
\multiput(4788,-3350)(9.55882,-5.73529){7}{\makebox(2.2222,15.5556){\tiny.}}
\multiput(4845,-3385)(9.55882,-5.73529){7}{\makebox(2.2222,15.5556){\tiny.}}
\multiput(4902,-3420)(9.55882,-5.73529){7}{\makebox(2.2222,15.5556){\tiny.}}
\multiput(4959,-3455)(11.38236,-6.82942){6}{\makebox(2.2222,15.5556){\tiny.}}
\multiput(5016,-3489)(11.38236,-6.82942){6}{\makebox(2.2222,15.5556){\tiny.}}
\multiput(5073,-3523)(11.38236,-6.82942){6}{\makebox(2.2222,15.5556){\tiny.}}
\multiput(5130,-3557)(11.23530,-6.74118){6}{\makebox(2.2222,15.5556){\tiny.}}
\multiput(5186,-3591)(11.05882,-6.63529){6}{\makebox(2.2222,15.5556){\tiny.}}
\multiput(5242,-3623)(10.91176,-6.54706){6}{\makebox(2.2222,15.5556){\tiny.}}
\multiput(5297,-3655)(10.82352,-6.49411){6}{\makebox(2.2222,15.5556){\tiny.}}
\multiput(5352,-3686)(10.82352,-6.49411){6}{\makebox(2.2222,15.5556){\tiny.}}
\multiput(5407,-3717)(10.96000,-5.48000){6}{\makebox(2.2222,15.5556){\tiny.}}
\multiput(5461,-3746)(10.80000,-5.40000){6}{\makebox(2.2222,15.5556){\tiny.}}
\multiput(5514,-3775)(10.64000,-5.32000){6}{\makebox(2.2222,15.5556){\tiny.}}
\multiput(5567,-3802)(10.64000,-5.32000){6}{\makebox(2.2222,15.5556){\tiny.}}
\multiput(5620,-3829)(10.40000,-5.20000){6}{\makebox(2.2222,15.5556){\tiny.}}
\multiput(5672,-3855)(10.32000,-5.16000){6}{\makebox(2.2222,15.5556){\tiny.}}
\multiput(5724,-3880)(10.08000,-5.04000){6}{\makebox(2.2222,15.5556){\tiny.}}
\multiput(5775,-3904)(10.08000,-5.04000){6}{\makebox(2.2222,15.5556){\tiny.}}
\multiput(5826,-3928)(12.67243,-5.06897){5}{\makebox(2.2222,15.5556){\tiny.}}
\multiput(5876,-3950)(12.88792,-5.15517){5}{\makebox(2.2222,15.5556){\tiny.}}
\multiput(5927,-3972)(12.45690,-4.98276){5}{\makebox(2.2222,15.5556){\tiny.}}
\multiput(5976,-3994)(12.58620,-5.03448){5}{\makebox(2.2222,15.5556){\tiny.}}
\multiput(6026,-4015)(12.58620,-5.03448){5}{\makebox(2.2222,15.5556){\tiny.}}
}%
\end{picture}%
\nop{
             fully plastic
           /     ^        \                              \space
         /       |          \                            \space
 +-----------------------+    \                          \space
 |    |          |       |     |
 |    V          |       |     V
 | amnesic ------+       |  plastic
 |    |            \     |     |     \                   \space
 |    |              \ / |     |       \                 \space
 |    |              / \ |     |         \               \space
 |    |             |    |     |          |
 |    |             V    | \   V          V
 |    |         dogmatic | learnable   damascan
 |    |             |    |                |
 |    |             |    +----------------------+
 |    |             |                     |     |
 |    |             V                     V     |
 |    |         believer ----------> correcting |
 |    |        /                                |
 |    |      /                                  |
 |    |    /                                    |
 |    V   V                                     |
 |  equating                                    |
 |                                              |
 +----------------------------------------------+
}
\hcaption{Radical revision}
\label{figure-relations-radical}
\end{hfigure}

Radical revision is not learnable: Theorem~\ref{theorem-radical-learnable}. It
is not damascan by Theorem~\ref{theorem-radical-damascan}. It is amnesic by
Theorem~\ref{theorem-radical-dogmatic}. It is dogmatic by
Theorem~\ref{theorem-radical-dogmatic}.

\section{Conclusions}
\label{section-conclusions}


No new revision method is introduced. No new framework, no new approach, no
new point of view. Only a comparative study of the existing revisions. Which
ones suit a certain application? Which ones turn every doxastic state into
every other, like for the beachfront restaurant? Which ones arrive to a
dogmatic state of mind, like for the astronomer and the round Earth? Which ones
reach every doxastic state from the empty one, like for the Greek island? Which
ones equate beliefs, like for the eclipses? Which ones possess these abilities,
which ones can do these changes?

Studying the properties, telling what systems can do is unoriginal. Nothing
new, just a study of the old. One by one, each ability is proved or disproved
for each revision. Not only this is unoriginal, it is surprisingly new. Why was
it not done before? The first iterated revision methods were introduced more
than thirty years ago~\cite{spoh-88}. Several others have been added during the
years. None emerged as the sole winner, suited for all possible use cases.

Just classifying revisions on contingency and how they are epiphanic
spotlighted a hole, filled by deep severe revision. The distinction is not only
in principles, it leads to a technical conclusion: the existing non-epiphanic
revisions are not equating, and therefore not amnesic and not fully plastic.
They cannot make conditions equally believed. They cannot forget everything.
They cannot turn every doxastic state into every other.

The proof is even easy. Some others are not, such as the plasticity of severe
revisions. Every doxastic state turns into almost every other one by first
removing all previous beliefs. Concretely: to change mind completely, first
forget everything. A question arises: is this easier than gradually mutating
the old beliefs? Maybe, or maybe not. Maybe it is just a mathematical trick,
something that works for the technical proofs but does not make sense in
concrete. Maybe not: beliefs root with time, changing them may be harder
than obliterating them all and seeding new ones.


More generally, how natural sequences of revisions are? The proof that severe
revision is plastic again provides an example: the revisions are as many as the
target doxastic state is large even if it differs very little from the current
one. The number of revisions may provide additional hints about the suitability
of a revision to a certain application. Many revisions are justified for a
complete change of mind, not for a minor alteration.

Even more generally, a revision missing an ability does not suit a context
requiring it, but one possessing it may not either. Other factors such as the
number, explainability or complexity of the revisions may play an essential
role.

Speaking about open questions, another is extending the analysis to other
revisions, such as
the comparison-based ones~%
\cite{cant-97,rott-09}.
They still hinge around doxastic states that are connected preorders, contrary
to many other ones, based on a formula~%
\cite{arav-etal-18},
on a generalization of the lexicographic order~%
\cite{andr-etal-02},
or some other kind of structures like graphs~%
\cite{gura-kodi-18,souz-etal-19},
prioritized bases~%
\cite{brew-89,nebe-91,benf-etal-93},
conditionals~%
\cite{kuts-19,andr-etal-02,saue-etal-22}
or other methods from the preference reasoning field~%
\cite{doms-etal-11}.

Also open is the list of the abilities. Many are listed and analyzed, but many
others are certainly needed in other contexts. Changing mind completely like
the damascan ability allows is a rare event. More common is a change of mind on
a specific topic only. Damascan in confined form makes more sense than a
totally damascan.

Every revision lacks some abilities. Yet, what is missing may be provided by
other revisions. For example, lexicographic revision is not equating and full
meet revision is not damascan. Yet, they are plastic together.

None of the analyzed operators is fully plastic.

The lack of this ability is systemic, it follows from what revisions are:
belief revisions are belief introductions. The current beliefs are revised by
incorporating a new one. At a minimum, the new belief is believed. The state of
no belief is unreachable.

Dropping this requirement solves the problem. Natural revision becomes fully
plastic by altering redundant revisions: $C \nat(A) = \flatorder$ if $C(0)
\subset A$. The revision $C(0) \cup C(1)$ proves this variant amnesic.
Lemma~\ref{lemma-plastic-severe} still proves it learnable because it employs
only revisions contained in a single class of the current order, none properly
contains its class $C(0)$; they work as in the unchanged version of natural
revision. Amnesic plus learnable equals plastic.

Does such a revision mechanism make sense? Why the exception for this case
$C(0) \subset A$?

A less extravagant change is $C \nat(\true) = \flatorder$: believing $\true$ is
believing everything. Believing everything makes every condition possible,
equally possible. All models have the same strength of belief.

The variant has a justification. Yet, the mathematical definition still
separates two cases, suggesting artificiality: if believing $\true$ is just
believing everything equally, why the exception in the formal definition? If it
is not a special case, it should follow with the others from a uniform
definition.

The conclusion so far is that a fully plastic revision is easy to obtain when
disregarding its justification. However, this is only the conclusion so far. A
fully plastic revision following sensible principles may still exist.

\appendix

\section{General results}
\label{section-plastic}

A common requirement for a revision is believing the new information. Another
is that nothing changes when revising by something obvious, something that is
always the case.

In logical terms, a revision is believed: it is true in all maximally believed
situations. It is entailed by the result of revising.

In logical terms, a tautology does not change the state of belief.

These two properties alone negate full plasticity. The flat doxastic state
never results from revising another. Forgetting everything is impossible. The
state of complete ignorance is only possible initially. If a revision becomes
entailed and revising by a tautology changes nothing, the revision is not
amnesic.

\begin{theorem}
\label{theorem-amnesic}

No revision $\rev()$ satisfying
{} $(C \rev(A))(0) \models A$
and
{} $C \rev(\true) = C$
is amnesic.

\end{theorem}

\proof The proof assumes that $C \rev(A)$ is $\flatorder$ and concludes it the
same as $C$: the flat doxastic state only results from revising itself.

Class zero of $C \rev(A) = \flatorder$ contains all models. By the first
assumption of the theorem, it entails $A$. All formulae entailed by all models
are tautologic: $A$ is equivalent to $\true$. By the second assumption of the
theorem, $C \rev(\true)$ is equal to $C$. In the other way around, $C$ is equal
to $C \rev(\true)$, which is $C \rev(A) = \flatorder$.~\qed

\

Some revisions merge classes. All variants of radical and severe revisions do:
their definitions contain unions like $C(0) \cup \cdots \cup C(\imin(A))$, or
$C(\imin(A)) \cup \cdots \cup C(\imax(A))$.

Other revisions only divide classes, never merge them. Their definitions
contain $C(i) \cap A$, $C(\imin(A)) \backslash A$ and similar expressions
containing a single input class $C(i)$ each or $\min(A)$.

\begin{theorem}
\label{theorem-split}

If every class of $C \rev(A)$ is contained in a class of $C$,
the revision $\rev()$ is not equating.

\end{theorem}

\proof Equating is
{} $I \equiv_{C \rev(R_1) \ldots \rev(R_n)} J$
and $I \not\equiv_C J$ for some models $I$ and $J$ and some sequence of
revisions $R_1,\ldots,R_n$. Since $I$ is equivalent to $J$ at the end of the
sequence and not the beginning, there exists at least an index $i$ such that
{} $I \equiv_{C \rev(R_1) \ldots \rev(R_i)} J$
and
{} $I \not\equiv_{C \rev(R_1) \ldots \rev(R_{i-1})} J$.

The definition of
{} $I \equiv_{C \rev(R_1) \ldots \rev(R_i)} J$
is that $I$ and $J$ belong to same class of
{} $C \rev(R_1) \ldots \rev(R_i)$.
By assumption, this class of the revised order containing both $I$ and $J$ is
contained in a single class of the unrevised order
{} $C \rev(R_1) \ldots \rev(R_{i-1})$.
Since $I$ and $J$ are in the same class of this order, they are equivalent:
{} $I \equiv_{C \rev(R_1) \ldots \rev(R_{i-1})} J$.
This contradicts the assumption
{} $I \not\equiv_{C \rev(R_1) \ldots \rev(R_{i-1})} J$.~\qed

Same class is equality. Different classes is inequality. Merging turns
non-equality into equality, the equating property.

Plasticity is changing every order in any possible way. Including class merges.
Plasticity requires equalizing models.

\begin{theorem}
\label{theorem-plastic-equating}

Every plastic change operator is equating.

\end{theorem}

\proof A plastic revision turns every order into an arbitrary non-flat revision.
Therefore, it turns
{} $C = [I, J, \true \backslash \{I,J\}]$
into
{} $G = [\{I, J\}, \true \backslash \{I,J\}]$,
changing the order $I < J$ into $I \equiv J$.~\qed

\

A believer revision not only merges classes. It also believes in some
situations over others, possibly overruling a previous opposite belief.

\begin{theorem}
\label{theorem-believer-correcting}

Every believer revision is correcting.

\end{theorem}

\proof The order $I < J$ between two arbitrary models is inverted by
$\rev(\{J\})$. The believer revision is
{} $C \rev(\{J\})(0) = \{J\}$,
which implies that the class of $I$ is greater than $0$. The class of $J$ is
$0$, less than $i$. This defines $J < I$.~\qed

In most cases, believer revisions are also equating. The only exception is a
singleton alphabet, a single variable.

\begin{theorem}
\label{theorem-believer-equating}

Every believer revision is equating if the alphabet comprises at least two
variables.

\end{theorem}

\proof A believer revision separates an arbitrary target first class from the
other models. If that class comprises two given models, they are made equal.
The alphabet of one variable is excluded because the revision cannot separate
the only two models from the others, since no other model exists.~\qed

\section{Natural revision}
\label{section-natural}

\draft

\begin{enumerate}

\item relevant example 

\v
+-------------+
|C0           | S1        h1 h2 h3 h4
+-------------+
|C1 +------+  | S2           h2 h3 h4
+---|------|--+
|C2 |   A  |  | S3              h3 h4
+---|------|--+
|C3 +------+  | S4                 h4
+-------------+
|C4           |
+-------------+
\vv

\item numbers in the definition

definition: $A | h_{\leq \neg A} \vee A | h_{>A}$

the minimal index of $\neg \neg A$ is 1, therefore $i=1$

\item translation into spheres and differences

\v
R1 =    A | h1vA | h2 h3 h4  =  A h2 h3 h4
                             =  A S2
                             =  A C1           ==>  G0 = R1 - 0
                                                       = C1 A

R2 =        h1vA | h2 h3 h4  =  h1 h2 h3 h4 v A h2 h3 h4
                             =  S1 v A S2
                             =  C0 u (C0 u C1) A
                             =  C0 u C1 A      ==>  G1 = R2 - R1
                                                       = (C0 u C1 A) - (C1 A)
                                                       = C0
R3 =               h2 h3 h4  =  S2
                             =  C0 u C1        ==>  G2 = R3 - R2
                                                       = (C0 u C1) - (C0 u C1 A)
                                                       = C1 - C1 A
                                                       = C1 - A
R4 =                  h3 h4  =  S3
                             =  C0 u C1 u C2
                                               ==>  G3 = R4 - R3
                                                       = C0 u C1 u C2 - (C0 u C1)
                                                       = C2

R5 =                     h4  =  S4
                             =  C0 u C1 u C2 u C3
                                               ==>  G4 = R5 - R4
                                                       = C3
\vv

\item class drawing

\v
                              +------+
                              +------+        C1 A
+-------------+           +-------------+
|C0           |           |             |     C0
+-------------+           +-------------+
|C1 +------+  |           |   +------+  |     C1 - A
+---|------|--+           +---+      +--+
|C2 |   A  |  |     =>
+---|------|--+           +---+------+--+
|C3 +------+  |           |   |   A  |  |     C2
+-------------+           +---|------|--+
|C4           |           |   +------+  |     C3
+-------------+           +-------------+
                          |             |
                          +-------------+
\vv

\end{enumerate}

\begin{hfigure}
\long\def\ttytex#1#2{#1}
\ttytex{
\begin{tabular}{ccc}
\setlength{\unitlength}{3750sp}%
\begin{picture}(1224,1824)(5089,-4873)
\thinlines
{\color[rgb]{0,0,0}\put(5101,-3361){\line( 1, 0){1200}}
}%
{\color[rgb]{0,0,0}\put(5101,-4861){\framebox(1200,1800){}}
}%
{\color[rgb]{0,0,0}\put(5101,-3961){\line( 1, 0){1200}}
}%
{\color[rgb]{0,0,0}\put(5101,-4261){\line( 1, 0){1200}}
}%
{\color[rgb]{0,0,0}\put(5101,-4561){\line( 1, 0){1200}}
}%
{\color[rgb]{0,0,0}\put(5401,-4486){\framebox(750,1050){}}
}%
{\color[rgb]{0,0,0}\put(5101,-3661){\line( 1, 0){1200}}
}%
\put(5851,-3886){\makebox(0,0)[b]{\smash{\fontsize{9}{10.8}
\usefont{T1}{cmr}{m}{n}{\color[rgb]{0,0,0}$A$}%
}}}
\end{picture}%
&
\setlength{\unitlength}{3750sp}%
\begin{picture}(1104,1104)(5659,-4303)
\thinlines
{\color[rgb]{0,0,0}\multiput(6391,-3391)(-9.47368,4.73684){20}{\makebox(2.1167,14.8167){\tiny.}}
\put(6211,-3301){\line( 0,-1){ 45}}
\put(6211,-3346){\line(-1, 0){180}}
\put(6031,-3346){\line( 0,-1){ 90}}
\put(6031,-3436){\line( 1, 0){180}}
\put(6211,-3436){\line( 0,-1){ 45}}
\multiput(6211,-3481)(9.47368,4.73684){20}{\makebox(2.1167,14.8167){\tiny.}}
}%
\end{picture}%
&
\setlength{\unitlength}{3750sp}%
\begin{picture}(1224,2049)(5089,-4873)
\thinlines
{\color[rgb]{0,0,0}\put(5101,-3361){\line( 1, 0){1200}}
}%
{\color[rgb]{0,0,0}\put(5101,-3661){\line( 1, 0){1200}}
}%
{\color[rgb]{0,0,0}\put(5101,-4861){\framebox(1200,1800){}}
}%
{\color[rgb]{0,0,0}\put(5101,-3961){\line( 1, 0){1200}}
}%
{\color[rgb]{0,0,0}\put(5101,-4261){\line( 1, 0){1200}}
}%
{\color[rgb]{0,0,0}\put(5101,-4561){\line( 1, 0){1200}}
}%
{\color[rgb]{0,0,0}\put(5401,-3061){\framebox(750,225){}}
}%
{\color[rgb]{0,0,0}\put(5401,-4486){\framebox(750,1050){}}
}%
{\color[rgb]{0,0,0}\multiput(5401,-3586)(7.89474,7.89474){20}{\makebox(2.2222,15.5556){\tiny.}}
}%
{\color[rgb]{0,0,0}\multiput(5401,-3661)(8.03571,8.03571){29}{\makebox(2.2222,15.5556){\tiny.}}
}%
{\color[rgb]{0,0,0}\multiput(5401,-3511)(8.33333,8.33333){10}{\makebox(2.2222,15.5556){\tiny.}}
}%
{\color[rgb]{0,0,0}\multiput(5476,-3661)(8.03571,8.03571){29}{\makebox(2.2222,15.5556){\tiny.}}
}%
{\color[rgb]{0,0,0}\multiput(5551,-3661)(8.03571,8.03571){29}{\makebox(2.2222,15.5556){\tiny.}}
}%
{\color[rgb]{0,0,0}\multiput(5626,-3661)(8.03571,8.03571){29}{\makebox(2.2222,15.5556){\tiny.}}
}%
{\color[rgb]{0,0,0}\multiput(5701,-3661)(8.03571,8.03571){29}{\makebox(2.2222,15.5556){\tiny.}}
}%
{\color[rgb]{0,0,0}\multiput(5776,-3661)(8.03571,8.03571){29}{\makebox(2.2222,15.5556){\tiny.}}
}%
{\color[rgb]{0,0,0}\multiput(5851,-3661)(8.03571,8.03571){29}{\makebox(2.2222,15.5556){\tiny.}}
}%
{\color[rgb]{0,0,0}\multiput(5926,-3661)(8.03571,8.03571){29}{\makebox(2.2222,15.5556){\tiny.}}
}%
{\color[rgb]{0,0,0}\multiput(6001,-3661)(7.89474,7.89474){20}{\makebox(2.2222,15.5556){\tiny.}}
}%
{\color[rgb]{0,0,0}\multiput(6076,-3661)(8.33333,8.33333){10}{\makebox(2.2222,15.5556){\tiny.}}
}%
\end{picture}%
\end{tabular}
}{
                             +------+ min(A)
                             +------+
+-------------+          +-------------+
|             |          |             |
+-------------+          +-------------+
|   +------+  |          |   +------+  |
+---|------|--+          +---+      +--+
|   |   A  |  |    =>
+---|------|--+          +---+------+--+
|   |      |  |          |   |      |  |
+---|------|--+          +---|------|--+
|   +------+  |          |   |      |  |
+-------------+          +---|------|--+
|             |          |   +------+  |
+-------------+          +-------------+
|             |          |             |
+-------------+          +-------------+
|             |          |             |
+-------------+          +-------------+
                         |             |
                         +-------------+
  current.fig              natural.fig
}
\hcaption{Natural revision}
\end{hfigure}

\enddraft

Natural revision increases the strength of belief in the currently most
believed situations admitted by the revision. Formally, it makes the models of
$\min(A)$ the new minimal ones. Nothing else changes.

\begin{definition}
\label{definition-natural}

\[
C \nat(A) =
[
	\min(A),
	C(0) \backslash \min(A) , \ldots , C(\last) \backslash \min(A)
]
\]

\end{definition}

Natural revision is learnable~\cite{libe-23}. The proof is generalized to
damascan and to other kinds of revisions.

\begin{lemma}
\label{lemma-natural-singleclass}

For every two orders $C$ and $G$ such that every class of $G$ is contained in a
class of $C$, a sequence of single-class natural revisions turns $C$ into $G$.

\end{lemma}

\proof A graphical overview of the proof is given first.

The first revision is $G(\last)$. By assumption, its models are all in some
class of $C$ and are therefore all minimal: $\min(G(\last)) = G(\last)$.
The revised order is
{} $[G(\last),C(0) \backslash G(\last), \ldots, C(\last) \backslash G(\last)]$,
as shown in Figure~\ref{natural-first}.

\begin{hfigure}
\long\def\ttytex#1#2{#1}
\ttytex{
\begin{tabular}{ccc}
\setlength{\unitlength}{3750sp}%
\begin{picture}(1224,2424)(4789,-5173)
\thinlines
{\color[rgb]{0,0,0}\put(4801,-3361){\line( 1, 0){1200}}
}%
{\color[rgb]{0,0,0}\put(4801,-4561){\line( 1, 0){1200}}
}%
{\color[rgb]{0,0,0}\put(4801,-3961){\line( 1, 0){1200}}
}%
{\color[rgb]{0,0,0}\put(4801,-5161){\framebox(1200,2400){}}
}%
{\color[rgb]{0,0,0}\put(4951,-4411){\framebox(450,300){}}
}%
\put(5176,-4336){\makebox(0,0)[b]{\smash{\fontsize{9}{10.8}
\usefont{T1}{cmr}{m}{n}{\color[rgb]{0,0,0}$G(\last)$}%
}}}
\end{picture}%
&
\setlength{\unitlength}{3750sp}%
\begin{picture}(1104,744)(5659,-3943)
\thinlines
{\color[rgb]{0,0,0}\multiput(6391,-3391)(-9.47368,4.73684){20}{\makebox(2.1167,14.8167){\tiny.}}
\put(6211,-3301){\line( 0,-1){ 45}}
\put(6211,-3346){\line(-1, 0){180}}
\put(6031,-3346){\line( 0,-1){ 90}}
\put(6031,-3436){\line( 1, 0){180}}
\put(6211,-3436){\line( 0,-1){ 45}}
\multiput(6211,-3481)(9.47368,4.73684){20}{\makebox(2.1167,14.8167){\tiny.}}
}%
\end{picture}%
&
\setlength{\unitlength}{3750sp}%
\begin{picture}(1224,2874)(4789,-5173)
\thinlines
{\color[rgb]{0,0,0}\put(4801,-3361){\line( 1, 0){1200}}
}%
{\color[rgb]{0,0,0}\put(4801,-4561){\line( 1, 0){1200}}
}%
{\color[rgb]{0,0,0}\put(4801,-3961){\line( 1, 0){1200}}
}%
{\color[rgb]{0,0,0}\put(4801,-5161){\framebox(1200,2400){}}
}%
{\color[rgb]{0,0,0}\put(4951,-4411){\framebox(450,300){}}
}%
{\color[rgb]{0,0,0}\put(4951,-2611){\framebox(450,300){}}
}%
{\color[rgb]{0,0,0}\multiput(4951,-4261)(7.89474,7.89474){20}{\makebox(2.2222,15.5556){\tiny.}}
}%
{\color[rgb]{0,0,0}\put(4951,-4411){\line( 1, 1){300}}
}%
{\color[rgb]{0,0,0}\put(5101,-4411){\line( 1, 1){300}}
}%
{\color[rgb]{0,0,0}\multiput(5251,-4411)(7.89474,7.89474){20}{\makebox(2.2222,15.5556){\tiny.}}
}%
\put(5176,-2536){\makebox(0,0)[b]{\smash{\fontsize{9}{10.8}
\usefont{T1}{cmr}{m}{n}{\color[rgb]{0,0,0}$G(\last)$}%
}}}
\end{picture}%
\end{tabular}
}{
                           +--+
                           |Go|
                           +--+
+-------------+          +-------------+
|             |          |             |
|             |          |             |
|             |          |             |
+-------------+          +-------------+
|             |          |             |
|             |          |             |
|             |          |             |
+-------------+    =>    +-------------+
| +--+        |          | +--+        |
| |Go|        |          | |//|        |
| +--+        |          | +--+        |
+-------------+          +-------------+
|             |          |             |
|             |          |             |
|             |          |             |
+-------------+          +-------------+
natural-start.fig       natural-first.fig
}%
\label{natural-first}
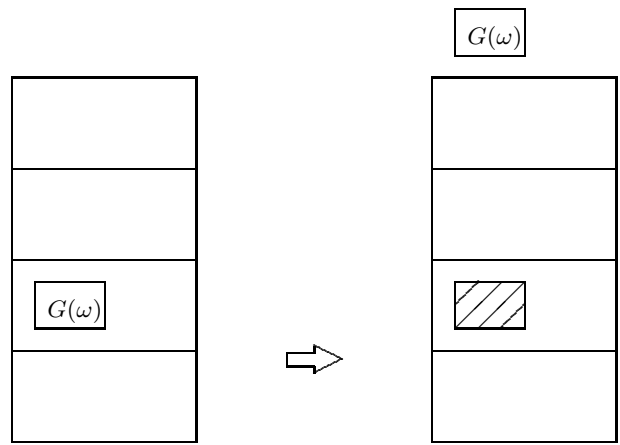
\hcaption{The first natural revision}
\end{hfigure}

The second revision is $G(\last-1)$. By assumption, it is all contained in some
class $C(i)$. It does not intersect $G(\last)$ because equivalence classes are
disjoint by definition. Therefore,
{} $G(\last-1) \subseteq C(i) \backslash G(\last)$.
The result of the second revision is in Figure~\ref{natural-second}.

\begin{hfigure}
\long\def\ttytex#1#2{#1}
\ttytex{
\begin{tabular}{ccc}
\setlength{\unitlength}{3750sp}%
\begin{picture}(1224,2874)(4789,-5173)
\thinlines
{\color[rgb]{0,0,0}\put(4801,-3361){\line( 1, 0){1200}}
}%
{\color[rgb]{0,0,0}\put(4801,-4561){\line( 1, 0){1200}}
}%
{\color[rgb]{0,0,0}\put(4801,-3961){\line( 1, 0){1200}}
}%
{\color[rgb]{0,0,0}\put(4801,-5161){\framebox(1200,2400){}}
}%
{\color[rgb]{0,0,0}\put(4951,-4411){\framebox(450,300){}}
}%
{\color[rgb]{0,0,0}\put(4951,-2611){\framebox(450,300){}}
}%
{\color[rgb]{0,0,0}\multiput(4951,-4261)(7.89474,7.89474){20}{\makebox(2.2222,15.5556){\tiny.}}
}%
{\color[rgb]{0,0,0}\put(4951,-4411){\line( 1, 1){300}}
}%
{\color[rgb]{0,0,0}\put(5101,-4411){\line( 1, 1){300}}
}%
{\color[rgb]{0,0,0}\multiput(5251,-4411)(7.89474,7.89474){20}{\makebox(2.2222,15.5556){\tiny.}}
}%
{\color[rgb]{0,0,0}\put(5251,-3811){\framebox(675,300){}}
}%
\put(5176,-2536){\makebox(0,0)[b]{\smash{\fontsize{9}{10.8}
\usefont{T1}{cmr}{m}{n}{\color[rgb]{0,0,0}$G(\last)$}%
}}}
\put(5551,-3736){\makebox(0,0)[b]{\smash{\fontsize{9}{10.8}
\usefont{T1}{cmr}{m}{n}{\color[rgb]{0,0,0}$G(\last-1)$}%
}}}
\end{picture}%
&
\setlength{\unitlength}{3750sp}%
\begin{picture}(1104,744)(5659,-3943)
\thinlines
{\color[rgb]{0,0,0}\multiput(6391,-3391)(-9.47368,4.73684){20}{\makebox(2.1167,14.8167){\tiny.}}
\put(6211,-3301){\line( 0,-1){ 45}}
\put(6211,-3346){\line(-1, 0){180}}
\put(6031,-3346){\line( 0,-1){ 90}}
\put(6031,-3436){\line( 1, 0){180}}
\put(6211,-3436){\line( 0,-1){ 45}}
\multiput(6211,-3481)(9.47368,4.73684){20}{\makebox(2.1167,14.8167){\tiny.}}
}%
\end{picture}%
&
\setlength{\unitlength}{3750sp}%
\begin{picture}(1224,3324)(4789,-5173)
\thinlines
{\color[rgb]{0,0,0}\put(4801,-3361){\line( 1, 0){1200}}
}%
{\color[rgb]{0,0,0}\put(4801,-4561){\line( 1, 0){1200}}
}%
{\color[rgb]{0,0,0}\put(4801,-3961){\line( 1, 0){1200}}
}%
{\color[rgb]{0,0,0}\put(4801,-5161){\framebox(1200,2400){}}
}%
{\color[rgb]{0,0,0}\put(4951,-4411){\framebox(450,300){}}
}%
{\color[rgb]{0,0,0}\put(4951,-2611){\framebox(450,300){}}
}%
{\color[rgb]{0,0,0}\multiput(4951,-4261)(7.89474,7.89474){20}{\makebox(2.2222,15.5556){\tiny.}}
}%
{\color[rgb]{0,0,0}\put(4951,-4411){\line( 1, 1){300}}
}%
{\color[rgb]{0,0,0}\put(5101,-4411){\line( 1, 1){300}}
}%
{\color[rgb]{0,0,0}\multiput(5251,-4411)(7.89474,7.89474){20}{\makebox(2.2222,15.5556){\tiny.}}
}%
{\color[rgb]{0,0,0}\put(5401,-3811){\line( 1, 1){300}}
}%
{\color[rgb]{0,0,0}\put(5551,-3811){\line( 1, 1){300}}
}%
{\color[rgb]{0,0,0}\put(5251,-3811){\framebox(675,300){}}
}%
{\color[rgb]{0,0,0}\multiput(5251,-3661)(7.89474,7.89474){20}{\makebox(2.2222,15.5556){\tiny.}}
}%
{\color[rgb]{0,0,0}\put(5251,-3811){\line( 1, 1){300}}
}%
{\color[rgb]{0,0,0}\multiput(5701,-3811)(8.03571,8.03571){29}{\makebox(2.2222,15.5556){\tiny.}}
}%
{\color[rgb]{0,0,0}\multiput(5851,-3811)(8.33333,8.33333){10}{\makebox(2.2222,15.5556){\tiny.}}
}%
{\color[rgb]{0,0,0}\put(5251,-2161){\framebox(675,300){}}
}%
\put(5176,-2536){\makebox(0,0)[b]{\smash{\fontsize{9}{10.8}
\usefont{T1}{cmr}{m}{n}{\color[rgb]{0,0,0}$G(\last)$}%
}}}
\put(5551,-2086){\makebox(0,0)[b]{\smash{\fontsize{9}{10.8}
\usefont{T1}{cmr}{m}{n}{\color[rgb]{0,0,0}$G(\last-1)$}%
}}}
\end{picture}%
\end{tabular}
}{
                                 +----+
                                 |Go-1|
                                 +----+
  +--+                     +--+
  |Go|                     |Go|
  +--+                     +--+
+-------------+          +-------------+
|             |          |             |
|             |          |             |
|             |          |             |
+-------------+          +-------------+
|      +----+ |          |      +----+ |
|      |Go-1| |          |      |////| |
|      +----+ |          |      +----+ |
+-------------+    =>    +-------------+
| +--+        |          | +--+        |
| |//|        |          | |//|        |
| +--+        |          | +--+        |
+-------------+          +-------------+
|             |          |             |
|             |          |             |
|             |          |             |
+-------------+          +-------------+
}%
\label{natural-second}
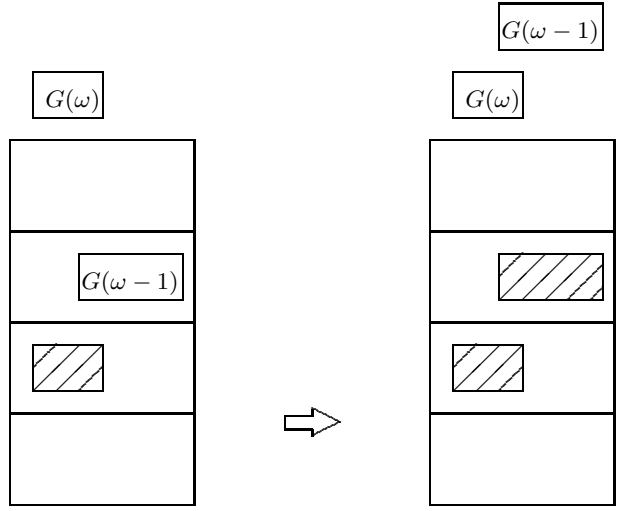
\hcaption{The second natural revision}
\end{hfigure}

The following revisions are $G(\last-2),\ldots,G(0)$. Like the first two ones,
each of them makes its models the most believed ones, more believed than all
others.

\

The inductive proof hinges on the invariant that the order after the revision
$G(i)$:

\begin{itemize}

\item begins with the classes from $G(i)$ to $G(\last)$ in this order;

\item the remaining classes are all subsets of some $C(j)$ each.

\end{itemize}

This is the induction assumption and claim.

\begin{eqnarray*}
\lefteqn{C [\nat(G(\last)), \ldots, \nat(G(i))] = }
\\
&=&
[
	G(i),\ldots,G(\last),
\\
&&
	C(0) \backslash (G(i) \cup \cdots \cup G(\last)),
		\ldots,
	C(\last) \backslash (G(i) \cup \cdots \cup G(\last))
]
\end{eqnarray*}

In the base case $i=\last$, the claim is
{} $\flatorder [\nat(G(\last))] = [
{}	G(\last),
{}	C(0) \backslash G(\last), \ldots, C(\last) \backslash G(\last)
{} ]$.
It follows from the definition of natural revision and from the assumption
{} $G(\last) \subseteq C(i)$ from some $i$,
which implies
{} $\min(G(\last)) = G(\last)$.

\begin{eqnarray*}
\lefteqn{C [\nat(G(\last))]}
\\
&=& [
	\min(G(\last)),
	C(0) \backslash \min(G(\last)),
		\ldots,
	C(\last) \backslash \min(G(\last))
]
\\
&=& [
	G(\last),
	C(0) \backslash G(\last),
		\ldots,
	C(\last) \backslash G(\last)
]
\end{eqnarray*}

The induction assumption is that the revision $G(i-1)$ applies to the following
order.

\begin{eqnarray*}
\lefteqn{C [\nat(G(\last)), \ldots, \nat(G(i))] = }
\\
&=&
[
	G(i),\ldots,G(\last),
\\
&&
	C(0) \backslash (G(i) \cup \cdots \cup G(\last)),
		\ldots,
	C(\last) \backslash (G(i) \cup \cdots \cup G(\last))
]
\end{eqnarray*}

The induction claim is that the following order results.

\begin{eqnarray*}
\lefteqn{C [\nat(G(\last)), \ldots, \nat(G(i)), \nat(G(i+1))] = }
\\
&=&
[
	G(i-1),G(i),\ldots,G(\last),
\\
&&
	C(0) \backslash (G(i-1) \cup G(i) \cup \cdots \cup G(\last)),
		\ldots,
	C(\last) \backslash (G(i-1) \cup G(i) \cup \cdots \cup G(\last))
]
\end{eqnarray*}

By the assumption of the lemma, all models of $G(i-1)$ are in some class
$C(j)$. None of them is in $G(i), \ldots, G(\last)$ since equivalence classes
are disjoint by definition. As a result, $G(i-1)$ is all contained in
{} $C(j) \backslash (G(i) \cup \cdots \cup G(\last))$
for some class $C(j)$. All its models are minimal: $\min(G(i-1)) = G(i-1)$.

\begin{eqnarray*}
\lefteqn{C [\nat(G(\last)), \ldots, \nat(G(i)),\nat(G(i-1))] =}
\\
&=& [
	\min(G(i-1)),
\\
&&
	G(i),\ldots,G(\last),
\\
&&
	C(0) \backslash (G(i) \cup \cdots \cup G(\last))
		\backslash \min(G(i-1)),
\\
&&
			\ldots,
\\
&&
	C(\last) \backslash (G(i) \cup \cdots \cup G(\last))
		\backslash \min(G(i-1))
]
\\
&=& [
	G(i-1),
\\
&&
	G(i),\ldots,G(\last),
\\
&&
	C(0) \backslash (G(i) \cup \cdots \cup G(\last))
		\backslash G(i-1)),
\\
&&
			\ldots,
\\
&&
	C(\last) \backslash (G(i) \cup \cdots \cup G(\last))
		\backslash G(i-1))
]
\\
&=& [
	G(i-1),
	G(i),\ldots,G(\last),
\\
&&
	C(0) \backslash (G(i-1) \cup G(i) \cup \cdots \cup G(\last))
\\
&&
			\ldots,
\\
&&
	C(\last) \backslash (G(i-1) \cup G(i) \cup \cdots \cup G(\last))
]
\end{eqnarray*}

This is the induction claim.

When applied to the last revision $G(0)$, it shows that the resulting order is
the following.

\begin{eqnarray*}
\lefteqn{C [\nat(G(\last)), \ldots, \nat(G(0))] =}
\\
&=& [
	G(0),\ldots,G(\last),
\\
&&
	C(0) \backslash (G(0) \cup \cdots \cup G(\last))
\\
&&
			\ldots,
\\
&&
	C(\last) \backslash (G(0) \cup \cdots \cup G(\last))
]
\\
&=& [
	G(0),\ldots,G(\last),
\\
&&
	C(0) \backslash \true,
\\
&&
			\ldots,
\\
&&
	C(\last) \backslash \true
]
\\
&=& [
	G(0),\ldots,G(\last),
\\
&&
	\emptyset, \ldots, \emptyset
]
\\
&=& [
	G(0),\ldots,G(\last)
]
\end{eqnarray*}

\qed

The requirement that every class of $G$ is contained in some class of $C$ is
met by every doxastic state $G$ when $C$ is the flat order $\flatorder =
[\true]$. The lemma proves that every doxastic state results from naturally
revising the flat order: natural revision is learnable.

\begin{theorem}
\label{theorem-natural-learnable}

Natural revision is learnable even when restricting to single-class revisions.

\end{theorem}

The requirement that every class of $G$ is contained in some class of $C$ is
also met when every class of $G$ coincides with some class of $C$. This is the
case when reversing an order: $G(i) = C(\last - i)$ for all classes $G(i)$.
This proves that natural revision is damascan.

\begin{theorem}
\label{theorem-natural-damascan}

Natural revision is damascan even when restricting to single-class revisions.

\end{theorem}

\

Natural revision is not equating.

\begin{theorem}
\label{theorem-natural-equating}

Natural revision is not equating.

\end{theorem}

\proof Theorem~\ref{theorem-split}, proves the claim when every class of $C
\rev(A)$ is contained in a class of $C$. This is shown the case for natural
revision.

\[
C \nat(A) =
[
	\min(A),
	C(0) \backslash \min(A) , \ldots , C(\last) \backslash \min(A)
]
\]

The class $\min(A)$ is a subset of $C(\imin(A))$. The other classes are each
$C(i) \backslash \min(A)$, a subset of $C(i)$.~\qed

Theorem~\ref{theorem-plastic-equating} proves that natural revision is not
plastic because it is not equating.

\begin{theorem}
\label{theorem-natural-plastic}

Natural revision is not plastic.

\end{theorem}

\section{Lexicographic revision}
\label{section-lexicographic}

Lexicographic revision believes the new information in all possible situations,
not just the most believed ones.

\begin{definition}
\label{definition-lexicographic}

\[
C \lex(A) = [
C(0) \cap A, \ldots, C(\last) \cap A,
C(0) \backslash A, \ldots, C(\last) \backslash A
]
\]

\end{definition}

Lexicographic revision coincides with natural when the revision is contained in
a class of the current doxastic state, like in
Lemma~\ref{theorem-natural-learnable} and Lemma~\ref{theorem-natural-damascan}.

\begin{lemma}
\label{lemma-lexicographic-natural}

Lexicographic and natural revision coincide when the revision is contained in a
class of the order: $C \lex(A) = C \nat(A)$ if $A \subseteq C(i)$ for some $i$.

\end{lemma}

\proof If $A$ is contained in a single class $C(i)$, then $\min(A) = A$ and $C(j) \cap
A = \emptyset$ for all $j \not= i$.

\begin{eqnarray*}
\lefteqn{C \lex(A)}
\\
&=& [
	C(0) \cap A, \ldots, C(\last) \cap A,
	C(0) \backslash A, \ldots , C(\last) \backslash A
]
\\
&=& [
	C(0) \cap A, \ldots, C(i-1) \cap A, C(i) \cap (A), C(i+1) \cap A, C(\last) \cap A,
	C(0) \backslash A, \ldots , C(\last) \backslash A
]
\\
&=& [
	C(i) \cap A,
	C(0) \backslash A, \ldots , C(\last) \backslash A
]
\\
&=& [
	C(i) \cap \min(A),
	C(0) \backslash \min(A), \ldots , C(\last) \backslash \min(A)
]
\\
&=&
	C \nat(A)
\end{eqnarray*}~\qed

As a consequence, lexicographic revision is learnable and damascan.

\begin{theorem}
\label{theorem-lexicographic-learnable}

Lexicographic revision is learnable.

\end{theorem}

\proof Lemma~\ref{theorem-natural-learnable} proves natural revision learnable even when
bounded to revisions all contained in a class of the current order.
Lemma~\ref{lemma-lexicographic-natural} proves that it gives the same result of
lexicographic revision in this case.~\qed

\begin{theorem}
\label{theorem-lexicographic-damascan}

Lexicographic revision is damascan.

\end{theorem}

\proof Lemma~\ref{theorem-natural-damascan} proves natural revision damascan even when
bounded to revisions all contained in a class of the current order.
Lemma~\ref{lemma-lexicographic-natural} proves that it gives the same result of
lexicographic revision in this case.~\qed

\

Lexicographic revision is not equating and therefore not amnesic nor plastic.

\begin{theorem}
\label{theorem-lexicographic-equating}

Lexicographic revision is not equating.

\end{theorem}

\proof The classes of $C \lex(A)$ are either $C(i) \cap A$ or $C(i) \backslash
A$ for some $i$. They are both contained in $C(i)$: every class of $C \lex(A)$
is contained in a class of $C$. Theorem~\ref{theorem-split} proves that such a
revision is not equating.~\qed

Theorem~\ref{theorem-split} proves that only equating revisions are plastic.

\begin{corollary}

Lexicographic revision is not plastic.

\end{corollary}

\section{Restrained revision}
\label{section-restrained}

Restrained revision strongly believes the new information in the currently most
believed situations and minimally in all others.

\begin{definition}
\label{definition-restrained}

\[
C \res(A) = [
	\min(A),
	C(0) \cap A \backslash \min(A), C(0) \backslash A,
	\ldots
	C(\last) \cap A \backslash \min(A), C(\last) \backslash A
]
\]

\end{definition}

Restrained revision coincides with natural revision when the revision is all
contained in a single class, like in
Lemma~\ref{theorem-natural-learnable}
and
Lemma~\ref{theorem-natural-damascan}.

\begin{lemma}
\label{lemma-restrained-natural}

Restrained and natural revision coincide when the revision is contained in a
class of the current order:
{} $C \res(A) = C \nat(A)$ if $A \subseteq C(i)$ for some $i$.

\end{lemma}

\proof If $A$ is contained in a single class $C(i)$, then $\min(A) = A$.

\begin{eqnarray*}
\lefteqn{C \res(A)}
\\
&=& [
	\min(A),
	C(0) \cap A \backslash \min(A), C(0) \backslash A,
	\ldots
	C(\last) \cap A \backslash \min(A), C(\last) \backslash A
]
\\
&=& [
	\min(A),
	C(0) \cap A \backslash A, C(0) \backslash A,
	\ldots
	C(\last) \cap A \backslash A, C(\last) \backslash A
]
\\
&=& [
	\min(A),
	C(0) \backslash A,
	\ldots
	C(\last) \backslash A
]
\\
&\equiv& [
	\min(A),
	C(0) \backslash \min(A),
	\ldots
	C(\last) \backslash \min(A)
]
\\
&=&
	C \nat(A)
\end{eqnarray*}~\qed

\begin{theorem}
\label{theorem-restrained-learnable}

Restrained revision is learnable.

\end{theorem}

\proof Lemma~\ref{theorem-natural-learnable} proves natural revision learnable
even when bounded to revisions all contained in a class of the current order.
Lemma~\ref{lemma-restrained-natural} proves that it coincides with restrained
revision in this case.~\qed

\begin{theorem}
\label{theorem-restrained-damascan}

Restrained revision is damascan.

\end{theorem}

\proof Lemma~\ref{theorem-natural-damascan} proves natural revision learnable
even when bounded to revisions all contained in a class of the current order.
Lemma~\ref{lemma-restrained-natural} proves that it gives the same result of
restrained revision in this case.~\qed

Restrained revision is not equating and therefore not plastic, as proved by
Theorem~\ref{theorem-split}.

\begin{theorem}
\label{theorem-restrained-equating}

Restrained revision is not equating.

\end{theorem}

\proof The classes of $C \res(A)$ are $\min(A)$, $C(i) \cap A \backslash
\min(A)$ and $C(i) \backslash A$ for some index $i$. By definition, $\min(A)$
is contained in $C(\imin(A))$. The other classes are each a subset of $C(i)$.
Every class of $C \res(A)$ is contained in a class of $C$.
Theorem~\ref{theorem-split} proves that such a revision is not equating.~\qed

Theorem~\ref{theorem-split} disproves the plasticity of restrained revision.

\begin{theorem}
\label{theorem-restrained-plastic}

Restrained revision is not plastic.

\end{theorem}

\section{Radical revision}
\label{section-radical}

\draft

definition in cite{rott-09} where n is the number of formulae

\[
	h_{<n} | h_n \wedge A
\]

expression of the classes:

\begin{enumerate}

\item significant example

\v
+-------------+
|C0           | S1        h1 h2 h3 h4
+-------------+
|C1 +------+  | S2           h2 h3 h4
+---|------|--+
|C2 |   A  |  | S3              h3 h4
+---|------|--+
|C3 |      |  | S4                 h4
+---|------|--+
|C4 +------+  |
+-------------+
\vv

differs from very radical only if A intersects the last class

\item calculations
$n = 4$

\item translation

\v
R1 =   h1 h2 h3 h4A  =  A S1  =  A C0 = 0  ==>  empty

R2 =      h2 h3 h4A  =  A S2  =  A (C0 u C1)  =  A C1
                                                   ==>  G0 = R2 - R1 = A C1

R3 =         h3 h4A  =  A S3  =  A (C0 u C1 u C2)
                              =  A C0 u A C1 u A C2
                              =  A C1 u A C2      ==>  G1 = R3 - R2 = A C2

R4 =            h4A  =  A S4  =  A (C0 u C1 u C2 u C3)
                              =  A C0 u A C1 u A C2 u A C3
                              =  A C1 u A C2 u A C3
                                                  ==>  G2 = R4 - R3 = A C3

                                                  ==>  G4 = T - R4
                                                          = T - A (C1 u C2 u C3)
                                                          = T - A (T - C4)
                                                          = T - AT u AC4
                                                          = T - A u AC4
\vv

\item drawing of classes $[A C_1, A C_2, A C_3, \true \backslash A]$

\v
                                      +------+ A C1
                                      |------|
                                      |   A  | A C2
                                      |------|
                                      |      | A C3
+-------------+         +-------------+------+
|C0           |         |             +------+ A C4
+-------------+         |             |
|C1 +------+  |         |   +------+  |
+---|------|--+         |   |      |  |
|C2 |   A  |  |         |   |      |  |
+---|------|--+   ==>   |   |      |  |
|C3 |      |  |         |   |      |  |
+---|------|--+         |   +------+  |
|C4 +------+  |         |    T - A    |
+-------------+         +-------------+
\vv

\end{enumerate}

\

the flat state is a special case

single sphere including all models

singleton sequence of formulae: only $h_1=\true$

one formula: n=1

the first part $h_{<n}$ of the definition is empty;
only $h_n \wedge A$ remains, with $h_n = h_1 = \true$

\[
	h_{<n} | h_n \wedge A    =    \true \wedge A    =    A
\]

one formula, $h_1' = A$;
first class contains the model of $A$

order is $[A, \true \backslash A]$

in this case, the models in the last class do not remain there

only the models of $A \cap C(\last) \backslash C(0)$ remain in the last class

models in the last class stay there, unless it is the only one

whether they are models of the revision A or not is irrelevant

\v
                                       +------+
                                       |------|
                                       |      |
                                       |------|
                                       |      |
+-------------+          +-------------+------|
|             |          |  T - A      :   J  |
+-------------+          |             +------+
|   +------+  |          |   +------+  |
+---|------|--+          |   |      |  |
|   |   A  |  |    =>    |   |      |  |
+---|------|--+          |   |      |  |
|   |      |  |          |   |      |  |
+---|------|--+          |   |      |  |
|   |   J  |  |          |   |      |  |
| I +------+  |          | I +------+  |
+-------------+          +-------------+
\vv

no way to obtain an order where I is in class zero and a class one exists

\begin{hfigure}
\long\def\ttytex#1#2{#1}
\ttytex{
\begin{tabular}{ccc}
\setlength{\unitlength}{3750sp}%
\begin{picture}(1224,1824)(5089,-4873)
\thinlines
{\color[rgb]{0,0,0}\put(5101,-3361){\line( 1, 0){1200}}
}%
{\color[rgb]{0,0,0}\put(5101,-4861){\framebox(1200,1800){}}
}%
{\color[rgb]{0,0,0}\put(5101,-3961){\line( 1, 0){1200}}
}%
{\color[rgb]{0,0,0}\put(5101,-4261){\line( 1, 0){1200}}
}%
{\color[rgb]{0,0,0}\put(5101,-4561){\line( 1, 0){1200}}
}%
{\color[rgb]{0,0,0}\put(5401,-4486){\framebox(750,1050){}}
}%
{\color[rgb]{0,0,0}\put(5101,-3661){\line( 1, 0){1200}}
}%
\put(5851,-3886){\makebox(0,0)[b]{\smash{\fontsize{9}{10.8}
\usefont{T1}{cmr}{m}{n}{\color[rgb]{0,0,0}$A$}%
}}}
\end{picture}%
&
\setlength{\unitlength}{3750sp}%
\begin{picture}(1104,1104)(5659,-4303)
\thinlines
{\color[rgb]{0,0,0}\multiput(6391,-3391)(-9.47368,4.73684){20}{\makebox(2.1167,14.8167){\tiny.}}
\put(6211,-3301){\line( 0,-1){ 45}}
\put(6211,-3346){\line(-1, 0){180}}
\put(6031,-3346){\line( 0,-1){ 90}}
\put(6031,-3436){\line( 1, 0){180}}
\put(6211,-3436){\line( 0,-1){ 45}}
\multiput(6211,-3481)(9.47368,4.73684){20}{\makebox(2.1167,14.8167){\tiny.}}
}%
\end{picture}%
&
\setlength{\unitlength}{3750sp}%
\begin{picture}(1974,2949)(5089,-4873)
\thinlines
{\color[rgb]{0,0,0}\put(6301,-2461){\line( 1, 0){750}}
}%
{\color[rgb]{0,0,0}\put(6301,-2761){\line( 1, 0){750}}
}%
{\color[rgb]{0,0,0}\put(6301,-3061){\line( 1, 0){750}}
}%
{\color[rgb]{0,0,0}\put(6301,-1936){\line( 1, 0){750}}
\put(7051,-1936){\line( 0,-1){1350}}
\put(7051,-3286){\line(-1, 0){750}}
\put(6301,-3286){\line( 0,-1){1575}}
\put(6301,-4861){\line(-1, 0){1200}}
\put(5101,-4861){\line( 0, 1){1800}}
\put(5101,-3061){\line( 1, 0){1200}}
\put(6301,-3061){\line( 0, 1){1125}}
}%
{\color[rgb]{0,0,0}\put(5401,-4786){\framebox(750,1350){}}
}%
{\color[rgb]{0,0,0}\put(6301,-2161){\line( 1, 0){750}}
}%
\put(6751,-2386){\makebox(0,0)[b]{\smash{\fontsize{9}{10.8}
\usefont{T1}{cmr}{m}{n}{\color[rgb]{0,0,0}$A$}%
}}}
\end{picture}%
\end{tabular}
}{
                                       +------+
                                       |------|
                                       |   A  |
                                       |------|
                                       |      |
+-------------+          +-------------+------|
|             |          | T - A + C(o)+------+
+-------------+          |             |
|   +------+  |          |   +------+  |
+---|------|--+          |   |      |  |
|   |   A  |  |    =>    |   |      |  |
+---|------|--+          |   |      |  |
|   |      |  |          |   |      |  |
+---|------|--+          |   |      |  |
|   +------+  |          |   +------+  |
+-------------+          +-------------+
  current.fig              radical.fig
}
\hcaption{radical revision}
\end{hfigure}

\enddraft

Radical revision believes the new information in all possible situations like
lexicographic revision. It rejects everything contradicting it. It also
considers the least believed situations impossible.

\begin{definition}
\label{definition-radical}

\begin{eqnarray*}
\lefteqn{C \rad(A)}
\\
&=& [
	C(\imin(A)) \cap A \backslash (C(\last) \backslash C(0)),
		\ldots,
	C(\imax(A)) \cap A \backslash (C(\last) \backslash C(0)),
\\
&&
	\true \backslash A \cup (C(\last) \backslash C(0))
]
\end{eqnarray*}

\end{definition}

Radical revision is amnesic: revising an order by its last class flattens it.
As a result, it is also equating since every strict order $I <_C J$ turns into
an equality in $C(i) \rad(C(\last)) = \flatorder$.

\begin{lemma}
\label{lemma-radical-last}

Every non-flat order is flattened by radically revising it by a model $I$ of
its last class.

\end{lemma}

\proof Every non-flat order $C$ is flattened by radically revising it by a
model $I$ of its last class $C(\last)$. The minimal and maximal indexes of
$\{I\}$ are both $\last$.

Since $C$ is not flat, it contains at least two classes. As a result, $C(0)$
and $C(\last)$ do not coincide. They do not intersect either since $C$ is a
partition. Therefore, $C(\last) \backslash C(0)$ is $C(\last)$, which contains
$I$.

\begin{eqnarray*}
\lefteqn{C \rad(A)}
\\
&=& [
	C(\imin(A)) \cap A \backslash (C(\last) \backslash C(0)),
		\ldots,
	C(\imax(A)) \cap A \backslash (C(\last) \backslash C(0)),
	\true \backslash A \cup (C(\last) \backslash C(0))
]
\\
&=& [
	C(\last) \cap \{I\} \backslash C(\last),
		\ldots,
	C(\last) \cap \{I\} \backslash C(\last),
	\true \backslash \{I\} \cup C(\last)
]
\\
&=& [
	C(\last) \cap A \backslash C(\last),
	\true \backslash \{I\} \cup C(\last)
]
\\
&=& [
	\emptyset,
	\true
]
\\
&=& [
	\true
]
\\
&=& \flatorder
\end{eqnarray*}~\qed

\begin{theorem}
\label{theorem-radical-amnesic}

Radical revision is amnesic.

\end{theorem}

\proof Every non-flat order $C$ is flattened by radically revising it by a
model $I$ of its last class by Lemma~\ref{lemma-radical-last}.

The flat order is the result of
{} $\flatorder []$ or
{} $\flatorder \rad(\true)$:
the empty sequence of radical revisions and the sequence comprising only the
revision $\true$.~\qed

Radical revision is not learnable, and therefore not plastic: no order of three
or more classes is obtained by revising an order of two classes or fewer.

The first step of the proof shows that the only single-class doxastic state
$\flatorder$ is never revised into an order comprising more than two classes.

\begin{lemma}
\label{theorem-radical-one-two}

$\flatorder \rad(A)$ comprises at most two classes.

\end{lemma}

\proof The claim is depicted in Figure~\ref{figure-two}.

\begin{hfigure}
\long\def\ttytex#1#2{#1}
\ttytex{
\begin{tabular}{ccc}
\setlength{\unitlength}{3750sp}%
\begin{picture}(1224,624)(4789,-3973)
\thinlines
{\color[rgb]{0,0,0}\put(4801,-3961){\framebox(1200,600){}}
}%
{\color[rgb]{0,0,0}\put(5176,-3811){\framebox(450,300){}}
}%
\put(5401,-3736){\makebox(0,0)[b]{\smash{\fontsize{9}{10.8}
\usefont{T1}{cmr}{m}{n}{\color[rgb]{0,0,0}$A$}%
}}}
\end{picture}%
&
\setlength{\unitlength}{3750sp}%
\begin{picture}(1104,744)(5659,-3943)
\thinlines
{\color[rgb]{0,0,0}\multiput(6391,-3391)(-9.47368,4.73684){20}{\makebox(2.1167,14.8167){\tiny.}}
\put(6211,-3301){\line( 0,-1){ 45}}
\put(6211,-3346){\line(-1, 0){180}}
\put(6031,-3346){\line( 0,-1){ 90}}
\put(6031,-3436){\line( 1, 0){180}}
\put(6211,-3436){\line( 0,-1){ 45}}
\multiput(6211,-3481)(9.47368,4.73684){20}{\makebox(2.1167,14.8167){\tiny.}}
}%
\end{picture}%
&
\setlength{\unitlength}{3750sp}%
\begin{picture}(1224,1074)(4789,-3973)
\thinlines
{\color[rgb]{0,0,0}\put(4801,-3961){\framebox(1200,600){}}
}%
{\color[rgb]{0,0,0}\put(5176,-3811){\framebox(450,300){}}
}%
{\color[rgb]{0,0,0}\put(5176,-3211){\framebox(450,300){}}
}%
{\color[rgb]{0,0,0}\multiput(5176,-3661)(7.89474,7.89474){20}{\makebox(2.2222,15.5556){\tiny.}}
}%
{\color[rgb]{0,0,0}\put(5176,-3811){\line( 1, 1){300}}
}%
{\color[rgb]{0,0,0}\put(5326,-3811){\line( 1, 1){300}}
}%
{\color[rgb]{0,0,0}\multiput(5476,-3811)(7.89474,7.89474){20}{\makebox(2.2222,15.5556){\tiny.}}
}%
\put(5401,-3136){\makebox(0,0)[b]{\smash{\fontsize{9}{10.8}
\usefont{T1}{cmr}{m}{n}{\color[rgb]{0,0,0}$A$}%
}}}
\end{picture}%
\end{tabular}
}{
                                +---+
                                | A |
+-------------+                 +---+
|     +---+   |           +-------------+
|     | A |   |           |     +---+   |
|     +---+   |    ==>    |     |///|   |
|             |           |     +---+   |
|             |           |             |
+-------------+           |             |
                          +-------------+
   flat-a.fig                  a.fig
}
\label{figure-two}
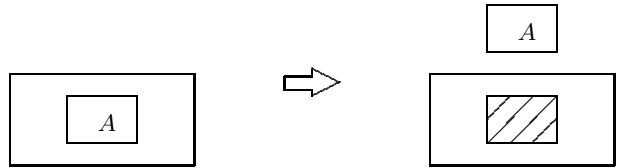
\hcaption{The radical revision of $\flatorder$ comprises two classes at most.}
\end{hfigure}

Since
{}	$\flatorder$
comprises a single class and $A$ is not empty,
the minimum and maximum classes of $A$ are zero:
{}	$\imin(A) = 0$
and
{}	$\imax(A) = 0$.
Another consequence of $\last=0$ is
{}	$\flatorder(\last) \backslash \flatorder(0) = 
{}		\flatorder(0) \backslash \flatorder(0) =
{}		\emptyset$.
These values are in the definition of radical revision.

\begin{eqnarray*}
\lefteqn{\flatorder \rad(A)}
\\
&=& [
	\flatorder(\imin(A)) \cap A \backslash (\flatorder(\last) \backslash \flatorder(0)),
		\ldots,
	\flatorder(\imax(A)) \cap A \backslash (\flatorder(\last) \backslash \flatorder(0)),
	\true \backslash A \cup (\flatorder(\last) \backslash \flatorder(0))
]
\\
&=& [
	\flatorder(0) \cap A \backslash \emptyset,
		\ldots,
	\flatorder(0) \cap A \backslash \emptyset,
	\true \backslash A \cup \emptyset
]
\\
&=& [
	\flatorder(0) \cap A,
	\true \backslash A
]
\end{eqnarray*}

Two classes result, or one if $A = \flatorder(0) = \true$.~\qed

\

The second and last step is the proof that radical revision does not increase
the number of classes of a two-class order.

\begin{lemma}
\label{theorem-radical-two-two}

$[C(0),C(1)] \rad(A)$ comprises at most two classes.

\end{lemma}

\proof The last class of $C$ is $C(1)$, its index $\last$ is $1$.

Since every revision is not-contradictory by assumption, it either intersects
$C(0)$, $C(1)$ or both. Each of the three cases is analyzed.

\begin{description}

\item[$A \subseteq C(0)$]

\

\begin{hfigure}
\long\def\ttytex#1#2{#1}
\ttytex{
\begin{tabular}{ccc}
\setlength{\unitlength}{3750sp}%
\begin{picture}(1224,1224)(4789,-3973)
\thinlines
{\color[rgb]{0,0,0}\put(5176,-3211){\framebox(450,300){}}
}%
{\color[rgb]{0,0,0}\put(4801,-3961){\framebox(1200,1200){}}
}%
{\color[rgb]{0,0,0}\put(4801,-3361){\line( 1, 0){1200}}
}%
\put(5401,-3136){\makebox(0,0)[b]{\smash{\fontsize{9}{10.8}
\usefont{T1}{cmr}{m}{n}{\color[rgb]{0,0,0}$A$}%
}}}
\end{picture}%
&
\setlength{\unitlength}{3750sp}%
\begin{picture}(1104,744)(5659,-3943)
\thinlines
{\color[rgb]{0,0,0}\multiput(6391,-3391)(-9.47368,4.73684){20}{\makebox(2.1167,14.8167){\tiny.}}
\put(6211,-3301){\line( 0,-1){ 45}}
\put(6211,-3346){\line(-1, 0){180}}
\put(6031,-3346){\line( 0,-1){ 90}}
\put(6031,-3436){\line( 1, 0){180}}
\put(6211,-3436){\line( 0,-1){ 45}}
\multiput(6211,-3481)(9.47368,4.73684){20}{\makebox(2.1167,14.8167){\tiny.}}
}%
\end{picture}%
&
\setlength{\unitlength}{3750sp}%
\begin{picture}(1224,1674)(4789,-3973)
\thinlines
{\color[rgb]{0,0,0}\put(5176,-3211){\framebox(450,300){}}
}%
{\color[rgb]{0,0,0}\put(4801,-3961){\framebox(1200,1200){}}
}%
{\color[rgb]{0,0,0}\put(5176,-2611){\framebox(450,300){}}
}%
{\color[rgb]{0,0,0}\multiput(5176,-3061)(7.89474,7.89474){20}{\makebox(2.2222,15.5556){\tiny.}}
}%
{\color[rgb]{0,0,0}\put(5176,-3211){\line( 1, 1){300}}
}%
{\color[rgb]{0,0,0}\put(5326,-3211){\line( 1, 1){300}}
}%
{\color[rgb]{0,0,0}\multiput(5476,-3211)(7.89474,7.89474){20}{\makebox(2.2222,15.5556){\tiny.}}
}%
\put(5401,-2536){\makebox(0,0)[b]{\smash{\fontsize{9}{10.8}
\usefont{T1}{cmr}{m}{n}{\color[rgb]{0,0,0}$A$}%
}}}
\end{picture}%
\end{tabular}
}{
                                +---+
                                | A |
+-------------+                 +---+
|     +---+   |           +-------------+
|     | A |   |           |     +---+   |
+-----+---+---+    ==>    |     |///|   |
|             |           |     +---+   |
|             |           |             |
+-------------+           |             |
                          +-------------+
   a-zero.fig           a-zero-radical.fig
}
\label{figure-a-zero-radical}
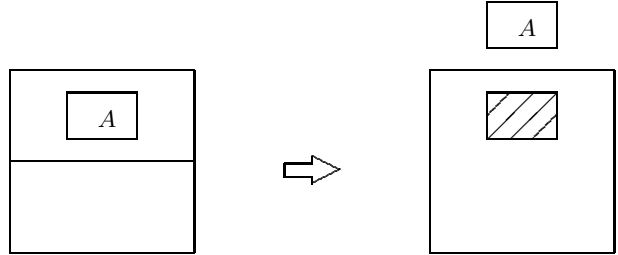
\hcaption{First case: $A$ included in the first class}
\end{hfigure}

The assumption
{} $A \subseteq C(0)$ implies
{}	$\imin(A) = 0$
and
{}	$\imax(A) = 0$.


\begin{eqnarray*}
\lefteqn{C \rad(A)}
\\
&=& [
	C(\imin(A)) \cap A \backslash (C(\last) \backslash C(0)),
		\ldots,
	C(\imax(A)) \cap A \backslash (C(\last) \backslash C(0)),
	\true \backslash A \cup (C(\last) \backslash C(0))
]
\\
&=& [
	C(0) \cap A \backslash (C(1) \backslash C(0)),
		\ldots,
	C(0) \cap A \backslash (C(1) \backslash C(0)),
	\true \backslash A \cup (C(1) \backslash C(0))
]
\\
&=& [
	C(0) \cap A \backslash (C(1) \backslash C(0)),
	\true \backslash A \cup (C(1) \backslash C(0))
]
\end{eqnarray*}

This order comprises at most two classes, one if either
{} $C(0) \cap A \backslash (C(1) \backslash C(0))$
or
{} $\true \backslash A \cup (C(1) \backslash C(0))$
is empty.

\

\item[$A \cap C(0) \not= \emptyset$ and $A \cap C(1) \not= \emptyset$]

\

\begin{hfigure}
\long\def\ttytex#1#2{#1}
\ttytex{
\begin{tabular}{ccc}
\setlength{\unitlength}{3750sp}%
\begin{picture}(1224,1224)(4789,-3973)
\thinlines
{\color[rgb]{0,0,0}\put(4801,-3961){\framebox(1200,1200){}}
}%
{\color[rgb]{0,0,0}\put(4801,-3361){\line( 1, 0){1200}}
}%
{\color[rgb]{0,0,0}\put(5176,-3811){\framebox(450,900){}}
}%
\put(5401,-3136){\makebox(0,0)[b]{\smash{\fontsize{9}{10.8}
\usefont{T1}{cmr}{m}{n}{\color[rgb]{0,0,0}$A$}%
}}}
\end{picture}%
&
\setlength{\unitlength}{3750sp}%
\begin{picture}(1104,744)(5659,-3943)
\thinlines
{\color[rgb]{0,0,0}\multiput(6391,-3391)(-9.47368,4.73684){20}{\makebox(2.1167,14.8167){\tiny.}}
\put(6211,-3301){\line( 0,-1){ 45}}
\put(6211,-3346){\line(-1, 0){180}}
\put(6031,-3346){\line( 0,-1){ 90}}
\put(6031,-3436){\line( 1, 0){180}}
\put(6211,-3436){\line( 0,-1){ 45}}
\multiput(6211,-3481)(9.47368,4.73684){20}{\makebox(2.1167,14.8167){\tiny.}}
}%
\end{picture}%
&
\setlength{\unitlength}{3750sp}%
\begin{picture}(1674,1674)(4789,-3973)
\thinlines
{\color[rgb]{0,0,0}\put(5176,-3811){\framebox(450,900){}}
}%
{\color[rgb]{0,0,0}\put(6001,-2761){\line( 1, 0){450}}
}%
{\color[rgb]{0,0,0}\put(6001,-2311){\line( 1, 0){450}}
\put(6451,-2311){\line( 0,-1){900}}
\put(6451,-3211){\line(-1, 0){450}}
\put(6001,-3211){\line( 0,-1){750}}
\put(6001,-3961){\line(-1, 0){1200}}
\put(4801,-3961){\line( 0, 1){1200}}
\put(4801,-2761){\line( 1, 0){1200}}
\put(6001,-2761){\line( 0, 1){450}}
}%
{\color[rgb]{0,0,0}\multiput(5176,-3061)(7.89474,7.89474){20}{\makebox(2.2222,15.5556){\tiny.}}
}%
{\color[rgb]{0,0,0}\put(5176,-3211){\line( 1, 1){300}}
}%
{\color[rgb]{0,0,0}\put(5176,-3361){\line( 1, 1){450}}
}%
{\color[rgb]{0,0,0}\put(5176,-3511){\line( 1, 1){450}}
}%
{\color[rgb]{0,0,0}\put(5176,-3661){\line( 1, 1){450}}
}%
{\color[rgb]{0,0,0}\put(5176,-3811){\line( 1, 1){450}}
}%
{\color[rgb]{0,0,0}\put(5326,-3811){\line( 1, 1){300}}
}%
{\color[rgb]{0,0,0}\multiput(5476,-3811)(7.89474,7.89474){20}{\makebox(2.2222,15.5556){\tiny.}}
}%
\put(6226,-2611){\makebox(0,0)[b]{\smash{\fontsize{9}{10.8}
\usefont{T1}{cmr}{m}{n}{\color[rgb]{0,0,0}$A$}%
}}}
\end{picture}%
\end{tabular}
}{
                                        +---+
+-------------+                         | A |
|     +---+   |           +-------------+---+
|     | A |   |           |     +---+       |
+-----|---|---+    ==>    |     |///|   +---+
|     |   |   |           |     |///|   |
|     +---+   |           |     |///|   |
+-------------+           |     +---+   |
                          +-------------+
   a-both.fig              a-both-radical.fig
}
\label{figure-a-both-radical}
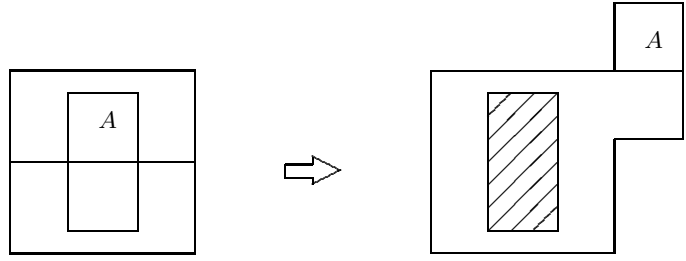
\hcaption{Second case: $A$ intersects both classes}
\end{hfigure}

The assumptions
{}	$A \cap C(0) \not= \emptyset$
and
{}	$A \cap C(1) \not= \emptyset$
imply
{}	$\imin(A) = 0$
and
{}	$\imax(A) = 1$.

\begin{eqnarray*}
\lefteqn{C \rad(A)}
\\
&=& [
	C(\imin(A)) \cap A \backslash (C(\last) \backslash C(0)),
		\ldots,
	C(\imax(A)) \cap A \backslash (C(\last) \backslash C(0)),
	\true \backslash A \cup (C(\last) \backslash C(0))
]
\\
&=& [
	C(0) \cap A \backslash (C(1) \backslash C(0)),
		\ldots,
	C(1) \cap A \backslash (C(1) \backslash C(0)),
	\true \backslash A \cup (C(1) \backslash C(0))
]
\\
&=& [
	C(0) \cap A \backslash (C(1) \backslash C(0)),
	C(1) \cap A \backslash (C(1) \backslash C(0)),
	\true \backslash A \cup (C(1) \backslash C(0))
]
\\
&=& [
	C(0) \cap A \backslash C(1),
	C(1) \cap A \backslash C(1),
	\true \backslash A \cup C(1)
]
\\
&=& [
	C(0) \cap A \backslash C(1),
	\emptyset,
	\true \backslash A \cup C(1)
]
\\
&=& [
	C(0) \cap A \backslash C(1),
	\true \backslash A \cup C(1)
]
\end{eqnarray*}

This order comprises two classes, one if the other is empty.

\

\item[$A \subseteq C(1)$]

\

\begin{hfigure}
\long\def\ttytex#1#2{#1}
\ttytex{
\begin{tabular}{ccc}
\setlength{\unitlength}{3750sp}%
\begin{picture}(1224,1224)(4789,-3973)
\thinlines
{\color[rgb]{0,0,0}\put(4801,-3961){\framebox(1200,1200){}}
}%
{\color[rgb]{0,0,0}\put(4801,-3361){\line( 1, 0){1200}}
}%
{\color[rgb]{0,0,0}\put(5176,-3811){\framebox(450,300){}}
}%
\put(5401,-3736){\makebox(0,0)[b]{\smash{\fontsize{9}{10.8}
\usefont{T1}{cmr}{m}{n}{\color[rgb]{0,0,0}$A$}%
}}}
\end{picture}%
&
\setlength{\unitlength}{3750sp}%
\begin{picture}(1104,744)(5659,-3943)
\thinlines
{\color[rgb]{0,0,0}\multiput(6391,-3391)(-9.47368,4.73684){20}{\makebox(2.1167,14.8167){\tiny.}}
\put(6211,-3301){\line( 0,-1){ 45}}
\put(6211,-3346){\line(-1, 0){180}}
\put(6031,-3346){\line( 0,-1){ 90}}
\put(6031,-3436){\line( 1, 0){180}}
\put(6211,-3436){\line( 0,-1){ 45}}
\multiput(6211,-3481)(9.47368,4.73684){20}{\makebox(2.1167,14.8167){\tiny.}}
}%
\end{picture}%
&
\setlength{\unitlength}{3750sp}%
\begin{picture}(1224,1674)(4789,-3973)
\thinlines
{\color[rgb]{0,0,0}\put(4801,-3961){\framebox(1200,1200){}}
}%
{\color[rgb]{0,0,0}\put(5176,-3811){\framebox(450,300){}}
}%
{\color[rgb]{0,0,0}\put(5176,-2611){\framebox(450,300){}}
}%
{\color[rgb]{0,0,0}\multiput(5176,-3661)(7.89474,7.89474){20}{\makebox(2.2222,15.5556){\tiny.}}
}%
{\color[rgb]{0,0,0}\put(5176,-3811){\line( 1, 1){300}}
}%
{\color[rgb]{0,0,0}\put(5326,-3811){\line( 1, 1){300}}
}%
{\color[rgb]{0,0,0}\multiput(5476,-3811)(7.89474,7.89474){20}{\makebox(2.2222,15.5556){\tiny.}}
}%
\put(5401,-2536){\makebox(0,0)[b]{\smash{\fontsize{9}{10.8}
\usefont{T1}{cmr}{m}{n}{\color[rgb]{0,0,0}$A$}%
}}}
\end{picture}%
\end{tabular}
}{
                                 +---+
                                 | A |
+-------------+                  +---+
|             |            +-------------+
|             |            |             |
+-----+---+---+    ==>     |             |
|     | A |   |            |     +---+   |
|     +---+   |            |     |///|   |
+-------------+            |     +---+   |
                           +-------------+
   a-one.fig              a-one-radical.fig
}
\label{figure-a-one-radical}
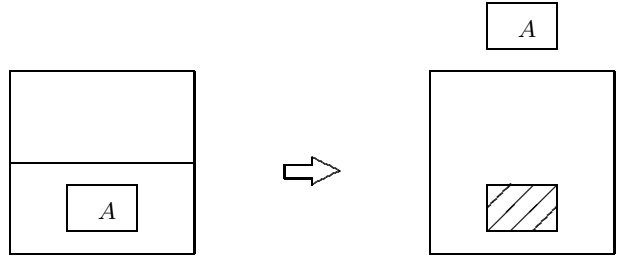
\hcaption{Third case: $A$ included in the second class}
\end{hfigure}

The assumption
{}	$A \subseteq C(1)$
implies
{}	$\imin(A) = 1$
and
{}	$\imax(A) = 1$.

\begin{eqnarray*}
\lefteqn{C \rad(A)}
\\
&=& [
	C(\imin(A)) \cap A \backslash (C(\last) \backslash C(0)),
		\ldots,
	C(\imax(A)) \cap A \backslash (C(\last) \backslash C(0)),
	\true \backslash A \cup (C(\last) \backslash C(0))
]
\\
&=& [
	C(1) \cap A \backslash (C(1) \backslash C(0)),
		\ldots,
	C(1) \cap A \backslash (C(1) \backslash C(0)),
	\true \backslash A \cup (C(1) \backslash C(0))
]
\\
&=& [
	C(1) \cap A \backslash (C(1) \backslash C(0)),
	\true \backslash A \cup (C(1) \backslash C(0))
]
\end{eqnarray*}

This order comprises at most two classes, as required. Actually, the first is
empty and the order is $\flatorder$, but proving it is not necessary.

\end{description}~\qed

Radical revision is not learnable because it never increases the number of
classes from zero to more than two.

\begin{theorem}
\label{theorem-radical-learnable}

Radical revision is not learnable.

\end{theorem}

\proof Follows from Lemma~\ref{theorem-radical-one-two} and
Lemma~\ref{theorem-radical-two-two}: revising the flat order gives at most two
classes; revising that gives at most two. No order of three classes is even
reached.~\qed

Radical revision is dogmatic: it turns every order into an arbitrary
two-classes order. It is therefore also believer.

\begin{theorem}
\label{theorem-radical-dogmatic}

Radical revision is dogmatic.

\end{theorem}

\proof Every two-class order $G = [G(0),G(1)]$ results from radically revising
an arbitrary order $C$ twice. The first revision flattens $C$, the second
separates the two classes.

The first revision is provided by Lemma~\ref{lemma-radical-last}: every
non-flat order is flattened by radically revising it by its last class.

The second revision is the first class $G(0)$ of $G$. The minimal and maximal
classes of every formula in the flat order are zero:
{} $\imin(G(0)) = \imax(G(0)) = 0$.

\begin{eqnarray*}
\lefteqn{\flatorder \rad(G(0)}
\\
&=& [
	\flatorder(\imin(G(0))) \cap G(0) \backslash (\flatorder(\last) \backslash \flatorder(0)),
		\ldots,
	\flatorder(\imax(G(0))) \cap G(0) \backslash (\flatorder(\last) \backslash \flatorder(0)),
	\true \backslash A \cup (\flatorder(\last) \backslash \flatorder(0))
]
\\
&=& [
	\flatorder(0) \cap G(0) \backslash (\flatorder(0) \backslash \flatorder(0)),
		\ldots,
	\flatorder(0) \cap G(0) \backslash (\flatorder(0) \backslash \flatorder(0)),
	\true \backslash A \cup (\flatorder(0) \backslash \flatorder(0))
]
\\
&=& [
	\flatorder(0) \cap G(0) \backslash (\flatorder(0) \backslash \flatorder(0)),
	\true \backslash A \cup (\flatorder(0) \backslash \flatorder(0))
]
\\
&=& [
	\true \cap G(0) \backslash (\true \backslash \true),
	\true \backslash A \cup (\true \backslash \true)
]
\\
&=& [
	\true \cap G(0) \backslash \emptyset,
	\true \backslash A \cup \emptyset
]
\\
&=& [
	\true \cap G(0),
	\true \backslash G(0)
]
\\
&=& [
	G(0),
	G(1)
]
\end{eqnarray*}~\qed

Radical revision is not damascan. While it can invert a two-classes order, it
cannot invert any order of three classes or more.

\begin{theorem}
\label{theorem-radical-damascan}

Radical revision is not damascan.

\end{theorem}

\proof A damascan revision inverts a doxastic state: the last class becomes the
first and vice versa.

A sequence of revisions moves a model $I$ of the last class to the first. Let
$A$ be one of the revisions that moves $I$ out the last class and $C$ the
doxastic state it is applied to: $I$ is in $C(\last)$ and not in $C
\rad(A)(\last)$. The latter class is defined as follows.

\[
C \rad(A)(\last) = \true \backslash A \cup (C(\last) \backslash C(0))
\]

This class contains $C(\last) \backslash C(0)$. Since equivalence classes are
disjoint by definition, this difference contains all of $C(\last)$, including
$I$, unless $C(\last) = C(0)$, which implies that $\last$ of $C$ is zero: $C$
is the flat doxastic state $\flatorder$.

By Theorem~\ref{theorem-radical-one-two}, the flat order only revises in a
two-class order at most, which only revises in a two-classes order by 
Theorem~\ref{theorem-radical-two-two}. No order of three classes is ever
generated. An order of the three classes is not inverted by any sequence of
radical revisions.~\qed

\section{Very radical revision}
\label{section-veryradical}

\draft

definition in rott-09:

\[
	h | A
\]

expression as classes

\begin{enumerate}

\item significant example

\v
+-------------+
|C0           | S1        h1 h2 h3 h4
+-------------+
|C1 +------+  | S2           h2 h3 h4
+---|------|--+
|C2 |   A  |  | S3              h3 h4
+---|------|--+
|C3 +------+  | S4                 h4
+-------------+
|C4           |
+-------------+
\vv

\item no calculation necessary

\item translation in classes

\v
R1 =   h1 h2 h3 h4 A  =  A S1  =  A C0 = 0  ==>  empty

R2 =      h2 h3 h4 A  =  A S2  =  A (C0 u C1)  =  A C1
                                                   ==>  G0 = R2 - R1 = A C1

R3 =         h3 h4 A  =  A S3  =  A (C0 u C1 u C2)
                               =  A C0 u A C1 u A C2
                               =  A C1 u A C2      ==>  G1 = R3 - R2 = A C2

R4 =            h4 A  =  A S4  =  A (C0 u C1 u C2 u C3)
                               =  A C0 u A C1 u A C2 u A C3
                               =  A C1 u A C2 u A C3
                                                   ==>  G2 = R4 - R3 = A C3

R5 =               A
                                                   ==>  R5 - R4
                                                           = A - (C1 u C2 u C3)
                                                           empty

                                                   ==>  G5 = T - R5
                                                           = T - A
\vv

\item drawing of classes $[A C_1, A C_2, A C_3, \true \backslash A]$

\v
                            +------+ A C1
                            |------|
                            |   A  | A C2
                            |------|
                            +------+ A C3
+-------------+         +-------------+
|C0           |         |             |
+-------------+         |             | T - A
|C1 +------+  |         |   +------+  |
+---|------|--+         |   |      |  |
|C2 |   A  |  |         |   |      |  |
+---|------|--+   ==>   |   |      |  |
|C3 +------+  |         |   +------+  |
+-------------+         |             |
|C4           |         |             |
+-------------+         +-------------+
\vv

\end{enumerate}

\begin{hfigure}
\long\def\ttytex#1#2{#1}
\ttytex{
\begin{tabular}{ccc}
\setlength{\unitlength}{3750sp}%
\begin{picture}(1224,1824)(5089,-4873)
\thinlines
{\color[rgb]{0,0,0}\put(5101,-3361){\line( 1, 0){1200}}
}%
{\color[rgb]{0,0,0}\put(5101,-4861){\framebox(1200,1800){}}
}%
{\color[rgb]{0,0,0}\put(5101,-3961){\line( 1, 0){1200}}
}%
{\color[rgb]{0,0,0}\put(5101,-4261){\line( 1, 0){1200}}
}%
{\color[rgb]{0,0,0}\put(5101,-4561){\line( 1, 0){1200}}
}%
{\color[rgb]{0,0,0}\put(5401,-4486){\framebox(750,1050){}}
}%
{\color[rgb]{0,0,0}\put(5101,-3661){\line( 1, 0){1200}}
}%
\put(5851,-3886){\makebox(0,0)[b]{\smash{\fontsize{9}{10.8}
\usefont{T1}{cmr}{m}{n}{\color[rgb]{0,0,0}$A$}%
}}}
\end{picture}%
&
\setlength{\unitlength}{3750sp}%
\begin{picture}(1104,1104)(5659,-4303)
\thinlines
{\color[rgb]{0,0,0}\multiput(6391,-3391)(-9.47368,4.73684){20}{\makebox(2.1167,14.8167){\tiny.}}
\put(6211,-3301){\line( 0,-1){ 45}}
\put(6211,-3346){\line(-1, 0){180}}
\put(6031,-3346){\line( 0,-1){ 90}}
\put(6031,-3436){\line( 1, 0){180}}
\put(6211,-3436){\line( 0,-1){ 45}}
\multiput(6211,-3481)(9.47368,4.73684){20}{\makebox(2.1167,14.8167){\tiny.}}
}%
\end{picture}%
&
\setlength{\unitlength}{3750sp}%
\begin{picture}(1224,2949)(5089,-4873)
\thinlines
{\color[rgb]{0,0,0}\put(5101,-4861){\framebox(1200,1800){}}
}%
{\color[rgb]{0,0,0}\put(5401,-4486){\framebox(750,1050){}}
}%
{\color[rgb]{0,0,0}\put(5401,-2986){\framebox(750,1050){}}
}%
{\color[rgb]{0,0,0}\put(5401,-2161){\line( 1, 0){750}}
}%
{\color[rgb]{0,0,0}\put(5401,-2461){\line( 1, 0){750}}
}%
{\color[rgb]{0,0,0}\put(5401,-2761){\line( 1, 0){750}}
}%
\put(5851,-2386){\makebox(0,0)[b]{\smash{\fontsize{9}{10.8}
\usefont{T1}{cmr}{m}{n}{\color[rgb]{0,0,0}$A$}%
}}}
\end{picture}%
\end{tabular}
}{
                             +------+
                             |------|
                             |   A  |
                             |------|
                             |      |
                             |------|
                             +------+
+-------------+          +-------------+
|             |          |             |
+-------------+          |             |
|   +------+  |          |   +------+  |
+---|------|--+          |   |      |  |
|   |   A  |  |    =>    |   |      |  |
+---|------|--+          |   |      |  |
|   |      |  |          |   |      |  |
+---|------|--+          |   |      |  |
|   +------+  |          |   +------+  |
+-------------+          |             |
|             |          |             |
+-------------+          |             |
|             |          |             |
+-------------+          |             |
|             |          |             |
+-------------+          +-------------+
  current.fig            veryradical.fig
}
\hcaption{Very radical revision}
\end{hfigure}

\enddraft

Very radical revision believes the new information in all possible situations
like lexicographic revision and rejects everything contradicting it. It differs
from radical revision in that no situation is impossible. Even the least
believed situations become believed if new information support them.

\begin{definition}
\label{definition-veryradical}

\[
C \vrad(A) = [
	C(\imin(A)) \cap A, \ldots, C(\imax(A)) \cap A,
	\true \backslash A
]
\]

\end{definition}

Very radical revision is plastic but not fully plastic. Every doxastic state
results from a sequence of radical revisions but the state of total ignorance.
Revisions cancel information, but not all of it.

The following lemma proves that every non-flat order $G$ is the result of a
sequence of very radical revisions that do not depend on the current order. The
independence is not required by the plastic property, which is therefore a
consequence.

\begin{lemma}
\label{lemma-plastic-veryradical}

Every non-flat arbitrary order $G$ has a sequence of very radical revisions
that turns every order $C$ into $G$.

\end{lemma}

\proof No class of $G$ contains all models since the only order with such a
class is the flat order $\flatorder$, and $G$ is assumed not flat.

The first class $G(0)$ does not contain some models. The first revision is one
of these models $I$. Being a single-model formula, it is all contained in a
class. Therefore, its minimal and maximal class coincide:
{} $C(\imax(I\})) = C(\imin(\{I\}))$.

\begin{eqnarray*}
\lefteqn{C \vrad(I)}
\\
&=& [
	C(\imin(I)) \cap \{I\}, \ldots, C(\imax(I)) \cap \{I\},
	\true \backslash \{I\}
]
\\
&=& [
	C(\imin(I)) \cap \{I\}, \ldots, C(\imin(I)) \cap \{I\},
	\true \backslash \{I\}
]
\\
&=& [
	C(\imin(I)) \cap \{I\},
	\true \backslash \{I\}
]
\\
&=& [
	\{I\},
	\true \backslash \{I\}
]
\end{eqnarray*}

The first revision makes $I$ strictly more believed than all other models, as
shown in Figure~\ref{figure-n-i}.

\begin{hfigure}
\long\def\ttytex#1#2{#1}
\ttytex{
\begin{tabular}{ccc}
\setlength{\unitlength}{3750sp}%
\begin{picture}(1224,3624)(4789,-5173)
\thinlines
{\color[rgb]{0,0,0}\put(4801,-3361){\line( 1, 0){1200}}
}%
{\color[rgb]{0,0,0}\put(4801,-2161){\line( 1, 0){1200}}
}%
{\color[rgb]{0,0,0}\put(4801,-4561){\line( 1, 0){1200}}
}%
{\color[rgb]{0,0,0}\put(4801,-5161){\framebox(1200,3600){}}
}%
{\color[rgb]{0,0,0}\put(4801,-3961){\line( 1, 0){1200}}
}%
{\color[rgb]{0,0,0}\put(4801,-2761){\line( 1, 0){1200}}
}%
{\color[rgb]{0,0,0}\put(5551,-3811){\framebox(300,300){}}
}%
\put(5701,-3736){\makebox(0,0)[b]{\smash{\fontsize{9}{10.8}
\usefont{T1}{cmr}{m}{n}{\color[rgb]{0,0,0}$I$}%
}}}
\end{picture}%
&
\setlength{\unitlength}{3750sp}%
\begin{picture}(1104,1104)(5659,-4303)
\thinlines
{\color[rgb]{0,0,0}\multiput(6391,-3391)(-9.47368,4.73684){20}{\makebox(2.1167,14.8167){\tiny.}}
\put(6211,-3301){\line( 0,-1){ 45}}
\put(6211,-3346){\line(-1, 0){180}}
\put(6031,-3346){\line( 0,-1){ 90}}
\put(6031,-3436){\line( 1, 0){180}}
\put(6211,-3436){\line( 0,-1){ 45}}
\multiput(6211,-3481)(9.47368,4.73684){20}{\makebox(2.1167,14.8167){\tiny.}}
}%
\end{picture}%
&
\setlength{\unitlength}{3750sp}%
\begin{picture}(1224,4074)(4789,-5173)
\thinlines
{\color[rgb]{0,0,0}\put(4801,-5161){\framebox(1200,3600){}}
}%
{\color[rgb]{0,0,0}\put(5551,-3811){\framebox(300,300){}}
}%
{\color[rgb]{0,0,0}\put(5551,-1411){\framebox(300,300){}}
}%
{\color[rgb]{0,0,0}\multiput(5551,-3661)(7.89474,7.89474){20}{\makebox(2.2222,15.5556){\tiny.}}
}%
{\color[rgb]{0,0,0}\put(5551,-3811){\line( 1, 1){300}}
}%
{\color[rgb]{0,0,0}\multiput(5701,-3811)(7.89474,7.89474){20}{\makebox(2.2222,15.5556){\tiny.}}
}%
\put(5701,-1336){\makebox(0,0)[b]{\smash{\fontsize{9}{10.8}
\usefont{T1}{cmr}{m}{n}{\color[rgb]{0,0,0}$I$}%
}}}
\end{picture}%
\end{tabular}
}{
                                    +-+
                                    |I|
+-------------+           +---------+-+-+
|             |           |             |
|             |           |             |
|             |           |             |
+-------------+           |             |
|             |           |             |
|             |           |             |
|             |           |             |
+-------------+           |             |
|         +-+ |           |         +-+ |
|         |I| |           |         |/| |
|         +-+ |           |         +-+ |
+-------------+           |             |
|             |           |             |
|             |           |             |
|             |           |    T \ I    |
+-------------+           |             |
|             |           |             |
|             |           |             |
|             |           |             |
+-------------+           +-------------+
   C rad(I)              C rad(I) rad(G(0))
}
\label{figure-n-i}
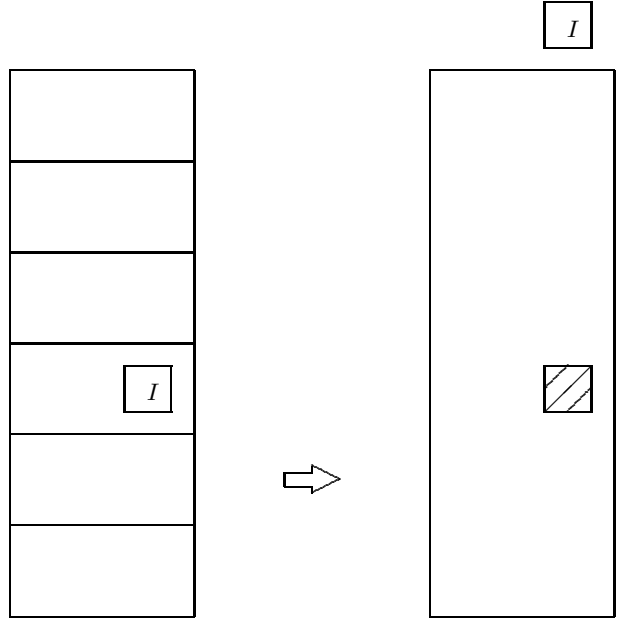
\hcaption{The first very radical revision}
\end{hfigure}

The second class of this order contains all of $G(0)$ because it comprises all
models but $I$, which is not in $G(0)$ as shown in
Figure~\ref{figure-n-i-zero}.

\begin{hfigure}
\setlength{\unitlength}{3750sp}%
\begin{picture}(1224,1074)(4789,-3973)
\thinlines
{\color[rgb]{0,0,0}\put(4801,-3961){\framebox(1200,600){}}
}%
{\color[rgb]{0,0,0}\put(5551,-3811){\framebox(300,300){}}
}%
{\color[rgb]{0,0,0}\put(5551,-3211){\framebox(300,300){}}
}%
{\color[rgb]{0,0,0}\multiput(5551,-3661)(7.89474,7.89474){20}{\makebox(2.2222,15.5556){\tiny.}}
}%
{\color[rgb]{0,0,0}\put(5551,-3811){\line( 1, 1){300}}
}%
{\color[rgb]{0,0,0}\multiput(5701,-3811)(7.89474,7.89474){20}{\makebox(2.2222,15.5556){\tiny.}}
}%
{\color[rgb]{0,0,0}\put(4951,-3811){\framebox(450,300){}}
}%
\put(5701,-3136){\makebox(0,0)[b]{\smash{\fontsize{9}{10.8}
\usefont{T1}{cmr}{m}{n}{\color[rgb]{0,0,0}$I$}%
}}}
\put(5176,-3736){\makebox(0,0)[b]{\smash{\fontsize{9}{10.8}
\usefont{T1}{cmr}{m}{n}{\color[rgb]{0,0,0}$G(0)$}%
}}}
\end{picture}%
\nop{
          +-+
          |I|
+---------+-+-+
|  +----+ +-+ |
|  |G(0)| |/| |
|  +----+ +-+ |
|    T \ I    |
+-------------+
C rad(I) rad(G(0))
  n-i-zero.fig
}%
\label{figure-n-i-zero}
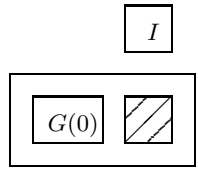
\hcaption{The first class of the target order in
	the result of the first revision}
\end{hfigure}

The second revision is $G(0)$. Since it is all contained in the second class of
the current order $C = [\{I\}, \true \backslash \{I\}]$, its minimal and maximal
classes
{} $\imin(G(0))$ and $\imax(G(0))$ are both $1$.
A consequence of the choice $I \not\in G(0)$ is
{} $\true \backslash \{I\} \cap G(0) = G(0)$.

\begin{eqnarray*}
\lefteqn{C \vrad(G(0))}
\\
&=& [
	C(\imin(G(0))) \cap G(0), \ldots, C(\imax(G(0))) \cap G(0),
	\true \backslash G(0)
]
\\
&=& [
	C(1) \cap G(0), \ldots, C(1) \cap G(0),
	\true \backslash G(0)
]
\\
&=& [
	C(1) \cap G(0),
	\true \backslash G(0)
]
\\
&=& [
	\true \backslash \{I\} \cap G(0), 
	\true \backslash G(0)
]
\\
&=& [
	G(0),
	\true \backslash G(0)
]
\end{eqnarray*}

This doxastic state believes $G(0)$ more than all other models, like the
doxastic state $G$ does. The following revisions set the strength of beliefs in
all situations of the other classes of $G$ in their order.

\begin{hfigure}
\setlength{\unitlength}{3750sp}%
\begin{picture}(1224,1524)(4789,-3973)
\thinlines
{\color[rgb]{0,0,0}\put(5551,-3811){\framebox(300,300){}}
}%
{\color[rgb]{0,0,0}\multiput(5551,-3661)(7.89474,7.89474){20}{\makebox(2.2222,15.5556){\tiny.}}
}%
{\color[rgb]{0,0,0}\put(5551,-3811){\line( 1, 1){300}}
}%
{\color[rgb]{0,0,0}\multiput(5701,-3811)(7.89474,7.89474){20}{\makebox(2.2222,15.5556){\tiny.}}
}%
{\color[rgb]{0,0,0}\put(4951,-3811){\framebox(450,300){}}
}%
{\color[rgb]{0,0,0}\put(4951,-2761){\framebox(450,300){}}
}%
{\color[rgb]{0,0,0}\put(5551,-2911){\line( 1, 0){300}}
\put(5851,-2911){\line( 0,-1){450}}
\put(5851,-3361){\line( 1, 0){150}}
\put(6001,-3361){\line( 0,-1){600}}
\put(6001,-3961){\line(-1, 0){1200}}
\put(4801,-3961){\line( 0, 1){600}}
\put(4801,-3361){\line( 1, 0){750}}
\put(5551,-3361){\line( 0, 1){450}}
}%
{\color[rgb]{0,0,0}\multiput(4951,-3661)(7.89474,7.89474){20}{\makebox(2.2222,15.5556){\tiny.}}
}%
{\color[rgb]{0,0,0}\multiput(5251,-3811)(7.89474,7.89474){20}{\makebox(2.2222,15.5556){\tiny.}}
}%
{\color[rgb]{0,0,0}\put(4951,-3811){\line( 1, 1){300}}
}%
{\color[rgb]{0,0,0}\put(5101,-3811){\line( 1, 1){300}}
}%
\put(5701,-3136){\makebox(0,0)[b]{\smash{\fontsize{9}{10.8}
\usefont{T1}{cmr}{m}{n}{\color[rgb]{0,0,0}$I$}%
}}}
\put(5176,-2686){\makebox(0,0)[b]{\smash{\fontsize{9}{10.8}
\usefont{T1}{cmr}{m}{n}{\color[rgb]{0,0,0}$G(0)$}%
}}}
\end{picture}%
\nop{
   +----+
   |G(0)|
   +----+
          +-+
          |I|
+---------+ +-+
|  +----+ +-+ |
|  |////| |/| |
|  +----+ +-+ |
|   T \ G(0)  |
+-------------+
C rad(I) rad(G(0))
   zero-i.fig
}%
\label{figure-zero-i}
\hcaption{The order after the second very radical revision}
\end{hfigure}

The current doxastic state is
{} $C = [G(0), \true \backslash G(0)]$.
The model $I$ chosen at the beginning is irrelevant from this point on. It is
no longer depicted in Figure~\ref{figure-flat-zero} and the following ones.

\begin{hfigure}
\setlength{\unitlength}{3750sp}%
\begin{picture}(1224,1074)(4789,-3973)
\thinlines
{\color[rgb]{0,0,0}\put(4951,-3811){\framebox(450,300){}}
}%
{\color[rgb]{0,0,0}\multiput(4951,-3661)(7.89474,7.89474){20}{\makebox(2.2222,15.5556){\tiny.}}
}%
{\color[rgb]{0,0,0}\multiput(5251,-3811)(7.89474,7.89474){20}{\makebox(2.2222,15.5556){\tiny.}}
}%
{\color[rgb]{0,0,0}\put(4951,-3811){\line( 1, 1){300}}
}%
{\color[rgb]{0,0,0}\put(5101,-3811){\line( 1, 1){300}}
}%
{\color[rgb]{0,0,0}\put(4801,-3961){\framebox(1200,600){}}
}%
{\color[rgb]{0,0,0}\put(4951,-3211){\framebox(450,300){}}
}%
\put(5176,-3136){\makebox(0,0)[b]{\smash{\fontsize{9}{10.8}
\usefont{T1}{cmr}{m}{n}{\color[rgb]{0,0,0}$G(0)$}%
}}}
\end{picture}%
\nop{
   +----+
   |G(0)|
   +----+
+-------------+
|  +----+     |
|  |////|     |
|  +----+     |
|   T \ G(0)  |
+-------------+
C rad(I) rad(G(0))
flat-zero.fig
}%
\label{figure-flat-zero}
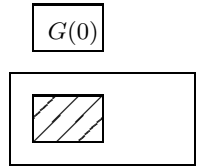
\hcaption{The order after the second very radical revision,
	different depiction}
\end{hfigure}

The following revisions are $G(0) \cup \cdots \cup G(i)$ for $i=1,\ldots,\last-1$.
The proof is based on the following induction assumption and claim:

\begin{quote}
the doxastic state produced by the revision
{} $G(0) \cup \cdots \cup G(i)$
is
{} $[G(0), \ldots, G(i), \true \backslash (G(0) \cup \cdots G(i))]$
\end{quote}

\begin{hfigure}
\setlength{\unitlength}{3750sp}%
\begin{picture}(1524,2424)(4639,-3973)
\thinlines
{\color[rgb]{0,0,0}\put(4951,-3211){\framebox(450,300){}}
}%
{\color[rgb]{0,0,0}\put(4951,-2311){\framebox(450,300){}}
}%
{\color[rgb]{0,0,0}\put(4951,-1861){\framebox(450,300){}}
}%
{\color[rgb]{0,0,0}\put(4651,-3961){\framebox(1500,600){}}
}%
\put(5176,-1786){\makebox(0,0)[b]{\smash{\fontsize{9}{10.8}
\usefont{T1}{cmr}{m}{n}{\color[rgb]{0,0,0}$G(0)$}%
}}}
\put(5176,-2236){\makebox(0,0)[b]{\smash{\fontsize{9}{10.8}
\usefont{T1}{cmr}{m}{n}{\color[rgb]{0,0,0}$G(1)$}%
}}}
\put(5176,-3136){\makebox(0,0)[b]{\smash{\fontsize{9}{10.8}
\usefont{T1}{cmr}{m}{n}{\color[rgb]{0,0,0}$G(i)$}%
}}}
\put(5176,-2686){\makebox(0,0)[b]{\smash{\fontsize{9}{10.8}
\usefont{T1}{cmr}{m}{n}{\color[rgb]{0,0,0}$\ldots$}%
}}}
\put(5401,-3736){\makebox(0,0)[b]{\smash{\fontsize{9}{10.8}
\usefont{T1}{cmr}{m}{n}{\color[rgb]{0,0,0}$G(i+1)\cup\cdots\cup G(\last)$}%
}}}
\end{picture}%
\nop{
   +----+
   |G(0)|
   +----+
   |G(1)|
   +----+
   |... |
   +----+
   |G(i)|
   +----+
+-------------+
|             |
|G(i+1)..G(o) |
|             |
|             |
+-------------+
}%
\label{figure-partial-sequence}
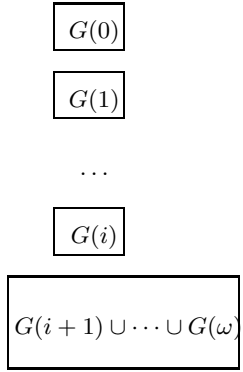
\hcaption{The order after a partial sequence of very radical revisions}
\end{hfigure}

The claim is depicted in Figure~\ref{figure-partial-sequence}.

The base case is the doxastic state $C = [\{I\}, \true \backslash \{I\}]$ and
the revision $G(0)$. The result is proved above to be $[G(0), \true \backslash
G(0)]$, which meets the assumption claim with $i=0$.

The induction case is the doxastic state
{} $C = [G(0), \ldots, G(i-1), \true \backslash (G(0) \cup \cdots \cup G(i-1))]$
revised by
{} $A = G(0) \cup \cdots \cup G(i)$.

\begin{hfigure}
\long\def\ttytex#1#2{#1}
\ttytex{
\begin{tabular}{ccc}
\setlength{\unitlength}{3750sp}%
\begin{picture}(1524,2424)(4639,-3973)
\thinlines
{\color[rgb]{0,0,0}\put(4951,-2311){\framebox(450,300){}}
}%
{\color[rgb]{0,0,0}\put(4951,-1861){\framebox(450,300){}}
}%
{\color[rgb]{0,0,0}\put(4651,-3961){\framebox(1500,600){}}
}%
{\color[rgb]{0,0,0}\put(4876,-3736){\framebox(600,300){}}
}%
{\color[rgb]{0,0,0}\put(4876,-3211){\framebox(600,300){}}
}%
\put(5176,-1786){\makebox(0,0)[b]{\smash{\fontsize{9}{10.8}
\usefont{T1}{cmr}{m}{n}{\color[rgb]{0,0,0}$G(0)$}%
}}}
\put(5176,-2236){\makebox(0,0)[b]{\smash{\fontsize{9}{10.8}
\usefont{T1}{cmr}{m}{n}{\color[rgb]{0,0,0}$G(1)$}%
}}}
\put(5176,-2686){\makebox(0,0)[b]{\smash{\fontsize{9}{10.8}
\usefont{T1}{cmr}{m}{n}{\color[rgb]{0,0,0}$\ldots$}%
}}}
\put(5401,-3886){\makebox(0,0)[b]{\smash{\fontsize{9}{10.8}
\usefont{T1}{cmr}{m}{n}{\color[rgb]{0,0,0}$G(i+1)\cup\cdots\cup G(\last)$}%
}}}
\put(5176,-3136){\makebox(0,0)[b]{\smash{\fontsize{9}{10.8}
\usefont{T1}{cmr}{m}{n}{\color[rgb]{0,0,0}$G(i\neg 1)$}%
}}}
\put(5176,-3661){\makebox(0,0)[b]{\smash{\fontsize{9}{10.8}
\usefont{T1}{cmr}{m}{n}{\color[rgb]{0,0,0}$G(i)$}%
}}}
\end{picture}%
&
\setlength{\unitlength}{3750sp}%
\begin{picture}(1104,1464)(5659,-4663)
\thinlines
{\color[rgb]{0,0,0}\multiput(6391,-3391)(-9.47368,4.73684){20}{\makebox(2.1167,14.8167){\tiny.}}
\put(6211,-3301){\line( 0,-1){ 45}}
\put(6211,-3346){\line(-1, 0){180}}
\put(6031,-3346){\line( 0,-1){ 90}}
\put(6031,-3436){\line( 1, 0){180}}
\put(6211,-3436){\line( 0,-1){ 45}}
\multiput(6211,-3481)(9.47368,4.73684){20}{\makebox(2.1167,14.8167){\tiny.}}
}%
\end{picture}%
&
\setlength{\unitlength}{3750sp}%
\begin{picture}(1524,2874)(4639,-4423)
\thinlines
{\color[rgb]{0,0,0}\put(4951,-2311){\framebox(450,300){}}
}%
{\color[rgb]{0,0,0}\put(4951,-1861){\framebox(450,300){}}
}%
{\color[rgb]{0,0,0}\put(4876,-3661){\framebox(600,300){}}
}%
{\color[rgb]{0,0,0}\put(4651,-4411){\framebox(1500,600){}}
}%
{\color[rgb]{0,0,0}\put(4876,-4186){\framebox(600,300){}}
}%
{\color[rgb]{0,0,0}\multiput(4876,-4036)(7.89474,7.89474){20}{\makebox(2.2222,15.5556){\tiny.}}
}%
{\color[rgb]{0,0,0}\put(4876,-4186){\line( 1, 1){300}}
}%
{\color[rgb]{0,0,0}\put(5026,-4186){\line( 1, 1){300}}
}%
{\color[rgb]{0,0,0}\put(5176,-4186){\line( 1, 1){300}}
}%
{\color[rgb]{0,0,0}\multiput(5326,-4186)(7.89474,7.89474){20}{\makebox(2.2222,15.5556){\tiny.}}
}%
{\color[rgb]{0,0,0}\put(4876,-3211){\framebox(600,300){}}
}%
\put(5176,-1786){\makebox(0,0)[b]{\smash{\fontsize{9}{10.8}
\usefont{T1}{cmr}{m}{n}{\color[rgb]{0,0,0}$G(0)$}%
}}}
\put(5176,-2236){\makebox(0,0)[b]{\smash{\fontsize{9}{10.8}
\usefont{T1}{cmr}{m}{n}{\color[rgb]{0,0,0}$G(1)$}%
}}}
\put(5176,-2686){\makebox(0,0)[b]{\smash{\fontsize{9}{10.8}
\usefont{T1}{cmr}{m}{n}{\color[rgb]{0,0,0}$\ldots$}%
}}}
\put(5401,-4336){\makebox(0,0)[b]{\smash{\fontsize{9}{10.8}
\usefont{T1}{cmr}{m}{n}{\color[rgb]{0,0,0}$G(i+1)\cup\cdots\cup G(\last)$}%
}}}
\put(5176,-3136){\makebox(0,0)[b]{\smash{\fontsize{9}{10.8}
\usefont{T1}{cmr}{m}{n}{\color[rgb]{0,0,0}$G(i\neg 1)$}%
}}}
\put(5176,-3586){\makebox(0,0)[b]{\smash{\fontsize{9}{10.8}
\usefont{T1}{cmr}{m}{n}{\color[rgb]{0,0,0}$G(i)$}%
}}}
\end{picture}%
\end{tabular}
}{
   +------+                  +------+
   | G(0) |                  | G(0) |
   +------+                  +------+
   | G(1) |                  | G(1) |
   +------+                  +------+
   | ...  |                  | ...  |
   +------+                  +------+
   |G(i-1)|          ==>     |G(i-1)|
   +------+                  +------+
+-------------+              | G(i) |
|  +------+   |              +------+
|  | G(i) |   |           +-------------+
|  +------+   |           |  +------+   |
|G(i-1)..G(o) |           |  |//////|   |
+-------------+           |  +------+   |
                          | G(i)..G(o)  |
                          +-------------+
 partial-i.fig       partial-i-veryradical.fig
}
\label{figure-partial-i-veryradical}
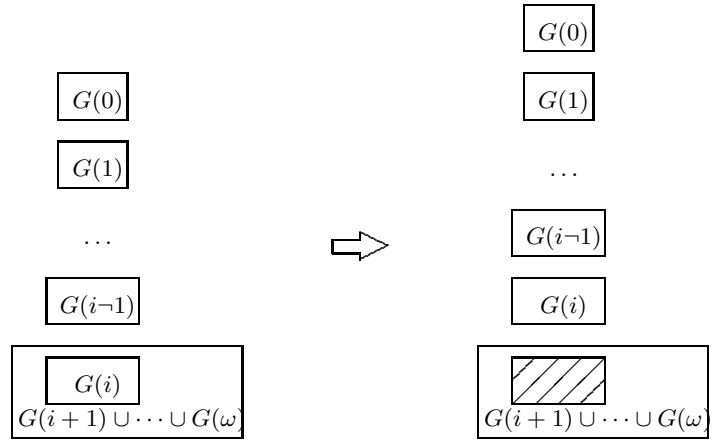
\hcaption{Yet another very radical revision}
\end{hfigure}

\long\def\ttytex#1#2{#1}
\ttytex{
}{
}

Since $G$ is a partition, the union of its classes comprises all models:
{} $G(0) \cup \cdots \cup G(\last) = \true$.
This equation is the same as 
{} $G(0) \cup \cdots G(i-1) \cup G(i) \cup G(\last) = \true$.
A consequence is the expression
{} $\true \backslash G(0) \cup \cdots \cup G(i-1) =
    G(i) \cup G(i+1) \cup \cdots \cup G(\last)$
of the last class of the current order $C$.

The intersections of $A = G(0) \cup \cdots \cup G(i)$ with the classes
{} $G(0), \ldots, G(i-1), \true \backslash (G(0) \cup \cdots \cup G(i-1))$
of the current order $C$ are therefore
{} $G(0), \ldots, G(i-1), G(i)$.

Since these classes are not empty, they are the minimal and maximal classes of
the revision:
{}	$\imin(A) = 0$
and
{}	$\imax(A) = i$.

\begin{eqnarray*}
\lefteqn{C \vrad(A)}
\\
&=& [
	C(\imin(A)) \cap A, \ldots, C(\imax(A)) \cap A,
	\true \backslash A
]
\\
&=& [
	C(0) \cap A, \ldots, C(i) \cap A
	\true \backslash A
]
\\
&=& [
	G(0) \cap A, \ldots, G(i-1) \cap A,
	(G(i) \cup \cdots \cup G(\last)) \cap A,
	\true \backslash A
]
\\
&=& [
	G(0), \ldots, G(i-1),
	G(i),
	\true \backslash (G(0) \cup \cdots \cup G(i)
]
\end{eqnarray*}

This doxastic state meets the induction claim.

\

Because of the induction claim, the last revision
{} $G(0) \cup \cdots \cup G(\last-1)$
produces the following order.

\begin{eqnarray*}
\lefteqn{[G(0), \ldots, G(\last-1), \true \backslash (G(0) \cup \cdots G(\last-1)]}
\\
&=&
	[G(0), \ldots, G(\last-1), G(\last) \cup \cdots \cup G(\last)]
\\
&=&
	[G(0), \ldots, G(\last-1), G(\last)]
\\
&=&
	G
\end{eqnarray*}

This is the target order in the claim of the lemma.~\qed

Apart from the unnecessary condition that the revisions are independent of the
initial doxastic state, this lemma proves very radical revision plastic.

\begin{theorem}
\label{theorem-plastic-veryradical}

Very radical revision is plastic.

\end{theorem}

\proof The previous Lemma~\ref{lemma-plastic-veryradical} proves that every
doxastic state is the result of a sequence of very radical revisions applied to
a non-empty doxastic state. This is the plastic ability.~\qed

\

Very radical revision is not amnesic and therefore not fully plastic either.

\begin{theorem}
\label{theorem-veryradical-amnesic}

Very radical revision is not amnesic.

\end{theorem}

\proof Very radical revision has the two required properties of
Theorem~\ref{theorem-amnesic}:
{} $C \vrad(A)(0) \models A$
and
{} $C \vrad(\true) = C$.

The first property holds because the first class of $C \vrad(A)$ is
$C(\imin(A)) \cap A$, which is a subset of $A$.

The proof of the second property follows from
{}	$\imin(\true) = 0$
and
{}	$\imax(\true) = \last$.

\begin{eqnarray*}
\lefteqn{C \vrad(\true)}
\\
&=& [
	C(\imin(\true)) \cap \true, \ldots, C(\imax(\true)) \cap \true,
	\true \backslash \true
]
\\
&=& [
	C(0) \cap \true, \ldots, C(\last) \cap \true,
	\emptyset
]
\\
&=& [
	C(0), \ldots, C(\last)
]
\\
&=&
	C
\end{eqnarray*}~\qed

\begin{corollary}
\label{corollary-veryradical-full}

Very radical revision is not fully plastic.

\end{corollary}

\section{Full meet revision}
\label{section-fullmeet}

\draft

defined in mixed and in icdt-97

\begin{hfigure}
\long\def\ttytex#1#2{#1}
\ttytex{
\begin{tabular}{ccc}
\setlength{\unitlength}{3750sp}%
\begin{picture}(1224,1824)(5089,-4873)
\thinlines
{\color[rgb]{0,0,0}\put(5101,-3361){\line( 1, 0){1200}}
}%
{\color[rgb]{0,0,0}\put(5101,-4861){\framebox(1200,1800){}}
}%
{\color[rgb]{0,0,0}\put(5101,-3961){\line( 1, 0){1200}}
}%
{\color[rgb]{0,0,0}\put(5101,-4261){\line( 1, 0){1200}}
}%
{\color[rgb]{0,0,0}\put(5101,-4561){\line( 1, 0){1200}}
}%
{\color[rgb]{0,0,0}\put(5401,-4486){\framebox(750,1050){}}
}%
{\color[rgb]{0,0,0}\put(5101,-3661){\line( 1, 0){1200}}
}%
\put(5851,-3886){\makebox(0,0)[b]{\smash{\fontsize{9}{10.8}
\usefont{T1}{cmr}{m}{n}{\color[rgb]{0,0,0}$A$}%
}}}
\end{picture}%
&
\setlength{\unitlength}{3750sp}%
\begin{picture}(1104,1104)(5659,-4303)
\thinlines
{\color[rgb]{0,0,0}\multiput(6391,-3391)(-9.47368,4.73684){20}{\makebox(2.1167,14.8167){\tiny.}}
\put(6211,-3301){\line( 0,-1){ 45}}
\put(6211,-3346){\line(-1, 0){180}}
\put(6031,-3346){\line( 0,-1){ 90}}
\put(6031,-3436){\line( 1, 0){180}}
\put(6211,-3436){\line( 0,-1){ 45}}
\multiput(6211,-3481)(9.47368,4.73684){20}{\makebox(2.1167,14.8167){\tiny.}}
}%
\end{picture}%
&
\setlength{\unitlength}{3750sp}%
\begin{picture}(1224,2049)(5089,-4873)
\thinlines
{\color[rgb]{0,0,0}\put(5101,-4861){\framebox(1200,1800){}}
}%
{\color[rgb]{0,0,0}\put(5401,-3061){\framebox(750,225){}}
}%
{\color[rgb]{0,0,0}\put(5401,-3661){\framebox(750,225){}}
}%
{\color[rgb]{0,0,0}\multiput(5401,-3511)(8.33333,8.33333){10}{\makebox(2.2222,15.5556){\tiny.}}
}%
{\color[rgb]{0,0,0}\multiput(6076,-3661)(8.33333,8.33333){10}{\makebox(2.2222,15.5556){\tiny.}}
}%
{\color[rgb]{0,0,0}\multiput(5401,-3586)(7.89474,7.89474){20}{\makebox(2.2222,15.5556){\tiny.}}
}%
{\color[rgb]{0,0,0}\multiput(5401,-3661)(8.03571,8.03571){29}{\makebox(2.2222,15.5556){\tiny.}}
}%
{\color[rgb]{0,0,0}\multiput(5476,-3661)(8.03571,8.03571){29}{\makebox(2.2222,15.5556){\tiny.}}
}%
{\color[rgb]{0,0,0}\multiput(5551,-3661)(8.03571,8.03571){29}{\makebox(2.2222,15.5556){\tiny.}}
}%
{\color[rgb]{0,0,0}\multiput(5626,-3661)(8.03571,8.03571){29}{\makebox(2.2222,15.5556){\tiny.}}
}%
{\color[rgb]{0,0,0}\multiput(5701,-3661)(8.03571,8.03571){29}{\makebox(2.2222,15.5556){\tiny.}}
}%
{\color[rgb]{0,0,0}\multiput(5776,-3661)(8.03571,8.03571){29}{\makebox(2.2222,15.5556){\tiny.}}
}%
{\color[rgb]{0,0,0}\multiput(5851,-3661)(8.03571,8.03571){29}{\makebox(2.2222,15.5556){\tiny.}}
}%
{\color[rgb]{0,0,0}\multiput(5926,-3661)(8.03571,8.03571){29}{\makebox(2.2222,15.5556){\tiny.}}
}%
{\color[rgb]{0,0,0}\multiput(6001,-3661)(7.89474,7.89474){20}{\makebox(2.2222,15.5556){\tiny.}}
}%
\end{picture}%
\end{tabular}
}{
                             +------+ min(A)
                             +------+
+-------------+          +-------------+
|             |          |             |
+-------------+          |             |
|   +------+  |          |   ////////  |
+---|------|--+          |   ////////  |
|   |   A  |  |    =>    |             |
+---|------|--+          |             |
|   |      |  |          |             |
+---|------|--+          |             |
|   +------+  |          |             |
+-------------+          |             |
|             |          |             |
+-------------+          |             |
|             |          |             |
+-------------+          |             |
|             |          |             |
+-------------+          +-------------+
  current.fig             fullmeet.fig
}
\hcaption{full meet revision}
\end{hfigure}

contingent, contingent version di very radical

maximally epiphanic

\enddraft

Full meet revision~\cite{alch-gard-maki-85} believes only the most believed
situations supported by the new information, and nothing else. Its original
definition on plain belief bases extends this principle to full doxastic states
by disbelieving all other situations the same.

\begin{definition}
\label{definition-fullmeet}

\[
C \full(A) = [
	\min(A),
	\true \backslash \min(A)
]
\]

\end{definition}

Full meet revision is not equating only in the corner case of an alphabet of
one variable.

\begin{theorem}
\label{theorem-fullmeet-amnesic}

Full meet revisions is amnesic if and only if the alphabet comprises at least
two symbols.

\end{theorem}

\proof If the alphabet comprises one variable $a$ only, the models are only
two: $a$ is true in $I$ and false in $J$. The equating property requires full
meet revision to turn $I <_C J$ into $I \equiv_{C \full(A)} J$. The strict
comparison $I <_C J$ implies $C = [\{I\},\{J\}]$. The equality is proved false
for the three possible values of a consistent formula $A$ of a single variable.

\begin{description}

\item[$A = \{I\}$]

\begin{eqnarray*}
\lefteqn{C \full(\{I\})}
\\
&=& [
	\min(\{I\}),
	\true \backslash \min(\{I\})
]
\\
&=& [
	\{I\},
	\true \backslash \{I\}
]
\\
&=& [
	\{I\},
	\{J\}
]
\end{eqnarray*}

\item[$A = \{J\}$]

\begin{eqnarray*}
\lefteqn{C \full(\{J\})}
\\
&=& [
	\min(\{J\}),
	\true \backslash \min(\{J\})
]
\\
&=& [
	\{J\},
	\true \backslash \{J\}
]
\\
&=& [
	\{J\},
	\{I\}
]
\end{eqnarray*}

\item[$A = \{I,J\}$]

\begin{eqnarray*}
\lefteqn{C \full(\{I,J\})}
\\
&=& [
	\min(\{I,J\}),
	\true \backslash \min(\{I,J\})
]
\\
&=& [
	\{I\},
	\true \backslash \{I\}
]
\\
&=& [
	\{I\},
	\{J\}
]
\end{eqnarray*}

\end{description}

In none of the possible outcomes $I$ is in the same class of $J$.

\

Two variables have four models. More variables have more models. Equating $I$
and $J$ is the result of a full meet revision by a different model $K$.

\begin{eqnarray*}
\lefteqn{C \full(\{K\})}
\\
&=& [
	\min(\{K\}),
	\true \backslash \min(\{K\})
]
\\
&=& [
	\{K\},
	\true \backslash \{K\}
]
\end{eqnarray*}

The second class of $C \full(\{K\})$ contains both $I$ and $J$ since both
differ from $K$. This proves that they are equal no matter of how $C$ sorts
them.~\qed

Full meet revision only separates situations in two classes. It never produces
an order of three or more.

\begin{theorem}
\label{theorem-fullmeet-learnable}

Full meet revision is not learnable.

\end{theorem}

\proof By definition, full meet revision only produces orders of two classes:
$C \full(A)$ comprises the classes $\min(A)$ and $\true \backslash \min(A)$.
Learnability requires the generation of every order, including orders of three
classes like $[a, \neg a b, \neg a \neg b]$.~\qed

Full meet revision is correcting: it can invert any order.

\begin{theorem}
\label{theorem-fullmeet-correcting}

Full meet revision is correcting.

\end{theorem}

\proof Full meet revision makes $I$ less than $J$ by a revision $I$. A
singleton $\{I\}$ is always all minimal.

\begin{eqnarray*}
\lefteqn{C \full(\{I\})}
\\
&=& [
	\min(\{I\}),
	\true \backslash \min(\{I\})
]
\\
&=& [
	\{I\},
	\true \backslash \{I\}
]
\end{eqnarray*}

Since $J$ is less than $I$ in $C$, it differs from $I$. It is therefore in the
second class $\true \backslash \{I\}$ of $C \full(\{I\})$.~\qed

Full meet revision can invert the order between two arbitrary models, but not
among all models.

\begin{theorem}
\label{theorem-fullmeet-damascan}

Full meet revision is not damascan.

\end{theorem}

\proof Full meet revision never generates the opposite of a three-class order
because it always separates models in two classes: $C \full(A)$ comprises
$\min(A)$ and $\true \backslash \min(A)$.~\qed

Full meet revision is dogmatic and therefore believer.

\begin{theorem}
\label{theorem-fullmeet-dogmatic}

Full meet revision is dogmatic.

\end{theorem}

\proof Dogmatic is generating a given two-class order $[A,\neg A]$. Full meet
revision can do that, but not in a single step if the models of $A$ are not
already in the same class. Otherwise, it first needs to equate them.

\begin{hfigure}
\long\def\ttytex#1#2{#1}
\ttytex{
\begin{tabular}{ccc}
\setlength{\unitlength}{3750sp}%
\begin{picture}(1224,3624)(4789,-5173)
\thinlines
{\color[rgb]{0,0,0}\put(4801,-3361){\line( 1, 0){1200}}
}%
{\color[rgb]{0,0,0}\put(4801,-2161){\line( 1, 0){1200}}
}%
{\color[rgb]{0,0,0}\put(4801,-4561){\line( 1, 0){1200}}
}%
{\color[rgb]{0,0,0}\put(4801,-5161){\framebox(1200,3600){}}
}%
{\color[rgb]{0,0,0}\put(4801,-3961){\line( 1, 0){1200}}
}%
{\color[rgb]{0,0,0}\put(4801,-2761){\line( 1, 0){1200}}
}%
{\color[rgb]{0,0,0}\put(5551,-3811){\framebox(300,300){}}
}%
\put(5701,-3736){\makebox(0,0)[b]{\smash{\fontsize{9}{10.8}
\usefont{T1}{cmr}{m}{n}{\color[rgb]{0,0,0}$I$}%
}}}
\end{picture}%
&
\setlength{\unitlength}{3750sp}%
\begin{picture}(1104,1104)(5659,-4303)
\thinlines
{\color[rgb]{0,0,0}\multiput(6391,-3391)(-9.47368,4.73684){20}{\makebox(2.1167,14.8167){\tiny.}}
\put(6211,-3301){\line( 0,-1){ 45}}
\put(6211,-3346){\line(-1, 0){180}}
\put(6031,-3346){\line( 0,-1){ 90}}
\put(6031,-3436){\line( 1, 0){180}}
\put(6211,-3436){\line( 0,-1){ 45}}
\multiput(6211,-3481)(9.47368,4.73684){20}{\makebox(2.1167,14.8167){\tiny.}}
}%
\end{picture}%
&
\setlength{\unitlength}{3750sp}%
\begin{picture}(1224,4074)(4789,-5173)
\thinlines
{\color[rgb]{0,0,0}\put(4801,-5161){\framebox(1200,3600){}}
}%
{\color[rgb]{0,0,0}\put(5551,-3811){\framebox(300,300){}}
}%
{\color[rgb]{0,0,0}\put(5551,-1411){\framebox(300,300){}}
}%
{\color[rgb]{0,0,0}\multiput(5551,-3661)(7.89474,7.89474){20}{\makebox(2.2222,15.5556){\tiny.}}
}%
{\color[rgb]{0,0,0}\put(5551,-3811){\line( 1, 1){300}}
}%
{\color[rgb]{0,0,0}\multiput(5701,-3811)(7.89474,7.89474){20}{\makebox(2.2222,15.5556){\tiny.}}
}%
\put(5701,-1336){\makebox(0,0)[b]{\smash{\fontsize{9}{10.8}
\usefont{T1}{cmr}{m}{n}{\color[rgb]{0,0,0}$I$}%
}}}
\end{picture}%
\end{tabular}
}{
                                    +-+
                                    |I|
+-------------+           +---------+-+-+
|             |           |             |
|             |           |             |
|             |           |             |
+-------------+           |             |
|             |           |             |
|             |           |             |
|             |           |             |
+-------------+           |             |
|         +-+ |           |         +-+ |
|         |I| |           |         |/| |
|         +-+ |           |         +-+ |
+-------------+           |             |
|             |           |             |
|             |           |             |
|             |           |    T \ I    |
+-------------+           |             |
|             |           |             |
|             |           |             |
|             |           |             |
+-------------+           +-------------+
      S                      S full(I)
}
\label{figure-n-i-full}
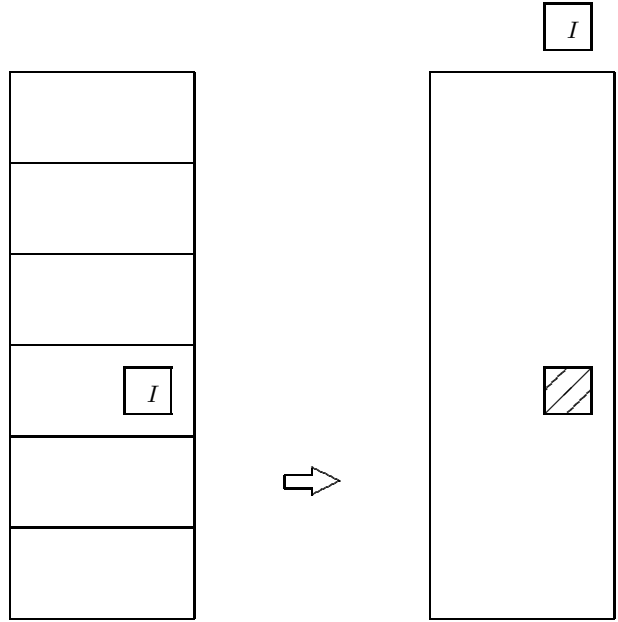
\hcaption{Full meet revision of a single model}
\end{hfigure}

The first step is achieved by revising the current order $C$ by a model $I$
that falsifies $A$. Being a single model, its only minimal is itself:
{} $\min(\{I\}) = \{I\}$.

\begin{eqnarray*}
\lefteqn{C \full(\{I\})}
\\
&=& [
	\min(A),
	\true \backslash \min(\{\{I\})
]
\\
&=& [
	I,
	\true \backslash I
]
\end{eqnarray*}

Since $I$ is not a model of $A$, all models of $A$ are in the second class
$\true \backslash I$ of this order.

\begin{hfigure}
\long\def\ttytex#1#2{#1}
\ttytex{
\begin{tabular}{ccc}
\setlength{\unitlength}{3750sp}%
\begin{picture}(1224,4074)(4789,-5173)
\thinlines
{\color[rgb]{0,0,0}\put(4801,-5161){\framebox(1200,3600){}}
}%
{\color[rgb]{0,0,0}\put(5551,-3811){\framebox(300,300){}}
}%
{\color[rgb]{0,0,0}\put(5551,-1411){\framebox(300,300){}}
}%
{\color[rgb]{0,0,0}\multiput(5551,-3661)(7.89474,7.89474){20}{\makebox(2.2222,15.5556){\tiny.}}
}%
{\color[rgb]{0,0,0}\put(5551,-3811){\line( 1, 1){300}}
}%
{\color[rgb]{0,0,0}\multiput(5701,-3811)(7.89474,7.89474){20}{\makebox(2.2222,15.5556){\tiny.}}
}%
{\color[rgb]{0,0,0}\put(4951,-4411){\framebox(600,300){}}
}%
\put(5701,-1336){\makebox(0,0)[b]{\smash{\fontsize{9}{10.8}
\usefont{T1}{cmr}{m}{n}{\color[rgb]{0,0,0}$I$}%
}}}
\put(5251,-4336){\makebox(0,0)[b]{\smash{\fontsize{9}{10.8}
\usefont{T1}{cmr}{m}{n}{\color[rgb]{0,0,0}$A$}%
}}}
\end{picture}%
&
\setlength{\unitlength}{3750sp}%
\begin{picture}(1104,1104)(5659,-4303)
\thinlines
{\color[rgb]{0,0,0}\multiput(6391,-3391)(-9.47368,4.73684){20}{\makebox(2.1167,14.8167){\tiny.}}
\put(6211,-3301){\line( 0,-1){ 45}}
\put(6211,-3346){\line(-1, 0){180}}
\put(6031,-3346){\line( 0,-1){ 90}}
\put(6031,-3436){\line( 1, 0){180}}
\put(6211,-3436){\line( 0,-1){ 45}}
\multiput(6211,-3481)(9.47368,4.73684){20}{\makebox(2.1167,14.8167){\tiny.}}
}%
\end{picture}%
&
\setlength{\unitlength}{3750sp}%
\begin{picture}(1224,4524)(4789,-5173)
\thinlines
{\color[rgb]{0,0,0}\put(5551,-3811){\framebox(300,300){}}
}%
{\color[rgb]{0,0,0}\multiput(5551,-3661)(7.89474,7.89474){20}{\makebox(2.2222,15.5556){\tiny.}}
}%
{\color[rgb]{0,0,0}\put(5551,-3811){\line( 1, 1){300}}
}%
{\color[rgb]{0,0,0}\multiput(5701,-3811)(7.89474,7.89474){20}{\makebox(2.2222,15.5556){\tiny.}}
}%
{\color[rgb]{0,0,0}\put(4951,-4411){\framebox(600,300){}}
}%
{\color[rgb]{0,0,0}\put(4951,-961){\framebox(600,300){}}
}%
{\color[rgb]{0,0,0}\put(4951,-4411){\line( 1, 1){300}}
}%
{\color[rgb]{0,0,0}\put(5101,-4411){\line( 1, 1){300}}
}%
{\color[rgb]{0,0,0}\put(5251,-4411){\line( 1, 1){300}}
}%
{\color[rgb]{0,0,0}\multiput(5401,-4411)(7.89474,7.89474){20}{\makebox(2.2222,15.5556){\tiny.}}
}%
{\color[rgb]{0,0,0}\multiput(4951,-4261)(7.89474,7.89474){20}{\makebox(2.2222,15.5556){\tiny.}}
}%
{\color[rgb]{0,0,0}\put(5551,-1111){\line( 1, 0){300}}
\put(5851,-1111){\line( 0,-1){450}}
\put(5851,-1561){\line( 1, 0){150}}
\put(6001,-1561){\line( 0,-1){3600}}
\put(6001,-5161){\line(-1, 0){1200}}
\put(4801,-5161){\line( 0, 1){3600}}
\put(4801,-1561){\line( 1, 0){750}}
\put(5551,-1561){\line( 0, 1){450}}
}%
\put(5701,-1336){\makebox(0,0)[b]{\smash{\fontsize{9}{10.8}
\usefont{T1}{cmr}{m}{n}{\color[rgb]{0,0,0}$I$}%
}}}
\put(5251,-886){\makebox(0,0)[b]{\smash{\fontsize{9}{10.8}
\usefont{T1}{cmr}{m}{n}{\color[rgb]{0,0,0}$A$}%
}}}
\end{picture}%
\end{tabular}
}{
          +-+                 +-----+
          |I|                 |  A  |
          +-+                 +-----+
+-------------+           +-------------+
|             |           |             |
|             |           |             |
|             |           |             |
|             |           |             |
|             |           |             |
|             |           |             |
|             |           |             |
|             |           |             |
|         +-+ |           |             |
|         |/| |    ==>    |          I  |
|         +-+ |           |             |
|             |           |             |
|    T \ I    |           |  T \ A      |
|             |           |             |
|   +-----+   |           |   +-----+   |
|   |  A  |   |           |   |/////|   |
|   +-----+   |           |   +-----+   |
|             |           |             |
|             |           |             |
+-------------+           +-------------+
   C full(I)            C full(I) full(G(0)
    i-a.fig                  a-true.fig
}
\label{figure-i-a}
\hcaption{Full meet revision of $A$}
\end{hfigure}

The second revision is $A$. Since all its models are in $\true \backslash
\{I\}$, all its models are minimal:%
{} $\min(A) = A$.

\begin{eqnarray*}
\lefteqn{[I, \true \backslash I] \full(A)}
\\
&=& [
	\min(A),
	\true \backslash \min(A)
]
\\
&=& [
	A,
	\true \backslash A
]
\\
&=& [
	A,
	\neg A
]
\end{eqnarray*}

This is the required order $[A, \neg A]$.~\qed

\section{Severe revision}
\label{section-severe}

\draft

definition in rott-09:

\[
A | h_{>\neg A}
\]

translation as classes

\begin{enumerate}

\item significant example

\v
+-------------+
|C0           | S1        h1 h2 h3 h4
+-------------+
|C1 +------+  | S2           h2 h3 h4
+---|------|--+
|C2 |   A  |  | S3              h3 h4
+---|------|--+
|C3 +------+  | S4                 h4
+-------------+
|C4           |
+-------------+
\vv

\item calculations

the largest sphere implicating $\neg A$ is $S_1$; therefore, $i=1$

in a different way: the first class consistent with $A$ is $C(1)$; therefore,
$i=1$

\[
A | h_2 h_3 h_4
\]

\item translation

\v
R1 =    A h2 h3 h4  =  A S2           ==>  G0 = R1 - 0
                                              = A S2 = A C1
                                              = A (C0 u C1)
                                              = A C1

R2 =      h2 h3 h4  =  S2             ==>  G1 = R2 - R1
                                              = S2 - A S2
                                              = S2 (T - A)
                                              = (C0 u C1) - A

R3 =         h3 h4  =  S3             ==>  G2 = R3 - R2
                                              = S3 - S2
                                              = C2

R3 =            h4  =  S4             ==>  G4 = R4 - R3
                                              = C3

                                      ==>  G5 = T - R4
                                              = C4
\vv

\item drawing of classes $[A C(1), (C(0) \cup C(1)) \backslash A, C(2), C(3), C(4)]$

\v
                              +------+ A C1
                              +------+
                          +-------------+
                          |             |
+-------------+           |             |  (C0 u C1) - A
|C0           |           |   +------+  |
+-------------+           +---+      +--+
|C1 +------+  |
+---|------|--+           +---+------+--+
|C2 |   A  |  |    ==>    |   |   A  |  |  C2
+---|------|--+           +---|------|--+
|C3 +------+  |           |   +------+  |  C3
+-------------+           +-------------+
|C4           |           |             |  C4
+-------------+           +-------------+
\vv

\end{enumerate}

\separator

severe revision lifts $\min(A)$ to the top and merges all classes from the
previous class zero to the class that contained $\min(A)$

\begin{hfigure}
\long\def\ttytex#1#2{#1}
\ttytex{
\begin{tabular}{ccc}
\setlength{\unitlength}{3750sp}%
\begin{picture}(1224,1824)(5089,-4873)
\thinlines
{\color[rgb]{0,0,0}\put(5101,-3361){\line( 1, 0){1200}}
}%
{\color[rgb]{0,0,0}\put(5101,-4861){\framebox(1200,1800){}}
}%
{\color[rgb]{0,0,0}\put(5101,-3961){\line( 1, 0){1200}}
}%
{\color[rgb]{0,0,0}\put(5101,-4261){\line( 1, 0){1200}}
}%
{\color[rgb]{0,0,0}\put(5101,-4561){\line( 1, 0){1200}}
}%
{\color[rgb]{0,0,0}\put(5401,-4486){\framebox(750,1050){}}
}%
{\color[rgb]{0,0,0}\put(5101,-3661){\line( 1, 0){1200}}
}%
\put(5851,-3886){\makebox(0,0)[b]{\smash{\fontsize{9}{10.8}
\usefont{T1}{cmr}{m}{n}{\color[rgb]{0,0,0}$A$}%
}}}
\end{picture}%
&
\setlength{\unitlength}{3750sp}%
\begin{picture}(1104,1104)(5659,-4303)
\thinlines
{\color[rgb]{0,0,0}\multiput(6391,-3391)(-9.47368,4.73684){20}{\makebox(2.1167,14.8167){\tiny.}}
\put(6211,-3301){\line( 0,-1){ 45}}
\put(6211,-3346){\line(-1, 0){180}}
\put(6031,-3346){\line( 0,-1){ 90}}
\put(6031,-3436){\line( 1, 0){180}}
\put(6211,-3436){\line( 0,-1){ 45}}
\multiput(6211,-3481)(9.47368,4.73684){20}{\makebox(2.1167,14.8167){\tiny.}}
}%
\end{picture}%
&
\setlength{\unitlength}{3750sp}%
\begin{picture}(1224,2049)(5089,-4873)
\thinlines
{\color[rgb]{0,0,0}\put(5101,-3961){\line( 1, 0){1200}}
}%
{\color[rgb]{0,0,0}\put(5101,-4261){\line( 1, 0){1200}}
}%
{\color[rgb]{0,0,0}\put(5101,-4561){\line( 1, 0){1200}}
}%
{\color[rgb]{0,0,0}\put(5101,-4861){\framebox(1200,1200){}}
}%
{\color[rgb]{0,0,0}\put(5101,-3661){\line( 0, 1){600}}
\put(5101,-3061){\line( 1, 0){1200}}
\put(6301,-3061){\line( 0,-1){600}}
\put(6301,-3661){\line(-1, 0){150}}
\put(6151,-3661){\line( 0, 1){225}}
\put(6151,-3436){\line(-1, 0){750}}
\put(5401,-3436){\line( 0,-1){225}}
\put(5401,-3661){\line(-1, 0){300}}
}%
{\color[rgb]{0,0,0}\put(5401,-3061){\framebox(750,225){}}
}%
{\color[rgb]{0,0,0}\put(5401,-4486){\framebox(750,825){}}
}%
{\color[rgb]{0,0,0}\multiput(5401,-3511)(8.33333,8.33333){10}{\makebox(2.2222,15.5556){\tiny.}}
}%
{\color[rgb]{0,0,0}\multiput(5401,-3586)(7.89474,7.89474){20}{\makebox(2.2222,15.5556){\tiny.}}
}%
{\color[rgb]{0,0,0}\multiput(5401,-3661)(8.03571,8.03571){29}{\makebox(2.2222,15.5556){\tiny.}}
}%
{\color[rgb]{0,0,0}\multiput(6076,-3661)(8.33333,8.33333){10}{\makebox(2.2222,15.5556){\tiny.}}
}%
{\color[rgb]{0,0,0}\multiput(6001,-3661)(7.89474,7.89474){20}{\makebox(2.2222,15.5556){\tiny.}}
}%
{\color[rgb]{0,0,0}\multiput(5476,-3661)(8.03571,8.03571){29}{\makebox(2.2222,15.5556){\tiny.}}
}%
{\color[rgb]{0,0,0}\multiput(5551,-3661)(8.03571,8.03571){29}{\makebox(2.2222,15.5556){\tiny.}}
}%
{\color[rgb]{0,0,0}\multiput(5626,-3661)(8.03571,8.03571){29}{\makebox(2.2222,15.5556){\tiny.}}
}%
{\color[rgb]{0,0,0}\multiput(5701,-3661)(8.03571,8.03571){29}{\makebox(2.2222,15.5556){\tiny.}}
}%
{\color[rgb]{0,0,0}\multiput(5776,-3661)(8.03571,8.03571){29}{\makebox(2.2222,15.5556){\tiny.}}
}%
{\color[rgb]{0,0,0}\multiput(5851,-3661)(8.03571,8.03571){29}{\makebox(2.2222,15.5556){\tiny.}}
}%
{\color[rgb]{0,0,0}\multiput(5926,-3661)(8.03571,8.03571){29}{\makebox(2.2222,15.5556){\tiny.}}
}%
\end{picture}%
\end{tabular}
}{
                              +------+ min(A)
                              +------+
                                                        \space
+-------------+           +---+------+--+
|             |           |             |
+-------------+           |             |
|   +------+  |           |   +------+  |
+---|------|--+           +---+      +--+
|   |   A  |  |
+---|------|--+           +---+------+--+
|   |      |  |           |   |      |  |
+---|------|--+           +---|------|--+
|   +------+  |           |   |      |  |
+-------------+           +---|------|--+
|             |           |   +------+  |
+-------------+           +-------------+
|             |           |             |
+-------------+           +-------------+
|             |           |             |
+-------------+           +-------------+
                          |             |
                          +-------------+
  current.fig               severe.fig
}%
\label{figure-severe-revision}
\hcaption{severe revision}
\end{hfigure}

\enddraft

Severe revision believes the new information only in the current situation, not
in all possible situations. The acquisition sparks doubts on the previous
beliefs. Not all of them, however. Only the ones previously believed as much as
the new ones.

\begin{definition}
\label{definition-severe}

\[
C \sev(A) = [
	\min(A),
	C(0) \cup \cdots \cup C(\imin(A)) \backslash A,
	C(\imin(A)+1),
	\ldots,
	C(\last)
]
\]

\end{definition}

Severe revision is plastic like very radical revision. The proof however
requires knowledge of the order for its first step, which flattens the order by
revising it by a model of its last class.

\begin{lemma}
\label{lemma-plastic-severe}

For every order $S$, a sequence of severe revisions each revising an order
containing it in a single class turns $S$ into an arbitrary non-flat order $G$.

\end{lemma}

\proof
\displaytext{summary}
In summary, the first two revisions flatten most of the order, the following
isolates the last class of the target order, the following do the same with the
others.

\begin{hfigure}
\setlength{\unitlength}{3750sp}%
\begin{picture}(1224,1074)(4789,-4423)
\thinlines
{\color[rgb]{0,0,0}\put(4801,-3961){\framebox(1200,600){}}
}%
{\color[rgb]{0,0,0}\put(4876,-3811){\framebox(450,300){}}
}%
{\color[rgb]{0,0,0}\multiput(4876,-3661)(7.89474,7.89474){20}{\makebox(2.2222,15.5556){\tiny.}}
}%
{\color[rgb]{0,0,0}\put(4876,-3811){\line( 1, 1){300}}
}%
{\color[rgb]{0,0,0}\put(5026,-3811){\line( 1, 1){300}}
}%
{\color[rgb]{0,0,0}\multiput(5176,-3811)(7.89474,7.89474){20}{\makebox(2.2222,15.5556){\tiny.}}
}%
{\color[rgb]{0,0,0}\put(4876,-4411){\framebox(450,300){}}
}%
\put(5101,-4336){\makebox(0,0)[b]{\smash{\fontsize{9}{10.8}
\usefont{T1}{cmr}{m}{n}{\color[rgb]{0,0,0}$G(\last)$}%
}}}
\end{picture}%
\nop{
+-------------+
|       +--+  |
|       |//|  |
|       +--+  |
|   T \ G(o)  |
+-------------+
        +--+
        |Go|
        +--+
}%
\label{figure-true-last}
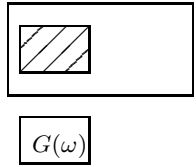
\hcaption{The result of the first revisions}
\end{hfigure}

The result of the first revision has $G(\last)$ in its target position, the
last, as shown in Figure~\ref{figure-true-last}. The following steps do the
same with $G(\last-1)$, $G(\last-2)$, $G(\last-3)$ and so on.

\

\displaytext{first revision}

The first revision is a model $I$ of the last class of the current order $C$.
It is all contained in a single class, as required.

Severe revision merges the classes from the first until the minimal class of
the revision, the last in this case.

\begin{hfigure}
\long\def\ttytex#1#2{#1}
\ttytex{
\begin{tabular}{ccc}
\setlength{\unitlength}{3750sp}%
\begin{picture}(1224,3624)(4789,-5173)
\thinlines
{\color[rgb]{0,0,0}\put(4801,-3361){\line( 1, 0){1200}}
}%
{\color[rgb]{0,0,0}\put(4801,-2161){\line( 1, 0){1200}}
}%
{\color[rgb]{0,0,0}\put(4801,-4561){\line( 1, 0){1200}}
}%
{\color[rgb]{0,0,0}\put(4801,-5161){\framebox(1200,3600){}}
}%
{\color[rgb]{0,0,0}\put(4801,-3961){\line( 1, 0){1200}}
}%
{\color[rgb]{0,0,0}\put(4801,-2761){\line( 1, 0){1200}}
}%
{\color[rgb]{0,0,0}\put(5551,-5011){\framebox(300,300){}}
}%
\put(5701,-4936){\makebox(0,0)[b]{\smash{\fontsize{9}{10.8}
\usefont{T1}{cmr}{m}{n}{\color[rgb]{0,0,0}$I$}%
}}}
\end{picture}%
&
\setlength{\unitlength}{3750sp}%
\begin{picture}(1104,2184)(5659,-5383)
\thinlines
{\color[rgb]{0,0,0}\multiput(6391,-3391)(-9.47368,4.73684){20}{\makebox(2.1167,14.8167){\tiny.}}
\put(6211,-3301){\line( 0,-1){ 45}}
\put(6211,-3346){\line(-1, 0){180}}
\put(6031,-3346){\line( 0,-1){ 90}}
\put(6031,-3436){\line( 1, 0){180}}
\put(6211,-3436){\line( 0,-1){ 45}}
\multiput(6211,-3481)(9.47368,4.73684){20}{\makebox(2.1167,14.8167){\tiny.}}
}%
\end{picture}%
&
\setlength{\unitlength}{3750sp}%
\begin{picture}(1224,4074)(4789,-5173)
\thinlines
{\color[rgb]{0,0,0}\put(4801,-5161){\framebox(1200,3600){}}
}%
{\color[rgb]{0,0,0}\put(5551,-5011){\framebox(300,300){}}
}%
{\color[rgb]{0,0,0}\put(5551,-1411){\framebox(300,300){}}
}%
{\color[rgb]{0,0,0}\multiput(5551,-4861)(7.89474,7.89474){20}{\makebox(2.2222,15.5556){\tiny.}}
}%
{\color[rgb]{0,0,0}\put(5551,-5011){\line( 1, 1){300}}
}%
{\color[rgb]{0,0,0}\multiput(5701,-5011)(7.89474,7.89474){20}{\makebox(2.2222,15.5556){\tiny.}}
}%
\put(5701,-1336){\makebox(0,0)[b]{\smash{\fontsize{9}{10.8}
\usefont{T1}{cmr}{m}{n}{\color[rgb]{0,0,0}$I$}%
}}}
\end{picture}%
\end{tabular}
}{
                                 +-+
                                 |I|
                                 +-+
+-------------+           +-------------+
|             |           |             |
|             |           |             |
|             |           |             |
+-------------+           |             |
|             |           |             |
|             |           |             |
|             |           |             |
+-------------+           |             |
|             |    ==>    |             |
|             |           |             |
|             |           |   T \ I     |
+-------------+           |             |
|             |           |             |
|             |           |             |
|             |           |             |
+-------------+           |             |
|      +-+    |           |      +-+    |
|      |I|    |           |      |/|    |
|      +-+    |           |      +-+    |
+-------------+           +-------------+
       C                      C sev(I)
  i-last.fig                 i-last-true.fig
}
\label{figure-ilast-itrue}
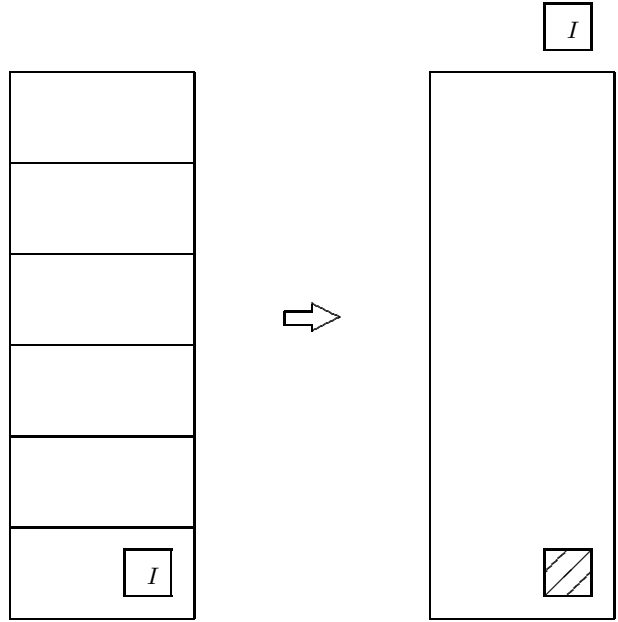
\hcaption{The first severe revision isolates a model of the last class}
\end{hfigure}

Since $I$ is a single model and is in the last class, its minimal model is
itself and is in the last class:%
{} $\min(\{I\}) = \{I\}$ and $\imin(\{I\}) = \last$.

\begin{eqnarray*}
C \sev(\{I\}) &=& [
	\min(\{I\}),
	C(0) \cup \cdots \cup C(\imin(\{I\})) \backslash \{I\},
	C(\imin(\{I\})+1), \ldots, C(\last)
]
\\
&=& [
	\{I\},
	C(0) \cup \cdots \cup C(\last) \backslash \{I\},
	C(\last+1), \ldots, C(\last)]
]
\\
&=& [
	\{I\},
	C(0) \cup \cdots \cup C(\last) \backslash \{I\}
]
\\
&=& [
	I,
	\true \backslash I
]
\end{eqnarray*}

Severe revision is not amnesic, it never generates the flat order, but this
order is close to it.

\

\displaytext{second revision}

The second revision is a model $J$ of $G(\last)$.
Being a single model, it is all contained in a single class, as required.

Being a single model, it is minimal. Its minimal index is either $0$ or $1$
since the current order
{} $C' = C \sev(\{I\}) = [\{I\}, \true \backslash \{I\}]$
only comprises two classes. The first case is $\imin(\{J\})=0$.

\begin{eqnarray*}
\lefteqn{C' \sev(\{J\})}
\\
&=& [
	\min(\{J\}),
	C'(0) \cup \cdots \cup C'(\imin(\{J\})) \backslash \{J\},
	C'(\imin(\{J\})+1),
	\ldots,
	C'(\last)
]
\\
&=& [
	\{J\},
	C'(0) \cup \cdots \cup C'(0) \backslash \{J\},
	C'(0+1),
	\ldots,
	C'(1)
]
\\
&=& [
	\{J\},
	C'(0) \backslash \{J\},
	C'(1),
]
\\
&=& [
	\{J\},
	\{J\} \backslash \{J\},
	\true \backslash \{J\}
]
\\
&=& [
	\{J\},
	\emptyset,
	\true \backslash \{J\}
]
\\
&=& [
	\{J\},
	\true \backslash \{J\}
]
\end{eqnarray*}

The second case is $\imin(\{J\}) = 1$.

\begin{eqnarray*}
\lefteqn{C' \sev(\{J\})}
\\
&=& [
	\min(\{J\}),
	C'(0) \cup \cdots \cup C'(\imin(\{J\})) \backslash \{J\},
	C'(\imin(\{J\})+1),
	\ldots,
	C'(\last)
]
\\
&=& [
	\{J\},
	C'(0) \cup \cdots \cup C'(1) \backslash \{J\},
	C'(1+1),
	\ldots,
	C'(1)
]
\\
&=& [
	\{J\},
	C'(0) \cup C'(1) \backslash \{J\},
	C'(2),
	\ldots,
	C'(1)
]
\\
&=& [
	\{J\},
	\true \backslash \{J\}
]
\end{eqnarray*}

The resulting order is $[\{J\}, \true \backslash \{J\}]$ in both cases.

\begin{hfigure}
\long\def\ttytex#1#2{#1}
\ttytex{
\begin{tabular}{ccc}
\setlength{\unitlength}{3750sp}%
\begin{picture}(1224,1074)(4789,-5173)
\thinlines
{\color[rgb]{0,0,0}\put(5551,-5011){\framebox(300,300){}}
}%
{\color[rgb]{0,0,0}\multiput(5551,-4861)(7.89474,7.89474){20}{\makebox(2.2222,15.5556){\tiny.}}
}%
{\color[rgb]{0,0,0}\put(5551,-5011){\line( 1, 1){300}}
}%
{\color[rgb]{0,0,0}\multiput(5701,-5011)(7.89474,7.89474){20}{\makebox(2.2222,15.5556){\tiny.}}
}%
{\color[rgb]{0,0,0}\put(4801,-5161){\framebox(1200,600){}}
}%
{\color[rgb]{0,0,0}\put(5551,-4411){\framebox(300,300){}}
}%
\put(5701,-4336){\makebox(0,0)[b]{\smash{\fontsize{9}{10.8}
\usefont{T1}{cmr}{m}{n}{\color[rgb]{0,0,0}$I$}%
}}}
\end{picture}%
&
\setlength{\unitlength}{3750sp}%
\begin{picture}(1104,744)(5659,-3943)
\thinlines
{\color[rgb]{0,0,0}\multiput(6391,-3391)(-9.47368,4.73684){20}{\makebox(2.1167,14.8167){\tiny.}}
\put(6211,-3301){\line( 0,-1){ 45}}
\put(6211,-3346){\line(-1, 0){180}}
\put(6031,-3346){\line( 0,-1){ 90}}
\put(6031,-3436){\line( 1, 0){180}}
\put(6211,-3436){\line( 0,-1){ 45}}
\multiput(6211,-3481)(9.47368,4.73684){20}{\makebox(2.1167,14.8167){\tiny.}}
}%
\end{picture}%
&
\setlength{\unitlength}{3750sp}%
\begin{picture}(1224,1074)(4789,-5173)
\thinlines
{\color[rgb]{0,0,0}\put(4801,-5161){\framebox(1200,600){}}
}%
{\color[rgb]{0,0,0}\put(5101,-4411){\framebox(300,300){}}
}%
{\color[rgb]{0,0,0}\put(5101,-5011){\framebox(300,300){}}
}%
{\color[rgb]{0,0,0}\multiput(5101,-4861)(7.89474,7.89474){20}{\makebox(2.2222,15.5556){\tiny.}}
}%
{\color[rgb]{0,0,0}\put(5101,-5011){\line( 1, 1){300}}
}%
{\color[rgb]{0,0,0}\multiput(5251,-5011)(7.89474,7.89474){20}{\makebox(2.2222,15.5556){\tiny.}}
}%
\put(5251,-4336){\makebox(0,0)[b]{\smash{\fontsize{9}{10.8}
\usefont{T1}{cmr}{m}{n}{\color[rgb]{0,0,0}$J$}%
}}}
\end{picture}%
\end{tabular}
}{
       +-+                          +-+
       |I|                          |J|
       +-+                          +-+
+-------------+    ==>    +-------------+
|      +-+    |           |         +-+ |
| T-I  |/|    |           | T-J     |/| |
|      +-+    |           |         +-+ |
+-------------+           +-------------+
    C sev(I)              C sev(I) sev(J)
   i-true.fig                j-true.fig
}
\label{figure-itrue-jtrue}
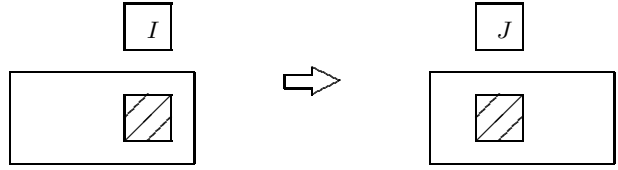
\hcaption{The second severe revision isolates a model of the target last class}
\end{hfigure}

\displaytext{third revision}

The order%
{} $C'' = C \sev(\{I\}) \sev(\{J\}) = [\{J\}, \true \backslash \{J\}]$
is revised by%
{} $A = \true \backslash G(\last)$.

\begin{hfigure}
\setlength{\unitlength}{3750sp}%
\begin{picture}(1224,1149)(4789,-5173)
\thinlines
{\color[rgb]{0,0,0}\put(4801,-5161){\framebox(1200,600){}}
}%
{\color[rgb]{0,0,0}\put(5101,-4411){\framebox(300,300){}}
}%
{\color[rgb]{0,0,0}\put(5101,-5011){\framebox(300,300){}}
}%
{\color[rgb]{0,0,0}\multiput(5101,-4861)(7.89474,7.89474){20}{\makebox(2.2222,15.5556){\tiny.}}
}%
{\color[rgb]{0,0,0}\put(5101,-5011){\line( 1, 1){300}}
}%
{\color[rgb]{0,0,0}\multiput(5251,-5011)(7.89474,7.89474){20}{\makebox(2.2222,15.5556){\tiny.}}
}%
{\color[rgb]{0,0,0}\put(4951,-5086){\line( 0, 1){1050}}
\put(4951,-4036){\line( 1, 0){600}}
\put(5551,-4036){\line( 0,-1){675}}
\put(5551,-4711){\line( 1, 0){300}}
\put(5851,-4711){\line( 0,-1){375}}
\put(5851,-5086){\line(-1, 0){900}}
}%
\put(5251,-4336){\makebox(0,0)[b]{\smash{\fontsize{9}{10.8}
\usefont{T1}{cmr}{m}{n}{\color[rgb]{0,0,0}$J$}%
}}}
\put(5626,-4936){\makebox(0,0)[b]{\smash{\fontsize{9}{10.8}
\usefont{T1}{cmr}{m}{n}{\color[rgb]{0,0,0}$G(\last)$}%
}}}
\end{picture}%
\nop{
          +-+
          |J|
          | |
+---------| |-+
| +-------+ + |
| |G(o)    /| |
| +---------+ |
|     T \ J   |
+-------------+
 j-true-go.fig
}%
\label{figure-case-first}
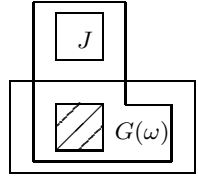
\hcaption{The position of the target last class in the current order}
\end{hfigure}

Since $J$ is in $G(\last)$, it is it not a model of the revision
{}$ A = \true \backslash G(\last) $.
Since $J$ is the only model of the first class of the current order
{}$ C'' = [\{J\}, \true \backslash \{J\}] $,
the revision does not intersect the first class. It is therefore all contained
in the second, the only other one:
{}$ \min(A) = A$ and $\imin(A) = 1$.

\begin{eqnarray*}
\lefteqn{C'' \sev(A)}
\\
&=& [
	\min(A),
	C''(0) \cup \cdots \cup C''(\imin(A)) \backslash A,
	C''(\imin(A)+1), \ldots, C''(\last)
]
\\
&=& [
	A,
	C''(0) \cup \cdots \cup C''(1) \backslash A,
	C''(1+1), \ldots, C''(1)
]
\\
&=& [
	\true \backslash G(\last),
	C''(0) \cup \cdots \cup C''(1) \backslash (\true \backslash G(\last))
]
\\
&=& [
	\true \backslash G(\last),
	(C(0) \cup C(1)) \cap G(\last)
]
\\
&=& [
	\true \backslash G(\last),
	\true \cap G(\last)
]
\\
&=& [
	\true \backslash G(\last),
	G(\last)
]
\end{eqnarray*}

The revision $\true \backslash G(\last)$ moves $G(\last)$ in its final
position, the last.

\begin{hfigure}
\long\def\ttytex#1#2{#1}
\ttytex{
\begin{tabular}{ccc}
\setlength{\unitlength}{3750sp}%
\begin{picture}(1224,1149)(4789,-5173)
\thinlines
{\color[rgb]{0,0,0}\put(4801,-5161){\framebox(1200,600){}}
}%
{\color[rgb]{0,0,0}\put(5101,-4411){\framebox(300,300){}}
}%
{\color[rgb]{0,0,0}\put(5101,-5011){\framebox(300,300){}}
}%
{\color[rgb]{0,0,0}\multiput(5101,-4861)(7.89474,7.89474){20}{\makebox(2.2222,15.5556){\tiny.}}
}%
{\color[rgb]{0,0,0}\put(5101,-5011){\line( 1, 1){300}}
}%
{\color[rgb]{0,0,0}\multiput(5251,-5011)(7.89474,7.89474){20}{\makebox(2.2222,15.5556){\tiny.}}
}%
{\color[rgb]{0,0,0}\put(4951,-5086){\line( 0, 1){1050}}
\put(4951,-4036){\line( 1, 0){600}}
\put(5551,-4036){\line( 0,-1){675}}
\put(5551,-4711){\line( 1, 0){300}}
\put(5851,-4711){\line( 0,-1){375}}
\put(5851,-5086){\line(-1, 0){900}}
}%
\put(5251,-4336){\makebox(0,0)[b]{\smash{\fontsize{9}{10.8}
\usefont{T1}{cmr}{m}{n}{\color[rgb]{0,0,0}$J$}%
}}}
\put(5626,-4936){\makebox(0,0)[b]{\smash{\fontsize{9}{10.8}
\usefont{T1}{cmr}{m}{n}{\color[rgb]{0,0,0}$G(\last)$}%
}}}
\end{picture}%
&
\setlength{\unitlength}{3750sp}%
\begin{picture}(1104,1104)(5659,-4303)
\thinlines
{\color[rgb]{0,0,0}\multiput(6391,-3391)(-9.47368,4.73684){20}{\makebox(2.1167,14.8167){\tiny.}}
\put(6211,-3301){\line( 0,-1){ 45}}
\put(6211,-3346){\line(-1, 0){180}}
\put(6031,-3346){\line( 0,-1){ 90}}
\put(6031,-3436){\line( 1, 0){180}}
\put(6211,-3436){\line( 0,-1){ 45}}
\multiput(6211,-3481)(9.47368,4.73684){20}{\makebox(2.1167,14.8167){\tiny.}}
}%
\end{picture}%
&
\setlength{\unitlength}{3750sp}%
\begin{picture}(1224,1674)(4789,-5098)
\thinlines
{\color[rgb]{0,0,0}\put(5101,-4411){\framebox(300,300){}}
}%
{\color[rgb]{0,0,0}\put(5101,-5011){\framebox(300,300){}}
}%
{\color[rgb]{0,0,0}\multiput(5101,-4861)(7.89474,7.89474){20}{\makebox(2.2222,15.5556){\tiny.}}
}%
{\color[rgb]{0,0,0}\put(5101,-5011){\line( 1, 1){300}}
}%
{\color[rgb]{0,0,0}\multiput(5251,-5011)(7.89474,7.89474){20}{\makebox(2.2222,15.5556){\tiny.}}
}%
{\color[rgb]{0,0,0}\put(4951,-5086){\line( 0, 1){1050}}
\put(4951,-4036){\line( 1, 0){600}}
\put(5551,-4036){\line( 0,-1){675}}
\put(5551,-4711){\line( 1, 0){300}}
\put(5851,-4711){\line( 0,-1){375}}
\put(5851,-5086){\line(-1, 0){900}}
}%
{\color[rgb]{0,0,0}\put(4951,-3436){\line(-1, 0){150}}
\put(4801,-3436){\line( 0,-1){600}}
\put(4801,-4036){\line( 1, 0){1200}}
\put(6001,-4036){\line( 0, 1){600}}
\put(6001,-3436){\line(-1, 0){450}}
\put(5551,-3436){\line( 0,-1){150}}
\put(5551,-3586){\line( 1, 0){300}}
\put(5851,-3586){\line( 0,-1){375}}
\put(5851,-3961){\line(-1, 0){900}}
\put(4951,-3961){\line( 0, 1){525}}
}%
\put(5251,-4336){\makebox(0,0)[b]{\smash{\fontsize{9}{10.8}
\usefont{T1}{cmr}{m}{n}{\color[rgb]{0,0,0}$J$}%
}}}
\put(5626,-4936){\makebox(0,0)[b]{\smash{\fontsize{9}{10.8}
\usefont{T1}{cmr}{m}{n}{\color[rgb]{0,0,0}$G(\last)$}%
}}}
\end{picture}%
\end{tabular}
}{
                          +---------+-+-+
                          | +-------+ + |
                          | |/////////| |
                          | +---------+ |
          +-+             |             |
          |J|             +-------------+
          | |                       +-+
+---------| |-+                     |J|
| +-------+ + |    ==>              | |
| |G(o)    /| |                     | |
| +---------+ |             +-------+ +
|     T \ J   |             |G(o)    /|
+-------------+             +---------+  
 j-true-go.fig             true-j-go.fig
}
\label{figure-case-first-severe}
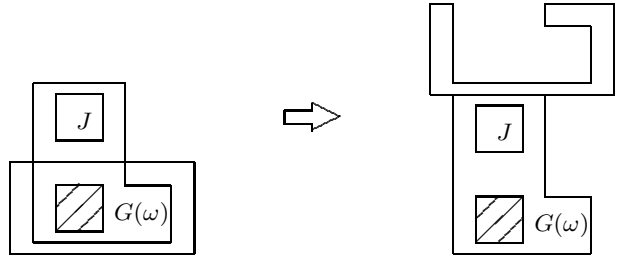
\hcaption{The third severe revision settles the last class of the target order}
\end{hfigure}

\displaytext{following revisions, summary}

This order
{} $[\true \backslash G(\last), G(\last)]$
is revised by
{} $G(0) \cup \cdots \cup G(\last-1)$,
then by
{} $G(0) \cup \cdots \cup G(\last-2)$,
and so on until
{} $G(0) \cup G(1)$.

\begin{hfigure}
\long\def\ttytex#1#2{#1}
\ttytex{
\begin{tabular}{ccc}
\setlength{\unitlength}{3750sp}%
\begin{picture}(1224,1074)(4789,-4423)
\thinlines
{\color[rgb]{0,0,0}\put(4801,-3961){\framebox(1200,600){}}
}%
{\color[rgb]{0,0,0}\put(4876,-3811){\framebox(450,300){}}
}%
{\color[rgb]{0,0,0}\multiput(4876,-3661)(7.89474,7.89474){20}{\makebox(2.2222,15.5556){\tiny.}}
}%
{\color[rgb]{0,0,0}\put(4876,-3811){\line( 1, 1){300}}
}%
{\color[rgb]{0,0,0}\put(5026,-3811){\line( 1, 1){300}}
}%
{\color[rgb]{0,0,0}\multiput(5176,-3811)(7.89474,7.89474){20}{\makebox(2.2222,15.5556){\tiny.}}
}%
{\color[rgb]{0,0,0}\put(4876,-4411){\framebox(450,300){}}
}%
{\color[rgb]{0,0,0}\put(5401,-3811){\framebox(525,300){}}
}%
\put(5101,-4336){\makebox(0,0)[b]{\smash{\fontsize{9}{10.8}
\usefont{T1}{cmr}{m}{n}{\color[rgb]{0,0,0}$G(\last)$}%
}}}
\put(5701,-3736){\makebox(0,0)[b]{\smash{\fontsize{9}{10.8}
\usefont{T1}{cmr}{m}{n}{\color[rgb]{0,0,0}$G(\last-1)$}%
}}}
\end{picture}%
&
\setlength{\unitlength}{3750sp}%
\begin{picture}(1104,1104)(5659,-4303)
\thinlines
{\color[rgb]{0,0,0}\multiput(6391,-3391)(-9.47368,4.73684){20}{\makebox(2.1167,14.8167){\tiny.}}
\put(6211,-3301){\line( 0,-1){ 45}}
\put(6211,-3346){\line(-1, 0){180}}
\put(6031,-3346){\line( 0,-1){ 90}}
\put(6031,-3436){\line( 1, 0){180}}
\put(6211,-3436){\line( 0,-1){ 45}}
\multiput(6211,-3481)(9.47368,4.73684){20}{\makebox(2.1167,14.8167){\tiny.}}
}%
\end{picture}%
&
\setlength{\unitlength}{3750sp}%
\begin{picture}(1224,1524)(4789,-4873)
\thinlines
{\color[rgb]{0,0,0}\put(4801,-3961){\framebox(1200,600){}}
}%
{\color[rgb]{0,0,0}\put(4876,-3811){\framebox(450,300){}}
}%
{\color[rgb]{0,0,0}\multiput(4876,-3661)(7.89474,7.89474){20}{\makebox(2.2222,15.5556){\tiny.}}
}%
{\color[rgb]{0,0,0}\put(4876,-3811){\line( 1, 1){300}}
}%
{\color[rgb]{0,0,0}\put(5026,-3811){\line( 1, 1){300}}
}%
{\color[rgb]{0,0,0}\multiput(5176,-3811)(7.89474,7.89474){20}{\makebox(2.2222,15.5556){\tiny.}}
}%
{\color[rgb]{0,0,0}\put(5401,-3811){\framebox(525,300){}}
}%
{\color[rgb]{0,0,0}\put(5401,-4411){\framebox(525,300){}}
}%
{\color[rgb]{0,0,0}\put(4876,-4861){\framebox(450,300){}}
}%
{\color[rgb]{0,0,0}\multiput(5401,-3661)(7.89474,7.89474){20}{\makebox(2.2222,15.5556){\tiny.}}
}%
{\color[rgb]{0,0,0}\put(5401,-3811){\line( 1, 1){300}}
}%
{\color[rgb]{0,0,0}\put(5551,-3811){\line( 1, 1){300}}
}%
{\color[rgb]{0,0,0}\multiput(5701,-3811)(8.03571,8.03571){29}{\makebox(2.2222,15.5556){\tiny.}}
}%
{\color[rgb]{0,0,0}\multiput(5851,-3811)(8.33333,8.33333){10}{\makebox(2.2222,15.5556){\tiny.}}
}%
\put(5701,-4336){\makebox(0,0)[b]{\smash{\fontsize{9}{10.8}
\usefont{T1}{cmr}{m}{n}{\color[rgb]{0,0,0}$G(\last-1)$}%
}}}
\put(5101,-4786){\makebox(0,0)[b]{\smash{\fontsize{9}{10.8}
\usefont{T1}{cmr}{m}{n}{\color[rgb]{0,0,0}$G(\last)$}%
}}}
\end{picture}%
\end{tabular}
}{
                             +-----------------+
                             | +----+ +------+ |
                             | |////| |//////| |
                             | +----+ +------+ |
+-----------------+          +-----------------+
| +----+ +------+ |                   +------+
| |////| |G(o-1)| |    =>             |G(o-1)|
| +----+ +------+ |                   +------+
+-----------------+
  +----+                       +----+
  |G(o)|                       |G(o)|
  +----+                       +----+
true-secondin-last.fig      true-second-last.fig
}
\label{figure-true-second-last}
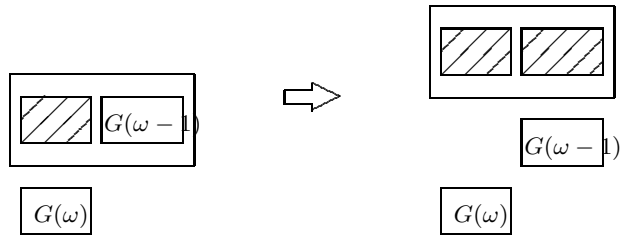
\hcaption{The fourth severe revision settles the second-to-last target class}
\end{hfigure}

\long\def\ttytex#1#2{#1}
\ttytex{
}{
}

The claim is proved by the following induction assumption and claim.

\begin{quote}
The revision
{}  $G(0) \cup \cdots \cup G(i)$
is applied to the order
{} $[G(0) \cup \cdots \cup G(i) \cup G(i+1), G(i+2), \ldots, G(\last)]$.
\end{quote}

\begin{hfigure}
\long\def\ttytex#1#2{#1}
\ttytex{
\begin{tabular}{ccc}
\setlength{\unitlength}{3750sp}%
\begin{picture}(1524,1974)(4639,-5323)
\thinlines
{\color[rgb]{0,0,0}\put(4651,-3961){\framebox(1500,600){}}
}%
{\color[rgb]{0,0,0}\put(4876,-3736){\framebox(600,300){}}
}%
{\color[rgb]{0,0,0}\put(4876,-4411){\framebox(600,300){}}
}%
{\color[rgb]{0,0,0}\put(4876,-5311){\framebox(600,300){}}
}%
\put(5176,-3661){\makebox(0,0)[b]{\smash{\fontsize{9}{10.8}
\usefont{T1}{cmr}{m}{n}{\color[rgb]{0,0,0}$G(i+1)$}%
}}}
\put(5401,-3886){\makebox(0,0)[b]{\smash{\fontsize{9}{10.8}
\usefont{T1}{cmr}{m}{n}{\color[rgb]{0,0,0}$G(0)\cup\cdots\cup G(i)$}%
}}}
\put(5176,-4336){\makebox(0,0)[b]{\smash{\fontsize{9}{10.8}
\usefont{T1}{cmr}{m}{n}{\color[rgb]{0,0,0}$G(i+2)$}%
}}}
\put(5176,-4786){\makebox(0,0)[b]{\smash{\fontsize{9}{10.8}
\usefont{T1}{cmr}{m}{n}{\color[rgb]{0,0,0}$\ldots$}%
}}}
\put(5176,-5236){\makebox(0,0)[b]{\smash{\fontsize{9}{10.8}
\usefont{T1}{cmr}{m}{n}{\color[rgb]{0,0,0}$G(\last)$}%
}}}
\end{picture}%
&
\setlength{\unitlength}{3750sp}%
\begin{picture}(1104,1104)(5659,-4303)
\thinlines
{\color[rgb]{0,0,0}\multiput(6391,-3391)(-9.47368,4.73684){20}{\makebox(2.1167,14.8167){\tiny.}}
\put(6211,-3301){\line( 0,-1){ 45}}
\put(6211,-3346){\line(-1, 0){180}}
\put(6031,-3346){\line( 0,-1){ 90}}
\put(6031,-3436){\line( 1, 0){180}}
\put(6211,-3436){\line( 0,-1){ 45}}
\multiput(6211,-3481)(9.47368,4.73684){20}{\makebox(2.1167,14.8167){\tiny.}}
}%
\end{picture}%
&
\setlength{\unitlength}{3750sp}%
\begin{picture}(1524,2424)(4639,-5323)
\thinlines
{\color[rgb]{0,0,0}\put(4876,-4411){\framebox(600,300){}}
}%
{\color[rgb]{0,0,0}\put(4876,-5311){\framebox(600,300){}}
}%
{\color[rgb]{0,0,0}\put(4876,-3961){\framebox(600,300){}}
}%
{\color[rgb]{0,0,0}\put(4651,-3511){\framebox(1500,600){}}
}%
{\color[rgb]{0,0,0}\put(4876,-3286){\framebox(600,300){}}
}%
{\color[rgb]{0,0,0}\multiput(4876,-3136)(7.89474,7.89474){20}{\makebox(2.2222,15.5556){\tiny.}}
}%
{\color[rgb]{0,0,0}\put(4876,-3286){\line( 1, 1){300}}
}%
{\color[rgb]{0,0,0}\put(5026,-3286){\line( 1, 1){300}}
}%
{\color[rgb]{0,0,0}\put(5176,-3286){\line( 1, 1){300}}
}%
{\color[rgb]{0,0,0}\multiput(5326,-3286)(7.89474,7.89474){20}{\makebox(2.2222,15.5556){\tiny.}}
}%
\put(5176,-4336){\makebox(0,0)[b]{\smash{\fontsize{9}{10.8}
\usefont{T1}{cmr}{m}{n}{\color[rgb]{0,0,0}$G(i+2)$}%
}}}
\put(5176,-4786){\makebox(0,0)[b]{\smash{\fontsize{9}{10.8}
\usefont{T1}{cmr}{m}{n}{\color[rgb]{0,0,0}$\ldots$}%
}}}
\put(5176,-5236){\makebox(0,0)[b]{\smash{\fontsize{9}{10.8}
\usefont{T1}{cmr}{m}{n}{\color[rgb]{0,0,0}$G(\last)$}%
}}}
\put(5401,-3436){\makebox(0,0)[b]{\smash{\fontsize{9}{10.8}
\usefont{T1}{cmr}{m}{n}{\color[rgb]{0,0,0}$G(0)\cup\cdots\cup G(i)$}%
}}}
\put(5176,-3886){\makebox(0,0)[b]{\smash{\fontsize{9}{10.8}
\usefont{T1}{cmr}{m}{n}{\color[rgb]{0,0,0}$G(i+1)$}%
}}}
\end{picture}%
\end{tabular}
}{
+-------------+           +-------------+
| G(0)..G(i)  |           | G(0)..G(i)  |
| +------+    |           | +------+    |
| |G(i+1)|    |           | |//////|    |
| +------+    |           | +------+    |
+-------------+           +-------------+
  +------+         ==>      +------+
  |G(i+2)|                  |G(i+1)|
  +------+                  +------+
  | ..   |                  |G(i+2)|
  +------+                  +------+
  | G(o) |                  |  ..  |
  +------+                  +------+
                            | G(o) |
                            +------+
assumption.fig              claim.fig
}
\label{figure-}
\hcaption{A severe revision settles the last class of the target order}
\end{hfigure}

Since
{}  $G(0) \cup \cdots \cup G(i)$
is applied to the order
{} $[G(0) \cup \cdots \cup G(i) \cup G(i+1), G(i+2), \ldots, G(\last)]$,
it is contained in its first class, proving that all revisions are applied to
an order containing them in their first class.

\

\displaytext{following revisions, base case}

The base case is the revision
{} $G(0) \cup \cdots \cup G(\last-1)$
applied to the order
{} $[\true \backslash G(\last), G(\last)]$.
This order is rewritten by replacing $\true$ with the union of the classes of
its partition $G$.

\begin{eqnarray*}
\lefteqn{[\true \backslash G(\last), G(\last)]}
\\
&=& [
	G(0) \cup \cdots \cup G(\last-1) \cup G(\last) \backslash G(\last),
	G(\last)
]
\\
&=& [
	G(0) \cup \cdots \cup G(\last-1), G(\last)
	]
\end{eqnarray*}

This is the inductive claim.

\

\displaytext{following revisions, induction case}

The induction case is the revision
{} $G(0) \cup \cdots \cup G(i)$
applied to the order
{} $[G(0) \cup \cdots \cup G(i) \cup G(i+1), G(i+2) \ldots, G(\last)]$.
They are denoted respectively $A$ and $C$.
The revision
{} $A = G(0) \cup \cdots \cup G(i)$
is all contained in the first class
{} $G(0) \cup \cdots \cup G(i) \cup G(i+1)$
of $C$.
All its models are therefore minimal and in the first class:
{} $\min(A) = A$ and $\imin(A) = 0$.

\begin{eqnarray*}
\lefteqn{C \sev(A)}
\\
&=& [
	\min(A),
	C(0) \cup \cdots \cup C(\imin(A)) \backslash A,
	C(\imin(A)+1),
	\ldots,
	C(\last)
]
\\
&=& [
	A,
	C(0) \cup \cdots \cup C(0) \backslash A,
	C(0+1),
	\ldots,
	C(\last)
]
\\
&=& [
	A,
	C(0) \backslash A,
	C(1),
	\ldots,
	C(\last)
]
\\
&=& [
	G(0) \cup \cdots \cup G(i),
	G(0) \cup \cdots \cup G(i) \cup G(i+1) \backslash (G(0) \cup \cdots \cup G(i)),
	G(i+2),
	\ldots,
	G(\last)
]
\\
&=& [
	G(0) \cup \cdots \cup G(i),
	G(i+1),
	G(i+2),
	\ldots,
	G(\last)
]
\end{eqnarray*}

This is the inductive claim.

\

\displaytext{following revisions, conclusion}

The induction claim applied to the last step is that $G(0)$ revises
{} $[G(0) \cup G(1), G(2), \ldots, G(\last)]$.
As proved in the induction case, it generates
{} $[G(0), G(1), G(2), \ldots, G(\last)]$,
the target doxastic state.~\qed

The lemma proves severe revision plastic.

\begin{theorem}
\label{theorem-severe-plastic}

Severe revision is plastic.

\end{theorem}

Severe revision is plastic, but not full plastic. It is indeed not amnesic:
beliefs cannot be completely erased.

\begin{corollary}
\label{corollary-severe-amnesic}

Severe revision is not amnesic.

\end{corollary}

\proof Severe revision possesses the two properties required by
Theorem~\ref{theorem-amnesic}:
{} $C \sev(A)(0) \models A$
and
{} $C \sev(\true) = C$.

The models of $C \sev(A)(0)$ are $\min(A)$ by definition. The minimal models of
$A$ are all models of $A$, and therefore imply it.

Revising by $\true$ does not change the order. The minimal models of $\true$
are the minimal of all models. Since classes are assumed non-empty, they are
the whole first class:
{} $\min(\true) = C(0)$
and
{} $\imin(\true) = 0$.

\begin{eqnarray*}
\lefteqn{C \sev(\true)}
\\
&=& [
	\min(\true),
	C(0) \cup \cdots \cup C(\imin(\true)) \backslash \true,
	C(\imin(\true)+1),
	\ldots,
	C(\last)
]
\\
&=& [
	C(0),
	C(0) \cup \cdots \cup C(0) \backslash \true,
	C(0+1),
	\ldots,
	C(\last)
]
\\
&=& [
	C(0),
	C(0) \backslash \true,
	C(1),
	\ldots,
	C(\last)
]
\\
&=& [
	C(0),
	\emptyset,
	C(1),
	\ldots,
	C(\last)
]
\\
&=& [
	C(0),
	C(1),
	\ldots,
	C(\last)
]
\end{eqnarray*}~\qed

Amnesic negates full plastic.

\begin{corollary}
\label{corollary-severe-fully}

Severe revision is not fully plastic.

\end{corollary}

\section{Moderate severe revision}
\label{section-moderatesevere}

\draft

definition in rott-09:

\[
h_{>\neg A} | A | h_{>\neg A} \vee A
\]

translation as classes

\begin{enumerate}

\item significant example

\v
+-------------+
|C0           | S1        h1 h2 h3 h4
+-------------+
|C1 +------+  | S2           h2 h3 h4
+---|------|--+
|C2 |   A  |  | S3              h3 h4
+---|------|--+
|C3 +------+  | S4                 h4
+-------------+
|C4           |
+-------------+
\vv

\item calculations:

the largest sphere implicating $\neg A$ is $S_1$; therefore, $i=1$

in a different way: the first class consistent with $A$ is $C_1$;
therefore, $i=1$

\[
h2 h3 h4 A h2vA h3vA h4vA
\]

\item translation

\v
R1 = h2 h3 h4 A h2vA h3vA h4vA = S2 A   ==>  G0 = R1 - 0
                                                = S2 A
                                                = (C0 u C1) A
                                                = C1 A

R2 =    h3 h4 A h2vA h3vA h4vA = S3 A   ==>  G1 = R2 - R1
                                                = S3 A - S2 A
                                                = (S3 - S2) A
                                                = C2 A

R3 =       h4 A h2vA h3vA h4vA = S4 A   ==>  G2 = R3 - R2
                                                = S4 A - S3 A
                                                = (S4 - S3) A
                                                = C3 A

R4 =          A h2vA h3vA h4vA = A      ==>       R4 - R3
                                                = A - S4 A
                                                = (T - S4) A
                                                = C4 A        (empty)

R5 =            h2vA h3vA h4vA = S2 v A ==>  G3 = R5 - R4
                                                = S2 v A - A
                                                = S2 - A
                                                = (C0 u C1) - A

R6 =                 h3vA h4vA = S3 v A ==>  G4 = R6 - R5
                                                = (S3 u A) - (S2 u A)
                                                = S3 u A - S2 - A
                                                = S3 - S2 - A
                                                = C2 - A

R7 =                      h4vA = S4 v A ==>  G5 = R7 - R6
                                                = (S4 u A) - (S3 u A)
                                                = S4 u A - S3 - A
                                                = S4 - S3 - A
                                                = C3 - A

                                             G6 = T - R7
                                                = C4 - A
                                                = C4
\vv

\item drawing of classes
{} $[A C1, A C2,
{}   A C3, (C0 \cup C1) \backslash A,
{}   C2 \backslash A, C3 \backslash A, C4]$

\v
                            +------+  C1 A
                            |------|
                            |   A  |  C2 A
                            |------|
                            +------+  C3 A

+-------------+         +-------------+
|C0           |         |             |
+-------------+         |             | (C0 u C1) - A
|C1 +------+  |         |   +------+  |
+---|------|--+         +---+      +--+
|C2 |   A  |  |   ==>
+---|------|--+         +---+      +--+
|C3 +------+  |         |   |      |  | C2 - A
+-------------+         +---|      |--+
|C4           |         |   +------+  | C3 - A
+-------------+         +-------------+
                        |             | C4 - A
                        +-------------+
\vv

\end{enumerate}

\separator

\begin{hfigure}
\long\def\ttytex#1#2{#1}
\ttytex{
\begin{tabular}{ccc}
\setlength{\unitlength}{3750sp}%
\begin{picture}(1224,1824)(5089,-4873)
\thinlines
{\color[rgb]{0,0,0}\put(5101,-3361){\line( 1, 0){1200}}
}%
{\color[rgb]{0,0,0}\put(5101,-4861){\framebox(1200,1800){}}
}%
{\color[rgb]{0,0,0}\put(5101,-3961){\line( 1, 0){1200}}
}%
{\color[rgb]{0,0,0}\put(5101,-4261){\line( 1, 0){1200}}
}%
{\color[rgb]{0,0,0}\put(5101,-4561){\line( 1, 0){1200}}
}%
{\color[rgb]{0,0,0}\put(5401,-4486){\framebox(750,1050){}}
}%
{\color[rgb]{0,0,0}\put(5101,-3661){\line( 1, 0){1200}}
}%
\put(5851,-3886){\makebox(0,0)[b]{\smash{\fontsize{9}{10.8}
\usefont{T1}{cmr}{m}{n}{\color[rgb]{0,0,0}$A$}%
}}}
\end{picture}%
&
\setlength{\unitlength}{3750sp}%
\begin{picture}(1104,1104)(5659,-4303)
\thinlines
{\color[rgb]{0,0,0}\multiput(6391,-3391)(-9.47368,4.73684){20}{\makebox(2.1167,14.8167){\tiny.}}
\put(6211,-3301){\line( 0,-1){ 45}}
\put(6211,-3346){\line(-1, 0){180}}
\put(6031,-3346){\line( 0,-1){ 90}}
\put(6031,-3436){\line( 1, 0){180}}
\put(6211,-3436){\line( 0,-1){ 45}}
\multiput(6211,-3481)(9.47368,4.73684){20}{\makebox(2.1167,14.8167){\tiny.}}
}%
\end{picture}%
&
\setlength{\unitlength}{3750sp}%
\begin{picture}(1224,2949)(5089,-4873)
\thinlines
{\color[rgb]{0,0,0}\put(5101,-4861){\framebox(1200,1800){}}
}%
{\color[rgb]{0,0,0}\put(5101,-4561){\line( 1, 0){1200}}
}%
{\color[rgb]{0,0,0}\put(5401,-2161){\line( 1, 0){750}}
}%
{\color[rgb]{0,0,0}\put(5401,-2461){\line( 1, 0){750}}
}%
{\color[rgb]{0,0,0}\put(5401,-2761){\line( 1, 0){750}}
}%
{\color[rgb]{0,0,0}\put(5401,-2986){\framebox(750,1050){}}
}%
{\color[rgb]{0,0,0}\put(5401,-4486){\framebox(750,1050){}}
}%
{\color[rgb]{0,0,0}\put(5101,-3661){\line( 1, 0){300}}
}%
{\color[rgb]{0,0,0}\put(6151,-3661){\line( 1, 0){150}}
}%
{\color[rgb]{0,0,0}\put(5101,-3961){\line( 1, 0){300}}
}%
{\color[rgb]{0,0,0}\put(5101,-4261){\line( 1, 0){300}}
}%
{\color[rgb]{0,0,0}\put(6151,-3961){\line( 1, 0){150}}
}%
{\color[rgb]{0,0,0}\put(6151,-4261){\line( 1, 0){150}}
}%
\put(5851,-2386){\makebox(0,0)[b]{\smash{\fontsize{9}{10.8}
\usefont{T1}{cmr}{m}{n}{\color[rgb]{0,0,0}$A$}%
}}}
\end{picture}%
\end{tabular}
}{
                             +------+
                             |------|
                             |   A  |
                             |------|
                             |      |
                             |------|
                             +------+
+-------------+          +-------------+
|             |          |             |
+-------------+          |             |
|   +------+  |          |   +------+  |
+---|------|--+          +---+      +--+
|   |   A  |  |          |   |      |  |
+---|------|--+          +---|      |--+
|   |      |  |    =>    |   |      |  |
+---|------|--+          +---|      |--+
|   +------+  |          |   +------+  |
+-------------+          +-------------+
|             |          |             |
+-------------+          +-------------+
|             |          |             |
+-------------+          +-------------+
|             |          |             |
+-------------+          +-------------+
  current.fig           moderatesevere.fig
}
\hcaption{moderate severe revision}
\end{hfigure}

\enddraft

Moderate severe revision differs from severe in believing the new information
in all possible situations, not only the present ones. Like severe revision,
the change sparks doubt on the other situations of comparable strenght of
belief with the new information.

\begin{definition}
\label{definition-moderatesevere}

\begin{eqnarray*}
\lefteqn{C \msev(A)}
\\
&=& [
	C(\imin(A)) \cap A, \ldots, C(\imax(A)) \cap A,
\\
&&
	C(0) \cup \cdots \cup C(\imin(A)) \backslash A,
\\
&&
	C(\imin(A)+1) \backslash A, \ldots, C(\last) \backslash A
]
\end{eqnarray*}

\end{definition}

The similarity with severe revision allows for reducing to it in a relevant
case, when the new belief is all contained in a class. The situations it
supports are already believed the same.

\begin{lemma}
\label{lemma-moderatesevere-severe}

Moderate severe revision and severe revision coincide when the revision is
contained in a class of the order: $C \sev(A) = C \msev(A)$ if $A \subseteq
C(i)$ for some $i$.

\end{lemma}

\proof If $A$ is contained in a single class of $C$, then
{} $\min(A) = A$
and
{} $\imin(A) = \imax(A)$.

\begin{eqnarray*}
\lefteqn{C \msev(A)}
\\
&=& [
	C(\imin(A)) \cap A, \ldots, C(\imax(A)) \cap A,
\\
&&
	C(0) \cup \cdots \cup C(\imin(A)) \backslash A,
\\
&&
	C(\imin(A)+1) \backslash A, \ldots, C(\last) \backslash A
]
\\
&=& [
	C(\imin(A)) \cap A, \ldots, C(\imin(A)) \cap A,
\\
&&
	C(0) \cup \cdots \cup C(\imin(A)) \backslash A,
\\
&&
	C(\imax(A)+1) \backslash A, \ldots, C(\last) \backslash A
]
\\
&=& [
	C(\imin(A)) \cap A,
\\
&&
	C(0) \cup \cdots \cup C(\imin(A)) \backslash A,
\\
&&
	C(\imax(A)+1), \ldots, C(\last)
]
\\
&=& [
	\min(A),
	C(0) \cup \cdots \cup C(\imin(A)) \backslash A,
	C(\imin(A)+1), \ldots, C(\last)
]
\\
&=&
	C \sev(A)
\end{eqnarray*}

The second step follows from
{} $C(i) \backslash A = C(i)$ when $i > \imax(A)$:
no class contains models of $A$ if all its models are greater than the maximal
models of $A$.~\qed

\begin{theorem}
\label{theorem-moderatesevere-plastic}

Moderate severe revision is plastic.

\end{theorem}

\proof Lemma~\ref{lemma-plastic-severe} proves the plasticity of severe
revision even when bounded to revisions all contained in a class of the current
order. Lemma~\ref{lemma-moderatesevere-severe} proves that moderate severe
revision gives the same results in this case.~\qed

Like severe revision, moderate severe revision is not amnesic.

\begin{theorem}
\label{theorem-moderatesevere-amnesic}

Moderate severe revision is not amnesic.

\end{theorem}

\proof Moderate severe revision possesses the two properties required by
Theorem~\ref{theorem-amnesic}:
{} $C \msev(A)(0) \models A$
and
{} $C \msev(\true) = C$.

The models of $C \msev(A)(0)$ are $C(\imin(A)) \cap A$ by definition. They are
all models of $A$, and therefore imply it.

Revising by $\true$ does not change the order. The minimal models of $\true$
are the minimal of all models. Since classes are assumed non-empty, they are
the whole first class:
{} $\imin(\true) = 0$.
The maximal models of $A$ are the maximal models among all:
{} $\imax(\true) = \last$.

\begin{eqnarray*}
\lefteqn{C \msev(\true)}
\\
&=& [
	C(\imin(\true)) \cap \true, \ldots, C(\imax(\true)) \cap \true,
\\
&&
	C(0) \cup \cdots \cup C(\imin(\true)) \backslash \true,
\\
&&
	C(\imin(\true)+1) \backslash \true, \ldots, C(\last) \backslash \true
]
\\
&=& [
	C(0) \cap \true, \ldots, C(\last) \cap \true,
\\
&&
	C(0) \cup \cdots \cup C(0) \backslash \true,
\\
&&
	C(\last+1) \backslash \true, \ldots, C(\last) \backslash \true
]
\\
&=& [
	C(0) \cap \true, \ldots, C(\last) \cap \true,
	C(0) \backslash \true
]
\\
&=& [
	C(0), \ldots, C(\last)
]
\\
&=& C
\end{eqnarray*}

Theorem~\ref{theorem-amnesic} proves that a revision with these two properties
is not amnesic.~\qed

\begin{corollary}
\label{corollary-moderatesevere-fully}

Moderate severe revision is not fully plastic.

\end{corollary}

\section{Deep severe revision}
\label{section-deepsevere}

\draft

the principle of epiphanic revisions like severe revision is that a change in
the strength of belief in a situation hints lack of knowledge about it, it
sparks doubts on situations that were previously more strongly believed

example: the order I < J < Z is revised by Z

Z becomes the most strongly believed situation

I and J were previously believed more strongly, but this turns out to be
wrong

maybe I < J is wrong as well

when revision is not contingent and the principle is applied to all such
situations, deep severe revision results

\v
                            +------+
                            |------|
                            |   A  |
                            |------|
                            +------+
+-------------+         +-------------+
|C0           |         |             |
+-------------+         |             |
|C1 +------+  |         |   +------+  |
+---|------|--+         |   |      |  |
|C2 |   A  |  |   ==>   |   |      |  |
+---|------|--+         |   |      |  |
|C3 +------+  |         |   +------+  |
+-------------+         +-------------+
|C4           |
+-------------+         +-------------+
|C5           |         |             |
+-------------+         +-------------+
                        |             |
                        +-------------+
\vv

moderate severe revision is actually a mix of severe and deep severe: it is not
contingent in that it trusts all situations of A more than all others; at the
same time, doubts are only cast on situations more strongly believed by the
minimal ones of A; it is non-contingent but kind of contingent in the way it is
epiphanic

\

in terms of formulae: the index of the maximal class consistent with A is
needed, or the maximal sphere consistent with A

rott-09 only defines the index of the maximal implying sphere

with this index i, it is h1 h2 h3 h4 A h3vA h4vA

therefore: h | A | h>=i v A

\separator

\begin{hfigure}
\long\def\ttytex#1#2{#1}
\ttytex{
\begin{tabular}{ccc}
\setlength{\unitlength}{3750sp}%
\begin{picture}(1224,1824)(5089,-4873)
\thinlines
{\color[rgb]{0,0,0}\put(5101,-3361){\line( 1, 0){1200}}
}%
{\color[rgb]{0,0,0}\put(5101,-4861){\framebox(1200,1800){}}
}%
{\color[rgb]{0,0,0}\put(5101,-3961){\line( 1, 0){1200}}
}%
{\color[rgb]{0,0,0}\put(5101,-4261){\line( 1, 0){1200}}
}%
{\color[rgb]{0,0,0}\put(5101,-4561){\line( 1, 0){1200}}
}%
{\color[rgb]{0,0,0}\put(5401,-4486){\framebox(750,1050){}}
}%
{\color[rgb]{0,0,0}\put(5101,-3661){\line( 1, 0){1200}}
}%
\put(5851,-3886){\makebox(0,0)[b]{\smash{\fontsize{9}{10.8}
\usefont{T1}{cmr}{m}{n}{\color[rgb]{0,0,0}$A$}%
}}}
\end{picture}%
&
\setlength{\unitlength}{3750sp}%
\begin{picture}(1104,1104)(5659,-4303)
\thinlines
{\color[rgb]{0,0,0}\multiput(6391,-3391)(-9.47368,4.73684){20}{\makebox(2.1167,14.8167){\tiny.}}
\put(6211,-3301){\line( 0,-1){ 45}}
\put(6211,-3346){\line(-1, 0){180}}
\put(6031,-3346){\line( 0,-1){ 90}}
\put(6031,-3436){\line( 1, 0){180}}
\put(6211,-3436){\line( 0,-1){ 45}}
\multiput(6211,-3481)(9.47368,4.73684){20}{\makebox(2.1167,14.8167){\tiny.}}
}%
\end{picture}%
&
\setlength{\unitlength}{3750sp}%
\begin{picture}(1224,2949)(5089,-4873)
\thinlines
{\color[rgb]{0,0,0}\put(5101,-4561){\line( 1, 0){1200}}
}%
{\color[rgb]{0,0,0}\put(5401,-2161){\line( 1, 0){750}}
}%
{\color[rgb]{0,0,0}\put(5401,-2461){\line( 1, 0){750}}
}%
{\color[rgb]{0,0,0}\put(5401,-2761){\line( 1, 0){750}}
}%
{\color[rgb]{0,0,0}\put(5401,-2986){\framebox(750,1050){}}
}%
{\color[rgb]{0,0,0}\put(5401,-4486){\framebox(750,1050){}}
}%
{\color[rgb]{0,0,0}\put(5101,-4861){\framebox(1200,1800){}}
}%
\put(5851,-2386){\makebox(0,0)[b]{\smash{\fontsize{9}{10.8}
\usefont{T1}{cmr}{m}{n}{\color[rgb]{0,0,0}$A$}%
}}}
\end{picture}%
\end{tabular}
}{
                             +------+
                             |------|
                             |   A  |
                             |------|
                             |      |
                             |------|
                             +------+
+-------------+          +-------------+
|             |          |             |
+-------------+          |             |
|   +------+  |          |   +------+  |
+---|------|--+          |   |      |  |
|   |   A  |  |          |   |      |  |
+---|------|--+          |   |      |  |
|   |      |  |          |   |      |  |
+---|------|--+          |   |      |  |
|   +------+  |    =>    |   +------+  |
+-------------+          +-------------+
|             |          |             |
+-------------+          +-------------+
|             |          |             |
+-------------+          +-------------+
|             |          |             |
+-------------+          +-------------+
  current.fig             deepsevere.fig
}
\hcaption{deep severe revision}
\end{hfigure}

\enddraft

Like moderate severe revision, the new information is believed in all
situations and not only in the currently most believed ones. Like all severe
forms of revision, the other situations of comparable strength of belief are
distrusted the same.

\begin{definition}
\label{definition-deepsevere}

\begin{eqnarray*}
\lefteqn{C \dsev(A)}
\\
&=& [
	C(\imin(A)) \cap A, \ldots, C(\imax(A)) \cap A,
\\
&&
	C(0) \cup \cdots \cup C(\imax(A)) \backslash A,
\\
&&
	C(\imax(A)+1), \ldots, C(\last)
]
\end{eqnarray*}

\end{definition}

Like moderate severe revision, deep severe revision reduces to severe revision
when the revision is all contained in a class.

\begin{lemma}
\label{lemma-deepsevere-severe}

Deep severe revision and severe revision coincide when the revision is
contained in a class of the order: $C \sev(A) = C \dsev(A)$ if $A \subseteq
C(i)$ for some $i$.

\end{lemma}

\proof If $A$ is contained in a single class of $C$, then $\imax(A) =
\imin(A)$. Two replacements in the definition of deep severe revision proves
that deep and moderate severe revisions coincide in this case.

\begin{eqnarray*}
\lefteqn{C \dsev(A)}
\\
&=& [
	C(\imin(A)) \cap A, \ldots, C(\imax(A)) \cap A,
\\
&&
	C(0) \cup \cdots \cup C(\imax(A)) \backslash A,
\\
&&
	C(\imax(A)+1), \ldots, C(\last)
]
\\
&=& [
	C(\imin(A)) \cap A, \ldots, C(\imin(A)) \cap A,
\\
&&
	C(0) \cup \cdots \cup C(\imin(A)) \backslash A,
\\
&&
	C(\imin(A)+1), \ldots, C(\last)
]
\\
&=& [
	C(\imin(A)) \cap A,
\\
&&
	C(0) \cup \cdots \cup C(\imin(A)) \backslash A,
\\
&&
	C(\imin(A)+1), \ldots, C(\last)
]
\\
&=& [
	\min(A),
	C(0) \cup \cdots \cup C(\imin(A)) \backslash A,
	C(\imin(A)+1), \ldots, C(\last)
]
\\
&=&
	C \sev(A)
\end{eqnarray*}~\qed

Since severe revision allows reaching an arbitrary non-flat doxastic state by a
sequence of revisions like in the previous lemma, so does deep severe revision.

\begin{theorem}
\label{theorem-deepsevere-plastic}

\end{theorem}

\proof Lemma~\ref{lemma-plastic-severe} proves the plasticity of severe
revision even when bounded to revisions all contained in a class of the current
order. Lemma~\ref{lemma-deepsevere-severe} proves that it gives the same result
of deep severe revision in this case.~\qed

Like most revisions, deep severe is plastic but not fully plastic. The flat
doxastic state is never reached from a non-flat one: beliefs cannot be
completely erased.

\begin{theorem}
\label{theorem-deepsevere-amnesic}

Deep severe revision is not amnesic.

\end{theorem}

\proof The claim follows from Theorem~\ref{theorem-amnesic}: no change operator
satisfying
{} $C \dsev(A)(0) \models A$
and
{} $C \dsev(\true) = C$
is amnesic.

The first class of $C \dsev(A)$ is
{} $C(\imin(A)) \cap A$.
Being the intersection of a set with $A$, it only contains models of $A$. It
therefore implies $A$.

The minimal models of $\true$ are the minimal among all models. Same for
maximal:
{} $\imin(\true) = 0$;
{} $\imax(\true) = \last$.

\begin{eqnarray*}
\lefteqn{C \dsev(\true)}
\\
&=& [
	C(\imin(\true)) \cap \true, \ldots, C(\imax(\true)) \cap \true,
\\
&&
	C(0) \cup \cdots \cup C(\imax(\true)) \backslash \true,
\\
&&
	C(\imax(\true)+1), \ldots, C(\last)
]
\\
&=& [
	C(0) \cap \true, \ldots, C(\last) \cap \true,
\\
&&
	C(0) \cup \cdots \cup C(\last) \backslash \true,
\\
&&
	C(\last)+1), \ldots, C(\last)
]
\\
&=& [
	C(0) \cap \true, \ldots, C(\last) \cap \true,
	\true \backslash \true
]
\\
&=& [
	C(0), \ldots, C(\last),
	\emptyset
]
\\
&=& [
	C(0), \ldots, C(\last)
]
\\
&=& C
\end{eqnarray*}~\qed

\begin{corollary}
\label{corollary-deepsevere-fully}

Deep severe revision is not fully plastic.

\end{corollary}

\section{Plain severe}
\label{section-plainsevere}

\draft

definition in rott-09:

\[
	h_{=\neg A + 1} \wedge A | h_{>\neg A + 1}
\]

expression in classes

\begin{enumerate}

\item significant example 

\v
+-------------+
|C0           | S1        h1 h2 h3 h4
+-------------+
|C1 +------+  | S2           h2 h3 h4
+---|------|--+
|C2 |   A  |  | S3              h3 h4
+---|------|--+
|C3 +------+  | S4                 h4
+-------------+
|C4           |
+-------------+
\vv

\item calculations

the largest sphere where $\neg A$ is true is S1; therefore, $i=1$

in a different way: the first class where A is true is C1;
therefore, i=1

i+1 = 2

\[
        h_2 \wedge A | h3 h4
\]

\item translation

\v
R1 =  Ah2 h3 h4  =  A h2 h3 h4  =  A S2  =  A (C0 u C1)  =  AC1
                                             ==>  G0 = R1 - 0 = AC1

R2 =      h3 h4  =  S3  =  C0 u C1 u C2      ==>  G1 = R2 - R1
                                                     = C0uC1uC2 - AC1

R3 =         h4  =  S4  =  C0 u C1 u C2 u C3 ==>  G2 = R3 - R2
                                                     = C3

                                             ==>  G4 = T - R3
                                                     = T - C0uC1uC2uC3
                                                     = C4
\vv

\item drawing of classes
{} $[A C1,
{}   C0 \cup C1 \cup C2 \backslash A \wedge C1,
{}   C3, C4]$

\v
                              +------+ A C1
                              +------+
+-------------+           +-------------+
|C0           |           |             |
+-------------+           |             |
|C1 +------+  |           |   +//////+  | C0 u C1 u C2 - A C1 = C0 u (C1-A) u C2
+---|------|--+           |   +------+  |
|C2 |   A  |  |    ==>    |   |      |  |
+---|------|--+           +---+------+--+
|C3 +------+  |
+-------------+           +---+------+--+
|C4           |           |   +------+  | C3
+-------------+           +-------------+
                          |             | C4
                          +-------------+
\vv

\end{enumerate}

merges a class more than severe revision;
including its part satisfyingly A

\separator

\begin{hfigure}
\long\def\ttytex#1#2{#1}
\ttytex{
\begin{tabular}{ccc}
\setlength{\unitlength}{3750sp}%
\begin{picture}(1224,1824)(5089,-4873)
\thinlines
{\color[rgb]{0,0,0}\put(5101,-3361){\line( 1, 0){1200}}
}%
{\color[rgb]{0,0,0}\put(5101,-4861){\framebox(1200,1800){}}
}%
{\color[rgb]{0,0,0}\put(5101,-3961){\line( 1, 0){1200}}
}%
{\color[rgb]{0,0,0}\put(5101,-4261){\line( 1, 0){1200}}
}%
{\color[rgb]{0,0,0}\put(5101,-4561){\line( 1, 0){1200}}
}%
{\color[rgb]{0,0,0}\put(5401,-4486){\framebox(750,1050){}}
}%
{\color[rgb]{0,0,0}\put(5101,-3661){\line( 1, 0){1200}}
}%
\put(5851,-3886){\makebox(0,0)[b]{\smash{\fontsize{9}{10.8}
\usefont{T1}{cmr}{m}{n}{\color[rgb]{0,0,0}$A$}%
}}}
\end{picture}%
&
\setlength{\unitlength}{3750sp}%
\begin{picture}(1104,1104)(5659,-4303)
\thinlines
{\color[rgb]{0,0,0}\multiput(6391,-3391)(-9.47368,4.73684){20}{\makebox(2.1167,14.8167){\tiny.}}
\put(6211,-3301){\line( 0,-1){ 45}}
\put(6211,-3346){\line(-1, 0){180}}
\put(6031,-3346){\line( 0,-1){ 90}}
\put(6031,-3436){\line( 1, 0){180}}
\put(6211,-3436){\line( 0,-1){ 45}}
\multiput(6211,-3481)(9.47368,4.73684){20}{\makebox(2.1167,14.8167){\tiny.}}
}%
\end{picture}%
&
\setlength{\unitlength}{3750sp}%
\begin{picture}(1224,2049)(5089,-5473)
\thinlines
{\color[rgb]{0,0,0}\put(5101,-4561){\line( 1, 0){1200}}
}%
{\color[rgb]{0,0,0}\put(5101,-4861){\line( 1, 0){1200}}
}%
{\color[rgb]{0,0,0}\put(5101,-5161){\line( 1, 0){1200}}
}%
{\color[rgb]{0,0,0}\put(5101,-5461){\framebox(1200,1800){}}
}%
{\color[rgb]{0,0,0}\put(5401,-3661){\framebox(750,225){}}
}%
{\color[rgb]{0,0,0}\put(5401,-4261){\framebox(750,225){}}
}%
{\color[rgb]{0,0,0}\multiput(5401,-4186)(7.89474,7.89474){20}{\makebox(2.2222,15.5556){\tiny.}}
}%
{\color[rgb]{0,0,0}\multiput(5401,-4261)(8.03571,8.03571){29}{\makebox(2.2222,15.5556){\tiny.}}
}%
{\color[rgb]{0,0,0}\multiput(5476,-4261)(8.03571,8.03571){29}{\makebox(2.2222,15.5556){\tiny.}}
}%
{\color[rgb]{0,0,0}\multiput(5551,-4261)(8.03571,8.03571){29}{\makebox(2.2222,15.5556){\tiny.}}
}%
{\color[rgb]{0,0,0}\multiput(5626,-4261)(8.03571,8.03571){29}{\makebox(2.2222,15.5556){\tiny.}}
}%
{\color[rgb]{0,0,0}\multiput(5701,-4261)(8.03571,8.03571){29}{\makebox(2.2222,15.5556){\tiny.}}
}%
{\color[rgb]{0,0,0}\multiput(5776,-4261)(8.03571,8.03571){29}{\makebox(2.2222,15.5556){\tiny.}}
}%
{\color[rgb]{0,0,0}\multiput(5851,-4261)(8.03571,8.03571){29}{\makebox(2.2222,15.5556){\tiny.}}
}%
{\color[rgb]{0,0,0}\multiput(5926,-4261)(8.03571,8.03571){29}{\makebox(2.2222,15.5556){\tiny.}}
}%
{\color[rgb]{0,0,0}\multiput(5401,-4111)(8.33333,8.33333){10}{\makebox(2.2222,15.5556){\tiny.}}
}%
{\color[rgb]{0,0,0}\multiput(6001,-4261)(7.89474,7.89474){20}{\makebox(2.2222,15.5556){\tiny.}}
}%
{\color[rgb]{0,0,0}\multiput(6076,-4261)(8.33333,8.33333){10}{\makebox(2.2222,15.5556){\tiny.}}
}%
{\color[rgb]{0,0,0}\put(5401,-5086){\framebox(750,525){}}
}%
\put(5851,-4786){\makebox(0,0)[b]{\smash{\fontsize{9}{10.8}
\usefont{T1}{cmr}{m}{n}{\color[rgb]{0,0,0}$A$}%
}}}
\put(5776,-3586){\makebox(0,0)[b]{\smash{\fontsize{9}{10.8}
\usefont{T1}{cmr}{m}{n}{\color[rgb]{0,0,0}$A$}%
}}}
\end{picture}%
\end{tabular}
}{
                              +------+
                              +------+
+-------------+           +-------------+
|             |           |             |
+-------------+           |             |
|   +------+  |           |   +//////+  |
+---|------|--+           |   +------+  |
|   |   A  |  |           |   |      |  |
+---|------|--+    ==>    +---+------+--+
|   |      |  |         
+---|------|--+           +---+------+--+
|   +------+  |           |   |      |  |
+-------------+           +---+------+--+
|             |           |   +------+  |
+-------------+           +-------------+
                          |             |
                          +-------------+
  current.fig             plainsevere.fig
}
\hcaption{plainsevere revision}
\end{hfigure}

\enddraft

Plain severe revision closely matches severe revision, differing only in the
situations that are currently believed slightly more than the new most believed
ones: the definition replaces $\imin(A)$ with $\imin(A)+1$.

\begin{definition}
\label{definition-plainsevere}

\[
C \psev(A) = [
	\min(A),
	C(0) \cup \cdots \cup C(\imin(A)+1) \backslash \min(A),
	C(\imin(A)+2), \ldots, C(\last)
]
\]

\end{definition}

A central lemma is about revising an order by a subset of its last class. The
result is the order of the revision, as shown in
Figure~\ref{figure-plainsevere-last}.

\begin{hfigure}
\long\def\ttytex#1#2{#1}
\ttytex{
\begin{tabular}{ccc}
\setlength{\unitlength}{3750sp}%
\begin{picture}(1224,3624)(4789,-5173)
\thinlines
{\color[rgb]{0,0,0}\put(4801,-3361){\line( 1, 0){1200}}
}%
{\color[rgb]{0,0,0}\put(4801,-2161){\line( 1, 0){1200}}
}%
{\color[rgb]{0,0,0}\put(4801,-4561){\line( 1, 0){1200}}
}%
{\color[rgb]{0,0,0}\put(4801,-5161){\framebox(1200,3600){}}
}%
{\color[rgb]{0,0,0}\put(4801,-3961){\line( 1, 0){1200}}
}%
{\color[rgb]{0,0,0}\put(4801,-2761){\line( 1, 0){1200}}
}%
{\color[rgb]{0,0,0}\put(5551,-5011){\framebox(300,300){}}
}%
\put(5701,-4936){\makebox(0,0)[b]{\smash{\fontsize{9}{10.8}
\usefont{T1}{cmr}{m}{n}{\color[rgb]{0,0,0}$A$}%
}}}
\end{picture}%
&
\setlength{\unitlength}{3750sp}%
\begin{picture}(1104,2184)(5659,-5383)
\thinlines
{\color[rgb]{0,0,0}\multiput(6391,-3391)(-9.47368,4.73684){20}{\makebox(2.1167,14.8167){\tiny.}}
\put(6211,-3301){\line( 0,-1){ 45}}
\put(6211,-3346){\line(-1, 0){180}}
\put(6031,-3346){\line( 0,-1){ 90}}
\put(6031,-3436){\line( 1, 0){180}}
\put(6211,-3436){\line( 0,-1){ 45}}
\multiput(6211,-3481)(9.47368,4.73684){20}{\makebox(2.1167,14.8167){\tiny.}}
}%
\end{picture}%
&
\setlength{\unitlength}{3750sp}%
\begin{picture}(1224,4074)(4789,-5173)
\thinlines
{\color[rgb]{0,0,0}\put(4801,-5161){\framebox(1200,3600){}}
}%
{\color[rgb]{0,0,0}\put(5551,-5011){\framebox(300,300){}}
}%
{\color[rgb]{0,0,0}\put(5551,-1411){\framebox(300,300){}}
}%
{\color[rgb]{0,0,0}\multiput(5551,-4861)(7.89474,7.89474){20}{\makebox(2.2222,15.5556){\tiny.}}
}%
{\color[rgb]{0,0,0}\put(5551,-5011){\line( 1, 1){300}}
}%
{\color[rgb]{0,0,0}\multiput(5701,-5011)(7.89474,7.89474){20}{\makebox(2.2222,15.5556){\tiny.}}
}%
\put(5701,-1336){\makebox(0,0)[b]{\smash{\fontsize{9}{10.8}
\usefont{T1}{cmr}{m}{n}{\color[rgb]{0,0,0}$A$}%
}}}
\end{picture}%
\end{tabular}
}{
                              +---+
                              | A |
                              +---+
+-------------+           +-------------+
|             |           |             |
|             |           |             |
|             |           |             |
+-------------+           |             |
|             |           |             |
|             |           |             |
|             |           |             |
+-------------+    ==>    |             |
|             |           |             |
|             |           |             |
|             |           |             |
+-------------+           |             |
|   +---+     |           |   +---+     |
|   | A |     |           |   |///|     |
|   +---+     |           |   +---+     |
+-------------+           +-------------+
    last.fig                 last-a.fig
}
\label{figure-plainsevere-last}
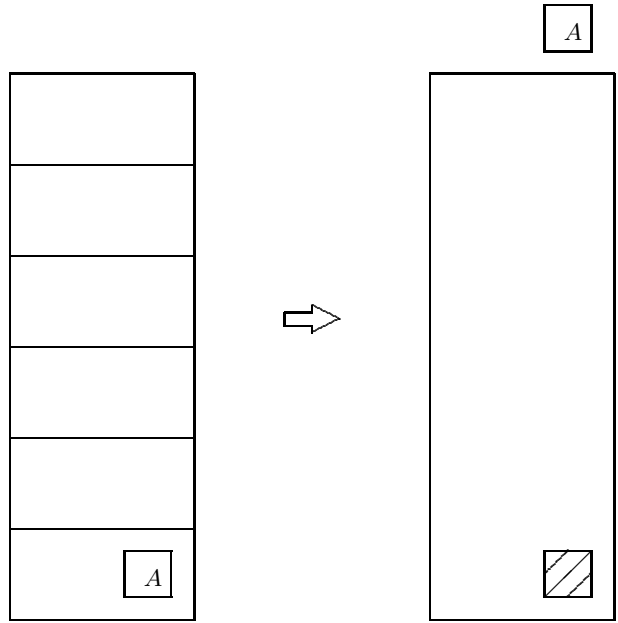
\hcaption{Plain severe revision by a subset of the last class}
\end{hfigure}

\begin{lemma}
\label{lemma-plainsevere-last}

Plainly severely revising an order $C$ by a subset $A \subseteq C(\last)$ of
its last class produces $\formulaorder A = [A, \true \backslash A]$.

\end{lemma}

\proof
Since all models of $A$ are in the last class $C(\last)$,
they are all minimal and their indexes is the last:
{} $\min(A) = A$ and $\imin(A) = \last$.

\begin{eqnarray*}
\lefteqn{C \psev(A)}
\\
&=&
[
	\min(A),
	C(0) \cup \cdots \cup C(\imin(A)+1) \backslash \min(A),
	C(\imin(A)+2), \ldots, C(\last)
]
\\
&=&
[
	A,
	C(0) \cup \cdots \cup C(\last+1) \backslash A,
	C(\last+2), \ldots, C(\last)
]
\\
&=&
[
	A,
	C(0) \cup \cdots \cup C(\last) \backslash A
]
\\
&=&
[
	A,
	\true \backslash A
]
\\
&=&
	\formulaorder A
\end{eqnarray*}

Classes $\last + 1$ and greater are empty because
$\last$ is the greatest index of $C$.~\qed

A consequence is that revising the flat order by a formula produces the order
of the formula, as shown in Figure~\ref{figure-two-plainsevere}.

\begin{hfigure}
\long\def\ttytex#1#2{#1}
\ttytex{
\begin{tabular}{ccc}
\setlength{\unitlength}{3750sp}%
\begin{picture}(1224,624)(4789,-3973)
\thinlines
{\color[rgb]{0,0,0}\put(4801,-3961){\framebox(1200,600){}}
}%
{\color[rgb]{0,0,0}\put(5176,-3811){\framebox(450,300){}}
}%
\put(5401,-3736){\makebox(0,0)[b]{\smash{\fontsize{9}{10.8}
\usefont{T1}{cmr}{m}{n}{\color[rgb]{0,0,0}$A$}%
}}}
\end{picture}%
&
\setlength{\unitlength}{3750sp}%
\begin{picture}(1104,744)(5659,-3943)
\thinlines
{\color[rgb]{0,0,0}\multiput(6391,-3391)(-9.47368,4.73684){20}{\makebox(2.1167,14.8167){\tiny.}}
\put(6211,-3301){\line( 0,-1){ 45}}
\put(6211,-3346){\line(-1, 0){180}}
\put(6031,-3346){\line( 0,-1){ 90}}
\put(6031,-3436){\line( 1, 0){180}}
\put(6211,-3436){\line( 0,-1){ 45}}
\multiput(6211,-3481)(9.47368,4.73684){20}{\makebox(2.1167,14.8167){\tiny.}}
}%
\end{picture}%
&
\setlength{\unitlength}{3750sp}%
\begin{picture}(1224,1074)(4789,-3973)
\thinlines
{\color[rgb]{0,0,0}\put(4801,-3961){\framebox(1200,600){}}
}%
{\color[rgb]{0,0,0}\put(5176,-3811){\framebox(450,300){}}
}%
{\color[rgb]{0,0,0}\put(5176,-3211){\framebox(450,300){}}
}%
{\color[rgb]{0,0,0}\multiput(5176,-3661)(7.89474,7.89474){20}{\makebox(2.2222,15.5556){\tiny.}}
}%
{\color[rgb]{0,0,0}\put(5176,-3811){\line( 1, 1){300}}
}%
{\color[rgb]{0,0,0}\put(5326,-3811){\line( 1, 1){300}}
}%
{\color[rgb]{0,0,0}\multiput(5476,-3811)(7.89474,7.89474){20}{\makebox(2.2222,15.5556){\tiny.}}
}%
\put(5401,-3136){\makebox(0,0)[b]{\smash{\fontsize{9}{10.8}
\usefont{T1}{cmr}{m}{n}{\color[rgb]{0,0,0}$A$}%
}}}
\end{picture}%
\end{tabular}
}{
                                +---+
                                | A |
                                +---+
+-------------+           +-------------+
|     +---+   |           |     +---+   |
|     | A |   |           |     |///|   |
|     +---+   |    ==>    |     +---+   |
+-------------+           +-------------+
   flat-a.fig                  a.fig
}
\label{figure-two-plainsevere}
\hcaption{A plain severe revision of $\flatorder$}
\end{hfigure}

\begin{lemma}
\label{lemma-plainsevere-flatorder}

Plainly severely revising $\flatorder$ by $A$ produces
{} $\flatorder \psev(A) = [A, \true \backslash A]$.

\end{lemma}

\proof The flat order $\flatorder$ only contains one class $\flatorder(0) =
\true$, which is therefore also its last. The set of models of every formula is
contained in $\true$. Lemma~\ref{lemma-plainsevere-last} applies: $\flatorder
\psev(A) = [A, \true \backslash A] = \formulaorder A$.~\qed

Contrary to the other severe revisions, plain severe revision is not learnable:
the flat doxastic state is never changed in a state of more than two classes.
Revising the flat doxastic state produces two classes at most, which remains
two.

\begin{lemma}
\label{lemma-plainsevere-one-two}

The order $\flatorder \psev(A)$ comprises at most two classes.

\end{lemma}

\proof Lemma~\ref{lemma-plainsevere-flatorder} proves that
{} $\flatorder \psev(A)$
is
{} $\formulaorder A = [A, \true \backslash A]$.
This order comprises two classes, or one if the other is empty.~\qed

No third class is ever generated out of two.

\begin{lemma}
\label{lemma-plainsevere-two-two}

If $C$ comprises two classes, so does $C \psev(A)$.

\end{lemma}

\proof The classes of $C$ are two:
{} $\last=1$.

\begin{eqnarray*}
\lefteqn{C \psev(A)} \\
&=& [
	\min(A),
	C(0) \cup \cdots \cup C(\imin(A)+1) \backslash \min(A),
	C(\imin(A)+2), \ldots, C(\last)
]
\\
&=& [
	\min(A),
	C(0) \cup \cdots \cup C(\imin(A)+1) \backslash \min(A),
	C(\imin(A)+2), \ldots, C(1)
]
\\
&=& [
	\min(A),
	C(0) \cup \cdots \cup C(\imin(A)+1) \backslash \min(A)
]
\end{eqnarray*}

The part
{} $C(\imin(A)+2), \ldots, C(1)$
is empty since its first index $\imin(A) + 2$ is greater than its last $1$.
Only two classes remain.

The first is not empty because revisions by contradictions are excluded. The
second is not empty because that would imply that the first comprises all
models. Since $A$ is a superset of $\min(A)$, it comprises all models as well.
Including all models of $C(0)$. The conclusion $C(0) = \true$ implies $C(1) =
\emptyset$, contradicting the assumption that $C$ comprises two classes.~\qed

A sequence of plain severe revisions does not increase the number of classes of
the flat order over two. No order of more classes comes from revising the flat
order.

\begin{theorem}
\label{theorem-plainsevere-learnable}

Plain severe revision is not learnable.

\end{theorem}

\proof

No sequence of plain severe revisions turns $\flatorder$ into an order
comprising two or more classes:

\begin{itemize}

\item $\flatorder \psev(A)$ comprises either one or two
classes by Lemma~\ref{lemma-plainsevere-one-two};

\item the only one-class order is $\flatorder$ itself;

\item two-class orders are always revised in two-class orders by
Lemma~\ref{lemma-plainsevere-two-two}.

\end{itemize}

\qed

Plain severe revision does not turn a two-class order into the flat one.

\begin{theorem}
\label{theorem-plainsevere-amnesic}

Plain severe revision is not amnesic.

\end{theorem}

\proof Plain severe revision turns a two-class order into another two-class
order by Lemma~\ref{lemma-plainsevere-two-two}. Iteratively, the order produced
by a sequence of revisions comprises two classes.~\qed

\begin{theorem}
\label{theorem-plainsevere-dogmatic}

Plain severe revision is dogmatic.

\end{theorem}

\proof The first step into turning $C$ into $[G(0),G(1)]$ is to revise it by a
model $I$ of $C(\last)$. By Lemma~\ref{lemma-plainsevere-last}, the result is
$[\{I\}, \true \backslash \{I\}]$.

\begin{hfigure}
\setlength{\unitlength}{3750sp}%
\begin{picture}(1224,1074)(4789,-5173)
\thinlines
{\color[rgb]{0,0,0}\put(4801,-5161){\framebox(1200,600){}}
}%
{\color[rgb]{0,0,0}\put(5251,-4411){\framebox(300,300){}}
}%
{\color[rgb]{0,0,0}\put(5251,-5011){\framebox(300,300){}}
}%
{\color[rgb]{0,0,0}\multiput(5251,-4861)(7.89474,7.89474){20}{\makebox(2.2222,15.5556){\tiny.}}
}%
{\color[rgb]{0,0,0}\put(5251,-5011){\line( 1, 1){300}}
}%
{\color[rgb]{0,0,0}\multiput(5401,-5011)(7.89474,7.89474){20}{\makebox(2.2222,15.5556){\tiny.}}
}%
\put(5401,-4336){\makebox(0,0)[b]{\smash{\fontsize{9}{10.8}
\usefont{T1}{cmr}{m}{n}{\color[rgb]{0,0,0}$I$}%
}}}
\end{picture}%
\nop{
      +-+
      |I|
      +-+
+-------------+
|     +-+     |
|     |/|     |
|     +-+     |
+-------------+
   flat-i.fig
}%
\label{figure-plainsevere-lastmodel}
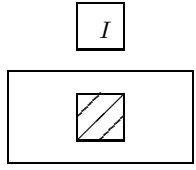
\hcaption{Plain severe revision by a model of the last class}
\end{hfigure}

This order is then revised by a model $J$ of $G(1)$. If $J$ differs from $I$,
it is in the last class and the result is $\formulaorder{\{J\}} = [\{J\}, \true
\{J\}]$ by Lemma~\ref{lemma-plainsevere-last}. If it coincides with $I$, then
its only minimal model is $I$ itself: $\min(\{J\}) = \{J\}$ and $\imin(\{J\}) =
0$.

\begin{eqnarray*}
\lefteqn{\formulaorder{\{J\}} \psev(\{J\})}
\\
&=&
[
	\min(\{J\}),
	C(0) \cup \cdots \cup C(\imin(\{J\})+1) \backslash \min(\{J\}),
	C(\imin(\{J\})+2), \ldots, C(\last)
]
\\
&=&
[
	\{J\},
	C(0) \cup \cdots \cup C(0+1) \backslash \{J\},
	C(0+2), \ldots, C(1)
]
\\
&=&
[
	\{J\},
	C(0) \cup \cdots \cup C(1) \backslash \{J\}
]
\\
&=&
[
	\{J\},
	\true \backslash \{J\}
]
\\
&=&
	\formulaorder{\{J\}}
\end{eqnarray*}

\begin{hfigure}
\long\def\ttytex#1#2{#1}
\ttytex{
\begin{tabular}{ccc}
\setlength{\unitlength}{3750sp}%
\begin{picture}(1224,1074)(4789,-5173)
\thinlines
{\color[rgb]{0,0,0}\put(4801,-5161){\framebox(1200,600){}}
}%
{\color[rgb]{0,0,0}\put(5251,-4411){\framebox(300,300){}}
}%
{\color[rgb]{0,0,0}\put(5251,-5011){\framebox(300,300){}}
}%
{\color[rgb]{0,0,0}\multiput(5251,-4861)(7.89474,7.89474){20}{\makebox(2.2222,15.5556){\tiny.}}
}%
{\color[rgb]{0,0,0}\put(5251,-5011){\line( 1, 1){300}}
}%
{\color[rgb]{0,0,0}\multiput(5401,-5011)(7.89474,7.89474){20}{\makebox(2.2222,15.5556){\tiny.}}
}%
\put(5401,-4336){\makebox(0,0)[b]{\smash{\fontsize{9}{10.8}
\usefont{T1}{cmr}{m}{n}{\color[rgb]{0,0,0}$I$}%
}}}
\end{picture}%
&
\setlength{\unitlength}{3750sp}%
\begin{picture}(1104,1104)(5659,-4303)
\thinlines
{\color[rgb]{0,0,0}\multiput(6391,-3391)(-9.47368,4.73684){20}{\makebox(2.1167,14.8167){\tiny.}}
\put(6211,-3301){\line( 0,-1){ 45}}
\put(6211,-3346){\line(-1, 0){180}}
\put(6031,-3346){\line( 0,-1){ 90}}
\put(6031,-3436){\line( 1, 0){180}}
\put(6211,-3436){\line( 0,-1){ 45}}
\multiput(6211,-3481)(9.47368,4.73684){20}{\makebox(2.1167,14.8167){\tiny.}}
}%
\end{picture}%
&
\setlength{\unitlength}{3750sp}%
\begin{picture}(1224,1074)(4789,-5173)
\thinlines
{\color[rgb]{0,0,0}\put(4801,-5161){\framebox(1200,600){}}
}%
{\color[rgb]{0,0,0}\put(5551,-5011){\framebox(300,300){}}
}%
{\color[rgb]{0,0,0}\multiput(5551,-4861)(7.89474,7.89474){20}{\makebox(2.2222,15.5556){\tiny.}}
}%
{\color[rgb]{0,0,0}\put(5551,-5011){\line( 1, 1){300}}
}%
{\color[rgb]{0,0,0}\multiput(5701,-5011)(7.89474,7.89474){20}{\makebox(2.2222,15.5556){\tiny.}}
}%
{\color[rgb]{0,0,0}\put(5551,-4411){\framebox(300,300){}}
}%
{\color[rgb]{0,0,0}\put(4951,-5011){\framebox(450,300){}}
}%
\put(5701,-4336){\makebox(0,0)[b]{\smash{\fontsize{9}{10.8}
\usefont{T1}{cmr}{m}{n}{\color[rgb]{0,0,0}$J$}%
}}}
\put(5176,-4936){\makebox(0,0)[b]{\smash{\fontsize{9}{10.8}
\usefont{T1}{cmr}{m}{n}{\color[rgb]{0,0,0}$G(0)$}%
}}}
\end{picture}%
\end{tabular}
}{
      +-+                           +-+  
      |I|                           |J|  
      +-+                           +-+  
+-------------+    ==>    +-------------+
|     +-+     |           | +----+  +-+ |
|     |/|     |           | |G(0)|  |/| |
|     +-+     |           | +----+  +-+ |
+-------------+           +-------------+
   flat-i.fig              flat-j-last.fig
}
\label{figure-plainsevere-last-j}
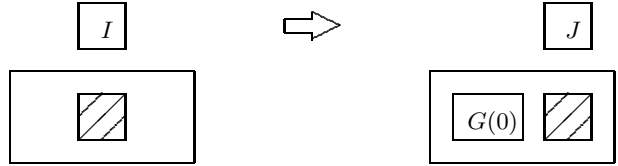
\hcaption{Plain severe revision
	by a model in the second class of the target order}
\end{hfigure}

Either way, the result is
{} $\formulaorder{\{J\}} = [\{J\}, \true \backslash \{J\}]$.

\begin{hfigure}
\long\def\ttytex#1#2{#1}
\ttytex{
\begin{tabular}{ccc}
\setlength{\unitlength}{3750sp}%
\begin{picture}(1224,1074)(4789,-5173)
\thinlines
{\color[rgb]{0,0,0}\put(4801,-5161){\framebox(1200,600){}}
}%
{\color[rgb]{0,0,0}\put(5551,-5011){\framebox(300,300){}}
}%
{\color[rgb]{0,0,0}\multiput(5551,-4861)(7.89474,7.89474){20}{\makebox(2.2222,15.5556){\tiny.}}
}%
{\color[rgb]{0,0,0}\put(5551,-5011){\line( 1, 1){300}}
}%
{\color[rgb]{0,0,0}\multiput(5701,-5011)(7.89474,7.89474){20}{\makebox(2.2222,15.5556){\tiny.}}
}%
{\color[rgb]{0,0,0}\put(5551,-4411){\framebox(300,300){}}
}%
{\color[rgb]{0,0,0}\put(4951,-5011){\framebox(450,300){}}
}%
\put(5701,-4336){\makebox(0,0)[b]{\smash{\fontsize{9}{10.8}
\usefont{T1}{cmr}{m}{n}{\color[rgb]{0,0,0}$J$}%
}}}
\put(5176,-4936){\makebox(0,0)[b]{\smash{\fontsize{9}{10.8}
\usefont{T1}{cmr}{m}{n}{\color[rgb]{0,0,0}$G(0)$}%
}}}
\end{picture}%
&
\setlength{\unitlength}{3750sp}%
\begin{picture}(1104,1104)(5659,-4303)
\thinlines
{\color[rgb]{0,0,0}\multiput(6391,-3391)(-9.47368,4.73684){20}{\makebox(2.1167,14.8167){\tiny.}}
\put(6211,-3301){\line( 0,-1){ 45}}
\put(6211,-3346){\line(-1, 0){180}}
\put(6031,-3346){\line( 0,-1){ 90}}
\put(6031,-3436){\line( 1, 0){180}}
\put(6211,-3436){\line( 0,-1){ 45}}
\multiput(6211,-3481)(9.47368,4.73684){20}{\makebox(2.1167,14.8167){\tiny.}}
}%
\end{picture}%
&
\setlength{\unitlength}{3750sp}%
\begin{picture}(1224,1074)(4789,-5173)
\thinlines
{\color[rgb]{0,0,0}\put(4801,-5161){\framebox(1200,600){}}
}%
{\color[rgb]{0,0,0}\put(4951,-5011){\framebox(450,300){}}
}%
{\color[rgb]{0,0,0}\put(4951,-4411){\framebox(450,300){}}
}%
{\color[rgb]{0,0,0}\multiput(4951,-4861)(7.89474,7.89474){20}{\makebox(2.2222,15.5556){\tiny.}}
}%
{\color[rgb]{0,0,0}\put(4951,-5011){\line( 1, 1){300}}
}%
{\color[rgb]{0,0,0}\put(5101,-5011){\line( 1, 1){300}}
}%
{\color[rgb]{0,0,0}\multiput(5251,-5011)(7.89474,7.89474){20}{\makebox(2.2222,15.5556){\tiny.}}
}%
\put(5176,-4336){\makebox(0,0)[b]{\smash{\fontsize{9}{10.8}
\usefont{T1}{cmr}{m}{n}{\color[rgb]{0,0,0}$G(0)$}%
}}}
\end{picture}%
\end{tabular}
}{
          +-+               +----+
          |J|               |G(0)|
          +-+               +----+
+-------------+           +-------------+
| +----+  +-+ |    ==>    | +----+      |
| |G(0)|  |/| |           | |////|      |
| +----+  +-+ |           | +----+      |
+-------------+           +-------------+
flat-j-last.fig           first-true.fig
}
\label{figure-plainsevere-first}
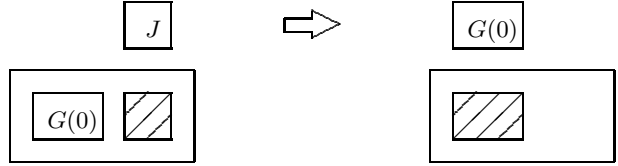
\hcaption{Plain severe revision by the first class of the target order}
\end{hfigure}

By construction, $J$ is a model of $G(1)$. It is therefore not a model of
$G(0)$ because equivalence classes are disjoint by definition. Since $J$ is the
only model of the first class, $G(0)$ is all contained in the second.
Lemma~\ref{lemma-plainsevere-last} proves that the revision $G(0)$
produces $[G(0), \true \backslash G(0)]$, the target order.~\qed

Plain severe revision is not damascan: it does not invert an order of three
classes or more.

\begin{theorem}
\label{theorem-plainsevere-damascan}

Plain severe revision is not damascan.

\end{theorem}

\proof A plain severe revision either leaves the last class untouched or
shortens the order to two classes and never elongate it.

\begin{eqnarray*}
\lefteqn{C \psev(A)}
\\
&=&
[
	\min(A),
	C(0) \cup \cdots \cup C(\imin(A)+1) \backslash \min(A),
	C(\imin(A)+2), \ldots, C(\last)
]
\end{eqnarray*}

The definition ends with $C(\last)$, suggesting that plain severe revision
never changes the last class of the order. The caveat is that the final part
{} $C(\imin(A)+2), \ldots, C(\last)$
of this order is empty if $\imin(A) + 2 > \last$, that is, $\imin(A)$ is either
$\last-1$ or $\last$. Regardless, the resulting order is what remains:
{} $[\min(A), C(0) \cup \cdots \cup C(\imin(A)+1) \backslash \min(A)]$.
This order only comprises two classes, one if $\min(A) = \true$. As proved by
Lemma~\ref{lemma-plainsevere-one-two} and
Lemma~\ref{lemma-plainsevere-two-two}, plain severe revisions of such order
never create a third classes.

Plain severe revisions do not invert a three-class order $C = [C(0), C(1),
C(2)]$ because they either leave $C(2)$ as the last class or produce a
one-class or two-class order.~\qed


\bibliographystyle{alpha}

\begin{thebibliography}{DHKP11}

\bibitem[AGM85]{alch-gard-maki-85}
C.~E. Alchourr\'on, P.~G{\"a}rdenfors, and D.~Makinson.
\newblock On the logic of theory change: Partial meet contraction and revision
  functions.
\newblock {\em Journal of Symbolic Logic}, 50:510--530, 1985.

\bibitem[APW18]{arav-etal-18}
T.I. Aravanis, P.~Peppas, and M.-A. Williams.
\newblock Iterated belief revision and {D}alal's operator.
\newblock In {\em Proceedings of the 10th Hellenic Conference on Artificial
  Intelligence ({SETN}~2018)}, pages 26:1--26:4. {ACM} Press, 2018.

\bibitem[ARS02]{andr-etal-02}
H.~Andr{\'{e}}ka, M.~Ryan, and P.-Y. Schobbens.
\newblock Operators and laws for combining preference relations.
\newblock {\em Journal of Logic and Computation}, 12(1):13--53, 2002.

\bibitem[BCD{\etalchar{+}}93]{benf-etal-93}
S.~Benferhat, C.~Cayrol, D.~Dubois, J.~Lang, and H.~Prade.
\newblock Inconsistency management and prioritized syntax-based entailment.
\newblock In {\em Proceedings of the Thirteenth International Joint Conference
  on Artificial Intelligence (IJCAI'93)}, pages 640--647, 1993.

\bibitem[Bre89]{brew-89}
G.~Brewka.
\newblock Preferred subtheories: an extended logical framework for default
  reasoning.
\newblock In {\em Proceedings of the Eleventh International Joint Conference on
  Artificial Intelligence (IJCAI'89)}, pages 1043--1048, 1989.

\bibitem[Can97]{cant-97}
J.~Cantwell.
\newblock On the logic of small changes in hypertheories.
\newblock {\em Theoria}, 63(1-2):54--89, 1997.

\bibitem[DHKP11]{doms-etal-11}
C.~Domshlak, E.~H{\"{u}}llermeier, S.~Kaci, and H.~Prade.
\newblock Preferences in {AI:} an overview.
\newblock {\em Artificial Intelligence}, 175(7-8):1037--1052, 2011.

\bibitem[DP97]{darw-pear-97}
A.~Darwiche and J.~Pearl.
\newblock On the logic of iterated belief revision.
\newblock {\em Artificial Intelligence Journal}, 89(1--2):1--29, 1997.

\bibitem[G{\"a}r88]{gard-88}
P.~G{\"a}rdenfors.
\newblock {\em Knowledge in Flux: Modeling the Dynamics of Epistemic States}.
\newblock Bradford Books, MIT Press, Cambridge, MA, 1988.

\bibitem[GK18]{gura-kodi-18}
D.P. Guralnik and D.E. Koditschek.
\newblock Iterated belief revision under resource constraints: {L}ogic as
  geometry.
\newblock {\em Computing Research Repository (CoRR)}, abs/1812.08313, 2018.

\bibitem[KM91]{kats-mend-91}
H.~Katsuno and A.~O. Mendelzon.
\newblock On the difference between updating a knowledge base and revising it.
\newblock In {\em Proceedings of the Second International Conference on the
  Principles of Knowledge Representation and Reasoning (KR'91)}, pages
  387--394, 1991.

\bibitem[Kut19]{kuts-19}
S.~Kutsch.
\newblock {InfOCF-Lib}: {A} {J}ava library for {OCF}-based conditional
  inference.
\newblock In {\em Proceedings of the eighth Workshop on Dynamics of Knowledge
  and Belief {(DKB-2019)}}, volume 2445 of {\em {CEUR} Workshop Proceedings},
  pages 47--58, 2019.

\bibitem[Lib97]{libe-97-c}
P.~Liberatore.
\newblock The complexity of iterated belief revision.
\newblock In {\em Proceedings of the Sixth International Conference on Database
  Theory (ICDT'97)}, pages 276--290, 1997.

\bibitem[Lib23]{libe-23}
P.~Liberatore.
\newblock Mixed iterated revisions: Rationale, algorithms and complexity.
\newblock {\em {ACM} Transactions on Computational Logic}, 24(3), 2023.

\bibitem[Lib25]{libe-25}
P.~Liberatore.
\newblock Natural revision is contingently-conditionalized revision.
\newblock {\em Journal of Automated Reasoning}, 186, 2025.

\bibitem[Nay94]{naya-94}
A.~Nayak.
\newblock Iterated belief change based on epistemic entrenchment.
\newblock {\em Erkenntnis}, 41:353--390, 1994.

\bibitem[Neb91]{nebe-91}
B.~Nebel.
\newblock Belief revision and default reasoning: Syntax-based approaches.
\newblock In {\em Proceedings of the Second International Conference on the
  Principles of Knowledge Representation and Reasoning (KR'91)}, pages
  417--428, 1991.

\bibitem[Par99]{pari-99}
R.~Parikh.
\newblock Beliefs, belief revision, and splitting languages.
\newblock {\em Logic, language and computation}, 2(96):266--268, 1999.

\bibitem[Rot09]{rott-09}
H.~Rott.
\newblock Shifting priorities: Simple representations for twenty-seven iterated
  theory change operators.
\newblock In D.~Makinson, J.~Malinowski, and H.~Wansing, editors, {\em Towards
  Mathematical Philosophy}, volume~28 of {\em Trends in Logic}, pages 269--296.
  Springer Netherlands, 2009.

\bibitem[SKB22]{saue-etal-22}
K.~Sauerwald, G.~Kern{-}Isberner, and C.~Beierle.
\newblock A conditional perspective on the logic of iterated belief
  contraction.
\newblock {\em Computing Research Repository (CoRR)}, abs/2202.03196, 2022.

\bibitem[SMV19]{souz-etal-19}
M.~Souza, A.F. Moreira, and R.~Vieira.
\newblock Iterated belief base revision: {A} dynamic epistemic logic approach.
\newblock In {\em Proceedings of the Thirdy-Third AAAI Conference on Artificial
  Intelligence (AAAI~2019)}, pages 3076--3083. {AAAI} Press/The {MIT} Press,
  2019.

\bibitem[Spo88]{spoh-88}
W.~Spohn.
\newblock Ordinal conditional functions: A dynamic theory of epistemic states.
\newblock In {\em Causation in Decision, Belief Change, and Statistics}, pages
  105--134. Kluwer Academics, 1988.

\end{thebibliography}
\newcommand{\etalchar}[1]{$^{#1}$}

\end{document}